# Human Rights, Democracy, and the Rule of Law Assurance Framework for AI Systems:

## A proposal prepared for the Council of Europe's Ad hoc Committee on Artificial Intelligence

David Leslie, Christopher Burr, Mhairi Aitken, Michael Katell, Morgan Briggs, and Cami Rincon

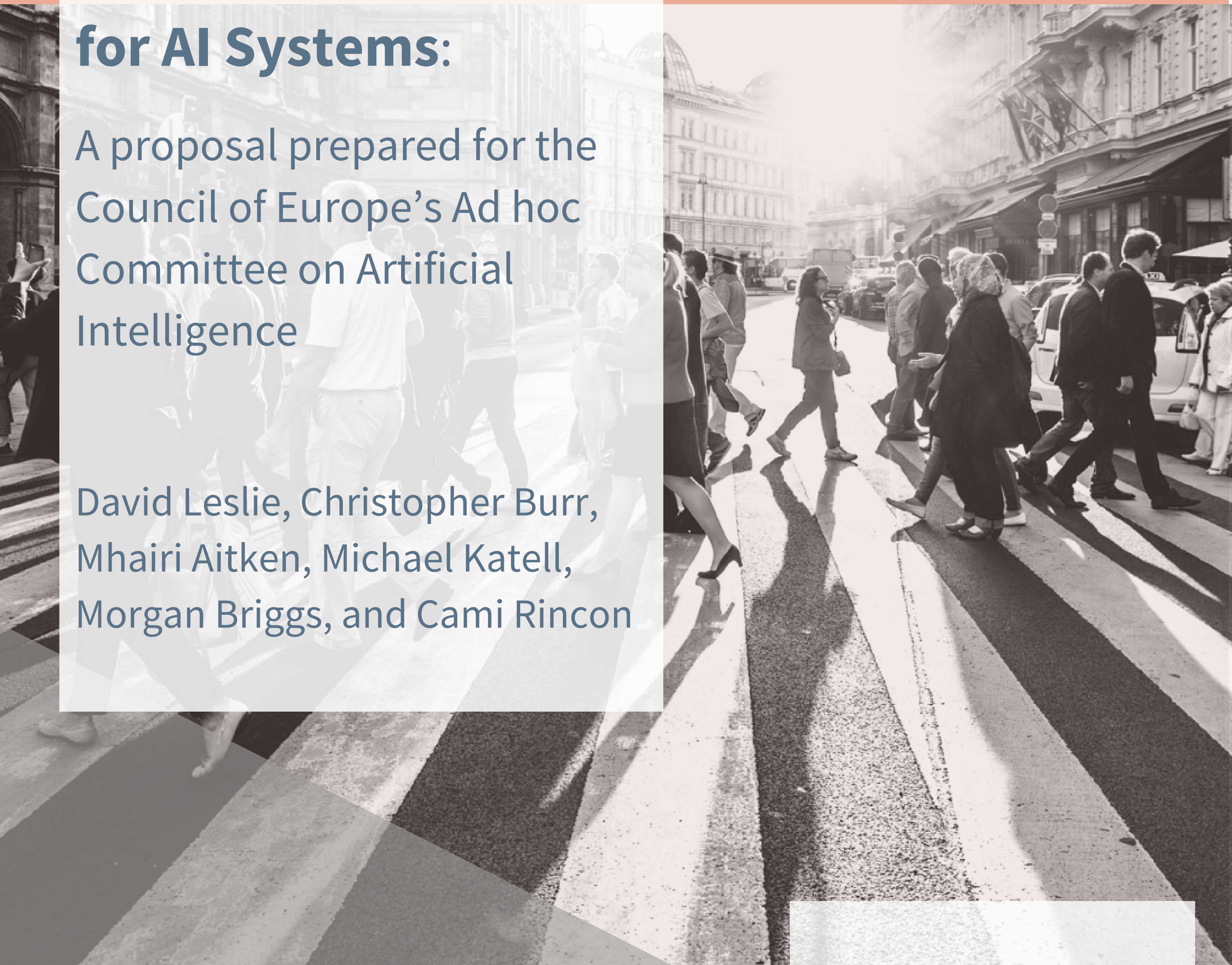

The Public Policy Programme at The Alan Turing Institute was set up in May 2018 with the aim of developing research, tools, and techniques that help governments innovate with data-intensive technologies and improve the quality of people's lives. We work alongside policy makers to explore how data science and artificial intelligence can inform public policy and improve the provision of public services. We believe that governments can reap the benefits of these technologies only if they make considerations of ethics and safety a first priority.

Please note, that this proposed framework is a living document that will evolve and improve with input from users, affected stakeholders, and interested parties. We need your participation. Please share feedback with us at policy@turing.ac.uk
This research was supported, in part, by ESRC Grant ES/T007354/1, EPSRC Grant EP/T001569/1, EPSRC Grant EP/W006022/1, and from the public funds that make the Turing's Public Policy Programme possible.

https://www.turing.ac.uk/research/research-programmes/public-policy

Cite this work as:

- Leslie, D., Burr, C., Aitken, M., Katell, M., Briggs, M., Rincon, C. (2021). *Human rights, democracy, and the rule of law assurance framework for AI systems: A proposal*. The Alan Turing Institute. https://doi.org/10.5281/zenodo.5981676

# Table of Contents

# Abbreviations

| | |
|---|---|
| CAHAI | Council of Europe’s Ad hoc Committee on Artificial Intelligence |
| CoE | Council of Europe |
| DHRI | Danish Human Rights Institute |
| HUDERAC | Human Rights, Democracy, and the Rule of Law Assurance Case |
| HUDERAF | Human Rights, Democracy, and the Rule of Law Assurance Framework |
| HUDERIA | Human Rights, Democracy, and the Rule of Law Impact Assessment |
| HRIA | Human Rights Impact Assessment |
| IMP | Impact Mitigation Plan |
| PCRA | Preliminary Context Based Risk Assessment |
| PS Report | Project Summary Report |
| RIN | Risk Index Number |
| SEP | Stakeholder Engagement Process |
| UNGP | *United Nations Guiding Principles on Business and Human Rights* |
| UNHROHC | UN Human Rights Office of the High Commissioner |

# Executive Summary

Following on from the publication of its *Feasibility Study* in December 2020, the Council of Europe's Ad Hoc Committee on Artificial Intelligence (and its subgroups) initiated efforts to formulate and draft its *Possible elements of a legal framework on artificial intelligence, based on the Council of Europe's standards on human rights, democracy and the rule of law*. This document was ultimately adopted by the CAHAI plenary in December 2021. To support this effort, The Alan Turing Institute undertook a programme of research that explored the governance processes and practical tools needed to operationalise the integration of human right due diligence with the assurance of trustworthy AI innovation practices.

The resulting framework was completed and submitted to the Council of Europe in September 2021. It presents an end-to-end approach to the assurance of AI project lifecycles that integrates context-based risk analysis and appropriate stakeholder engagement with comprehensive impact assessment, and transparent risk management, impact mitigation, and innovation assurance practices. Taken together, these interlocking processes constitute a *Human Rights, Democracy and the Rule of Law Assurance Framework* (HUDERAF). The HUDERAF combines the procedural requirements for principles-based human rights due diligence with the governance mechanisms needed to set up technical and socio-technical guardrails for responsible and trustworthy AI innovation practices. Its purpose is to provide an accessible and user-friendly set of mechanisms for facilitating compliance with a binding legal framework on artificial intelligence, based on the Council of Europe's standards on human rights, democracy and the rule of law, and to ensure that AI innovation projects are carried out with appropriate levels of public accountability, transparency, and democratic governance.

The HUDERAF encompasses four interrelated elements:

(1) The Preliminary Context-Based Risk Analysis (PCRA) provides an initial indication of the context-based risks that an AI system could pose to human rights, fundamental freedoms, and elements of democracy and the rule of law. The main purpose of the PCRA is to help AI project teams establish a proportionate approach to risk management and assurance practices and to the level of stakeholder engagement that is needed across the project lifecycle.

(2) The Stakeholder Engagement Process (SEP) helps project teams to identify stakeholder salience and to facilitate proportionate stakeholder involvement and input throughout the project workflow. This process safeguards the equity and the contextual accuracy of HUDERAF governance processes through stakeholder involvement, revisitation, and evaluation.

(3) The Human Rights, Democracy, and the Rule of Law Impact Assessment (HUDERIA) provides an opportunity for project teams and engaged stakeholders to collaboratively produce detailed evaluations of the potential and actual impacts that the design, development, and use of an AI system could have on human rights, fundamental freedoms and elements of democracy and the rule of law. This process contextualizes and corroborates potential harms that have been previously identified, enables the discovery of further harms through the integration of stakeholder perspectives, makes possible the collaborative assessment of the severity of potential adverse impacts identified, facilitates the co-design of an impact mitigation plan, sets up access to remedy, and establishes monitoring and impact re-assessment protocols.

(4) The Human Rights, Democracy, and Rule of Law Assurance Case (HUDERAC) enables AI project teams to build a structured argument that provides demonstrable assurance to stakeholders that claims about the attainment of goals established in the HUDERIA and other HUDERAF governance processes are warranted given available evidence. The process of developing an assurance case assists internal reflection and deliberation, promoting the adoption of best practices and integrating these into design, development, and deployment lifecycles. It also provides an accessible way to convey to impacted stakeholders information about actions taken across the project workflow to mitigate risks and to ensure the ascertainment of relevant normative goals. A diligently executed assurance case offers a clear and understandable mechanism of risk management and impact mitigation that supports appropriate levels of social licence, accountability, and transparency.

The HUDERAF instruments and processes have been built to steward multi-stakeholder involvement, engagement, and consultation across project lifecycles in accordance with the democratic priorities and values of the Council of Europe. These instruments and processes are meant to be adopted by the designers, developers, and users of AI systems as an innovation-enabling way to facilitate better, more responsible innovation practices, which conform to human rights, democracy, and the rule of law, so that these technologies can optimally promote individual, community, and planetary wellbeing.

The HUDERAF has also been designed to be as "algorithm neutral" and practice based as possible so that it can remain maximally future proof and inclusive of different AI applications. The HUDERAF model will, however, need to stay responsive to the development of novel AI innovations and use-cases. Unanticipated breakthroughs, unforeseen application scenarios, and exponential increases in the efficacy of existing AI technologies could pose new and unexpected risks to fundamental rights and freedoms that transform the project governance and impact mitigation requirements for assuring the conformity of these systems to human rights standards and elements of democracy and the rule of law. For this reason, the HUDERAF model should be seen as dynamic and in need of regular revisitation and re-evaluation. This will require that inclusive and multi-stakeholder processes be set in place for the intermittent re-assessment and updating of the model.

# Introduction and Overview

Dear members of the CAHAI-PDG and other esteemed colleagues and excellencies,

What follows is a proposal for a Human Rights, Democracy, and the Rule of Law Assurance Framework (HUDERAF) for AI systems that endeavours to operationalize the outline of a Model for a Human Rights, Democracy, and the Rule of Law Impact Assessment (HUDERIA) presented in CAHAI-PDG(2021)05rev. The purpose of the latter document was to "define a methodology to carry out impact assessments of Artificial Intelligence (AI) applications from the perspective of human rights, democracy, and the rule of law, based on relevant Council of Europe (CoE) standards and the work already undertaken in this field at the international and national level…, and to develop an impact assessment model."

The challenge put forward in CAHAI-PDG(2021)05rev was fourfold:

1. To develop an impact assessment model that integrates the practices of human right due diligence contained in general Human Rights Impact Assessments (HRIAs) with AI-centred approaches to algorithmic impact assessment and the assurance of trustworthy AI innovation practices,

2. To apply a risk-based approach guided by the precautionary principles to determine use-cases that trigger pre-emptive measures where high risks to human rights, democracy and the rule of law are present and cannot be mitigated,

3. To formulate a methodology of impact assessment that follows the proportionality principle, namely, that helps AI innovators and project teams to establish a proportionate approach to the impact assessment process and to the practices of stakeholder engagement, risk management, impact mitigation, and innovation assurance that derive therefrom, and

4. To develop a methodology for assessing and grading the likelihood and extent of risks associated with an AI system that takes into account the use-contexts and purposes of the system, its underlying technology, the actors involved in the production of the system and its stage of development, and the views of potentially impacted stakeholders.

As stated in CAHAI-PDG(2021)05rev, taken together, these desired elements of the HUDERIA demand a "coherent and integrated" approach: "The model for a HUDERIA should provide a coherent and integrated approach for assessing adverse impact on human rights, democracy and the rule of law generated by AI systems, addressing simultaneously the risks arising from the specific and inherent characteristics of AI systems and the impact of such systems on human rights, rule of law and democracy." A four-step process is suggested to achieve this:

**STEP ONE:** Identify relevant human rights—and rights proxies for democracy and the rule of law—that could be adversely impacted

**STEP TWO:** Assess the impact on those rights, encompassing both technical and socio-technical dimensions

**STEP THREE:** Assess governance mechanisms to ensure the mitigation of risks, appropriate stakeholder engagement, effective remedy, accountability, and transparency

**STEP FOUR:** Continuously monitor and evaluate the system to ensure that impact assessment, impact mitigation, and governance mechanisms are sufficiently responsive to changes in context and operating environment.

The framework we present here builds on and further develops these important ideas in the ends of constructing a cohesive set of practicable tools and processes that bring the HUDERIA to life. One of the main challenges we have faced in trying to do this has been that the "coherent and integrated" approach needed to bring all of the HUDERIA functions and purposes together requires that several different streams of activity be developed independently but, at the same time, remain functionally interwoven. We have focused much of our efforts on working out how best to integrate distinctive tasks of risk analysis, stakeholder engagement, impact assessment, risk management, impact mitigation, and innovation governance and assurance into a coherent and integrated whole.

Another related challenge has been figuring out, in accordance with the goals set out in CAHAI-PDG(2021)05rev, how to combine technical and socio-technical desiderata for responsible AI innovation with the procedural requirements for principles-based human rights due diligence. There is no clear precedence for this kind of synthesis, and building the connective tissue between the evaluative and dialogical processes needed to establish human rights due diligence and the practical principles and governance mechanisms needed to set up guardrails for responsible AI innovation practices is a complex undertaking.

### *A Proposal for a Coherent and Integrated Approach*

Our strategy in confronting these challenges has been to build several clearly articulated but sensibly interrelated processes (and instruments) that cover all of the functions and goals that were sketched out by the PDG. The result is the proposed Human Rights, Democracy, and the Rule of Law Assurance Framework (HUDERAF). There are four main elements of the HUDERAF:

1. The **Preliminary Context-Based Risk Analysis (PCRA)** provides an initial indication of the context-based risks that an AI system could pose to human rights, fundamental freedoms, and elements of democracy and the rule of law. The main purpose of the PCRA is to help project teams establish a proportionate approach to risk management and assurance practices and to the level of stakeholder engagement that is needed across the project lifecycle. The PCRA process begins with a set of project scoping and planning activities in which members of the project team (a) consolidate information about the project-, use-, domain-, and data-contexts of the prospective system, (b) identify relevant stakeholders, (c) begin to scope the potential impacts of their systems on human rights, democracy, and the rule of law by identifying salient rights and freedoms, and (d) map the governance workflow.

After these activities are recorded in a Project Summary Report, the project team is ready to complete its PCRA. The PCRA is a set of questions and prompts that help project teams to pinpoint risk factors and potential adverse impacts of their systems on human rights, fundamental freedoms, and elements of democracy and the rule of law. When the PCRA is finished, it automatically generates a report (through triggers built on simple conditional rules) that directs project teams to *specific actions* that they then need to take in their impact assessment process (HUDERIA) so that sufficient consideration can be given to each of the risk factors detected and to salient rights and freedoms that could be impacted by the prospective AI system. It also directs them to *specific goals, properties, and areas* that they should focus on in their subsequent risk management and assurance processes (HUDERAC) to reduce and mitigate associated risks. The report additionally assists project teams to understand, in a provisional way, the risk level of their system, so that they can take a proportionate approach to risk management actions and stakeholder engagement.

2. The **Stakeholder Engagement Process (SEP)** helps project teams to identify stakeholder salience and to facilitate proportionate stakeholder involvement and input throughout the project workflow. This process safeguards the equity and the contextual accuracy of the PCRA and other subsequent governance processes through stakeholder involvement, revisitation, and evaluation. The SEP involves a set of steps that include (a) carrying out stakeholder analysis, (b)engaging in positionality reflection, (c) establishing engagement objectives, (d) determining engagement methods, and (e) involving rights-holders in workflow revisitation and reporting.

3. The **Human Rights, Democracy, and the Rule of Law Impact Assessment (HUDERIA)** provides an opportunity for project teams and engaged stakeholders to come together to produce detailed evaluations of the potential and actual impacts that the design, development, and use of an AI system could have on human rights, fundamental freedoms and elements of democracy and the rule of law. This process (a) contextualizes and corroborates potential harms that have been previously identified, (b) enables the discovery of further harms through the dialogical integration of stakeholder perspectives, (c) makes possible the collaborative and context-sensitive assessment of the severity of potential adverse impacts identified, (d) facilitates the co-design of an impact mitigation plan and sets up access to remedy, and (e) establishes monitoring and impact re-assessment protocols.

4. The **Human Rights, Democracy, and Rule of Law Assurance Case (HUDERAC)** enables project teams build a structured argument that

provides demonstrable assurance to stakeholders that claims about the attainment of top-level normative goals (like safety, sustainability, accountability, and fairness) in processes of building and using an AI system are warranted given available evidence. Building an assurance case involves three iterative steps: (a) establishing the goals that are requisite for risk management and impact mitigation and determining the properties needed to assure these goals, (b) taking actions to operationalize these properties in the design, development, and use of the system, and (c) compiling evidence of these actions. The process of developing an assurance case assists internal reflection and deliberation, promoting the adoption of best practices and integrating these into design, development, and deployment lifecycles. It also provides an accessible way to convey all of this information to impacted stakeholders. A diligently executed assurance case offers a clear and understandable mechanism of risk management and impact mitigation that supports appropriate levels of public accountability and transparency.

### *Shape and Delivery of the Framework*

Seen as a single project governance regime—which provides assurance that the use of an AI system, as well as the processes behind its design, development and deployment, respect human rights, fundamental freedoms, democratic principles, and the rule of law—the proposed HUDERAF can be viewed as a collection of interrelated processes, practical steps, and user activities:

## Summary of Processes, Steps, and User Activities

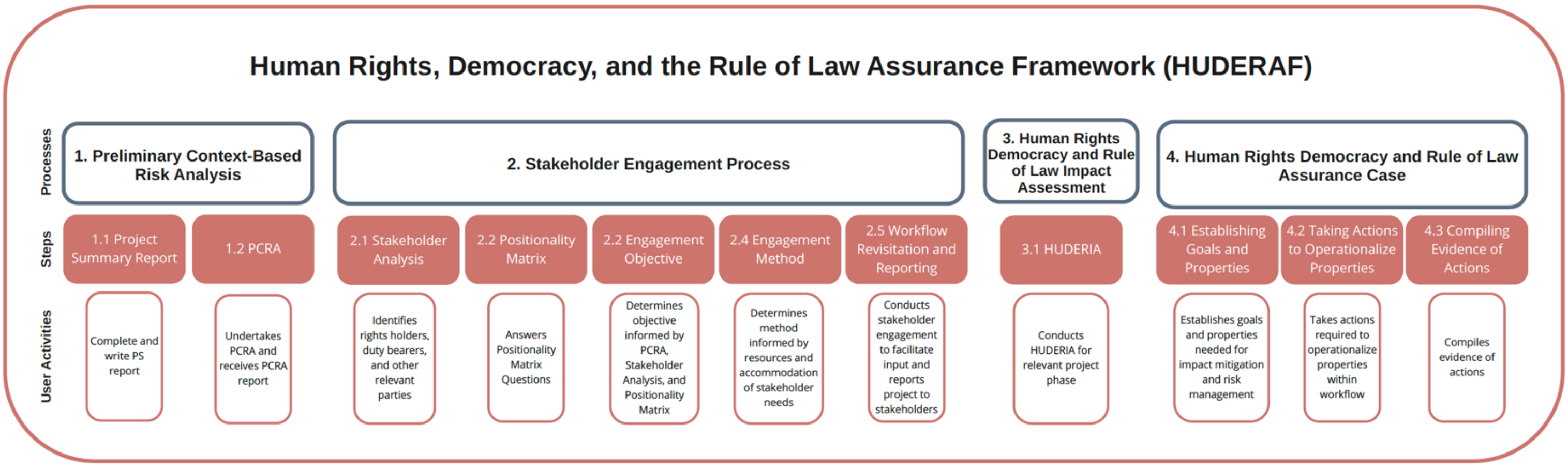

A larger image of this Summary of Processes can be found in Appendix 1.

Beyond such a summary view, this set of processes, steps, and user activities can also be seen as extending across the AI design, development, and deployment lifecycle. The production and implementation of AI technologies is a multi-phased, iterative, and often cyclical undertaking, and so effectively safeguarding the alignment of that complex enterprise with human rights, democracy, and the rule of law necessitates that risk analysis, impact assessment, and assurance methods are sufficiently responsive to the dynamic character of the AI innovation workflow.[1] The HUDERAF has been developed with this in mind. Stakeholder Engagement, HUDERIA, and HUDERAC processes are iterative and organised to respond appropriately to changes in production and implementation factors as well as in the operating environments of AI systems. Here is schematic mapping of what this looks like as such processes and steps move from the design to the development to the deployment phases of a project workflow:

## HUDERAF Project Lifecycle Process Map

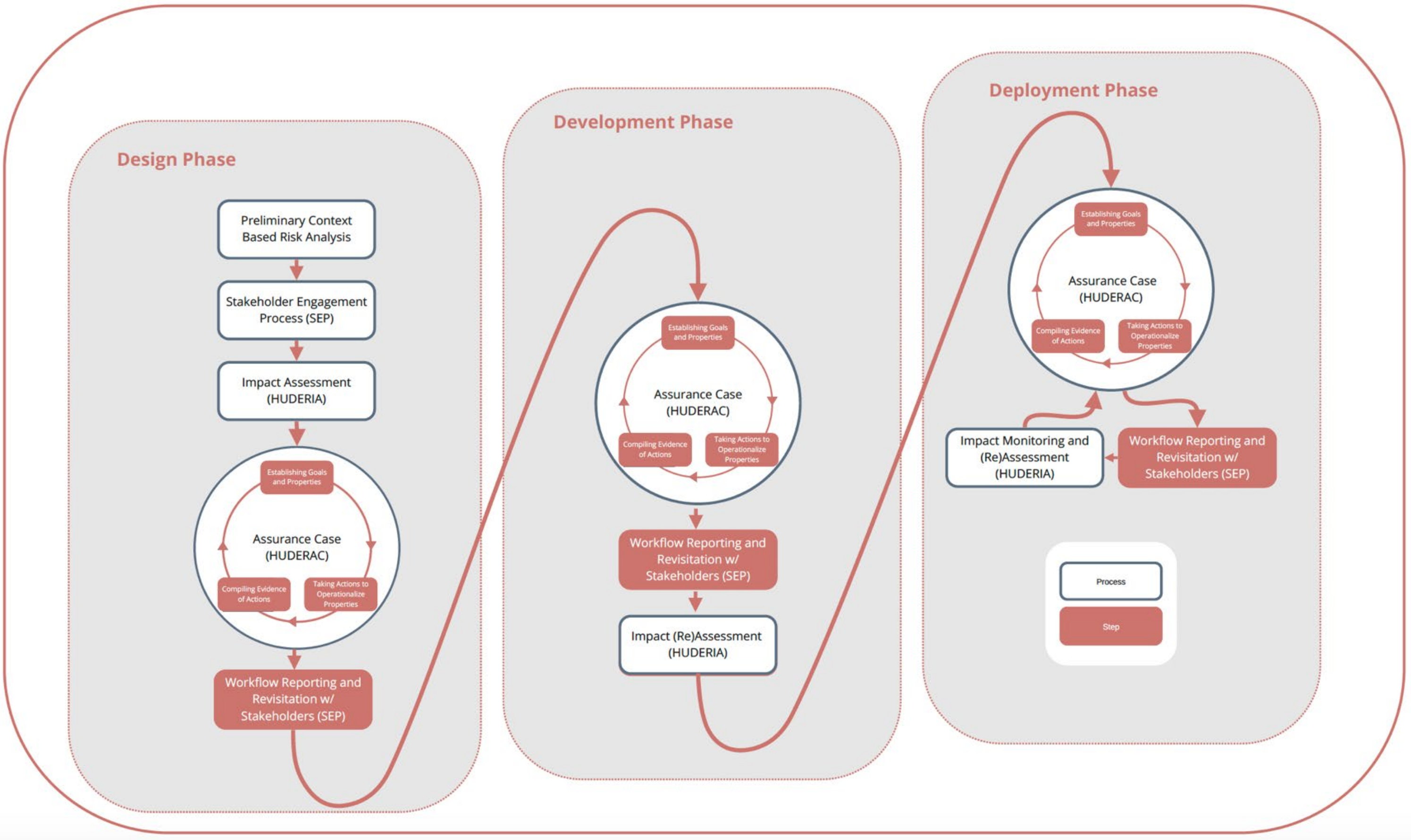

A larger image of this Project Lifecycle Process Map can be found in Appendix 2.

It should be noted that, as we have been constructing the elements of this framework, we have come to believe that the delivery of its processes and instruments would be best served in the medium of a single digital platform that allows for the myriad governance goals articulated by the CAHAI and its Feasibility Study to be all in one place and as user friendly as possible. As they are presented here, the description and explanation of each process (PCRA, SEP, HUDERIA, HUDERAC) is accompanied by a template that could be made into a simple and easy-to-use form. Each template consolidates the questions, prompts, and actions that need to be covered for the completion of the process at each step along the way. The PCRA Template, in particular, has been created as an interactive digital form that generates bespoke summary reports based upon conditional/branching logic and the compiling of results. These templates could be brought together and interconnected in one digital environment, so that users could have a streamlined way to ensure that their AI innovation practices properly align with human rights, fundamental freedoms, democracy, and the rule of law and to readily demonstrate this to potentially impacted stakeholders or their proxies.

[1] For an overview of the AI/ML project lifecycle through a socio-technical lens, see **Appendix 3.**

A final comment on our mode of presentation: This proposal has been written in the style of a user's guide or technical manual, so that all the elements of the framework could be presented with detailed explanations and elaborations of the rationale behind the mechanisms, tools, and concepts chosen. This reflects more of our attempt to convey the details of the framework to the CAHAI and CoE readership than a representation of what a finished product might look like. Pragmatically speaking, presenting the framework in this way will make the incorporation of policy-level decisions and adjustments from the CoE/PDG/LFG/plenary much easier for those elements of the framework that are deemed worthy of including.

### *Working Notes for the PDG*

As we have been working to develop these processes and instruments, we have hewed closely to the findings and policy content presented in the CAHAI's *Feasibility Study*, to the subsequent multi-stakeholder consultation, and to other CAHAI and Council of Europe documents. This notwithstanding, we ask reviewers and readers of this proposal to make sure that areas (where interpretive choices have had to be made) line up properly with the policy goals and developments of the PDG, the LFG, and the CAHAI, more generally. In building these processes and templates, our goal has been to provide the CAHAI with a coherent frame on which to hang its policy decisions and through which it could operationalize its governance goals, rather than to put forward any policy innovations as such.

Similarly, we have done our best to reflect the most recent positions articulated by the CAHAI on open policy issues (such as redlines and triggers of the precautionary principle). We ask that reviewers and readers keep this in mind, so that any necessary adjustments to current categories or policy content can be spotted and made.

Substantively, we ask that reviewers and readers look at Section 2 of the PCRA (Risks of Adverse Impacts on the Human Rights and Fundamental Freedoms of Persons, Democracy, and the Rule of Law) closely. In this section, we tried to build out a semi-quantitative method of risk analysis that generates provisional recommendations regarding proportionate risk management actions and stakeholder engagement. Coming up with a metric to support the determination of risk-based project governance was a thorny endeavour, and we ended up integrating a very straightforward and user-accessible approach to risk analysis (the calculation of a risk index number [RIN]) with (1) the risk management priorities articulated in the UNGP, the UNHROHC, and the CoE's Committee of Ministers and (2) the calibration mechanism of the precautionary principle. We also considered using sums and combinations of the risk factors identified in Section 1 of the PCRA for this metric, and this may still be an attractive option, should it be decided that the method used in Section 2 is not fit-for-purpose. In any case, we ask that you keep an eye on this part of the framework as it breaks ground into unexplored territory.

We would, finally, just like to apologize, in advance, to readers for the length of the present work and thank them for taking the time to look over the materials contained herein. We have had to cover much territory to animate the aspirations outlined by the PDG, and we are grateful to the reader for accompanying us on this journey to advance the CAHAI's historical mission.

# Preliminary Context-Based Risk Analysis (PCRA)

**Purpose**

The purpose of the Preliminary Context-Based Risk Analysis is to provide an initial indication of the context-based risks that an AI system could pose to human rights, fundamental freedoms, and elements of democracy and the rule of law. It also helps project teams establish a proportionate approach to risk management and assurance practices, and to the level of stakeholder engagement needed across the project lifecycle.

Inputs

- Desk Research

Steps

- 1.1 PS Report
- Outlining project-, use-, domain-, and data-contexts
- Identifying stakeholders
- Scoping human rights impacts
- PCRA
- Identifying risk factors and potential harms to human rights, democracy, and the rule of law

Outputs

- PS Report
- PCRA Report outlining preliminary analysis of circumstantial and modifiable risk factors and indicating proportionate risk management and stakeholder engagement practices

## Introduction

The Preliminary Context-Based Risk Analysis (PCRA) provides an initial indication of the context-based risks that an AI system could pose to human rights, fundamental freedoms, and elements of democracy and the rule of law. The main purpose of the PCRA is to help project teams establish a proportionate approach to risk management and assurance practices and to the level of stakeholder engagement that is needed across the project lifecycle. The PCRA process begins with a set of project scoping and planning activities in which members of the project team (a) consolidate information about the project-, use-, domain-, and data-contexts of the prospective system, (b) identify relevant stakeholders, (c) begin to scope the potential impacts of their systems on human rights, democracy, and the rule of law by identifying salient rights and freedoms, and (d) map the governance workflow.

After these activities are recorded in a Project Summary Report, the project team is ready to complete its PCRA. The PCRA is a set of questions and prompts that help project teams to pinpoint risk factors and potential adverse impacts of their systems on human rights, fundamental freedoms, and elements of democracy and the rule of law. When the PCRA is finished, it automatically generates a report (through triggers built on simple conditional rules) that directs project teams to *specific actions* that they then need to take in their impact assessment process (HUDERIA) so that sufficient consideration can be given to each of the risk factors detected and to salient rights and freedoms that could be impacted by the prospective AI system. It also directs them to *specific goals, properties, and areas* that they should focus on in their subsequent risk management and assurance processes (HUDERAC) to reduce and mitigate associated risks. The report additionally assists project teams to understand, in a provisional way, the risk level of their system, so that they can take a proportionate approach to risk management actions and stakeholder engagement.

## Preparing a Project Summary Report in the Planning and Scoping Phase

Deliberate and reflective reporting practices are a crucial component to assuring an AI system's alignment with human rights, democracy, and the rule of law. Project Summary (PS) reporting supports this goal by generating an ongoing synopsis of an AI project's contextual reflexivity, stakeholder awareness, governance resolve, and human rights diligence across its entire lifecycle. The first iteration of writing a PS Report serves, during the planning and scoping phase of a project, as preparation for answering the questions contained within the PCRA. It provides a reference point for your responses to the context-based questions about risk factors and the potential adverse impacts of AI on the human rights and fundamental freedoms of affected individuals and communities.

The first time that the PS Report is completed (at the very beginning of your HUDERAF), it should be informed by available organisational scoping and planning documents and by desk-based research pertaining to the system's use context, stakeholder contexts, and human rights, democracy, and the rule of law. It is later revised and updated at each iteration of the Stakeholder Engagement Process (drawing from stakeholder input). Changes to this document are recorded and registered over time, providing timestamped project summaries that are intended to help build public trust by operationalizing goals of accountability and transparency. The PS Report enables a dimension of traceability, illustrating how a project has been configured through time by the use of a Human Rights, Democracy, and Rule of Law Assurance Framework.

A Project Summary Report is comprised of four components: (1) a project overview, (2) a list of identified stakeholders, (3) a scoping of potential impacts on human rights, democracy, and rule of law, and (4) a project governance workflow. These components are developed through the following activities:

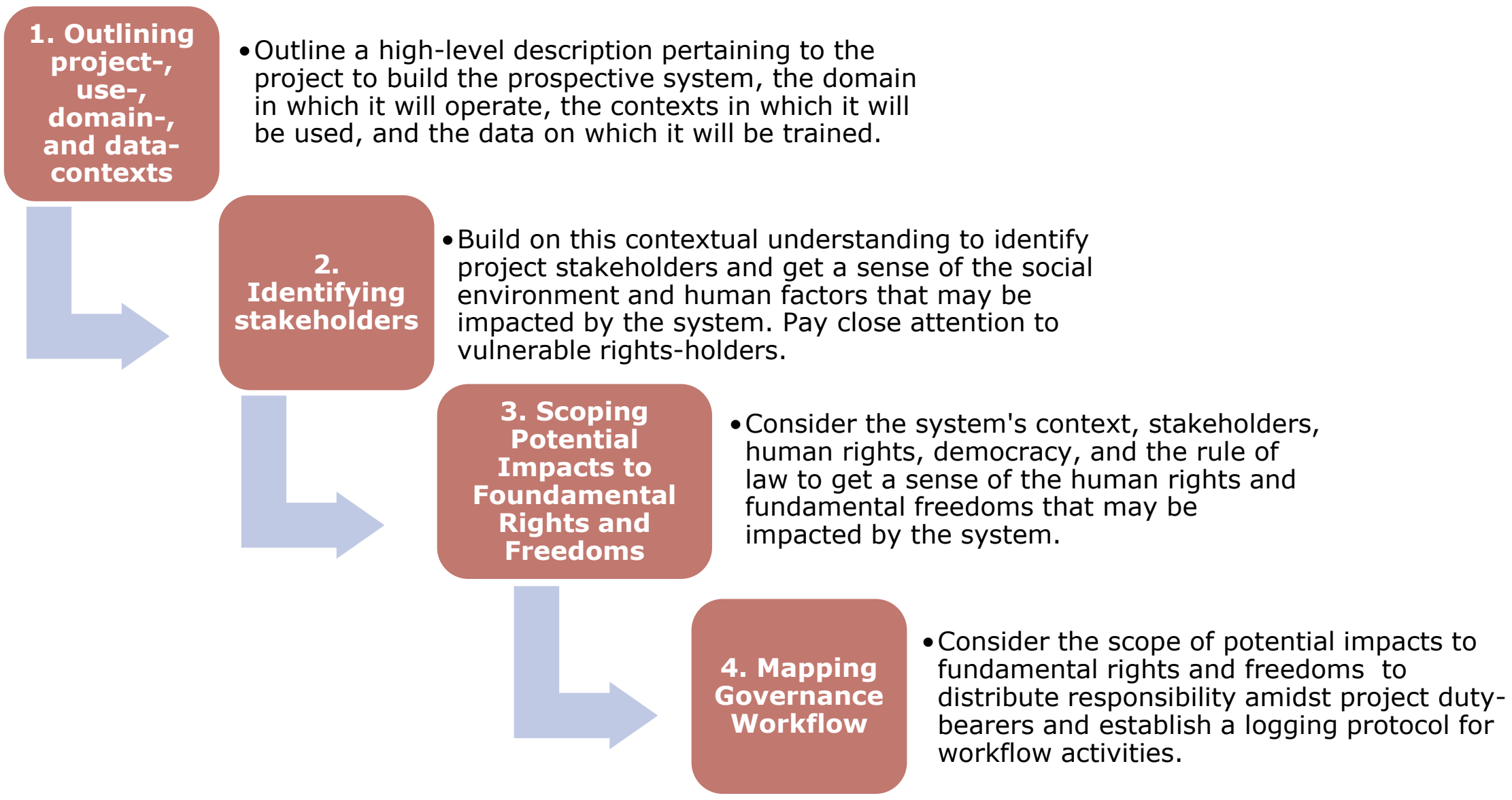

## *Activity 1: Outline Project-, Use-, Domain-, and Data-Contexts*

During the initial iteration of writing the PS Report, draw on organisational documents (i.e., the project business case, proof of concept, or project charter), project team collaboration, and desk research (if necessary) to answer the following questions[2]:

**Questions:**

**Project**

- *How would you describe your organisation and what sort of services or products do you typically provide (beyond the system under consideration)?*

---

[2] These questions draw on research in both the study of sociotechnical dimensions of AI governance and in human rights impact assessment. For exemplary resources in the first group, see: Arnold, M., Bellamy, R. K., Hind, M., Houde, S., Mehta, S., Mojsilović, A., ... & Varshney, K. R. (2019). FactSheets: Increasing trust in AI services through supplier's declarations of conformity. *IBM Journal of Research and Development*, 63(4/5), 6-1; Mitchell, M., Wu, S., Zaldivar, A., Barnes, P., Vasserman, L., Hutchinson, B., ... & Gebru, T. (2019, January). Model cards for model reporting. In *Proceedings of the conference on fairness, accountability, and transparency* (pp. 220-229); Gebru, T., Morgenstern, J., Vecchione, B., Vaughan, J. W., Wallach, H., Daumé III, H., & Crawford, K. (2018). Datasheets for datasets. *arXiv preprint* arXiv:1803.09010. For exemplary resources in the second group, see Danish Institute for Human Rights (2020). *Human Rights Impact Assessment: Guidance and Toolbox*. DIHR; OECD (2018), *OECD Due Diligence Guidance for Responsible Business Conduct;* Mantelero, A., and Esposito, S. (2021). An evidence-based methodology for human rights impact assessment (HRIA) in the development of AI data-intensive systems. Computer Law & Security Review.

- *What AI system is being built and what type of product or service will it offer?*
- *What benefits will the system bring to its users and customers, and will these benefits be widely accessible?*
- *Which organisation(s)—yours, other suppliers, or other providers—are responsible for building this AI system?*
- *Which parts or elements of the AI system, if any, will be procured from third-party vendors, suppliers, sub-contractors, or external developers?*
- *Which algorithms, techniques, and model types will be used in the AI system? (Provide links to technical papers where appropriate)*
- *In a scenario where your project optimally scales, how many people will it impact, for how long, and in what geographic range (local, national, global)? (Describe your rationale)*

**Use Context**

- *What is the purpose of this AI system and in which contexts will it be used? (Briefly describe a use-case that illustrates primary intended use)*
- *Is the AI system's processing output to be used in a fully automated way or will there be some degree of human control, oversight, or input before use? (Describe)*
- *Will the AI system evolve or learn continuously in its use context, or will it be static?*
- *To what degree will the use of the AI system be time-critical, or will users be able to evaluate outputs comfortably over time?*
- *What sort of out-of-scope uses could users attempt to apply the AI system, and what dangers may arise from this?*

**Domain**

- *In what domain will this AI system operate?*
- *Which, if any, domain experts have been or will be consulted in designing and developing the AI system?*

**Data**

- *What datasets are being used to build this AI system?*
- *Will any data being used in the production of the AI system be acquired from a vendor or supplier? (Describe)*
- *Will the data being used in the production of the AI system be collected for that purpose, or will it be re-purposed from existing datasets? (Describe)*
- *What quality assurance and bias mitigation processes do you have in place for the data lifecycle—for both acquired and collected data?*

## *Activity 2: Identify Stakeholders*

Stakeholders include[3]:

---

3 Danish Institute for Human Rights (2020). *Human Rights Impact Assessment: Guidance*

- Rights-Holders: All individuals are human rights-holders. In the impact assessment, the primary focus is on rights-holders who are, or may be, adversely affected by the project.
- Duty-Bearers: Duty-bearers are actors who have human rights duties or responsibilities towards rights-holders. These include the company or organization operating a project or conducting its activities, business suppliers and contractors, joint-venture and other business partners, state actors such as local government authorities and regional and national government departments and agencies.
- Other Relevant Parties: These may include individuals or organisations representing the interests of rights holders and official representations at international, national and local levels (e.g. the UN, national human rights institutions, NGOs or civil society organisations).

During stakeholder identification it is particularly important to consider which rights-holders are most vulnerable to potential impacts of the tool/system being developed and which stakeholders' views are underrepresented in development and deployment processes. During the initial iteration of writing the PS Report, draw on desk research pertaining to stakeholders and their contexts to identify stakeholders and to get a sense of the social environment and human factors that may be impacted by the system. Answer the following questions:

**Questions:**

- *Who are the rights-holders, duty bearers and other relevant parties that may be impacted by, or may impact, the project?*

*To further identify groups of rights-holders, answer the following questions:*

***Protected Characteristics:***

- *Do any of these rights-holders possess sensitive or protected characteristics that could increase their vulnerability to abuse or discrimination, or for reason of which they may require additional protection or assistance with respect to the impacts of the project? If so, what characteristics?*

***Contextual Vulnerability Characteristics:***

- *Could the outcomes of this project present significant concerns to specific groups of rights-holders given vulnerabilities caused or precipitated by their distinct circumstances?*
- *If so, what vulnerability characteristics expose them to being jeopardized by project outcomes?*

***For any characteristics identified in the previous answers, consider the questions:***

- *What affected group or groups of rights-holders could these protected or contextual vulnerability characteristics or qualities represent?*
- *How could the distribution of power relations (i.e., the relative advantages and disadvantages) between rights holders and duty bearers affect the way the benefits and risks of the project are allocated?*

## *Activity 3: Scope Impacts to Human Rights, Democracy, and Rule of Law*

The information contained below in the Table 1 (*Principles and Priorities, Corresponding Rights and Freedoms, and Elaborations*) serves as background material to provide you with a means of accessing and understanding some of the existing human rights and fundamental freedoms that could be impacted by the use of AI technologies. A thorough review of this table and an engagement of the links to the relevant Charters, Conventions, Declarations, and elaborations it contains is a critical first step that will help you identify the salient rights and freedoms that could be affected by your project. You should also explore whether your organisation has engaged in any previous Human Rights Impact Assessments or other impact assessments (data protection impact assessment, equality impact assessment, ethical and social impact assessment, environmental impact assessment, etc.)—and review these where they are present.

As a starting point, the rights and freedoms delineated within this material can be presented in summary form in the following principles and priorities, which are enumerated and elaborated upon in the CAHAI's *Feasibility Study* and in the multi-stakeholder consultation that followed the plenary's adoption of this text in December 2020 [4]:

[4] As has been indicated within and at the bottom of Table 1, this list of rights and freedoms also reflects important adjacent work done by the Council of Europe, UNESCO, the European Commission, OECD, and the European Agency for Fundamental Rights among others. For relevant sources, see: Council of Europe Commissioner for Human Rights (2019) – "Unboxing AI: 10 steps to protect Human Rights"; Council of Europe, "Recommendation CM/Rec (2020) of the Committee of Ministers to member States on the human rights impacts of algorithmic systems"; Muller, C. The Impact of Artificial Intelligence on Human Rights, Democracy and the Rule of Law. Council of Europe, CAHAI(2020)06-fin; Bergamini, D. (*Rapporteur)* (2020). Need for democratic governance of artificial intelligence. Council of Europe, Committee on Political Affairs and Democracy, Parliamentary Assembly; UNESCO (2021a/b) – Recommendations on the Ethics of Artificial Intelligence (two drafts, 1-25, 26-134); High-level Expert Group on Artificial Intelligence, European Commission (2020) – "The Assessment List for Trustworthy Artificial Intelligence (ALTAI) for self-assessment"; European Agency for Fundamental Rights (2020) – "Getting the Future Right: Artificial Intelligence and Fundamental Rights".

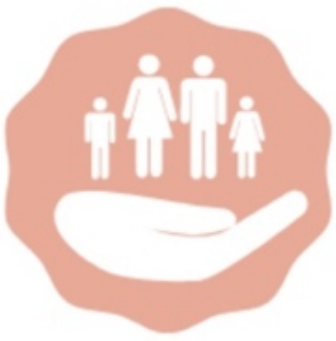

### Human Dignity

All individuals are inherently and inviolably worthy of respect by mere virtue of their status as human beings. Humans should be treated as moral subjects, and not as objects to be algorithmically scored or manipulated.

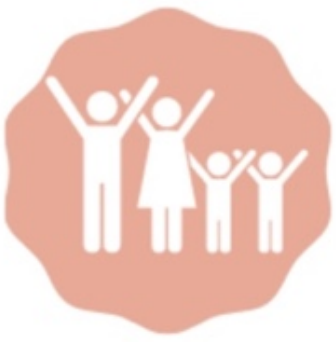

### Human Freedom and Autonomy

Humans should be empowered to determine in an informed and autonomous manner if, when, and how AI systems are to be used. These systems should not be employed to condition or control humans, but should rather enrich their capabilities.

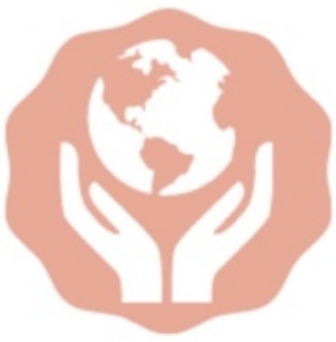

### Prevention of Harm

The physical and mental integrity of humans and the sustainability of the biosphere must be protected, and additional safeguards must be put in place to protect the vulnerable. AI systems must not be permitted to adversely impact human wellbeing or planetary health.

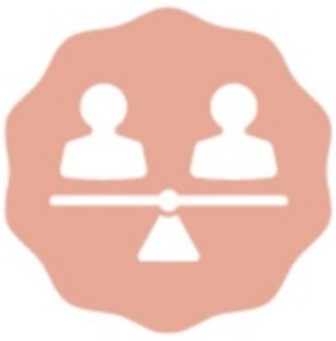

### Fairness, Non-Discrimination, Equality, Diversity, and Inclusiveness

All humans possess the right to non-discrimination and the right to equality and equal treatment under the law. AI systems must be designed to be fair, equitable, and inclusive in their beneficial impacts and in the distribution of their risks.

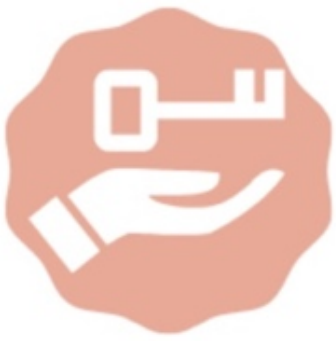

### Data Protection and the Right to Privacy

The design and use of AI systems that rely on the processing of personal data must secure a person's right to respect for private and family life, including the individual's right to control their own data. Informed, freely given, and unambiguous consent must play a role in this.

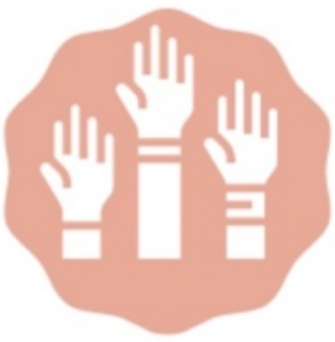

### Democracy

Individuals should enjoy the ability to freely form bonds of social cohesion, human connection, and solidarity through inclusive and regular democratic participation, whether in political life, work life, or social life. This requires informational plurality, the free and equitable flow of the legitimate and valid forms of information, and the protection of freedoms of expression, assembly, and association.

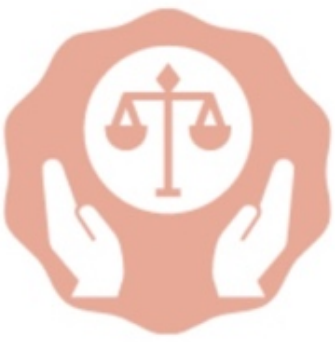

### Rule of Law

AI systems must not undermine judicial independence, effective remedy, the right to a fair trial, due process, or impartiality. To ensure this, the transparency, integrity, and fairness of the data, and data processing methods must be secured.

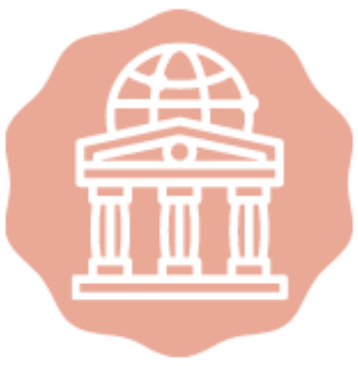

### Social and Economic Rights

Individuals must have access to the material means needed to participate fully in work life, social life, and creative life, and in the conduct of public affairs, through the provision of proper education, adequate living and working standards, health, safety, and social security. This means that AI systems should not infringe upon individuals' rights to work, to just, safe, and healthy working conditions, to social security, to the protection of health, and to social and medical assistance.

**Table 1. Principles and Priorities, Corresponding Rights and Freedoms, and Elaborations**

(Numbers in the first column indicate overlapping principles and priorities from other relevant sources listed at the bottom)

| Principles and Priorities | Corresponding Rights and Freedoms with Selected Elaborations | Resources for Principles and Corresponding Rights and Freedoms |
| --- | --- | --- |
| **Respect for and protection of human dignity (2, 3, 4, 5, 8)** | **-The right to human dignity, the right to life (Art. 2 ECHR), and the right to physical and mental integrity.**<br><br>**-The right to be informed of the fact that one is interacting with an AI system rather than with a human being.**<br><br>**-The right to refuse interaction with an AI system whenever this could adversely impact human dignity.**<br><br>~<br><br>"Human dignity is the foundation of all human rights. It recognises that all individuals are inherently worthy of respect by mere virtue of their status as human beings. Human dignity, as an absolute right, is inviolable. Hence, even when a human right is restricted – for instance when a balance of rights and interests must be made – human dignity must always be safeguarded. In the context of AI, this means that the design, development and use of AI systems must respect the dignity of the human beings interacting therewith or impacted thereby. Humans should be treated as moral subjects, and not as mere objects that are categorised, scored, predicted or manipulated.<br><br>AI applications can be used to foster human dignity and empower individuals, yet their use can also challenge it and (un)intentionally run counter to it. To safeguard | **European Convention on Human Rights (ECHR):**<br><br>-Article 2, European Convention on Human Rights – *Right to life*<br><br>-Article 2, 'Guide on Article 2 of the European Convention on Human Rights', Council of Europe – *Right to life*<br><br>**Charter of the Fundamental Rights of the EU:**<br><br>-Article 1, Charter of the Fundamental Rights of the EU – *Human dignity*<br><br>*-Article 1, Explanation on Article 1 — Human Dignity (Explanations Relating to the Charter of Fundamental Rights)* in the Official Journal of the European Union C 303/17 - 14.12.2007<br><br>-Article 2, Charter of the Fundamental Rights of the EU – *Right to life*<br><br>*-Article 2, Explanation on Article 2 — Right to life (Explanations Relating to the Charter of Fundamental Rights)* in the Official Journal of the European Union C 303/17 - 14.12.2007<br><br>**Universal Declaration of Human Rights:**<br><br>-Preamble, Universal Declaration of Human Rights – *Dignity* |
| | human dignity, it is essential that human beings are aware of the fact that they are interacting with an AI system and are not misled in this regard. Moreover, they should in principle be able to choose not to interact with it, and to not be subject to a decision informed or made by an AI system whenever this can significantly impact their lives, especially when this can violate rights related to their human dignity. Furthermore, the allocation of certain tasks may need to be reserved for humans rather than machines given their potential impact on human dignity."[5]<br><br>~<br><br>"As stated in the EU Charter's official guiding explanations (Official Journal of the European Union, 2007: C 303/17): 'dignity of the human person is not only a fundamental right in itself but constitutes the real basis of fundamental rights.' In this sense, dignity is a highly multifaceted concept… One way to grasp the broad range of its meanings is to explore what humans consider to be violations of their dignity, as done by Halbertal (2015), who points out the following three categories of violations: 1. **Humiliation**: being put in a state of helplessness, insignificance; losing autonomy over your own representation. 2. **Instrumentalization**: treating an individual as exchangeable and merely a means to an end. 3. **Rejection of one's gift**: making an individual superfluous, unacknowledging one's contribution, aspiration, and potential.[6] | **International Covenant on Civil and Political Rights:**<br><br>-Article 6, International Covenant on Civil and Political Rights – *Right to life*<br><br>**Council of Europe Resources:**<br><br>-Council of Europe, Convention for the Protection of Human Rights and Dignity of the Human Being with regard to the Application of Biology and Medicine: Convention on Human Rights and Biomedicine |

[5] CAHAI *Feasibility Study*, Council of Europe CAHAI (2020)23.

[6] Aizenberg, E., & van den Hoven, J. (2020). Designing for human rights in AI. *Big Data & Society*, *7*(2); Halbertal, M. (2015). Three concepts of human dignity. *2015 Dewey Lecture, University of Chicago*; Kaufmann, P., Kuch, H., Neuhaeuser, C., & Webster, E. (Eds.). (2010). *Humiliation, degradation, dehumanization: human dignity violated* (Vol. 24). Springer Science & Business Media.

| | | |
|---|---|---|
| | "The inviolable and inherent dignity of every human constitutes the foundation for the universal, indivisible, inalienable, interdependent and interrelated system of human rights and fundamental freedoms. Therefore, respect, protection and promotion of human dignity and rights as established by international law, including international human rights law, is essential throughout the life cycle of AI systems. Human dignity relates to the recognition of the intrinsic and equal worth of each individual human being, regardless of race, colour, descent, gender, age, language, religion, political opinion, national origin, ethnic origin, social origin, economic or social condition of birth, or disability, and any other grounds....Persons may interact with AI systems throughout their life cycle and receive assistance from them such as care for vulnerable people or people in vulnerable situations, including but not limited to children, older persons, persons with disabilities or the ill. Within such interactions, persons should never be objectified, nor should their dignity be otherwise undermined, or human rights and fundamental freedoms violated or abused."[7] | |
| **Protection of human freedom and autonomy (1, 2, 4, 6, 7, 8)** | **-The right to liberty and security (Art. 5 ECHR).**<br><br>**-The right to human autonomy and self-determination.**<br><br>**-The right not to be subject to a decision based solely on automated processing when this produces legal effects on groups or similarly significantly affects individuals.** | **European Convention on Human Rights (ECHR):**<br><br>-Article 5, European Convention on Human Rights – Right to liberty and security<br><br>-Article 5, 'Guide on Article 5 of the European Convention on Human Rights', Council of Europe – *Right to liberty and security* |

[7] UNESCO (2021a/b) – Recommendations on the Ethics of Artificial Intelligence (two drafts, 1-25, 26-134).

| | | |
|---|---|---|
| | **-The right to effectively contest and challenge decisions informed and/or made by an AI system and to demand that such decisions be reviewed by a person.**<br><br>**-The right to freely decide to be excluded from AI-enabled manipulation, individualized profiling, and predictions. This also applies to cases of non-personal data processing.**<br><br>**-The right to have the opportunity, when it is not overridden by competing legitimate grounds, to choose to have contact with a human being rather than a robot.**<br><br>~<br><br>"Individual self-determination and personal autonomy necessarily entail the ability to freely take decisions and have them respected by others. According to international and European human rights jurisprudence, individual personal autonomy also covers a further range of human behaviours, amongst which are the right to develop one's own personality and the right to establish and develop relationships with other people, the right to pursue one's own aspirations and to control one's own information."[8]<br><br>"Promoting autonomy means enabling individuals to make decisions about their lives on their own, which are not imposed upon them or manipulated by others. It requires having a range of genuine options available from which one can choose. A decision taken on the basis of | -Article 9, European Convention on Human Rights – *Freedom of thought, conscience, and religion*<br><br>-Article 9, 'Guide on Article 9 of the European Convention on Human Rights', Council of Europe – *Freedom of thought, conscience, and religion*<br><br>-Article 10, European Convention on Human Rights – *Freedom of expression*<br><br>*-Article 10,* 'Guide on Article 10 of the European Convention on Human Rights', Council of Europe – *Freedom of expression*<br><br>**Charter of the Fundamental Rights of the EU:**<br><br>-Article 6, Charter of the Fundamental Rights of the EU – *Right to liberty and security*<br><br>*-Article 6, Explanation on Article 6 — Right to liberty and security (Explanations Relating to the Charter of Fundamental Rights)* in the Official Journal of the European Union C 303/17 - 14.12.2007<br><br>-Article 10, *Charter of the Fundamental Rights of the EU* – *Freedom of thought, conscience, and religion*<br><br>*-Article 10, Explanation on Article 10 — Freedom of thought, conscience, and religion (Explanations Relating to the Charter of Fundamental Rights)* in the Official Journal of the European Union C 303/17 - 14.12.2007 |

[8] Mantelero, A., and Esposito, S. (2021). An evidence-based methodology for human rights impact assessment (HRIA) in the development of AI data-intensive systems. Computer Law & Security Review.

| | | |
|---|---|---|
| | inadequate information or deception is not considered autonomous."[9] | -Article 11, Charter of the Fundamental Rights of the EU – *Freedom expression and information*<br><br>*-Article 11, Explanation on Article 11 — Freedom of expression and information (Explanations Relating to the Charter of Fundamental Rights)* in the Official Journal of the European Union C 303/17 - 14.12.2007<br><br>**Universal Declaration of Human Rights:**<br><br>-Article 3, Universal Declaration of Human Rights – *Right to life, liberty, and the security of person*<br><br>-Article 18, Universal Declaration of Human Rights – *Right to freedom of thought, conscience, and religion*<br><br>-Article 19, Universal Declaration of Human Rights – *Right to freedom of opinion and expression*<br><br>**International Covenant on Civil and Political Rights:**<br><br>-Article 9, International Covenant on Civil and Political Rights – *Right to liberty and security of person*<br><br>-Article 18, International Covenant on Civil and Political Rights – *Right to freedom of thought, conscience, and religion*<br><br>-Article 19, International Covenant on Civil and Political Rights – *Freedom of expression*<br><br>**Council of Europe Resources:** |

[9] Loi, et al. (2021) *Algorithm Watch* – "Automated Decision-Making Systems in the Public Sector: An Impact Assessment Tool for Public Authorities"

| | | |
|---|---|---|
| | | -Commissioner for Human Rights, Human Rights and Disability (20 October, 2008) 'Equal rights for all', para. 5.2. – Council of Europe<br><br>-Bychawska-Siniarska, 'Protecting the right to freedom of expression under the European Convention on Human Rights' – Council of Europe, Strasbourg 2017 |
| **Prevention of harm and protection of the right to life and physical, psychological, and moral integrity (2, 5, 6, 8)** | **-The right to life (Art. 2 ECHR) and the right to physical and mental integrity.**<br><br>**-The right to the protection of the environment.**<br><br>**-The right to sustainability of the community and biosphere.**<br><br>~<br><br>"Throughout the life cycle of AI systems, human beings should not be harmed but rather their quality of life of human beings should be maintained or enhanced, while the definition of "quality of life" should be left open to individuals or groups, while respecting relevant laws, as long as there is no violation or abuse of human rights and fundamental freedoms, or the dignity of humans in terms of this definition....Environmental and ecosystem flourishing should be recognized, protected and promoted through the life cycle of AI systems. Furthermore, environment and ecosystems are the existential necessity for humanity and other living beings to be able to enjoy the benefits of advances in AI....All actors involved in the life cycle of AI systems must comply with applicable international law and domestic legislation, standards and practices, such as precaution, designed for environmental and ecosystem protection and restoration, and sustainable development. They should reduce the | **European Convention on Human Rights (ECHR):**<br><br>-Article 2, European Convention on Human Rights – *Right to life*<br><br>-Article 2, 'Guide on Article 2 of the European Convention on Human Rights', Council of Europe – *Right to life*<br><br>**Charter of the Fundamental Rights of the EU:**<br><br>-Article 2, Charter of the Fundamental Rights of the EU – *Right to life*<br><br>*-Article 2, Explanation on Article 2 — Right to life (Explanations Relating to the Charter of Fundamental Rights)* in the Official Journal of the European Union C 303/17 - 14.12.2007<br><br>-Article 3, Charter of the Fundamental Rights of the EU – *Right to integrity of the person*<br><br>*-Article 3, Explanation on Article 3 — Right to the integrity of the person (Explanations Relating to the Charter of Fundamental Rights)* in the Official Journal of the European Union C 303/17 - 14.12.2007 |
| | environmental impact of AI systems, including but not limited to, its carbon footprint, to ensure the minimization of climate change and environmental risk factors, and prevent the unsustainable exploitation, use and transformation of natural resources contributing to the deterioration of the environment and the degradation of ecosystems."[10] | |
| **Non-discrimination, fairness, and equality (1, 2, 3, 4, 5, 6, 7, 8)** | **-The right to non-discrimination (on the basis of the protected grounds set out in Article 14 of the ECHR and Protocol 12 to the ECHR), including intersectional discrimination.**<br><br>**-The right to non-discrimination and the right to equal treatment. This right must be ensured in relation to the entire lifecycle of an AI system (design, development, implementation, and use), as well as to the human choices concerning AI design, adoption, and use, whether used in the public or private sector.**<br><br>~<br><br>"In all circumstances, discrimination risks must be prevented and mitigated with special attention for groups that have an increased risk of their rights being disproportionately impacted by AI. This includes women, children, older people, economically disadvantaged persons, members of the LGBTI community, persons with disabilities, and "racial", ethnic or religious groups. Member states must refrain from using AI systems that discriminate or lead to discriminatory outcomes and, | **European Convention on Human Rights (ECHR):**<br><br>-Protocol No. 12, European Convention on Human Rights<br><br>-Article 14, European Convention on Human Rights – *Prohibition of discrimination*<br><br>-Article 14 and Article 12 of Protocol No. 12, 'Guide on Article 14 of the European Convention on Human Right and on Article 1 of Protocol No. 12 to the Convention', Council of Europe – *Prohibition of discrimination*<br><br>**Charter of the Fundamental Rights of the EU:**<br><br>-Articles 20 and 21, Charter of the Fundamental Rights of the EU – *Equality before the law and non-discrimination*<br><br>*-Article 20, Explanation on Article 20 — Equality before the law (Explanations Relating to the Charter of Fundamental Rights)* in the Official Journal of the European Union C 303/17 - 14.12.2007<br><br>*-Article 21, Explanation on Article 21 — Non-discrimination (Explanations Relating to the Charter of* |

[10] UNESCO (2021a/b) – Recommendations on the Ethics of Artificial Intelligence (two drafts, 1-25, 26-134).

| | | |
|---|---|---|
| | within their jurisdiction, protect individuals from the consequences of use of such AI systems by third parties.<br><br>The active participation of and meaningful consultation with a diverse community that includes effective representation from these groups in all stages of the AI lifecycle is an important component of prevention and mitigation of adverse human rights impacts. In addition, special attention needs to be paid to transparency and accessibility of information on the training data used for the development of an AI system."[11]<br><br>~<br><br>"AI actors should promote social justice, by safeguarding fairness and non-discrimination of any kind in compliance with international law. Fairness implies sharing benefits of AI technologies at local, national and international levels, while taking into consideration the specific needs of different age groups, cultural systems, different language groups, persons with disabilities, girls and women, and disadvantaged, marginalized and vulnerable populations. At the local level, it is a matter of working to give communities access to AI systems in the languages of their choice and respecting different cultures. At the national level, governments are obliged to demonstrate equity between rural and urban areas, and among all persons without distinction as to sex, gender, language, religion, political or other opinion, national, ethnic, indigenous or social origin, sexual orientation and gender identity, property, birth, disability, age or other status, in terms of access to and participation in the AI system life cycle. At the international level, the most technologically advanced countries have a responsibility of solidarity with | *Fundamental Rights)* in the Official Journal of the European Union C 303/17 - 14.12.2007<br><br>**Universal Declaration of Human Rights:**<br><br>-Article 7, Universal Declaration of Human Rights – *Equality before the law*<br><br>**International Covenant on Civil and Political Rights:**<br><br>-Article 6, International Covenant on Civil and Political Rights – *Right to life*<br><br>-Article 26, International Covenant on Civil and Political Rights – *Non-discrimination*<br><br>**Council of Europe Resources:**<br><br>-European Commission against Racism and Intolerance (ECRI) - Discrimination, artificial intelligence, and algorithmic decision-making (2018)<br>• AI applications have found ways to "escape current laws." The majority of non-discrimination statutes relate only to specific protected characteristics. There are other forms of discrimination that are not correlated with protected characteristics but can still reinforce social inequality.<br>• The idea of sector-specific rules for the protection of fairness and human rights in the area of AI is proposed, as different sectors necessitate different values and problems. |

[11] Council of Europe Commissioner for Human Rights (2019) – "Unboxing AI: 10 steps to protect Human Rights"

| | | |
|---|---|---|
| | the least advanced to ensure that the benefits of AI technologies are shared such that access to and participation in the AI system life cycle for the latter contributes to a fairer world order with regard to information, communication, culture, education, research, and socio-economic and political stability...AI actors should promote social justice and safeguard fairness and non-discrimination of any kind in compliance with international law. This implies an inclusive approach to ensuring the benefits of AI technologies are available and accessible to all taking into consideration the specific needs of different age groups, cultural systems, different language groups, persons with disabilities, girls and women, and disadvantaged, marginalized and vulnerable people or people in vulnerable situations."[12]<br><br>~<br><br>"AI is widely expected to improve the productivity of economies. However, these productivity gains are expected to be distributed unequally with most benefits accruing to the well-off. Similarly, data and design choices, combined with a lack of transparency of black box algorithms have shown to lead to a perpetuating unjust bias against already disadvantaged groups in society, such as women and ethnic minorities.28 AI could lead to inequality and segregation and thus threaten the necessary level of economic and social **equality** required for a thriving democracy."[13] | -Unboxing artificial intelligence: 10 steps to protect human rights (2019)<br>• Recommendations are to be used to mitigate or prevent negative impacts of AI systems on human rights.<br>• Practical recommendations are given with 10 areas of action: human rights impact assessments; public consultations; human rights standards in the private sector; information and transparency; independent monitoring; non-discrimination and equality; data protection and privacy; freedom of expression, freedom of assembly and association, and the right to work; avenues for redress; and promoting knowledge and understanding of AI.<br><br>**Office of the United Nations High Commissioner for Human Rights:**<br><br>-OHCHR, International Convention on the Elimination of All Forms of Racial Discrimination<br><br>-OHCHR, Convention on the Elimination of All Forms of Discrimination against Women |
| **Data protection and the right to respect of private** | **-The right to respect for private and family life and the protection of personal data (Art. 8 ECHR).** | **European Convention on Human Rights (ECHR):**<br><br>-Article 8, European Convention on Human Rights – *Right to respect for private and family life* |

[12] UNESCO (2021a/b) – Recommendations on the Ethics of Artificial Intelligence (two drafts, 1-25, 26-134).
[13] Muller, C. The Impact of Artificial Intelligence on Human Rights, Democracy and the Rule of Law. Council of Europe, CAHAI(2020)06-fin.

| | | |
|---|---|---|
| **and family life (1, 2, 3, 5, 7, 8)** | **-The right to physical, psychological, and moral integrity in light of AI-based profiling and emotion/personality recognition.**<br><br>**-All the rights enshrined in Convention 108+ and in its modernised version, and in particular with regard to AI-based profiling and location tracking.**<br><br>~<br><br>"The development, training, testing and use of AI systems that rely on the processing of personal data must fully secure a person's right to respect for private and family life under Article 8 of the European Convention on Human Rights, including the "right to a form of informational self-determination" in relation to their data.<br><br>Data processing in the context of AI systems shall be proportionate in relation to the legitimate purpose pursued through such processing, and should at all stages of the processing reflect a fair balance between the interests pursued through the development and deployment of the AI system and the rights and freedoms at stake.<br><br>Member states should effectively implement the modernised Council of Europe Convention for the Protection of Individuals with regard to Automatic Processing of Personal Data ("Convention 108+") as well as any other international instrument on data protection and privacy that is binding on the member state. The processing of personal data at any stage of an AI system lifecycle must be based on the principles set out under the Convention 108+, in particular (i) there must be a legitimate basis laid down by law for the processing of the personal data at the relevant stages of the AI system lifecycle; (ii) the personal data must be processed lawfully, fairly and in a transparent manner; (iii) the | -Article 8, 'Guide on Article 8 of the European Convention on Human Rights. Right to respect for private and family life, home and correspondence', Council of Europe – *Right to respect for private and family life*<br><br>**Charter of the Fundamental Rights of the EU:**<br><br>-Article 7, Charter of the Fundamental Rights of the EU – *Right for private and family life*<br><br>*-Article 7, Explanation on Article 7 — Right for private and family life (Explanations Relating to the Charter of Fundamental Rights)* in the Official Journal of the European Union C 303/17 - 14.12.2007<br><br>-Article 8, Charter of the Fundamental Rights of the EU – *Protection of personal data*<br><br>*-Article 8, Explanation on Article 8 — Protection of personal data (Explanations Relating to the Charter of Fundamental Rights)* in the Official Journal of the European Union C 303/17 - 14.12.2007<br><br>**Universal Declaration of Human Rights:**<br><br>-Article 12, Universal Declaration of Human Rights – *Right to respect for privacy, family, home, or correspondence*<br><br>**Council of Europe Resources:**<br><br>-Convention 108/108+ (1981/2018)<br>• Processing of sensitive data can only be allowed where appropriate guidelines are present<br>• Every individual has the right to know the purpose of processing their data. Along with this, they have a right to rectification and obtainment of |
| | personal data must be collected for explicit, specified and legitimate purposes and not processed in a way incompatible with those purposes; (iv) the personal data must be adequate, relevant and not excessive in relation to the purposes for which they are processed; (v) the personal data must be accurate and, where necessary, kept up to date; (vi) the personal data should be preserved in a form which permits identification of data subjects for no longer than is necessary for the purposes for which those data are processed."[14]<br><br>~<br><br>"The right to respect for private life and the protection of personal data (Articles 7 and 8 of the EU Charter) are at the core of fundamental rights discussions around the use of AI. While closely related, the rights to respect for private life and the protection of personal data are distinct, self-standing rights. They have been described as the 'classic' right to the protection of privacy and a more "modern" right, the right to data protection. Both strive to protect similar values, i.e. the autonomy and human dignity of individuals, by granting them a personal sphere in which they can freely develop their personalities, think and shape their opinions. They thus form an essential prerequisite for the exercise of other fundamental rights, such as the freedom of thought, conscience and religion (Article 10 of the EU Charter), freedom of expression and information (Article 11 of the EU Charter), and freedom of assembly and of association (Article 12 of the EU Charter). The concept of 'private life' or 'privacy' is complex and broad, and not susceptible to an exhaustive definition. It covers the physical and psychological integrity of a person, and can, therefore, embrace | knowledge where data are processed contrary to the Convention's provisions<br>• Transparency, proportionality, consent, data quality, data security, purpose limitation, accountability, data minimisation, data protection and privacy by design, fairness, and lawfulness are principles introduced<br>• Individuals should not be subjected to decisions made solely by automated processing of data without consideration of personal views<br>• "Legal framework built around Convention remains fully applicable to AI technology, as soon as the processed data fall within the scope of the Convention."<br>• Respect for the rights of data subjects and conformity to the additional obligations of data controllers and processors as set out in Articles 9 and 10 of Convention 108+<br><br>-Convention on Cybercrime ("Budapest Convention")(2001)<br>• "Criminalising offences against and by the means of computers, for procedural powers to investigate cybercrime and secure electronic evidence."<br>• Crimes include but are not limited to infringements of copyright, computer-related fraud, child pornography, and violations of a security network<br>• Investigation includes a series of powers and procedures including interception and the search of computer networks<br>• Primary objective is to "pursue a common criminal policy aimed at the protection of society against cybercrime, especially through appropriate legislation and international co-operation." |

[14] Council of Europe Commissioner for Human Rights (2019) – "Unboxing AI: 10 steps to protect Human Rights"

| | | |
|---|---|---|
| | multiple aspects of the person's physical and social identity. There is also a zone of interaction of a person with others, even in a public context, which may fall within the scope of 'privacy".'[15]<br><br>~<br><br>"Many AI-systems and uses have a broad and deep impact on the right to privacy. Privacy discussions around AI currently tend to focus primarily on data privacy and the indiscriminate processing of personal (and non-personal) data. It should however be noted that, while data privacy is indeed an important element, the impact of AI on our privacy goes well beyond our data. Art. 8 of the ECHR encompasses the protection of a wide range of elements of our private lives, that can be grouped into three broad categories namely: (i) a person's (general) privacy, (ii) a person's physical, psychological or moral integrity and (iii) a person's identity and autonomy. Different applications and uses of AI can have an impact on these categories, and have received little attention to date.<br><br>AI-driven (mass) surveillance, for example with facial recognition, involves the capture, storage and processing of personal (biometric) data (our faces), but it also affects our 'general' privacy, identity and autonomy in such a way that it creates a situation where we are (constantly) being watched, followed and identified. As a psychological 'chilling' effect, people might feel inclined to adapt their behaviour to a certain norm, which shifts the balance of power between the state or private organisation using facial recognition and the individual. In legal doctrine and precedent the chilling effect of surveillance can constitute a violation of the private space, which is necessary for | -Directorate General of Human Rights and Rule of Law, Guidelines on Artificial Intelligence and Data Protection, Council of Europe<br><br>-Roagna, 'Protecting the right to respect for private and family life under the European Convention on Human Rights' – Council of Europe, Strasbourg 2012 |

[15] European Agency for Fundamental Rights (2020) – "Getting the Future Right: Artificial Intelligence and Fundamental Rights".

| | | |
|---|---|---|
| | personal development and democratic deliberation. Even if our faces are immediately deleted after capturing, the technology still intrudes our psychological integrity."[16] | |
| **Economic and social rights (1, 7, 8)** | **-The right to just working conditions, the right to safe and healthy working conditions, the right to organize, the right to social security, and the rights to the protection of health and to social and medical assistance (Art. 2, 3, 5, 11, 13 ESC).**<br><br>~<br><br>"AI systems can have major benefits when used for hazardous, heavy, exhausting, unpleasant, repetitive or boring work. However, the wide adoption of AI systems in all domains of our lives also creates new risks to social and economic rights. AI systems are increasingly used to monitor and track workers, distribute work without human intervention and assess and predict worker potential and performance in hiring and firing situations. In some situations, this can also have detrimental consequences for workers' right to decent pay, as their pay can be determined by algorithms in a way that is irregular, inconsistent and insufficient. Furthermore, AI systems can also be used to detect and counter the unionisation of workers. These applications can jeopardise the right to just, safe and healthy working conditions, dignity at work as well as the right to organise. The discrimination capacity of AI systems that assess and predict the performance of job applications or workers can also undermine equality, including gender equality, in matters of employment and occupation.<br><br>In addition, AI systems can, for instance, be used in the context of social security decisions, in which case the | **European Social Charter:**<br><br>-Article 2, European Social Charter – *Dignity of work and the right to just working conditions*<br><br>-Article 3, European Social Charter – *Right to safe and healthy working conditions*<br><br>-Article 5, European Social Charter – *Right to organize*<br><br>-Article 11, European Social Charter – *Rights to the protection of health*<br><br>-Article 12, European Social Charter – *Right to social security*<br><br>-Article 13, European Social Charter – *Right to social and medical assistance*<br><br>**International Covenant on Economic, Social and Cultural Rights:**<br><br>-Article 6, International Covenant on Economic, Social, and Cultural Rights – *The right to work*<br><br>-Article 7, International Covenant on Economic, Social, and Cultural Rights – *Right to just and favourable conditions of work*<br><br>-Article 8, International Covenant on Economic, Social, and Cultural Rights – *Right to organise* |

[16] Muller, C. The Impact of Artificial Intelligence on Human Rights, Democracy and the Rule of Law. Council of Europe, CAHAI(2020)06-fin.

| | | |
|---|---|---|
| | right guaranteed by Article 12 of the European Social Charter – stating that all workers and their dependents have the right to social security – can be impacted. Indeed, AI systems are increasingly relied on in social welfare administration, and the decisions taken in that context can significantly impact individuals' lives. Similar issues arise where AI systems are deployed in the context of education or housing allocation administrations.<br><br>Moreover, whenever AI systems are used to automate decisions regarding the provision of healthcare and medical assistances, such use can also impact the rights enshrined in Articles 11 and 13 of the Charter, which respectively state that everyone has the right to benefit from measures that enable the enjoyment of the highest possible standard of health attainable, and that anyone without adequate resources has the right to social and medical assistance. AI systems can, for instance, be utilised to determine patients' access to health care services by analysing patients' personal data, such as their health care records, lifestyle data and other information. It is important that this occurs in line with not only the right to privacy and personal data protection, but also with all the social rights laid down in the aforementioned Charter, the impact on which has so far received less attention than the impact on civil and political rights."[17] | -Article 9, International Covenant on Economic, Social, and Cultural Rights – *Right to social security*<br><br>**Charter of the Fundamental Rights of the EU:**<br><br>-Article 15, Charter of the Fundamental Rights of the EU – *Freedom to choose an occupation and right to engage in work*<br><br>*-Article 15, Explanation on Article 15 — Freedom to choose an occupation and right to engage in work (Explanations Relating to the Charter of Fundamental Rights)* in the Official Journal of the European Union C 303/17 - 14.12.2007<br><br>-Article 16, Charter of the Fundamental Rights of the EU – *Freedom to conduct a business*<br><br>*-Article 16, Explanation on Article 16 — Freedom to conduct a business (Explanations Relating to the Charter of Fundamental Rights)* in the Official Journal of the European Union C 303/17 - 14.12.2007<br><br>-Article 17, Charter of the Fundamental Rights of the EU – *Right to property*<br><br>*-Article 17, Explanation on Article 17 — Right to property (Explanations Relating to the Charter of Fundamental Rights)* in the Official Journal of the European Union C 303/17 - 14.12.2007 |

[17] CAHAI *Feasibility Study*, Council of Europe CAHAI (2020)23.

| | | |
|---|---|---|
| | | -Article 27, Charter of the Fundamental Rights of the EU – *Workers' rights to information and consultation with the undertaking*<br><br>-Article 27, *Explanation on Article 27 — Workers' rights to information and consultation with the undertaking (Explanations Relating to the Charter of Fundamental Rights)* in the Official Journal of the European Union C 303/17 - 14.12.2007<br><br>-Article 31, Charter of the Fundamental Rights of the EU – *Fair and just working conditions*<br><br>-Article 31, *Explanation on Article 31 — Fair and just working conditions (Explanations Relating to the Charter of Fundamental Rights)* in the Official Journal of the European Union C 303/17 - 14.12.2007<br><br>-Article 34, Charter of the Fundamental Rights of the EU – *Social security and social assistance*<br><br>-Article 34, *Explanation on Article 34 — Social security and social assistance (Explanations Relating to the Charter of Fundamental Rights)* in the Official Journal of the European Union C 303/17 - 14.12.2007<br><br>-Article 35, Charter of the Fundamental Rights of the EU – *Healthcare*<br><br>-Article 35, *Explanation on Article 35 — Healthcare (Explanations Relating to the Charter of Fundamental Rights)* in the Official Journal of the European Union C 303/17 - 14.12.2007<br><br>**Universal Declaration of Human Rights:** |
| | | -Article 3, Universal Declaration of Human Rights – *Right to life, liberty, and the security of person*<br><br>-Article 12, Universal Declaration of Human Rights – *Right to private home life*<br><br>-Article 22, Universal Declaration of Human Rights – *Right to social security*<br><br>-Article 22, Universal Declaration of Human Rights – *Workers' rights* |
| **Accountability and effective remedy (1, 2, 3, 5, 6, 7)** | **-The right to an effective remedy for violation of rights and freedoms (Art. 13 ECHR). This should also include the right to effective and accessible remedies whenever the development or use of AI systems by private or public entities causes unjust harm or breaches an individual's legally protected rights.**<br><br>~<br><br>"AI systems must always remain under human control, even in circumstances where machine learning or similar techniques allow for the AI system to make decisions independently of specific human intervention. Member states must establish clear lines of responsibility for human rights violations that may arise at various phases of an AI system lifecycle. Responsibility and accountability for human rights violations that occur in the development, deployment or use of AI Systems must always lie with a natural or legal person, even in cases where the measure violating human rights was not directly ordered by a responsible human commander or operator. | **European Convention on Human Rights (ECHR):**<br><br>-Article 13, European Convention on Human Rights – *Right to an effective remedy*<br><br>-Article 13, 'Guide on Article 13 of the European Convention on Human Rights.', Council of Europe – *Right to an effective remedy*<br><br>**Charter of the Fundamental Rights of the EU:**<br><br>-Article 47, Charter of the Fundamental Rights of the EU – *Right to an effective remedy before a tribunal and to a fair trial*<br><br>*-Article 47, Explanation on Article 47 — Right to an effective remedy and a fair trial (Explanations Relating to the Charter of Fundamental Rights)* in the Official Journal of the European Union C 303/17 - 14.12.2007<br><br>**Universal Declaration of Human Rights:**<br><br>-Article 8, Universal Declaration of Human Rights – *Right to an effective remedy* |
| | Anyone who claims to be a victim of a human rights violation arising from the development, deployment or use by a public or private entity of an AI system should be provided with an effective remedy before a national authority. Moreover, member states should provide access to an effective remedy to those who suspect that they have been subjected to a measure that has been solely or significantly informed by the output of an AI system in a non-transparent manner and without their knowledge.<br><br>Effective remedies should involve prompt and adequate reparation and redress for any harm suffered by the development, deployment or use of AI systems, and may include measures under civil, administrative, or, where appropriate, criminal law. NHRSs can be such a source of redress, through rendering their own decisions in accordance with their respective mandates.<br><br>Member states should provide individuals with the right not to be subject to a decision significantly affecting them that is based on automated decision-making without meaningful human intervention. At the very least, an individual should be able to obtain human intervention in such automated decision-making and have their views taken into consideration before such a decision is implemented.<br><br>Member states must ensure that individuals have access to information in the possession of a defendant or a third party that is relevant to substantiating their claim that they are the victim of a human rights violation caused by an AI system, including, where relevant, training and testing data, information on how the AI system was used, meaningful and understandable information on how the AI system reached a recommendation, decision or | **International Covenant on Civil and Political Rights:**<br><br>-Article 2, International Covenant on Civil and Political Rights – *Right to effective remedy*<br><br>**Council of Europe Resources:**<br><br>-MSI-AUT Responsibility and AI: A study of the implications of advanced digital technologies (including AI systems) for the concept of responsibility within a human rights framework (2019)<br>• This report outlines what AI is and how task-specific technologies work, threats and harms associated with advanced digital technologies, and a range of 'responsibility models' for the adverse impacts of AI systems<br>• The main recommendations from this report are "effective and legitimate mechanisms that will prevent and forestall human rights violations", policy choices regarding responsibility models for AI systems, support of technical research involving human rights protections and 'algorithmic auditing', and the presence of legitimate governance mechanisms for the protection of human rights in the digital age |
| | prediction, and details of how the AI system's outputs were interpreted and acted on."[18] | |
| **Democracy (1, 2, 4, 5, 7, 8)** | **-The right to freedoms of expression, assembly, and association (Art. 10 and 11 ECHR).**<br><br>**-The right to vote and to be elected, the right to free and fair elections, and in particular universal, equal and free suffrage, including equality of opportunities and the freedom of voters to form an opinion. In this regard, individuals should not be subjected to any deception or manipulation.**<br><br>**-The right to (diverse) information, free discourse, and access to plurality of ideas and perspectives.**<br><br>**-The right to good governance.**<br><br>~<br><br>"Freedom of expression: Member states should be mindful of the obligation to create a diverse and pluralistic information environment and the adverse impact AI-driven content moderation and curation can have on the exercise of the right to freedom of expression, access to information, and freedom of opinion. Member states are also encouraged to consider taking appropriate measures to regulate technology monopolies to prevent the adverse effects of concentration of AI expertise and power on the free flow of information.<br><br>Freedom of assembly and association: Special attention should be paid to the impact the use of AI systems in content moderation can have on the freedom of assembly and association, especially in contexts where these freedoms are difficult to exercise offline. The use of facial | **European Convention on Human Rights (ECHR):**<br><br>-Article 3 of Protocol No.1, European Convention on Human Rights – *Right to free elections*<br><br>**-** Article 3 of Protocol No. 1, Guide on Article 3 of Protocol No. 1 to the European Convention of Human Rights – *Right to free elections*<br><br>-Article 10, European Convention on Human Rights – *Freedom of expression*<br><br>*-Article 10,* 'Guide on Article 10 of the European Convention on Human Rights', Council of Europe – *Freedom of expression*<br><br>-Article 11, European Convention on Human Rights – *Freedom of assembly and association*<br><br>-Article 11, 'Guide on Article 11 of the European Convention on Human Rights', Council of Europe – *Freedom of assembly and association*<br><br>**Charter of the Fundamental Rights of the EU:**<br><br>-Article 11, Charter of the Fundamental Rights of the EU – *Freedom of expression and information*<br><br>*-Article 11, Explanation on Article 11 — Freedom of expression and information (Explanations Relating to the Charter of Fundamental Rights)* in the Official Journal of the European Union C 303/17 - 14.12.2007 |

[18] Council of Europe Commissioner for Human Rights (2019) – "Unboxing AI: 10 steps to protect Human Rights"

| | | |
|---|---|---|
| | recognition technology should be strictly regulated by member states, including through legislation setting out clear limitations for its use, and public transparency to protect the effective exercise of the right to freedom of assembly."[19]<br><br>~<br><br>"Well-functioning democracies require a well-informed citizenry, an open social and political discourse and absence of opaque voter influence.<br><br>This requires a well-informed citizenry. In information societies citizens can only select to consume a small amount of all the available information. Search engines, social media feeds, recommender systems and many news sites employ AI to determine which content is created and shown to users (information personalization). If done well, this could help citizens to better navigate the flood of available information and improve their democratic competences, for instance by allowing them to access resources in other languages through translation tools. However, if AI determines which information is shown and consumed, what issues are suppressed in the flood of online information and which are virally amplified, this also brings risks of bias and unequal representation of opinions and voices.<br><br>**AI-driven** information personalisation is enabled by the constant monitoring and profiling of every individual. Driven by commercial or political motives this technologically-enabled informational infrastructure of our societies could amplify hyper-partisan content one is likely to agree with and provide an unprecedented | -Article 12, Charter of the Fundamental Rights of the EU – *Freedom of assembly and association*<br><br>*-Article 12, Explanation on Article 12 — Freedom of assembly and of association (Explanations Relating to the Charter of Fundamental Rights)* in the Official Journal of the European Union C 303/17 - 14.12.2007<br><br>**Universal Declaration of Human Rights:**<br><br>-Article 19, Universal Declaration of Human Rights – *Right to freedom of opinion and expression*<br><br>-Article 20, Universal Declaration of Human Rights – *Right to freedom of peaceful assembly and association*<br><br>**International Covenant on Civil and Political Rights:**<br><br>-Article 19, International Covenant on Civil and Political Rights – *Freedom of expression*<br><br>-Article 21, International Covenant on Civil and Political Rights – *Freedom of assembly*<br><br>-Article 22, International Covenant on Civil and Political Rights – *Freedom of association*<br><br>-Article 25, International Covenant on Civil and Political Rights – *Right to participate in public affairs, good governance, and elections*<br><br>**Council of Europe Resources:** |

[19] Council of Europe Commissioner for Human Rights (2019) – "Unboxing AI: 10 steps to protect Human Rights"

| | | |
|---|---|---|
| | powerful tool for individualised influence. As a consequence it may undermine the shared understanding, mutual respect and social cohesion required for democracy to thrive. If personal AI predictions become very powerful and effective, they may even threaten to undermine the human agency and autonomy required for meaningful decisions by voters.<br><br>Thirdly, AI can undermine a fair electoral process. Political campaigns or foreign actors can use (and have been using) personalised advertisements to send different messages to distinct voter groups without public accountability in the agora. However, it should be noted that it remains uncertain exactly how influential micro-targeted advertisement is. AI can also be used to create and spread misinformation and deep fakes, in the form of text, pictures, audio or video. Since these are hard to identify by citizens, journalists, or public institutions, misleading and manipulating the public becomes easier and the level of truthfulness and credibility of media and democratic discourse may deteriorate."[20]<br><br>~<br><br>"Democracy is government of the people by the people for the people. It provides checks against the concentration of power in the hands of a few and can function properly only if based on sound institutions which enjoy confidence of an active, committed, and informed citizenry and are able to provide for dynamic balance of interests of constituents. The crisis of modern democracies touches almost all elements of democratic order, including erosion of, and loss of confidence in institutions, mis- and disinformation of the public, break-up of cohesion and polarisation of society. Modern | -Venice Commission: Principles for a fundamental rights-compliant use of digital technologies in electoral processes (2020)<br>• Emphasized the need for a human rights-compliant approach to eight principles involving the use of digital technologies in elections<br><br>-Parliamentary Assembly, Need for democratic governance of artificial intelligence, Council of Europe |

[20] Muller, C. The Impact of Artificial Intelligence on Human Rights, Democracy and the Rule of Law. Council of Europe, CAHAI(2020)06-fin.

| | | |
|---|---|---|
| | technologies, including AI-based systems may both help resolve and aggravate this crisis.<br><br>The use of AI by humans is not neutral. It can be used to strengthen government accountability and can produce many benefits for democratic action, participation, and pluralism, making democracy more direct and responsive. However, it can also be used to strengthen repressive capabilities and for manipulation purposes. Indeed, the rapid integration of AI technologies into modern communication tools and social media platforms provides unique opportunities for targeted, personalised and often unnoticed influence on individuals and social groups, which different political actors may be tempted to use to their own benefit.<br><br>The experience of the last few years helps to identify some key areas where the use of AI-based technology can threaten to undermine and destabilise democracy, including, inter alia: (a)access to information (misinformation, "echo chambers" and erosion of critical thinking); (b)targeted manipulation of citizens; (c)interference in electoral processes; (d) erosion of civil rights; (e) shifts of financial and political power in the data economy.<br><br>Moreover, the broad use by States of AI-based technologies to control citizens such as automated filtering of information amounting to censorship, and mass surveillance using smartphones and closed-circuit television coupled with vast integrated databases, may lead to the erosion of political freedoms and the emergence of digital authoritarianism – a new social order competing with democracy."[21] | |

[21] Bergamini, D. (*Rapporteur)* (2020). Need for democratic governance of artificial intelligence. Council of Europe, Committee on Political Affairs and Democracy, Parliamentary Assembly.

| | | |
|---|---|---|
| **Rule of law (2)** | **-The right to a fair trial and due process (Art. 6 ECHR). This should also include the possibility of receiving insight into and challenging AI-informed decisions in the context of law enforcement or justice, including the right to review of such decisions by a human. The essential requirements that secure impacted individuals' access to the right of a fair trial must also be met: equality of arms, right to a natural judge established by law, the right to an independent and impartial tribunal, and respect for the adversarial process.**<br><br>**-The right to judicial independence and impartiality, and the right to legal assistance.**<br><br>**-The right to an effective remedy (Art. 13 ECHR), also in cases of unlawful harm or breach an individual's human rights in the context of AI systems.**<br><br>~<br><br>"The rule of law requires respect for principles such as legality, transparency, accountability, legal certainty, non-discrimination, equality and effective judicial protection – which can be at risk when certain decisions are delegated to AI systems. In addition, AI systems can also negatively affect the process of law-making and the application of the law by judges. Concerns have also been expressed on the possible negative effects of some AI applications used in judicial systems or connected areas. Such use could pose a challenge to the right to a fair trial enshrined in Article 6 of the ECHR, of which components such as the right to an independent and impartial judiciary, the right to a lawyer or the principle of equality of arms in judicial | **European Convention on Human Rights (ECHR):**<br><br>-Article 6, European Convention on Human Rights – *Right to a fair trial*<br><br>-Article 6, 'Guide on Article 6 of the European Convention on Human Rights.', Council of Europe – *Right to a fair trial*<br><br>-Article 13, European Convention on Human Rights – *Right to an effective remedy*<br><br>-Article 13, 'Guide on Article 13 of the European Convention on Human Rights.', Council of Europe – *Right to an effective remedy*<br><br>**Charter of the Fundamental Rights of the EU:**<br><br>-Article 47, Charter of the Fundamental Rights of the EU – *Right to an effective remedy before a tribunal and to a fair trial*<br><br>*-Article 47, Explanation on Article 47 — Right to an effective remedy and a fair trial (Explanations Relating to the Charter of Fundamental Rights)* in the Official Journal of the European Union C 303/17 - 14.12.2007<br><br>**Universal Declaration of Human Rights:**<br><br>-Article 8, Universal Declaration of Human Rights – *Right to an effective remedy*<br><br>- Article 10, Universal Declaration of Human Rights – *Right to a fair trial* |
| | proceedings are key elements that are also essential for the effective implementation of the rule of law.<br><br>Moreover, companies face increased pressure, including through regulation, to take decisions on the legality of content that is shown on their platform. Since social media platforms have become the new "public square", their own terms of service essentially set the rules of how freedom of expression manifests itself online, but with fewer safeguards than in more traditional public settings. It is, however, essential that states can and do continue to fulfil their responsibility for the protection of the rule of law."[22]<br><br>~<br><br>"The fact that AI can perpetuate or amplify existing biases, is particularly pertinent when used in law enforcement and the judiciary. In situations where physical freedom or personal security is at stake, such as with predictive policing, recidivism risk determination and sentencing, the right to liberty and security combined with the right to a fair trial are vulnerable**.** When an AI-system is used for recidivism prediction or sentencing it can have biased outcomes. When it is a black box, it becomes impossible for legal professionals, such as judges, lawyers and district attorneys to understand the reasoning behind the outcomes of the system and thus complicate the motivation and appeal of the judgement.<br><br>…Public institutions are held to a higher standard when it comes to their behaviour towards individuals and society, which is reflected in principles such as justification, proportionality and equality. AI can increase the efficiency | **International Covenant on Civil and Political Rights:**<br><br>-Article 2, International Covenant on Civil and Political Rights – *Right to effective remedy*<br><br>-Article 14, International Covenant on Civil and Political Rights – *Right to fair trial*<br><br>**Council of Europe Resources:**<br><br>-European Commission for the Efficiency of Justice, 'European ethical charter on the use of Artificial Intelligence in judicial systems and their environment' – Council of Europe |

[22] CAHAI *Feasibility Study*, Council of Europe CAHAI (2020)23.

| | | |
|---|---|---|
| | of institutions, yet on the other it can also erode the procedural legitimacy of and trust in democratic institutions and the authority of the law.<br><br>Courts, law enforcement and public administrations could become more efficient, yet at the cost of being more opaque and less human agency, autonomy and oversight.<br><br>Similarly, whereas previously courts were the only ones to determine what counts as illegal hate speech, today mostly private AI systems determine whether speech is taken down by social media platforms. These AI systems de facto compete for authority with judges and the law and In general, AI can contribute to developing judicial systems that operate outside the boundaries and protections of the rule of law.<br><br>Automated online dispute resolutions provided by private companies are governed by the terms of service rather than the law that do not award consumers the same rights and procedural protections in public courts."[23] | |

**Related enumerations of human rights, fundamental freedoms, and elements of democracy and the rule of law:**

1: Council of Europe Commissioner for Human Rights (2019) – "Unboxing AI: 10 steps to protect Human Rights"
2. OECD AI Principles and Classification Framework
3. European Agency for Fundamental Rights (2020) – "Getting the Future Right: Artificial Intelligence and Fundamental Rights"
4. Aizenberg & van den Hoven (2020) – "Designing for human rights in AI."
5. UNESCO (2021a/b) – Recommendations on the Ethics of Artificial Intelligence (two drafts, 1-25, 26-134)
6. Loi, et al. (2021) *Algorithm Watch* – "Automated Decision-Making Systems in the Public Sector: An Impact Assessment Tool for Public Authorities"
7. High-level Expert Group on Artificial Intelligence, European Commission (2020) – "The Assessment List for Trustworthy Artificial Intelligence (ALTAI) for self-assessment"
8. Mantelero & Esposito *(2021)* – An evidence-based methodology for human rights impact assessment (HRIA) in the development of AI data-intensive systems

[23] Muller, C. The Impact of Artificial Intelligence on Human Rights, Democracy and the Rule of Law. Council of Europe, CAHAI(2020)06-fin.

### *Determining the salience of the rights, freedoms, and elements of democracy and the rule of law to the AI system you are planning to build*

After reviewing Table 1—and carrying out further desk research (or consulting experts) to clarify questions pertaining to any of the rights or freedoms about which you are unsure—you are in a good position to begin considering the human rights, fundamental freedoms, and the elements of democracy and the rule of law that could be impacted by your system.

Go through each principle and priority listed in Table 1 (and the corresponding rights and freedoms enumerated in the adjacent columns) and ask:

- *How, if at all, are the rights, freedoms, or elements of democracy and the rule of law that are associated with this principle/priority salient to the AI system we are planning to build, given its intended purpose and the contexts in which it will be used?*
- *How could the rights, freedoms, or elements of democracy and the rule of law that are associated with this principle/priority be impacted by the AI system we are planning to build?*
- *If things go wrong in the implementation of our AI system or if it is used out-of-the-scope of its intended purpose and function, what harms could be done to impacted rights-holders in relation to the rights, freedoms, or elements of democracy and the rule of law that are associated with this principle/priority?*

Answering these questions as part of the PS Report will help you prepare for conducting the PCRA, where you will further assess the extent to which the AI system poses probable harms to specific rights and freedoms.

## *Activity 4: Map Governance Workflow*

The last activity of the planning and scoping phase of your PS Report involves mapping the governance workflow of your project. Governance is defined in the recently-adopted standard ISO 370000 as 'the system by which the whole organization is directed, controlled, and held accountable to achieve its core purpose in the long run'.[24] The governance workflow is the foundation of an anticipatory approach to your project that supports duty-bearers with ex-ante processes of accountability and auditing throughout the project lifecycle.[25]

[24] https://committee.iso.org/sites/tc309/home/projects/ongoing/ongoing-1.html

[25] The algorithmic impact assessment literature describes the value and importance of both ex-post and ex-ante assessments, including as they pertain to different features of organisational governance. However, recent work tends to place greater importance on the effectiveness of ex-ante assessments. See, for example, Moss, E., Watkins, E., Metcalf, J., & Elish, M. C. (2020). Governing with Algorithmic Impact Assessments: Six Observations. SSRN Electronic Journal. Leslie specifically argues that effective governance includes structures of accountability throughout the development and implementation of

Mapping a governance workflow, therefore, creates the baseline conditions for holding an organization actively accountable for the fulfilment of its goals and principles, including its duty to uphold human rights, democracy, and the rule of law.

Mapping the governance workflow is not only a process of articulating guiding principles and assigning responsibility; it is an opportunity to ensure a sufficiency of perspectives and understandings that extends beyond the organization to the societies and material conditions potentially impacted by an AI system. For example, as technical systems can significantly affect the social and economic conditions of human lives, systems of governance should be inclusive of a sufficiently diverse set of voices to increase the degree of critical reflection and the mitigation of institutional and individual biases that can lead to the reproduction and amplification of unjust social structures.[26]

Establishing a governance workflow is essential where the proposed system operates in an incomplete or uncertain regulatory environment, but an established governance workflow remains important even within tightly regulated contexts as a feature of *collaborative governance* in which structures of internal accountability complement regulatory structures and vice-versa.[27]

**Questions:**

- *What roles are involved in each of the project's phases?*
    - *Design*
    - *Development*
    - *Deployment*
- *What are the responsibilities associated with each of these roles?*
- *How are each of the duty-bearers involved in the project assigned responsibility for the system's potential impacts to human rights, democracy, and the rule of law?*
    - *Does this distribution establish a continuous chain of human accountability throughout the design, development, and deployment of this project? If so, how?*

AI systems rather than retrospectively. Leslie, D. (2019). Understanding artificial intelligence ethics and safety. The Alan Turing Institute. And McGregor et al. argues that human rights doctrine provides guidance for the governance of AI systems and organisations, including decision structures to guide the project lifecycle. McGregor, L., Murray, D., & Ng, V. (2019). International human rights law as a framework for algorithmic accountability. International and Comparative Law Quarterly, 68(2), 309–343.

[26] For further discussion on the components of an effective structure of internal governance for AI system development, including the requirements for critically assessing a system's social impact, see Raji et al. (2020). Closing the AI Accountability Gap: Defining an End-to-End Framework for Internal Algorithmic Auditing. Proceedings of the 2020 conference on fairness, accountability, and transparency. Similarly, Kemper & Kolkman argue for the importance of a *critical audience* to ensure that algorithmic systems are truly accountable beyond the firm. Kemper, J., & Kolkman, D. (2019). Transparent to whom? No algorithmic accountability without a critical audience. Information, Communication & Society, 22(14), 2081–2096.

[27] Complementary governance is described by Kaminski (2018). Binary Governance: Lessons from the GDPR's approach to algorithmic accountability. S. Cal. L. Rev., 92, 1529.

- *Do all duty-bearers understand the responsibility that has been assigned to them?*
- *What logging protocol will be established for documenting workflow activities?*
    - *Does this protocol enable external auditing and oversight of the design, development, and deployment of this project? If so, how?*
- *Can responsible duty bearers be traced in the event that the human rights or fundamental freedoms of rights-holders are harmed by this system? If so, how do the project's distribution of responsibilities and logging protocol enable this?*
- *If you are procuring parts or elements of the system from third-party vendors, suppliers, sub-contractors, or external developers, how are you instituting appropriate governance controls that will establish end-to-end accountability, traceability, and auditability?*
- *If any data being used in the production of the AI system will be acquired from a vendor or supplier, how are you instituting appropriate governance controls that will establish end-to-end accountability, traceability, and auditability across the data lifecycle?*

| **Project Summary Report Template** | |
|---|---|
| **Project Phases and Questions** | **Responses** |
| ***ACTIVITY 1: Outline Project, Use Context, Domain, and Data*** | |
| PROJECT | |
| *How would you describe your organisation and what sort of services or products do you typically provide (beyond the system under consideration)?* | |
| *What AI system is being built and what type of product or service will it offer?* | |
| *What benefits will the system bring to its users and customers, and will these benefits be widely accessible?* | |
| *Which organisation(s)—yours, other suppliers, or other providers—are responsible for building this AI system?* | |
| *Which parts or elements of the AI system, if any, will be procured from third-party vendors, suppliers, sub-contractors, or external developers?* | |
| *Which algorithms, techniques, and model types will be used in the AI system? (Provide links to technical papers where appropriate)* | |
| *In a scenario where your project optimally scales, how many people will it impact, for how long, and in what geographic range (local, national, global)? (Describe your rationale)* | |
| USE CONTEXT | |
| *What is the purpose of this AI system and in which contexts will it be used? (Briefly describe a use-case that illustrates primary intended use)* | |
| *Is the AI system's processing output to be used in a fully automated way or will there be some degree of human control, oversight, or input before use? (Describe)* | |
| *Will the AI system evolve or learn continuously in its use context, or will it be static?* | |
| *To what degree will the use of the AI system be time-critical, or will users be able to evaluate outputs comfortably over time?* | |
| *What sort of out-of-scope uses could users attempt to apply the AI system, and what dangers may arise from this?* | |
| DOMAIN | |
| *In what domain will this AI system operate?* | |
| *Which, if any, domain experts have been or will be consulted in designing and developing the AI system?* | |
| DATA | |
| *What datasets are being used to build this AI system?* | |
| *Will any data being used in the production of the AI system be acquired from a vendor or supplier? (Describe)* | |
| *Will the data being used in the production of the AI system be collected for that purpose, or will it be re-purposed from existing datasets? (Describe)* | |
| *What quality assurance and bias mitigation processes do you have in place for the data lifecycle—for both acquired and collected data?* | |
| ***ACTIVITY 2: Identify Stakeholders*** | |
| *Who are the rights-holders, duty bearers and other relevant parties that may be impacted by, or may impact, the project?* | |
| *Do any of these rights-holders possess sensitive or protected characteristics that could increase their vulnerability to abuse or discrimination, or for reason of which they may require* | |
| *additional protection or assistance with respect to the impacts of the project? If so, what characteristics?* | |
| *Could the outcomes of this project present significant concerns to specific groups of rights-holders given vulnerabilities caused or precipitated by their distinct circumstances?* | |
| *If so, what vulnerability characteristics expose them to being jeopardized by project outcomes?* | |
| *What affected group or groups of rights-holders could these protected or contextual vulnerability characteristics or qualities represent?* | |
| *How could the distribution of power relations (i.e., the relative advantages and disadvantages) between rights-holders and duty-bearers affect the way the benefits and risks of the project are allocated?* | |

**ACTIVITY 3: Scope Impacts to Human Rights, Democracy, and Rule of Law**

*Instructions: Go through each* ***principle and priority*** *listed in* ***Table 1*** *(and the corresponding rights and freedoms enumerated in the adjacent columns) and ask and respond to the following questions:*

1. *How, if at all, are   salient to the AI system we are planning to build, given its intended purpose and the contexts in which it will be used?*
2. *How could the rights, freedoms, or elements of democracy and the rule of law that are associated with this principle/priority be impacted by the AI system we are planning to build?*
3. *If things go wrong in the implementation of our AI system or if it is used out-of-the-scope of its intended purpose and function, what harms could be done to impacted persons in relation to the rights, freedoms, or elements of democracy and the rule of law that are associated with this principle/priority?*

<table>
<tr><td colspan="2"></td></tr>
<tr><td rowspan="3">Respect for and Protection of Human Dignity</td><td>1.</td></tr>
<tr><td>2.</td></tr>
<tr><td>3.</td></tr>
<tr><td rowspan="3">Protection of Human Freedom and Autonomy</td><td>1.</td></tr>
<tr><td>2.</td></tr>
<tr><td>3.</td></tr>
</table>

| | |
|---|---|
| **Prevention of Harm and Protection of the Right to Life and Physical, Psychological, and Moral Integrity** | 1. |
| | 2. |
| | 3. |
| **Non-discrimination, Fairness, and Equality** | 1. |
| | 2. |
| | 3. |
| **Data Protection and the Right to Respect of Private and Family Life** | 1. |
| | |
| | 2. |
| | 3. |
| **Economic and Social Rights** | 1. |
| | 2. |
| | 3. |
| **Accountability and Effective Remedy** | 1. |
| | 2. |
| | |
| | 3. |
| **Democracy** | 1. |
| | 2. |
| | 3. |
| **Rule of Law** | 1. |
| | 2. |
| | 3. |
| ***ACTIVITY 4: Map Governance Workflow*** | |
| *What roles are involved in each of the project phases?* | |
| Design Phase: | |
| Development Phase: | |
| Deployment Phase: | |
| *What are the responsibilities of each of these roles?* | |
| Design Roles: | |
| Development Roles: | |
| Deployment Roles: | |
| *How are each of these duty bearers assigned responsibility for the system's potential impacts to Human Rights, Democracy, and the Rule of Law?*<br>○ *Does this distribution establish a continuous chain of human accountability throughout the design, development, and deployment of this project? If so, how?* | |
| *What logging protocol is established for documenting workflow activities?*<br>○ *Does this protocol enable auditing and oversight of the design, development, and deployment of this project? If so, how?* | |
| *Can responsible duty bearers be traced in the event that the human rights or fundamental freedoms of rights-holders are harmed by this system? If so, how do the project's distribution of responsibilities and logging protocol enable this?* | |
| *If you are procuring parts or elements of the system from third-party vendors, suppliers, sub-contractors, or external developers, how are you instituting appropriate governance controls that will establish end-to-end accountability, traceability, and auditability?* | |
| *If any data being used in the production of the AI system will be acquired from a vendor or supplier, how are you instituting appropriate governance controls that will establish end-to-end accountability, traceability, and auditability across the data lifecycle?* | |

## PCRA: Document Overview

### *Introduction*

After you have done your preparatory desk research and completed the initial iteration of your PS Report, you will be ready to carry out your Preliminary Context-Based Risk Analysis (PCRA). The PCRA serves several important purposes:

1. It operates as an indicator of the pre-emptive actions needed to ensure non-harm for projects that pose unacceptable risk to impacted rights-holders. In this way, it facilitates the operationalization of the risk assessment component of the ***precautionary principle***.[28]
2. It helps you identify the ***risk factors*** that should be considered as you endeavour to reduce and mitigate the potential adverse impacts of your prospective project on the fundamental freedoms and human rights of affected rights-holders.
3. It directs you to ***specific actions*** that you then need to take in your impact assessment process (***HUDERIA***) so that sufficient consideration can be given to each of the risk factors detected and to the salient rights and freedoms that could be impacted by the prospective AI system. It also directs you to ***specific goals, properties, and areas*** that you should focus on in your subsequent risk management and assurance processes (***HUDERAC***) to reduce and mitigate associated risks.
4. Finally, it helps you establish a ***proportionate approach*** to your impact assessment and assurance practices, and to the level of stakeholder engagement that you will include across your project lifecycle. The proportionality principle counsels that efforts of impact assessment, risk management, and stakeholder involvement should be proportionate to the likelihood and extent of risks for adverse impacts on human rights, democracy, and the rule of law associated with an AI system. The report that is generated when you complete your PCRA will help you understand, in a provisional way, the risk level of your system, so that you can take appropriate risk management actions and engage stakeholders in a proportionate way.

[28] The European Commission guidance on the precautionary principles states that it "should be considered within a structured approach to the analysis of risk which comprises three elements: risk assessment, risk management, risk communication. The PCRA supports the risk assessment aspect, whereas—as is discussed below—the Human Rights, Democracy, and the Rule of Law Assurance Case (HUDERAC) primarily supports the risk management and risk communication elements. See: EU (2000). Communication from the commission on the precautionary principle, Communication. Brussels: Commission of the European Communities; Bourguignon, D. (2016) European Parliament, and Directorate-General for Parliamentary Research Services, The precautionary principle: definitions, applications and governance: in-depth analysis. Luxembourg: Publications Office.

## *PCRA Section One: Identifying Risk Factors*

In the first section of the PCRA, you will be asked to answer a series of questions that are intended to help you identify both ***modifiable and circumstantial risk factors*** to consider as you embark on the impact assessment and risk management and assurance elements of your HUDERAF. The term "risk factor" is here taken to mean *the characteristics of an AI innovation context at the technical, sociotechnical, historical, legal, economic, political, or practical levels that precede and are associated with a higher likelihood of outcomes that negatively impact human rights, democracy, and the rule of law*. [29] These factors are not necessarily to be treated as causes of adverse impacts but rather as conditions that are correlated with an increased chance of harm and that need to be anticipated and considered in risk mitigation efforts.

Those risk factors that emerge *externally* from the technical, sociotechnical historical, legal, economic, or political environments within which the design, development, and deployment of AI systems occur and that are thereby less controllable are called "circumstantial risk factors." Those risk factors that emerge *internally* from the actual practices of producing and using AI technologies, and that are thus more controllable, are called "modifiable risk factors".

To remain as "algorithm neutral"[30] as possible, the PCRA focuses on the spectrum of risk factors that surround the practical contexts of designing, developing, and deploying AI systems rather than on the technical details or underlying specifications of AI applications. Building on OECD's AI classification scheme[31] and related work, the questions in the first section focus on risk factors arising from both the use contexts and the design, development, and deployment contexts of the AI project lifecycle. In addition to this, the final fifteen questions focus on risk factors that arise in the immediate context human rights and fundamental freedoms. A summary of the question headings and sub-headings shows how this spectrum of concerns stretches from use contexts and design, development, and deployment contexts to rights and freedoms contexts:

[29] The concept of "risk factors," as used here, was first fully developed in medical and public health contexts but has since seen wide application in public policy and the social sciences. For background discussions, see Mercy, J.A., and O'Carroll, P.W. 1998. New directions in violence prevention: The public health arena. Violence and Victims 3(4):285–301.; Mrazek, P.J., and Haggerty, R.J., eds. 1994. Reducing Risks for Mental Disorders: Frontiers for Preventative Intervention Research. Washington, DC: National Academy Press.; Shader, M. (2001). Risk factors for delinquency: An overview. Washington, DC: US Department of Justice, Office of Justice Programs, Office of Juvenile Justice and Delinquency Prevention.; Sorooshian, S., & Mun, S. Y. (2020). Literature review: Critical risk factors affecting information-technology projects. *Calitatea*, *21*(175), 157-161.

[30] Following both the CAHAI *Feasibility Study*, Council of Europe CAHAI (2020)23 (p. 4) and CAHAI-PDG (2021)05rev.

[31] OECD (2021). Framework to Classify AI Systems. Public Consultation.

**Use context (Questions 1-20)**

- Sector or domain in which the system is being built
- Existing law and regulatory environment of the sector or domain
- Impact-level of the system
- Prohibited systems and uses
- Scope of deployment (breadth and temporality)
- Technological maturity
- Existing system (human or technological) that the application is replacing
- Bias and discrimination in sector or domain context
- Environmental context
- Cybersecurity context

**Data Lifecycle Context (Questions 21-40)**

- Data quality, integrity, and provenance
- Means and methods of data collection
- Data types
- Dataset linkage
- Data labelling and annotating practices

**Goal Setting and Problem Formulation Context (Questions 41-42)**

- Decision to design
- Definition of outcome

**Model Design & Development Context (Questions 43-46)**

- AI model characteristics
- Pre-processing and feature engineering
- Model selection

**Model Output & Implementation Context (Questions 47-52)**

- Model inference
- Model verification and validation
- Model accuracy and performance metrics

**System-User Interface and Human Factors Context (Questions 53-55)**

- Implementers or users of the system
- Level of automation/level of human involvement and choice

**Rights & Freedoms Context (Questions 56-71)**

- Respect for and protection of human dignity
- Protection of human freedom and autonomy
- Non-discrimination, fairness, and equality
- Data protection and privacy context
- Accountability and access to justice
- Social and economic rights

Depending on the form that the section one questions take, certain responses to each question will trigger one of three classes of risk factors, prohibitive, major, or moderate:

| | |
|---|---|
| **Prohibitive risk factor** | Prohibitive risk factors indicate the presence of determinants of potential harms that trigger the precautionary principle and precipitate pre-emptive measures to prevent adverse impacts on the human rights and fundamental freedoms of affected persons, democracy, and the rule of law. Pre-emptive measures are appropriate where the severity, scale, and irremediableness of the potential harm outweigh levels of risk reduction and mitigation. |
| **Major risk factor** | Major risk factors indicate the presence of determinants of potential harms that are directly or indirectly associated with significant risks of adverse impacts on the human rights and fundamental freedoms of affected persons, democracy, and the rule of law but that provide opportunities for risk reduction and mitigation that make the risks posed tolerable. |
| **Moderate risk factor** | Moderate risk factors indicate the presence of determinants of potential harms that are directly or indirectly associated with risks of adverse impacts on the human rights and fundamental freedoms of affected persons, democracy, and the rule of law but that provide opportunities for risk reduction and mitigation that make the risks posed broadly acceptable. |

Upon completion of the entire PCRA, a summary report will automatically be generated (See PCRA Output Example) which compiles and organizes the answers that flagged up risk factors into the three risk factor categories. The report provides prompts for each of these answers that direct you to specific actions to take in your impact assessment process and to specific goals, properties, and areas that you should focus on in your subsequent risk management and assurance processes to reduce and mitigate associated risks. For example, if you answered "yes" to Question 2, "Does the sector or domain in which the AI system will operate include vulnerable groups (or groups with protected characteristics) who may be significantly or disproportionately impacted by the design and use of the system?", the following message will be displayed in the summary report:

***Because you answered Yes to Question 2:***

**Major circumstantial risk factor**

Where the sector or domain in which the AI system will operate includes vulnerable groups (or groups with protected characteristics) who may be significantly or disproportionately impacted by the design and use of the system, this presents a ***major circumstantial risk factor*** for adverse impacts on the human rights and fundamental freedom of persons

- **Actions to take for your HUDERIA:** Make sure to focus upon considerations surrounding the potential impacts of your AI system on fairness, non-discrimination, equality, diversity, and inclusiveness
- **Goals, properties, and areas to focus on in your HUDERAC**: Fairness (non-discrimination, equality, bias mitigation, diversity, and inclusiveness)

The enumeration of risk factors and corresponding practical recommendations that are presented in the PCRA summary report is intended to support a deliberate, anticipatory, and reflective approach to your impact assessment, risk management, and assurance practices. It is meant to bolster safe and responsible AI innovation processes that respect human rights, democracy, and the rule of law.

### *PCRA Section Two: Risks of Adverse Impacts on the Human Rights and Fundamental Freedoms of Persons, Democracy, and the Rule of Law*

In the second section of your PCRA, you will be asked to provide a preliminary estimation of the likelihood of the potential adverse impacts that your system could have on human rights, democracy, and the rule of law. This will involve drawing on the stakeholder analysis and the determination of salient rights and freedoms you carried out as part of your initial PS reporting to think through the relevance and likelihood of the harms indicated in the prompts. Each prompt will be formed as a statement with likelihood/probability options below it.

These forty prompts encompass the categories of principles and priorities included in Table 1 of the Project Planning and Scoping Summary section of the PCRA. The range of potential adverse impacts covered in section two of the PCRA reflect the concerns raised in the CAHAI's *Feasibility Study* and in the multi-stakeholder consultation that followed the plenary's adoption of this text in December 2020.

It also reflects adjacent work done by the Council of Europe, UNESCO, the European Commission, OECD, and the European Agency for Fundamental Rights.[32]

The likelihood options that are provided for each of the section two prompts should be understood as follows[33]:

| Likelihood level | Description | Score |
|---|---|---|
| **Unlikely** | The risk of adverse impact is low, improbable, or highly improbable | **1** |
| **Possible** | The risk of adverse impact is moderate; the harm is possible and may occur | **2** |
| **Likely** | The risk of adverse impact is high; it is probable that the harm will occur | **3** |
| **Very Likely** | The risk of adverse impact is very high; it is highly probable that the harm will occur | **4** |
| **Not Applicable** | It can be claimed with certainty that the risk of adverse impact indicated in the prompt does not apply to the AI system. | **0** |

The likelihood score that is indicated in the far-right column is applied in the summary report as part of the calculation of a risk index number (RIN) that will be used to generate preliminary recommendations for proportionate risk management and assurance practices and stakeholder engagement given the risk level of the prospective AI system. After you have completed all the PCRA

[32] For other relevant sources, see: Council of Europe Commissioner for Human Rights (2019) – "Unboxing AI: 10 steps to protect Human Rights"; Council of Europe, "Recommendation CM/Rec (2020) of the Committee of Ministers to member States on the human rights impacts of algorithmic systems"; Muller, C. The Impact of Artificial Intelligence on Human Rights, Democracy and the Rule of Law. Council of Europe, CAHAI(2020)06-fin; Bergamini, D. (*Rapporteur)* (2020). Need for democratic governance of artificial intelligence. Council of Europe, Committee on Political Affairs and Democracy, Parliamentary Assembly; UNESCO (2021a/b) – Recommendations on the Ethics of Artificial Intelligence (two drafts, 1-25, 26-134); High-level Expert Group on Artificial Intelligence, European Commission (2020) – "The Assessment List for Trustworthy Artificial Intelligence (ALTAI) for self-assessment"; European Agency for Fundamental Rights (2020) – "Getting the Future Right: Artificial Intelligence and Fundamental Rights".

[33] Likelihood level descriptions are drawn partly from the probability table presented in (Mantelero & Esposito, 2021, p. 19), but whereas the latter includes the category of exposure in the description and calculation of likelihood levels, the PCRA counts only the probability. In the PCRA, the degree of human exposure is covered in the calculation of risk level by the variable of "number of rights-holders affected," the information for which is drawn from answers given to Question 10 in section 1 of the PCRA. Mantelero, A., and Esposito, S. (2021). An evidence-based methodology for human rights impact assessment (HRIA) in the development of AI data-intensive systems. Computer Law & Security Review.

questions, the likelihood scores from the answers you have given to the section 2 questions will be compiled in a table in the summary report that produces an estimated risk level for each of potential adverse impacts that have received a likelihood score of 1 or above:

| **Adverse Impacts:** | **Gravity Potential** | **Rights-Holders Affected** | **Severity** | **Likelihood** | **Risk Index Number (RIN)** | **Risk Level** |
|---|---|---|---|---|---|---|
| Anthropomorphic confusion over interaction with a computational system (dignity) | 3 | 1 | 4 | 2 | 6 | Moderate |
| Exposure to humiliation (dignity) | 3 | 1 | 4 | 1 | 5 | Low |
| Deprivation of rights-holders' abilities to make free, independent, and well-informed decisions about their lives or about the system's outputs, including their ability of rights-holders to effectively challenge decisions informed and/or made by that system and to demand that such decision be reviewed by a person (human freedom) | 3 | 1 | 4 | 1 | 5 | Low |
| Deprivation of the right to life or physical, psychological, or moral integrity (Prevention of harm) | 4 | 1 | 5 | 2 | 7 | High |
| Discrimination, discriminatory effects on impacted rights-holders, or differential performance (Non-discrimination) | 3 | 1 | 4 | 2 | 6 | Moderate |

(Table taken from PCRA Output Example). Note that numbers in the "Gravity Potential" and "Affected Rights-Holders Affected" columns reflect ordinal scales rather than real values and will be more fully described below)

The following formula is used to calculate the estimated RIN for potential harm event $i$:

$$RIN_i = Severity\ Potential_i + Likelihood_i$$ [34]

In this equation, the risk index number of a potential harm event $i$ is determined by an additive combination of the severity of the potential harm and its estimated likelihood.[35] This calculation works from the common risk analysis formulation of risk level as the product of severity (or impact/consequence) and likelihood (or probability/ occurrence), [36] but, in this case, severity is significantly upweighted in accordance with its priority as a "predominant factor" in human rights risk.[37] In the RIN formula used in the table, the Severity Potential number is determined by the addition of the Gravity Potential of the harm and the Number of Affected Rights-Holders[38]:

$$Severity\ Potential_i = Gravity\ Potential_i + Number\ of Affected\ RightsHolders_i$$

[34] Note that at this preliminary scoping phase, where risk analysis is helping to determine the proportionality of stakeholder engagement and risk management strategies, the variable of probability/likelihood figures in to impact estimation in a way that will differ from the approach taken in the HUDERIA phase of impact assessment, where severity assessment and impact management prioritization does not include probability considerations (following UNGP guidance) or incorporates them only as a secondary concern after severity assessment determines impact mitigation measures. As the DHRI notes, at the impact assessment stage, "severity does not include consideration of probability; instead, it prioritises a focus on the human rights consequences of the impact. This is not to say that consideration of probability is irrelevant. Consideration of probability will necessarily be involved in initial issues scoping." See See Figure 1 in Götzmann, N., Bansal, T., Wrzoncki, E., Veiberg, C. B., Tedaldi, J., & Høvsgaard, R. (2020). Human rights impact assessment guidance and toolbox. The Danish Institute for Human Rights. p. 90.
[35] The RIN calculation uses additive rather than multiplicative combination to reflect the approximately logarithmic relation of categories and the nature of the potential harms, following (Levine 2012), (Duijm 2015), and (Rausand and Haugen 2020). See: Levine, E.

Here, Gravity Potential is defined as the maximal or "worst-case scenario" degree of the gravity or seriousness of a potential harm's expected consequence, where gravity is categorized according to the extent of a potential harm's damage to human dignity or to the integrity of individual, collective, and planetary life.[39] Here is a table that maps out different levels of impact gravity (these gradients are

---

S. (2012). Improving risk matrices: the advantages of logarithmically scaled axes. *Journal of Risk Research*, 15(2), 209-222; Duijm, N. J. (2015). Recommendations on the use and design of risk matrices. *Safety science*, 76, 21-31. Rausand, M., & Haugen, S. (2020). *Risk Assessment: Theory, Methods, and Applications*. John Wiley & Sons.

[36] The quantification of risk is often represented by the general expression, $\sum_{i=1}^{n} l(C_i)p_i$, where consequence $C_i$ (quantified in loss function $l(C_i)$) is multiplied by the probability of the event $p_i$. However, where loss functions and probabilities are less quantifiable, ordinal approaches to determining severity and probability levels lead to the use of semi-quantitative methods. It should be noted, as well, that human rights risks cannot be reduced to any sort of quantified cost-benefit analysis, so conventional, quantitative approach to risk assessment are not applicable. For a discussion of the importance of combining quantitative and qualitative approaches in human rights impact assessments, see Nordic Trust Fund (2013). *Study on Human Rights Impact Assessments: A Review of the Literature, Differences with other Forms of Assessments and Relevance for Development*.

[37] UN Human Rights Office of the High Commissioner (UNHROHC) (2012) "The Corporate Responsibility to Protect Human Rights: An Interpretive Guide". p. 7.

[38] These two dimensions, gravity and number of affected rights-holders, line up with what the UN Human Rights Office of the High Commissioner term "scale" (how grave or serious the impact is) and "scope" (numbers of people impacted). Scale and scope are seen by the latter as constitutive elements of severity. N.B, the other crucial variable or "remediability" will be considered in the HUDERIA phase, where potentially affected stakeholders are included in the assessment of human rights impacts and these impacts are properly contextually specified. See UN Human Rights Office of the High Commissioner (2020) "Identifying and Assessing Human Rights Risks related to End-Use".

[39] It must be acknowledged that there is no predetermined ordering of human rights that indicates a prioritization of one over another, and that severity is a relative and contextually situated concept rather than an absolute one. The use of a notion of "gravity potential" here is not intended to imply otherwise, but rather to provide a provisional way to frame the human rights risks of AI through a precaution-centered approach so that proportionate methods of stakeholder engagement and risk management can be initiated. For guidance on characterizing severity, see: the UN Human Rights Office of the High Commissioner (UNHROHC) (2012) "The Corporate Responsibility to Protect Human Rights: An Interpretive Guide". pp. 83-84.

intended to orient you as you think through the harm potential of the system as it relates to the adverse impact on the right or freedom under consideration):

| Gravity Level | Description | Score |
|---|---|---|
| **Catastrophic Harm** | Catastrophic prejudices or impairments in the exercise of fundamental rights and freedoms that lead to the deprivation of the right to life; irreversible injury to physical, psychological, or moral integrity; deprivation of the welfare of entire groups or communities; catastrophic harm to democratic society, the rule of law, or to the preconditions of democratic ways of life and just legal order; deprivation of individual freedom and of the right to liberty and security; harm to the biosphere. | **4** |
| **Critical Harm** | Critical prejudices or impairments in the exercise of fundamental rights and freedoms that lead to the significant and enduring degradation of human dignity, autonomy, physical, psychological, or moral integrity, or the integrity of communal life, democratic society, or just legal order | **3** |
| **Serious Harm** | Serious prejudices or impairments in the exercise of fundamental rights and freedoms that lead to the temporary degradation of human dignity, autonomy, physical, psychological, or moral integrity, or the integrity of communal life, democratic society, or just legal order or that harm to the information and communication environment | **2** |
| **Moderate or Minor Harm** | Moderate or minor prejudices or impairments in the exercise of fundamental rights and freedoms that do not lead to any significant, enduring, or temporary degradation of human dignity, autonomy, physical, psychological, or moral integrity, or the integrity of communal life, democratic society, or just legal order | **1** |

For the purposes of a preliminary estimation of risk level, Gravity Potential should be treated as a heuristic that is based on the contextual assumption of a maximal degree of potential injury or damage for any given adverse impact—a heuristic that prioritizes the protection of rights-holders, communities, and the biosphere from potential irreversible, irreparable, catastrophic, critical, or serious harm. This follows the logic of the precautionary principle in working from the assumption that, in situations of uncertainty, where there are reasonable grounds for concern, levels of protection should prioritize human and environmental health and safety to facilitate sufficient preventative and anticipatory action.[40] It is important to note, however, that the Gravity Potential heuristic is *used only for the purpose of determining preliminary proportionality recommendations for risk management and assurance practices and stakeholder engagement—and is properly re-visited and re-evaluated in a contextually sensitive way (and with stakeholder input) in the Stakeholder Engagement Process and in the HUDERIA.*[41]

The Affected Rights-Holders Number is derived from the answer to question 10 from the first section of the PCRA, which asks you to estimate the number of affected rights-holders who would be directly or indirectly affected by the AI system over time in a scenario where your project optimally scales. In the calculation of the RIN, the four ranks of Affected Rights-Holders Numbers are assigned values that increase in intervals of .5, starting at .5:

| **Number of Affected Rights-Holders** | **Score** |
|---|---|
| **Between 1-10,000 Rights-Holders** | **.5** |
| **Between 10,001-100,000 Rights-Holders** | **1** |
| **Between 100,001-1,000,000 Rights-Holders** | **1.5** |
| **Over 1,000,000 Rights-Holders** | **2** |

This calibration reflects an up-weighting of the Gravity Potential and Likelihood scores in the RIN formula as they relate to the Affected Rights-Holders score, in keeping with their principal importance in the determination of the risk level. Whereas Gravity Potential and Likelihood are essential elements of the risk calculus and hence double-weighted, the Number of Affected Rights-Holders is added to the Severity Potential element of the calculation to provide a crucial but supplementary weight that incorporates information about the scope of the potential harm. The addition of the Affected Rights-Holders score to the Gravity Potential score, however, also means an overall upweighting of Severity Potential variable in keeping with its priority as a "predominant factor" in human rights risk.

[40] Following positions taken in the Rio Declaration (1992), the Maastricht Treaty (1992), the Wingspread Statement (1998), and Commission Communication (2000), but stressing the importance of "moderate precaution" in accordance with the Commission Communication (2000). For discussions, see: Garnett, K., & Parsons, D. J. (2017). Multi-case review of the application of the precautionary principle in European Union law and case law. *Risk Analysis*, 37(3), 502-516; Lofstedt, R. E. (2003). The precautionary principle: Risk, regulation and politics. *Process Safety and Environmental Protection*, 81(1), 36-43; Zander, J. (2010). *The application of the precautionary principle in practice: comparative dimensions.* Cambridge University Press; Sachs, N. M. (2011). Rescuing the strong precautionary principle from its critics. *U. Ill. L. Rev*., 1285.
[41] Another significant reason for using the precautionary principle to inform this heuristic is that it addresses, in part, problems surrounding the decision-making and "knowledge aspects" of risk characterization under conditions of uncertainty as emphasized in (Aven 2017) and (Krisper 2021). See: Aven, T. (2017). Improving risk characterisations in practical situations by highlighting knowledge aspects, with applications to risk matrices. *Reliability Engineering & System Safety*, 167, 42-48; Krisper, M. (2021). Problems with Risk Matrices Using Ordinal Scales. *arXiv preprint* arXiv:2103.05440.

The combination of Severity Potential (Gravity Potential + Number of Rights-Holders Affected) and Likelihood yields RINs that range from 2.5 to 10. Risk levels are established as intervals within this range[42]:

| Low Risk | Moderate Risk | High Risk | Very High Risk |
|---|---|---|---|
| $RIN \leq 5$ | $5.5 \leq RIN \leq 6$ | $6.5 \leq RIN \leq 7.5$ | $RIN \geq 8$ |

The distribution of RINs across these risk levels is reflected in the matrix below.[43] At the extremes, a minor or moderate harm from an AI system affecting between 1-10,000 rights-holders that is deemed unlikely yields a "low risk" output, whereas a catastrophic harm from an AI system affecting over 1,000,000 rights holders that is deemed very likely yields a "very high risk" output. Risk levels for borderline cases have been calibrated to accord with a precautionary approach that prioritizes sufficient levels of protection in risk reduction, mitigation, and management where substantial harm may occur. So, for instance, a serious harm from an AI system affecting 1 – 10,000 rights holders that is deemed likely yields a "high risk" output, and a catastrophic harm from the same system that is deemed possible also generates a "high risk" output.

**Distribution of RINs (risk index numbers)**

| | Severity | | | | | | | | |
|---|---|---|---|---|---|---|---|---|---|
| Likelihood | 1.5 | 2.5 | 3 | 3.5 | 4 | 4.5 | 5 | 5.5 | 6 |
| 1 | 2.5 | 3.5 | 4 | 4.5 | 5 | 5.5 | 6 | 6.5 | 7 |
| 2 | 3.5 | 4.5 | 5 | 5.5 | 6 | 6.5 | 7 | 7.5 | 8 |
| 3 | 4.5 | 5.5 | 6 | 6.5 | 7 | 7.5 | 8 | 8.5 | 9 |
| 4 | 5.5 | 6.5 | 7 | 7.5 | 8 | 8.5 | 9 | 9.5 | 10 |

[42] To explore the way this formula for risk level determination works, see the "Harms Table and RIN Distribution" Excel file that accompanies this proposal.

[43] The semi-quantitative methods employed here have taken into account concerns around consistency, resolution, input-output ambiguity, and aggregation long recognized in the critical study of risk matrices by (Cox 2008), (Smith et al 2009), (Levine 2012), (Duijm 2015), (Baybutt 2017), and (Krisper 2021). The distribution of RINs has been stress-tested and calibrated in accordance with the precautionary principle, but the provisional character and intermediate place of the results in a wider, dialogical and right-holder involving process must be stressed as a mitigating factor to possible coarseness in outputs. See: Anthony (Tony) Cox Jr, L. (2008). What's wrong with risk matrices?. *Risk Analysis: An International Journal*, 28(2), 497-512; Smith, E.D., Siefert, W.T., Drain, D., 2009. Risk matrix input data biases. *Syst. Eng.* 12 (4), 344–360. Levine, E. S. (2012). Improving risk matrices: the advantages of logarithmically scaled axes. *Journal of Risk Research*, 15(2), 209-222; Duijm, N. J. (2015). Recommendations on the use and design of risk matrices. *Safety science*, 76, 21-31; Baybutt, P. (2018). Guidelines for designing risk matrices. *Process safety progress*, 37(1), 49-55. Krisper, M. (2021). Problems with Risk Matrices Using Ordinal Scales. *arXiv preprint* arXiv:2103.05440.

In addition to providing a table which compiles the risks calculations for all the adverse impacts that have been identified in the PCRA (shown above), the PCRA Output Report (as demonstrated in the PCRA Output Example) will direct you to take specific ***risk management actions*** for each of these potential impacts:

| | |
|---|---|
| **For any $RIN \geq 8$ (very high)** | Further examination should be undertaken, through expert and stakeholder consultation, as to whether sufficient risk reduction is possible to make very high risks of harm tolerable or whether this is not feasible, and the risks are unacceptable. |
| **For any $6.5 \leq RIN \leq 7.5$ (high)** | Further examination should be undertaken, through expert and stakeholder consultation, as to whether the risks of harm indicated to be high are tolerable and can be appropriately reduced. Where likelihood = 1 (unlikely) and RPN indicates high risk, confirmation of low risk probability should also be undertaken through expert and stakeholder consultation. |
| **For any $5.5 \leq RIN \leq 6$ (moderate)** | Further examination should be undertaken, through expert and stakeholder consultation, as to whether the risks of harm indicated to be moderate are broadly accepted as such, are tolerable, and can be appropriately reduced. Where likelihood = 1 (unlikely) and RPN indicates moderate risk, confirmation of low risk probability should also be undertaken through expert and stakeholder consultation. |
| **For any $RIN \leq 5$ (low)** | Further examination should be undertaken, through expert and stakeholder consultation, as to whether the risks of harm indicated to be low are broadly accepted as such. |

The summary report will also produce more ***general recommendations for proportionate risk management practices and stakeholder engagement***. These recommendations are based on a prioritization of a degree of risk management and human right diligence that is proportionate to the highest level of risk identified across an AI system's potential impacts. Additional recommendations for establishing sufficient public transparency and accountability—as well as for ensuring adequate human rights diligence—are included for large-scale projects that could have macro-scale, long-term impacts on individuals and society. Here is table that includes all of the general recommendations:

| | |
|---|---|
| **Where the highest $RIN \geq 8$ (very high)** | Full diligence in risk management and assurance practices across all risks is recommended to prioritize risk reduction and mitigation. This should be informed by the risk factors identified in section 1 of the PCRA and by the HUDERIA impact assessment process; Comprehensive stakeholder engagement across the project lifecycle (e.g. partnering with or empowering rights-holders as determined by the Stakeholder Engagement Process) is also recommended. |
| **Where the highest $6.5 \leq RIN \leq 7.5$ (high)** | Full diligence in risk management and assurance practices is recommended to prioritize risk reduction and mitigation. This should be informed by the risk factors identified in section 1 of the PCRA and by the HUDERIA impact assessment process; Comprehensive stakeholder engagement across the project lifecycle (e.g. partnering with or empowering rights-holders as determined by the Stakeholder Engagement Process) is also recommended. |
| **Where the highest $5.5 \leq RIN \leq 6$ (moderate)** | Full diligence in risk management and assurance practices is recommended to prioritize risk reduction and mitigation but more targeted approaches to addressing specific risks may be acceptable, where experts and stakeholders have been appropriately consulted as part of this determination. Risk management and assurance practices should be informed by the risk factors identified in section 1 of the PCRA and by the HUDERIA impact assessment process; Appropriate stakeholder engagement across the project lifecycle is also recommended following the engagement objectives established by your Stakeholder Engagement Process. |
| **Where the highest $RIN \leq 5$ (low)** | Targeted approaches to addressing specific risks may be acceptable, where your highest risk index level is low and experts and stakeholders have been appropriately consulted to confirm this. Diligence in risk management and assurance practices across the project lifecycle is, however, still recommended. Risk management and assurance practices should be informed by the risk factors identified in section 1 of the PCRA and by the HUDERIA impact assessment process; Appropriate stakeholder engagement across the project lifecycle is also recommended following the engagement objectives established by your Stakeholder Engagement Process. |
| **Additional PCRA triggered recommendations for large-scale projects:** | If the AI system could have generational (10 to 20 years), long-term (20 to 60 years), or cross-generational (over 60 years) impacts on over 100,000 people and could pose risks to human rights, democracy, and the rule of law, you should |
| **If answer to PCRA question #10 = c or d and answer to PCRA question #11 = c, d, or e and answer to any PCRA questions #64-83 = likelihood 2, 3, or 4, then:** | prioritise public transparency and accountability by forming an oversight board for your project, which sufficiently represents the rights-holders and communities impacted by the system. You should also carry out a complete Human Rights Impact Assessment, following the recommendations of the Danish Institute for Human Rights or similar, to gauge the wider effects of your business activities on the communities and individuals they impact. |

## Preliminary Context Based Risk Analysis Template

**Key:**

| | |
|---|---|
| ***Example*** | Examples (hover cursor over text to display) |
| | No trigger message |
| | Moderate risk factor (considerable) |
| | Major risk factor (elevated) |
| | Prohibitive risk factor (triggers precautionary principle) |

| Context | Question | Trigger messages for summary report |
|---|---|---|
| **Section 1: Identifying risk factors**<br>In this section, you will be asked a series of questions that are intended to help you identify risk factors to consider as you embark on the impact assessment and risk management and assurance elements of your HUDERAF. Draw on your Project Summary Report to answer the questions to the best of your ability. If you are unsure about how to answer, choose that option and move forward to the next question. As part of the report generated when you complete this PCRA, you will receive an enumeration of the risk factors that have been identified. | | |
| **Use Context** | | |
| **Sector or domain for which the system is being built** | 1) Will the AI system serve an essential or primary function in a high impact or safety critical sector (e.g., transport, social care, healthcare, other divisions of the public sector)?<br><br>☐ YES<br>☐ NO<br>☐ UNSURE<br><br>**Example** | YES:<br><br>**Major circumstantial risk factor**<br><br>• Where AI systems serve primary or critical functions in high impact or safety critical sectors, this presents a ***major circumstantial risk factor*** for adverse impacts on the human rights and fundamental freedom of persons.<br><br>**Actions to take for your HUDERIA:**<br><br>• Make sure to focus upon considerations surrounding the prevention of harm and the respect of the right to life and to physical, psychological, and moral integrity |
| | | **Goals, properties, and areas to focus on in your HUDERAC**:<br><br>• Safety (accuracy and system performance, security, reliability, and robustness)<br>• Sustainability, (reflection on context and impacts, change monitoring) |
| | | NO: No message |
| | | UNSURE: Before taking any further steps in the proposed project, you should determine, through expert and stakeholder input where appropriate, whether the AI system will serve a primary or critical function in a high impact or safety critical sector. When this information is ascertained, you should return to your PCRA and revise it accordingly. |
| | 2) Does the sector or domain in which the AI system will operate include vulnerable groups (or groups with protected characteristics) who may be significantly or disproportionately impacted by the design and use of the system?<br><br>☐ YES<br><br>☐ NO<br><br>☐ UNSURE | YES:<br><br>**Major circumstantial risk factor**<br><br>• Where the sector or domain in which the AI system will operate includes vulnerable groups (or groups with protected characteristics) who may be significantly or disproportionately impacted by the design and use of the system, this presents a ***major circumstantial risk factor*** for adverse impacts on the human rights and fundamental freedom of persons. |
| | ***Example*** | **Actions to take for your HUDERIA:**<br><br>• Make sure to focus upon considerations surrounding the potential impacts of your AI system on fairness, non-discrimination, equality, diversity, and inclusiveness<br><br>**Goals, properties, and areas to focus on in your HUDERAC**:<br><br>• Fairness (non-discrimination, equality, bias mitigation, diversity, and inclusiveness) |
| | | NO: No message |
| | | UNSURE: Before taking any further steps in the proposed project, you should determine, through expert and stakeholder input where appropriate, whether the AI system will operate in a sector or domain which includes vulnerable groups (or groups with protected characteristics), who may be significantly or disproportionately impacted by the design and use of the system. When this information is ascertained, you should return to your PCRA and revise it accordingly. |
| **Existing law and regulatory environment of the sector or domain** | 3) Have you assessed existing law and regulation in the sector or domain in which the AI system will operate and determined that it has | YES: No message |
| | | NO: STOP. |
| | a legal basis and can be developed and deployed lawfully?<br><br>☐ YES<br>☐ NO<br>☐ UNSURE | **Prohibitive circumstantial risk factor**<br><br>If you have not been able to establish a legal basis for the AI system, you should not proceed. Where appropriate, you should consult with experts to establish the lawfulness of your processing objective and proposed project before taking any next steps. If you would like to continue with the PCRA anyway, select yes. If not, please reconsider your project and return when you have ascertained that your AI system will be lawful. |
| | | UNSURE: STOP. Before taking any further steps in the proposed project, you should determine, through expert and stakeholder input where appropriate, whether your project has a legal basis. When this information is ascertained, you should return to your PCRA and revise it accordingly. |
| | 4) Is the sector or domain in which the AI system will operate historically highly regulated?<br><br>☐ YES<br>☐ NO<br>☐ UNSURE<br><br>***Example*** | YES:<br><br>**Moderate circumstantial risk factor**<br><br>• Where the sector or domain in which the AI system will operate is historically highly regulated, this presents a ***moderate circumstantial risk factor*** for adverse impacts on the human rights and fundamental freedom of persons. |
| | | **Actions to take for your HUDERIA:**<br><br>• Make sure to consider the regulatory context of your project and to reflect on the expectations of compliant and reasonable practices that arise in that context.<br><br>**Goals, properties, and areas to focus on in your HUDERAC**:<br><br>• Accountability and process transparency, (traceability, accessibility, auditability, and responsible governance) |
| | | NO: No message |
| | | UNSURE: Before taking any further steps in the proposed project, you should examine, seeking expert and stakeholder input where appropriate, whether any relevant existing regulation in your sector or domain applies to the prospective AI system. When this information is ascertained, you should return to your PCRA and revise it accordingly. |
| | 5) Does statute or regulation in the sector or domain in which the AI system will operate require any other types of impact assessment for the specific use-case of the AI systems you are planning to | YES:<br><br>**Moderate circumstantial risk factor**<br><br>• Where statute or regulation in the sector or domain in which the AI system will |
| | develop (e.g., data protection impact assessment, equality impact assessment, human rights impact assessment, etc.)?<br><br>☐ YES<br><br>☐ NO<br><br>☐ UNSURE<br><br>***Example*** | operate requires other types of impact assessment for the specific use-case of the AI systems, this presents a ***moderate circumstantial risk factor*** for adverse impacts on the human rights and fundamental freedom of persons.<br><br>**Actions to take for your HUDERIA:**<br><br>• Make sure to integrate the completion of other compulsory impact assessments as supplementary material and evidentiary support for your HUDERIA.<br><br>**Goals, properties, and areas to focus on in your HUDERAC**:<br><br>• Accountability and process transparency (traceability, accessibility, auditability, and responsible governance) |
| | | NO: No message |
| | | UNSURE: Before taking any further steps in the proposed project, you should determine, through expert and stakeholder input where appropriate, whether any other types of impact assessment for the specific use-case of the AI systems you are planning to develop are required by law or regulation. When this information is ascertained, you should return to your PCRA and revise it accordingly. |
| | | YES: |

<table>
<tr>
<td rowspan="3"><u>Impact-level of the system</u></td>
<td rowspan="3">6) Will the AI system perform a safety critical or high impact function independent of the sector in which it operates?<br><br>☐ YES<br><br>☐ NO<br><br>☐ UNSURE<br><br>Example</td>
<td>Major circumstantial risk factor<br><br>• Where AI systems perform a safety critical or high impact function independent of the sector in which they operate, this presents a major circumstantial risk factor for adverse impacts on the human rights and fundamental freedom of persons.<br><br>Actions to take for your HUDERIA:<br><br>• Make sure to focus upon considerations surrounding the prevention of harm and the respect of the right to life and to physical, psychological, and moral integrity<br><br>Goals, properties, and areas to focus on in your HUDERAC:<br><br>• Safety (accuracy and system performance, security, reliability, and robustness) and sustainability</td>
</tr>
<tr>
<td>NO: No message</td>
</tr>
<tr>
<td>UNSURE: Before taking any further steps in the proposed project, you should determine, through expert and stakeholder input where appropriate, whether the AI system you are planning to build performs a safety critical or high impact function.</td>
</tr>
</table>

| | | |
|---|---|---|
| | | When this information is ascertained, you should return to your PCRA and revise it accordingly. |
| **Prohibited AI systems and uses** | 7) Is the AI system on the list of prohibited systems - EU AI Reg?<br><br>☐ YES<br>☐ NO<br>☐ UNSURE<br><br>***Example*** | YES: STOP.<br><br>**Prohibitive circumstantial risk factor**<br><br>If the AI system you are planning to build is on the list of prohibited systems, it poses too high a risk to impacted rights-holders and communities for you to proceed. Prohibited systems that may catastrophically, critically, or seriously harm the physical, psychological, or moral integrity or the human rights and fundamental freedoms of affected persons should not be developed or implemented in that form. If you would like to continue with the PCRA anyway at this time, select yes. If not, please reconsider your project and return when you have ascertained that your AI system will be lawful. If you would like to continue with the PCRA anyway, select yes. If not, please reconsider your project and return when you have ascertained that your AI system will be not fall under the list of prohibited systems. |
| | | NO: No message |
| | | UNSURE: STOP. Before taking any further steps in the proposed project, you should determine, through expert and stakeholder input where appropriate, whether the AI system you are planning to build is on the list of prohibited systems. |
| | 8) Could the AI system be repurposed or used in ways that fall under the list of prohibited systems?<br><br>☐ YES<br><br>☐ NO<br><br>☐ UNSURE<br><br>***Example*** | YES:<br><br>**Major circumstantial risk factor**<br><br>• Where AI systems could be repurposed or used in ways that fall under the list of prohibited systems, this presents a ***major circumstantial risk factor*** for adverse impacts on the human rights and fundamental freedom of persons.<br><br>**Actions to take for your HUDERIA:**<br><br>No trigger message<br><br>**Goal, properties, and areas to focus on in your HUDERAC**:<br><br>• You should take special precautions to put in place protections, processes, and mechanisms that ensure the limitation of the system's use to the purpose for which it was created. These protections, processes, and mechanisms should be demonstrated and evidenced in your HUDERAC. |
| | | NO: No message |
| | | UNSURE: Before taking any further steps in the proposed project, you should determine, through expert and stakeholder input where appropriate, |

<table>
<tr><td></td><td></td><td>whether the AI system could be used or repurposed in ways that fall under the list of prohibited systems are considered to pose significant risks. When this information is ascertained, you should return to your PCRA and revise it accordingly.</td></tr>
<tr><td rowspan="2">Scope of deployment (breadth and temporality)</td><td rowspan="2">9) In a scenario where your project optimally scales, will the AI system directly and/or indirectly affect rights-holders and groups within:<br><br>☐ a. Organisations<br>☐ b. Local populations<br>☐ c. National populations<br>☐ d. Global population?<br><br>Example a<br>Example b<br>Example c<br>Example d</td><td>a. Low risk factor<br>No message</td></tr>
<tr><td>b. Moderate circumstantial risk factor<br><br>• Where AI systems directly or indirectly affect rights-holders and groups within local populations, this presents a moderate circumstantial risk factor for adverse impacts on the human rights and fundamental freedom of persons.<br><br>Actions to take for your HUDERIA:<br><br>• In assessing the potential impacts of the system, you should pay special attention to the potential effects of the AI system on local populations and on the communities and groups that comprise them.<br><br>Goals, properties, and areas to focus on in your HUDERAC:<br>No trigger message</td></tr>
</table>

| | | |
|---|---|---|
| | | c. **Major circumstantial risk factor**<br><br>• Where AI systems directly or indirectly affect rights-holders and groups within national populations, this presents a ***major circumstantial risk factor*** for adverse impacts on the human rights and fundamental freedom of persons.<br><br>**Actions to take for your HUDERIA:**<br><br>• In assessing the potential impacts of the system, make sure to pay special attention to the potential effects of the AI system on local and national populations and on the communities and groups that comprise them.<br><br>**Goal, properties, and areas to focus on in your HUDERAC**:<br><br>No trigger message |
| | | d. **Major circumstantial risk factor**<br><br>• Where AI systems directly or indirectly affect rights-holders and groups within national populations, this presents a ***major circumstantial risk factor*** for adverse impacts on the human rights and fundamental freedom of persons.<br><br>**Actions to take for your HUDERIA:** |
| | | • In assessing the potential impacts of the system, make sure to pay special attention to the potential effects of the AI system on local, national, and global populations and on the communities and groups that comprise them.<br><br>**Goal, properties, and areas to focus on in your HUDERAC**:<br>No trigger message |
| | 10) In a scenario where your project optimally scales, will the AI system directly and/or indirectly affect:<br><br>☐ a. Between 1 – 10,000 rights-holders<br><br>☐ b. Between 10,001 – 100,000 rights holders<br><br>☐ c. Between 100,001 – 1,000,000 rights-holders<br><br>☐ d. Over 1,000,000 rights-holders?<br><br>***Example a*** | a. No message<br><br>b. **Moderate circumstantial risk factor**<br><br>• Where AI systems directly or indirectly affect between 10,001-100,000 rights-holders, this presents a ***moderate circumstantial risk factor*** for adverse impacts on the human rights and fundamental freedom of persons.<br><br>**Actions to take for your HUDERIA:**<br>No trigger message<br><br>**Goal, properties, and areas to focus on in your HUDERAC**:<br>No trigger message |
| | ***Example b***<br><br>***Example c***<br><br>***Example d*** | c. **Major circumstantial risk factor**<br><br>• Where AI systems directly or indirectly affect between 100,001-1,000,000 rights-holders, this presents a ***major circumstantial risk factor*** for adverse impacts on the human rights and fundamental freedom of persons.<br><br>**Actions to take for your HUDERIA:**<br><br>• In assessing the potential impacts of the system, make sure to pay special attention to the *at scale or mass-level effects* of the use of the AI system on social and political processes and institutions—in particular, as these relate to human rights, fundamental freedoms, democracy, and the rule of law.<br><br>**Goal, properties, and areas to focus on in your HUDERAC**:<br><br>No trigger message |
| | | d. **Major circumstantial risk factor**<br><br>• Where AI systems directly or indirectly affect over 1,000,000 rights-holders, this presents a ***major circumstantial risk factor*** for adverse impacts on the human rights and fundamental freedom of persons.<br><br>**Actions to take for your HUDERIA:**<br><br>• In assessing the potential impacts of the system, make sure to pay special attention to the *at scale or mass-level effects* of the use of the AI system on social and political processes and institutions—in particular, as these relate to human rights, fundamental freedoms, democracy, and the rule of law.<br><br>**Goal, properties, and areas to focus on in your HUDERAC**:<br>No trigger message |
| | 11) Considering the potential direct and indirect impacts of your project on individuals, communities, and the environment, which of the following is the widest timescale within which the AI system could affect rights-holders and groups: | a. No message |
| | | b. **Moderate circumstantial risk factor**<br><br>• Where AI systems directly or indirectly affect rights-holders medium term (1 to 10 years), this presents a ***moderate circumstantial risk factor*** for adverse |
| | ☐ a. Short term (less than a year)<br><br>☐ b. Medium term (1 to 10 years)<br><br>☐ c. Generationally (10 to 20 years)<br><br>☐ d. Long-term (20 to 60 years)<br><br>☐ e. Over lifetimes and across future generations (over 60 years)<br><br>☐ f. UNSURE<br><br>***Example a***<br><br>***Example b***<br><br>***Example c***<br><br>***Example d***<br><br>***Example e:1***<br><br>***Example e:2*** | impacts on the human rights and fundamental freedom of persons.<br><br>**Actions to take for your HUDERIA:**<br><br>• In assessing the potential impacts of the system, make sure to pay special attention to the *medium term (1-10 years) effects* of the use of the AI system on the individual identities of rights holders as well as on social and political processes and institutions—in particular, as these relate to human rights, fundamental freedoms, democracy, and the rule of law.<br><br>**Goal, properties, and areas to focus on in your HUDERAC**:<br><br>No trigger message |
| | | c. **Major circumstantial risk factor**<br><br>• Where AI systems directly or indirectly affect rights-holders generationally (10 to 20 years), this presents a ***major circumstantial risk factor*** for adverse impacts on the human rights and fundamental freedom of persons.<br><br>**Actions to take for your HUDERIA:** |
| | | <ul><li>In assessing the potential impacts of the system make sure to pay special attention to the *generational (10-20 years) effects* of the use of the AI system on the individual identities of rights holders, on socioeconomic relationships, and on social and political processes and institutions—in particular, as these relate to human rights, fundamental freedoms, democracy, and the rule of law.</li></ul>**Goal, properties, and areas to focus on in your HUDERAC**:<br>No trigger message |
| | | d. **Major circumstantial risk factor**<ul><li>Where AI systems directly or indirectly affect rights-holders long-term (20 to 60 years), this presents a ***major circumstantial risk factor*** for adverse impacts on the human rights and fundamental freedom of persons.</li></ul>**Actions to take for your HUDERIA:**<ul><li>In assessing the potential impacts of the system, make sure to pay special attention to the *long-term (20 to 60 years) and generational (10-20 years) effects* of the use of the AI system on the</li></ul> |
| | | individual identities of rights holders, on socioeconomic relationships, and on social and political processes and institutions—in particular, as these relate to human rights, fundamental freedoms, democracy, and the rule of law.<br><br>**Goal, properties, and areas to focus on in your HUDERAC**:<br><br>No trigger message |
| | | e. **Major circumstantial risk factor**<br><br>• Where AI systems directly or indirectly affect rights-holders over lifetimes and across future generations (over 60 years), this presents a ***major circumstantial risk factor*** for adverse impacts on the human rights and fundamental freedom of persons.<br><br>**Actions to take for your HUDERIA:**<br><br>• In assessing the potential impacts of the system, make sure to pay special attention to the *long-term (20 to 60 years), generational (10-20 years), and cross-generational (over 60 years) effects* of the use of the AI system on the individual identities of rights holders, on socioeconomic relationships, and on social and political processes and |
| | | institutions—in particular, as these relate to human rights, fundamental freedoms, democracy, and the rule of law.<br><br>**Goal, properties, and areas to focus on in your HUDERAC**:<br>No trigger message<br><br>**Proportionality trigger**: If the AI system could have cross-generational (over 60 years) impacts on over 100,000 people, you should prioritise public transparency and accountability by forming an oversight board for your project, which sufficiently represents the rights-holders and communities impacted by the system. You should also carry out a complete Human Rights Impact Assessment to gauge the wider effects of your business activities on the communities and individuals they impact. |
| | | UNSURE: Before taking any further steps in the proposed project, you should determine, seeking expert and stakeholder input where appropriate, the timescale of the prospective AI system's impacts on affected rights-holders and communities. When this information is ascertained, you should return to your PCRA and revise it accordingly. |
| **Technological maturity** | 12) Will the AI system's design be based on well-understood techniques that have previously been in operation and externally | YES: No message |
| | | NO: |
| | validated for a similar purpose and in the same sector?<br><br>☐ YES<br><br>☐ NO<br><br>☐ UNSURE<br><br>***Example*** | **Moderate modifiable risk factor**<br><br>• Where AI systems are not designed based on well-understood techniques that have previously been in operation and externally validated for a similar purpose and in the same sector, this presents a ***moderate modifiable risk factor*** for adverse impacts on the human rights and fundamental freedom of persons.<br><br>**Actions to take for your HUDERIA:**<br><br>No trigger message<br><br>**Goals, properties, and areas to focus on in your HUDERAC**:<br><br>• Safety (accuracy and system performance, security, reliability, and robustness) and sustainability (change monitoring). In particular, you should demonstrate in your assurance case that diligent processes of testing, verifying, and externally validating the performance of the system has occurred. You should also establish system monitoring and performance evaluation protocols that are proportionate to the system's technological maturity. |
| | | UNSURE: Before taking any further steps in the proposed project, you should determine, through expert and stakeholder input where appropriate, whether the techniques on which the AI system will be based have been successfully used in other |
| | | externally validated systems that have had similar purposes and were in the same sector. When this information is ascertained, you should return to your PCRA and revise it accordingly. |
| **Existing system (human or technological) that the application is replacing** | 13) If the AI system is replacing a human, technical, or hybrid system that serves the same or similar function, is a reason for replacement that the existing system is considered flawed or harmful?<br><br>☐ YES<br>☐ NO<br>☐ UNSURE<br>☐ NOT APPLICABLE<br><br>***Example*** | YES:<br><br>**Moderate circumstantial risk factor**<br><br>• Where AI systems are replacing a human, technical, or hybrid system that serves the same or similar function, because the existing system is considered flawed or harmful, this presents a ***moderate circumstantial risk factor*** for adverse impacts on the human rights and fundamental freedom of persons.<br><br>**Actions to take for your HUDERIA:**<br><br>• Make sure to pay special attention to assessing how your AI system could potentially repeat the harmful outcomes generated by the replaced system and how to minimise any potential replicated harms.<br><br>**Goals, properties, and areas to focus on in your HUDERAC**:<br><br>• Safety (accuracy and system performance, security, reliability, and robustness). Your assurance case should demonstrate how you have ensured that the flaws or harms generated by the previous system are redressed and |
| | | rectified across the design, development, and deployment lifecycle |
| | | NO: No message |
| | | UNSURE: Before taking any further steps in the proposed project, you should determine, through expert and stakeholder input where appropriate, whether the prospective AI system is replacing a human, technical, or hybrid system that serves the same or similar function, is a reason for replacement that the existing system is considered flawed or harmful. When this information is ascertained, you should return to your PCRA and revise it accordingly. |
| | | NOT APPLICABLE: No message |
| | 14) If the human, technical, or hybrid system being replaced is considered critical infrastructure or serves a safety-critical or high impact function, has assessment and planning been done and made public to transparently ensure that its updating/replacement does not cause unacceptable outage or harm?<br><br>☐ YES | YES: No message |
| | | NO: **Major modifiable risk factor**<br><br>• Where assessment and planning is not done and made public to transparently ensure that the updating/replacement of a human, technical, or hybrid system that is considered critical infrastructure or serves a safety-critical or high impact function does not cause unacceptable outage or harm, this presents a ***major modifiable risk factor*** for adverse impacts on the human rights and fundamental freedom of persons. |
| | ☐ NO<br><br>☐ UNSURE<br><br>☐ NOT APPLICABLE<br><br>***Example*** | **Actions to take for your HUDERIA:**<br><br>No trigger message<br><br>**Goals, properties, and areas to focus on in your HUDERAC**:<br><br>• You should incorporate publicly transparent assessment and planning into your assurance case which demonstrates that the updating/replacement of the previous system will not cause unacceptable outage or harm. |
| | | UNSURE: Before taking any further steps in the proposed project, you should determine, through expert and stakeholder input where appropriate, whether the AI system you are planning to build will replace a human, technical, or hybrid system that is considered critical infrastructure or serves a safety-critical or high impact function and whether the new system may cause unacceptable outage or harm. When this information is ascertained, you should return to your PCRA and revise it accordingly. |
| | | NOT APPLICABLE: No message |
| | 15) If the system is automating or replacing human labour, will assessment and planning be done and made public to transparently ensure that any corresponding job | YES, WE HAVE PLANS IN PLACE TO DO THIS: No message |
| | | WE HAD NOT CONSIDERED IT, BUT MAY DO THIS / WE HAD NOT CONSIDERED IT, BUT ARE |
| | loss, labour displacement, or redeployment is publicly acceptable, and any possible corollary harms are managed and mitigated?<br><br>☐ YES, WE HAVE PLANS IN PLACE TO DO THIS<br><br>☐ WE HAD NOT CONSIDERED IT, BUT MAY DO THIS<br><br>☐ WE HAD NOT CONSIDERED IT, BUT ARE UNLIKELY TO DO THIS<br><br>☐ NO, WE ARE NOT PLANNING TO DO THIS<br><br>☐ UNSURE<br><br>☐ NOT APPLICABLE<br><br>***Example 1***<br><br>***Example 2*** | UNLIKELY TO DO THIS / NO, WE ARE NOT PLANNING TO DO THIS:<br><br>**Moderate modifiable risk factor**<br><br>• Where assessment and planning is not done and made public to transparently ensure that job loss, labour displacement, or redeployment, which is the result of an AI system automating or replacing human labour, is publicly acceptable, and any possible corollary harms are managed and mitigated, this presents a ***moderate modifiable risk factor*** for adverse impacts on the human rights and fundamental freedom of persons.<br><br>**Actions to take for your HUDERIA:**<br><br>• Make sure to pay special attention to assessing how your AI system could potentially harm human dignity through labour displacement and infringe upon the right to just working conditions, the right to safe and healthy working conditions, and the right to organize as set out in the European Social Charter<br><br>**Goals, properties, and areas to focus on in your HUDERAC**:<br><br>• Sustainability (reflection on context and impacts and stakeholder engagement and involvement) and fairness (non-discrimination and equality). Make sure to |
| | | incorporate publicly transparent assessment and planning into your assurance case to ensure that any corresponding job loss, labour displacement, or redeployment is publicly acceptable and that any possible corollary harms are managed and mitigated. |
| | | UNSURE: Before taking any further steps in the proposed project, you should determine, through expert and stakeholder input where appropriate, whether the system you are planning to build will automate or replace human labour in a way that creates publicly unacceptable job loss, labour displacement, or redeployment. When this information is ascertained, you should return to your PCRA and revise it accordingly. |
| | | NOT APPLICABLE: No message |
| **Bias and discrimination in sector or domain context** | 16) Do the sector(s) or domain(s) in which the AI system will operate, and from which the data used to train it are drawn, contain historical legacies and patterns of discrimination, inequality, bias, racism, or unfair treatment of minority, marginalized, or vulnerable groups that could be replicated or augmented in the functioning of the system or in its outputs and short- and long-term impacts? | YES:<br><br>**Major circumstantial risk factor**<br><br>• Where the sector(s) or domain(s) in which the AI system will operate, and from which the data used to train it are drawn, contain historical legacies and patterns of discrimination, inequality, bias, racism, or unfair treatment of minority, marginalized, or vulnerable groups that could be replicated or augmented in the functioning of the system or in its outputs and short- and long-term impacts, this presents a ***major circumstantial risk*** |
| | ☐ YES<br><br>☐ NO<br><br>☐ UNSURE<br><br>☐ NOT APPLICABLE<br><br>***Example 1***<br><br>***Example 2*** | ***factor*** for adverse impacts on the human rights and fundamental freedom of persons.<br><br>**Actions to take for your HUDERIA:**<br><br>• Make sure to focus upon considerations surrounding the fairness, non-discrimination, equality, diversity, and inclusiveness<br><br>**Goals, properties, and areas to focus on in your HUDERAC**:<br><br>• Fairness (non-discrimination, equality, bias mitigation, diversity, and inclusiveness). Continued emphasis should be placed on data representativeness and bias mitigation across design, development, and deployment processes. |
| | | NO: No message |
| | | UNSURE: Before taking any further steps in the proposed project, you should determine, through expert and stakeholder input where appropriate, whether the sector(s) or domain(s) in which the AI system will operate, and from which the data used to train it are drawn, contain historical legacies and patterns of discrimination, inequality, bias, racism, or unfair treatment of minority, marginalized, or vulnerable groups that could be replicated or augmented in the functioning of the system or in its outputs and |
| | | short- and long-term impacts. When this information is ascertained, you should return to your PCRA and revise it accordingly. |
| | | NOT APPLICABLE: No message |
| **Environmental context** | 17) If the design, development, and deployment of the AI system will have potentially significant impacts on the environment, will sufficient and transparently reported processes be implemented throughout the project's lifecycle to ensure that the system, in both its production and use, complies with applicable environmental protection standards and supports the sustainability of the planet?<br><br>☐ YES, WE HAVE PLANS IN PLACE TO DO THIS<br><br>☐ WE HAD NOT CONSIDERED IT, BUT MAY DO THIS<br><br>☐ WE HAD NOT CONSIDERED IT, BUT ARE UNLIKELY TO DO THIS<br><br>☐ NO, WE ARE NOT PLANNING TO DO THIS<br><br>☐ UNSURE | YES, WE HAVE PLANS IN PLACE TO DO THIS: No message |
| | | WE HAD NOT CONSIDERED IT, BUT MAY DO THIS / WE HAD NOT CONSIDERED IT, BUT ARE UNLIKELY TO DO THIS / NO, WE ARE NOT PLANNING TO DO THIS:<br><br>**Major modifiable risk factor**<br><br>• Where the design, development, and deployment of the AI system will have potentially significant impacts on the environment and sufficient and transparently reported processes are not implemented throughout the project's lifecycle to ensure that the system, in both its production and use, complies with applicable environmental protection standards and supports the sustainability of the planet, this presents a ***major modifiable risk factor*** for adverse impacts on the human rights and fundamental freedom of persons.<br><br>**Actions to take for your HUDERIA:** |
| | ☐ NOT APPLICABLE<br><br>***Example*** | • Make sure to focus upon considerations surrounding the impacts of the AI system on the biosphere and planetary health. You should also explore in your HUDERIA, which communities may be adversely impacted by environmental harms—paying close attention to effects on vulnerable and marginalized groups.<br><br>**Goals, properties, and areas to focus on in your HUDERAC**:<br><br>• Sustainability (reflection on context and impacts and stakeholder engagement and involvement) |
| | | UNSURE: Before taking any further steps in the proposed project, you should determine, through expert and stakeholder input where appropriate, how to incorporate sufficient and transparently reported processes to ensure that the system, in both its production and use, does not adversely impact the environment or harm the biosphere and that the system complies with applicable environmental protection standards. When this information is ascertained, you should return to your PCRA and revise it accordingly. |
| | | NOT APPLICABLE: No message |
| **Cybersecurity context** | 18) Could the AI system present motivations or opportunities for malicious parties to hack or corrupt it to achieve substantial financial | YES:<br><br>**Moderate modifiable risk factor** |
| | gains, political goals, or other perceived benefits?<br><br>☐ YES<br><br>☐ NO<br><br>☐ UNSURE<br><br>☐ NOT APPLICABLE<br><br>***Example*** | • Where AI systems provide motivations or opportunities for malicious parties to hack or corrupt them to achieve substantial financial gains, political goals, or other perceived benefits, this presents a ***moderate modifiable risk factor*** for adverse impacts on the human rights and fundamental freedom of persons.<br><br>**Actions to take for your HUDERIA:**<br><br>No trigger message<br><br>**Goals, properties, and areas to focus on in your HUDERAC**:<br><br>• Safety (security and robustness) |
| | | NO: No message |
| | | UNSURE: Before taking any further steps in the proposed project, you should determine, through expert and stakeholder input where appropriate, whether the AI system present motivations or opportunities for malicious parties to hack or corrupt it to achieve substantial financial gains, political goals, or other perceived benefits. When this information is ascertained, you should return to your PCRA and revise it accordingly. |
| | | NOT APPLICABLE: No message |
| | 19) Will sufficient and transparently reported processes be | YES, WE HAVE PLANS IN PLACE TO DO THIS: No message |
| | implemented throughout the project's lifecycle to ensure that measures put in place to safeguard the system's safety, security, and robustness are appropriately proportional to potential risks of hacking, adversarial attack, data poisoning, model inversion, or other cybersecurity threats?<br><br>☐ YES, WE HAVE PLANS IN PLACE TO DO THIS<br><br>☐ WE HAD NOT CONSIDERED IT, BUT MAY DO THIS<br><br>☐ WE HAD NOT CONSIDERED IT, BUT ARE UNLIKELY TO DO THIS<br><br>☐ NO, WE ARE NOT PLANNING TO DO THIS<br><br>☐ UNSURE<br><br>☐ NOT APPLICABLE<br><br>***Example*** | |
| | | WE HAD NOT CONSIDERED IT, BUT MAY DO THIS / WE HAD NOT CONSIDERED IT, BUT ARE UNLIKELY TO DO THIS / NO, WE ARE NOT PLANNING TO DO THIS:<br><br>**Major modifiable risk factor**<br><br>• Where sufficient and transparently reported processes are not implemented throughout project lifecycles to ensure that measures put in place to safeguard AI systems' safety, security, and robustness are appropriately proportional to potential risks of hacking, adversarial attack, data poisoning, model inversion, or other cybersecurity threats, this presents a ***major modifiable risk factor*** for adverse impacts on the human rights and fundamental freedom of persons.<br><br>**Actions to take for your HUDERIA:**<br><br>No trigger message<br><br>**Goals, properties, and areas to focus on in your HUDERAC**:<br><br>• Safety (security and robustness) |
| | | UNSURE: Before taking any further steps in the proposed project, you should determine, through expert and stakeholder input where appropriate, how to incorporate sufficient and transparently reported processes throughout the project's lifecycle to ensure that measures put in place to |
| | | safeguard the system's safety, security, and robustness are appropriately proportional to potential risks of hacking, adversarial attack, data poisoning, model inversion, or other cybersecurity threats. When this information is ascertained, you should return to your PCRA and revise it accordingly. |
| | | NOT APPLICABLE: No message |
| | 20) Will sufficient and transparently reported processes be implemented throughout the project's lifecycle to stress test the AI system for cybersecurity vulnerabilities and resilience?<br><br>☐ YES, WE HAVE PLANS IN PLACE TO DO THIS<br><br>☐ WE HAD NOT CONSIDERED IT, BUT MAY DO THIS<br><br>☐ WE HAD NOT CONSIDERED IT, BUT ARE UNLIKELY TO DO THIS<br><br>☐ NO, WE ARE NOT PLANNING TO DO THIS<br><br>☐ UNSURE<br><br>☐ NOT APPLICABLE | YES, WE HAVE PLANS IN PLACE TO DO THIS: No message |
| | | WE HAD NOT CONSIDERED IT, BUT MAY DO THIS / WE HAD NOT CONSIDERED IT, BUT ARE UNLIKELY TO DO THIS / NO, WE ARE NOT PLANNING TO DO THIS:<br><br>**Moderate modifiable risk factor**<br><br>• Where sufficient and transparently reported processes are not implemented throughout project lifecycles to stress test AI systems for cybersecurity vulnerabilities and resilience, this presents a ***moderate modifiable risk factor*** for adverse impacts on the human rights and fundamental freedom of persons.<br><br>**Actions to take for your HUDERIA:**<br><br>No trigger message |
| | ***Example*** | **Goals, properties, and areas to focus on in your HUDERAC**:<br><br>• Safety (security and robustness) |
| | | UNSURE: Before taking any further steps in the proposed project, you should determine, through expert and stakeholder input where appropriate, how to incorporate sufficient and transparently reported processes throughout the project's lifecycle to stress test the AI system for cybersecurity vulnerabilities and resilience. When this information is ascertained, you should return to your PCRA and revise it accordingly. |
| | | NOT APPLICABLE: No message |
| **Data Lifecycle Context** | | |
| **Data quality, integrity, and provenance** | 21) Will sufficient and transparently reported processes be implemented throughout the project's lifecycle to ensure that all data used in producing the system are sufficiently balanced and representative of the individual rights-holders and groups it is affecting?<br><br>☐ YES, WE HAVE PLANS IN PLACE TO DO THIS | YES, WE HAVE PLANS IN PLACE TO DO THIS: No message |
| | | WE HAD NOT CONSIDERED IT, BUT MAY DO THIS / WE HAD NOT CONSIDERED IT, BUT ARE UNLIKELY TO DO THIS / NO, WE ARE NOT PLANNING TO DO THIS:<br><br>**Moderate modifiable risk factor**<br><br>• Where sufficient and transparently reported processes are not implemented throughout the project's lifecycle to ensure that all data used in producing the |
| | ☐ WE HAD NOT CONSIDERED IT, BUT MAY DO THIS<br><br>☐ WE HAD NOT CONSIDERED IT, BUT ARE UNLIKELY TO DO THIS<br><br>☐ NO, WE ARE NOT PLANNING TO DO THIS<br><br>☐ UNSURE<br><br>***Example*** | system are sufficiently balanced and representative of the individual rights-holders and groups it is affecting, this presents a ***moderate modifiable risk factor*** for adverse impacts on the human rights and fundamental freedom of persons.<br><br>**Actions to take for your HUDERIA:**<br><br>No trigger message<br><br>**Goals, properties, and areas to focus on in your HUDERAC**:<br><br>• Data quality (dataset balance and representativeness) and fairness (non-discrimination, and bias-mitigation) |
| | | UNSURE: Before taking any further steps in the proposed project, you should determine, through expert and stakeholder input where appropriate, how to incorporate into the design and development of your AI system the inclusion of datasets, which are sufficiently balanced and representative of the individual rights-holders and groups they are affecting. You should also determine how to demonstrate this in your assurance case. When this information is ascertained, you should return to your PCRA and revise it accordingly. |
| | 22) Will sufficient and transparently reported processes be implemented throughout the | YES, WE HAVE PLANS IN PLACE TO DO THIS: No message |
| | project's lifecycle to ensure that all data used in producing the system are accurate, reliable, relevant, appropriate, up-to-date, and of adequate quantity and quality for the use case, domain, function, and purpose of the system?<br><br>☐ YES, WE HAVE PLANS IN PLACE TO DO THIS<br><br>☐ WE HAD NOT CONSIDERED IT, BUT MAY DO THIS<br><br>☐ WE HAD NOT CONSIDERED IT, BUT ARE UNLIKELY TO DO THIS<br><br>☐ NO, WE ARE NOT PLANNING TO DO THIS<br><br>☐ UNSURE<br><br>***Example*** | WE HAD NOT CONSIDERED IT, BUT MAY DO THIS / WE HAD NOT CONSIDERED IT, BUT ARE UNLIKELY TO DO THIS / NO, WE ARE NOT PLANNING TO DO THIS:<br><br>**Moderate modifiable risk factor**<br><br>• Where sufficient and transparently reported processes are not implemented throughout the project's lifecycle to ensure that all data used in producing the system are accurate, reliable, relevant, appropriate, up-to-date, and of adequate quantity and quality for the use case, domain, function, and purpose of the system, this presents a ***moderate modifiable risk factor*** for adverse impacts on the human rights and fundamental freedom of persons.<br><br>**Actions to take for your HUDERIA:**<br><br>No trigger message<br><br>**Goals, properties, and areas to focus on in your HUDERAC**:<br><br>• Data quality (measurement accuracy and source integrity, data reliability, data relevance and appropriateness, data timeliness and recency, adequacy of data quantity and quality for the use case, domain, function, and purpose of the system, and responsible data management) |
| | | UNSURE: Before taking any further steps in the proposed project, you should determine, through expert and stakeholder input where appropriate, how to incorporate into the design and development of your AI system the inclusion of data, which are accurate, reliable, relevant, up-to-date, appropriate, and of adequate quantity and quality for the use case, domain, function, and purpose of the system. You should also determine how to demonstrate this in your assurance case. When this information is ascertained, you should return to your PCRA and revise it accordingly. |
| | 23) Will sufficient and transparently reported processes be implemented throughout the project's lifecycle to ensure that all data used in producing the system are attributable, consistent, complete, and contemporaneous with collection?<br><br>☐ YES, WE HAVE PLANS IN PLACE TO DO THIS<br><br>☐ WE HAD NOT CONSIDERED IT, BUT MAY DO THIS<br><br>☐ WE HAD NOT CONSIDERED IT, BUT ARE UNLIKELY TO DO THIS<br><br>☐ NO, WE ARE NOT PLANNING TO DO THIS | YES, WE HAVE PLANS IN PLACE TO DO THIS: No message |
| | | WE HAD NOT CONSIDERED IT, BUT MAY DO THIS / WE HAD NOT CONSIDERED IT, BUT ARE UNLIKELY TO DO THIS / NO, WE ARE NOT PLANNING TO DO THIS:<br><br>**Moderate modifiable risk factor**<br><br>• Where sufficient and transparently reported processes are not implemented throughout the project's lifecycle to ensure that all data used in producing the system are attributable, consistent, complete, and contemporaneous with collection, this presents a ***moderate modifiable risk factor*** for adverse impacts on the human rights and fundamental freedom of persons.<br><br>**Actions to take for your HUDERIA:** |
| | ☐ UNSURE<br><br>***Example*** | No trigger message<br><br>**Goals, properties, and areas to focus on in your HUDERAC**:<br><br>• Data integrity (data attributability, data consistency, data completeness, data contemporaneous, and responsible data management) |
| | | UNSURE: Before taking any further steps in the proposed project, you should determine, through expert and stakeholder input where appropriate, how to incorporate into the design and development of your AI system the inclusion of data, which are attributable, consistent, complete, and contemporaneous with collection. You should also determine how to demonstrate this in your assurance case. When this information is ascertained, you should return to your PCRA and revise it accordingly. |
| | 24) Will sufficient and transparently reported processes be implemented throughout the project's lifecycle to ensure the proper recording, traceability, and auditability of the provenance and lineage of all data used in producing the system, and any other data involved in the dynamic learning, tuning, or re-training of the system across its lifecycle? | YES, WE HAVE PLANS IN PLACE TO DO THIS: No message |
| | | WE HAD NOT CONSIDERED IT, BUT MAY DO THIS / WE HAD NOT CONSIDERED IT, BUT ARE UNLIKELY TO DO THIS / NO, WE ARE NOT PLANNING TO DO THIS:<br><br>**Moderate modifiable risk factor** |
| | ☐ YES, WE HAVE PLANS IN PLACE TO DO THIS<br><br>☐ WE HAD NOT CONSIDERED IT, BUT MAY DO THIS<br><br>☐ WE HAD NOT CONSIDERED IT, BUT ARE UNLIKELY TO DO THIS<br><br>☐ NO, WE ARE NOT PLANNING TO DO THIS<br><br>☐ UNSURE<br><br>***Example*** | • Where sufficient and transparently reported processes are not implemented throughout the project's lifecycle to ensure the proper recording, traceability, and auditability of the provenance and lineage of all data used in producing the system, and any other data involved in the dynamic learning, tuning, or re-training of the system across its lifecycle, , this presents a ***moderate modifiable risk factor*** for adverse impacts on the human rights and fundamental freedom of persons.<br><br>**Actions to take for your HUDERIA:**<br><br>No trigger message<br><br>**Goals, properties, and areas to focus on in your HUDERAC**:<br><br>• Accountability and process transparency, (accessibility, clear data provenance and lineage) and data integrity (responsible data management) |
| | | UNSURE: Before taking any further steps in the proposed project, you should determine, through expert and stakeholder input where appropriate, how to incorporate into the design and development of your AI system the inclusion mechanisms and processes to ensure the proper recording, traceability, and auditability of the provenance and lineage of all data used to train, test, and validate the system, and any other data |
| | | involved in the dynamic learning, tuning, or re-training of the system across its lifecycle. You should also determine how to demonstrate this in your assurance case. When this information is ascertained, you should return to your PCRA and revise it accordingly. |
| **Means and methods of data collection** | 25) Where there is human involvement in the data lifecycle, will transparent and publicly accessible measures be implemented to ensure mitigation of potential measurement errors or biases in collection, measurement, and recording processes?<br><br>☐ YES, WE HAVE PLANS IN PLACE TO DO THIS<br><br>☐ WE HAD NOT CONSIDERED IT, BUT MAY DO THIS<br><br>☐ WE HAD NOT CONSIDERED IT, BUT ARE UNLIKELY TO DO THIS<br><br>☐ NO, WE ARE NOT PLANNING TO DO THIS<br><br>☐ UNSURE<br><br>☐ NOT APPLICABLE<br><br>***Example*** | YES, WE HAVE PLANS IN PLACE TO DO THIS: No message |
| | | WE HAD NOT CONSIDERED IT, BUT MAY DO THIS / WE HAD NOT CONSIDERED IT, BUT ARE UNLIKELY TO DO THIS / NO, WE ARE NOT PLANNING TO DO THIS:<br><br>**Moderate modifiable risk factor**<br><br>• Where transparent and publicly accessible measures are not implemented to ensure mitigation of potential measurement errors or biases in collection, measurement, and recording processes, where there is human involvement in the data lifecycle, , this presents a ***moderate modifiable risk factor*** for adverse impacts on the human rights and fundamental freedom of persons.<br><br>**Actions to take for your HUDERIA:**<br><br>No trigger message |
| | | **Goals, properties, and areas to focus on in your HUDERAC**:<br><br>• Data quality, measurement accuracy and source integrity, and responsible data management |
| | | UNSURE: Before taking any further steps in the proposed project, you should determine, through expert and stakeholder input where appropriate, how to incorporate into the design and development of your AI system transparent and publicly accessible measures to ensure mitigation of the potential for measurement errors or biases in collection, measurement, and recording processes where there is human involvement in data collection. You should also determine how to demonstrate this in your assurance case. When this information is ascertained, you should return to your PCRA and revise it accordingly. |
| | | Not Applicable: No message |
| | 26) In the event that collected or procured datasets have missing or unusable data, will the methods used for addressing these deficiencies be transparent and made accessible to relevant stakeholders? | YES, WE HAVE PLANS IN PLACE TO DO THIS: No message |
| | | WE HAD NOT CONSIDERED IT, BUT MAY DO THIS / WE HAD NOT CONSIDERED IT, BUT ARE UNLIKELY TO DO THIS / NO, WE ARE NOT PLANNING TO DO THIS:<br><br>**Moderate modifiable risk factor** |
| | ☐ YES, WE HAVE PLANS IN PLACE TO DO THIS<br><br>☐ WE HAD NOT CONSIDERED IT, BUT MAY DO THIS<br><br>☐ WE HAD NOT CONSIDERED IT, BUT ARE UNLIKELY TO DO THIS<br><br>☐ NO, WE ARE NOT PLANNING TO DO THIS<br><br>☐ UNSURE<br><br>☐ NOT APPLICABLE<br><br>***Example*** | • Where the methods used for addressing missing or unusable data in collected or procured datasets is not transparent and made accessible to relevant stakeholders, this presents a ***moderate modifiable risk factor*** for adverse impacts on the human rights and fundamental freedom of persons.<br><br>**Actions to take for your HUDERIA:**<br><br>No trigger message<br><br>**Goals, properties, and areas to focus on in your HUDERAC**:<br><br>• Data quality (data appropriateness), data integrity (data completeness, and responsible data management), and accountability and process transparency (accessibility, and clear data provenance and lineage) |
| | | UNSURE: Before taking any further steps in the proposed project, you should determine, through expert and stakeholder input where appropriate, the methods and measure you will put into place to address issues arising from collected or procured datasets, which have missing or unusable data. You should also determine how to demonstrate this in your assurance case. When this information is ascertained, you should return to your PCRA and revise it accordingly. |
| | | NOT APPLICABLE: No message |
| | 27) Where personal data are used in the production of the AI system, will information be made available to impacted rights-holders and other relevant stakeholders about the consent or legitimate basis to use that data for the purpose of the system?<br><br>☐ YES, WE HAVE PLANS IN PLACE TO DO THIS<br><br>☐ WE HAD NOT CONSIDERED IT, BUT MAY DO THIS<br><br>☐ WE HAD NOT CONSIDERED IT, BUT ARE UNLIKELY TO DO THIS<br><br>☐ NO, WE ARE NOT PLANNING TO DO THIS<br><br>☐ UNSURE<br><br>☐ NOT APPLICABLE<br><br>***Example*** | YES, WE HAVE PLANS IN PLACE TO DO THIS: No message |
| | | WE HAD NOT CONSIDERED IT, BUT MAY DO THIS / WE HAD NOT CONSIDERED IT, BUT ARE UNLIKELY TO DO THIS / NO, WE ARE NOT PLANNING TO DO THIS:<br><br>**Moderate modifiable risk factor**<br><br>• Where information is not made available to impacted rights-holders and other relevant stakeholders about the consent or legitimate basis to use personal data for the purpose of the system, this presents a ***moderate modifiable risk factor*** for adverse impacts on the human rights and fundamental freedom of persons.<br><br>**Actions to take for your HUDERIA:**<br><br>No trigger message<br><br>**Goals, properties, and areas to focus on in your HUDERAC**:<br><br>• Data protection and privacy (consent and accountability) and process transparency |
| | | (accessibility, and clear data provenance and lineage) |
| | | UNSURE: Before taking any further steps in the proposed project, you should determine, through expert and stakeholder input where appropriate, how to make information available to impacted rights-holders and other relevant stakeholders about the consent or legitimate basis to use personal data for the purpose of the system. When this information is ascertained, you should return to your PCRA and revise it accordingly. |
| | | NOT APPLICABLE: No message |
| | 28) If consent or the legitimate basis to use personal data is implied, will rights-holders and other relevant stakeholders be consulted to identify acceptability of the data use or concerns that need to be addressed?<br><br>☐ YES, WE HAVE PLANS IN PLACE TO DO THIS<br><br>☐ WE HAD NOT CONSIDERED IT, BUT MAY DO THIS<br><br>☐ WE HAD NOT CONSIDERED IT, BUT ARE UNLIKELY TO DO THIS | YES, WE HAVE PLANS IN PLACE TO DO THIS: No message |
| | | WE HAD NOT CONSIDERED IT, BUT MAY DO THIS / WE HAD NOT CONSIDERED IT, BUT ARE UNLIKELY TO DO THIS / NO, WE ARE NOT PLANNING TO DO THIS:<br><br>**Moderate modifiable risk factor**<br><br>• Where rights-holders and other relevant stakeholders are not consulted to identify acceptability of the data use or concerns that need to be addressed when consent to use personal data is implied, this presents a ***moderate modifiable risk factor*** for adverse impacts on the human rights and fundamental freedom of persons. |
| | ☐ NO, WE ARE NOT PLANNING TO DO THIS<br><br>☐ UNSURE<br><br>☐ NOT APPLICABLE<br><br>***Example*** | **Actions to take for your HUDERIA:**<br><br>• Make sure to incorporate considerations surrounding the acceptability of the personal data use into your impact assessment, integrating affected rights-holders into relevant deliberations through your Stakeholder Engagement Process where appropriate.<br><br>**Goals, properties, and areas to focus on in your HUDERAC**:<br><br>• Sustainability (reflection on context and impacts, stakeholder engagement and involvement) |
| | | UNSURE: Before taking any further steps in the proposed project, you should determine, through expert and stakeholder input where appropriate, whether your project involves the use of personal data based on implied consent and, if so, how you can consult impacted rights-holders and other relevant stakeholders to identify acceptability of the data use or concerns that need to be addressed. When this information is ascertained, you should return to your PCRA and revise it accordingly. |
| | | NOT APPLICABLE: No message |
| **Data types** | 29) Will the AI system use dynamic data, collected and processed in | YES:<br><br>**Major modifiable risk factor** |
| | real time (or near real time), for continuous learning?<br><br>☐ YES<br>☐ NO<br>☐ UNSURE<br>☐ NOT APPLICABLE<br><br>***Example*** | • Where the AI system uses dynamic data, collected and processed in real time (or near real time), for continuous learning, this presents a ***major modifiable risk factor*** for adverse impacts on the human rights and fundamental freedom of persons.<br><br>**Actions to take for your HUDERIA:**<br><br>No trigger message<br><br>**Goals, properties, and areas to focus on in your HUDERAC**:<br><br>• Safety (security, reliability, and robustness), data quality, and data integrity as well as, where appropriate, non-discrimination and bias mitigation. You should also transparently report your use of dynamic data, and the measures you are taking to manage the risks surrounding them, to impacted rights-holders and relevant stakeholders. |
| | | NO: No message |
| | | UNSURE: Before taking any further steps in the proposed project, you should determine, through expert and stakeholder input where appropriate, whether your project involves the use of use dynamic data, collected and processed in real time (or near real time), for continuous learning and, if so, how to manage the risks surrounding this. You should also determine how to |
| | | demonstrate such risk management measures in the relevant parts of your assurance case—in particular, those involving the assurance of goals and properties of safety, security, reliability, robustness, data quality, and data integrity as well as, where appropriate, non-discrimination and bias mitigation. When this information is ascertained, you should return to your PCRA and revise it accordingly. |
| | | NOT APPLICABLE: No message |
| | 30) Do the domain in which the data are collected or procured, and the type of the data collected or procured, pose risks of rapid or unexpected distributional shifts or drifts that could adversely impact the accuracy and performance of the system?<br><br>☐ YES<br>☐ NO<br>☐ UNSURE<br>☐ NOT APPLICABLE<br><br>***Example*** | YES:<br><br>**Moderate circumstantial risk factor**<br><br>• Where the domain in which the data are collected or procured, and the type of the data collected or procured, pose risks of rapid or unexpected distributional shifts or drifts that could adversely impact the accuracy and performance of the system, this presents a ***moderate modifiable risk factor*** for adverse impacts on the human rights and fundamental freedom of persons.<br><br>**Actions to take for your HUDERIA:**<br><br>No trigger message<br><br>**Goals, properties, and areas to focus on in your HUDERAC**: |
| | | • Sustainability (change monitoring), safety (reliability and robustness), data quality, and data integrity as well as, where appropriate, non-discrimination and bias mitigation. You should also involve domain experts to determine the potential sources of distributional shifts or drifts and build processes of dynamic assessment, re-assessment, external validation, and monitoring into your project lifecycle. |
| | | NO: No message |
| | | UNSURE: Before taking any further steps in the proposed project, you should determine, through expert and stakeholder input where appropriate, whether the collected or procured data you will be using are subject to rapid or unexpected distributional shifts or drifts and, if so, how to manage the risks surrounding this. You should also determine how to demonstrate such risk management measures in the relevant parts of your assurance case—in particular, those involving the assurance of goals of safety, security, reliability, robustness, data quality, and data integrity as well as, where appropriate, non-discrimination and bias mitigation. In addition to this, you should also involve domain experts to determine the potential sources of distributional shifts or drifts and build processes of dynamic assessment, re-assessment, external validation, and monitoring into your project lifecycle. When this information is ascertained, you should return to your PCRA and revise it accordingly. |
| | | NOT APPLICABLE: No message |
| | 31) If the AI system will use unstructured data or a combination of structured and unstructured data, will the project lifecycle incorporate mechanisms and processes to ensure that the inferences generated from that data by the system are reasonable, fair, and do not contain lurking proxies or correlations that are discriminatory or inequitable?<br><br>☐ YES, WE HAVE PLANS IN PLACE TO DO THIS<br><br>☐ WE HAD NOT CONSIDERED IT, BUT MAY DO THIS<br><br>☐ WE HAD NOT CONSIDERED IT, BUT ARE UNLIKELY TO DO THIS<br><br>☐ NO, WE ARE NOT PLANNING TO DO THIS<br><br>☐ UNSURE<br><br>☐ NOT APPLICABLE<br><br>***Example*** | YES, WE HAVE PLANS IN PLACE TO DO THIS: No message |
| | | WE HAD NOT CONSIDERED IT, BUT MAY DO THIS / WE HAD NOT CONSIDERED IT, BUT ARE UNLIKELY TO DO THIS / NO, WE ARE NOT PLANNING TO DO THIS:<br><br>**Moderate modifiable risk factor**<br><br>• Where the project lifecycle does not incorporate mechanisms and processes to ensure that the inferences generated from the use of unstructured data, or a combination of structured and unstructured data, are reasonable, fair, and do not contain lurking proxies or correlations that are discriminatory or inequitable, this presents a ***moderate modifiable risk factor*** for adverse impacts on the human rights and fundamental freedom of persons.<br><br>**Actions to take for your HUDERIA:**<br><br>No trigger message<br><br>**Goals, properties, and areas to focus on in your HUDERAC**: |
| | | • Fairness (non-discrimination, and bias mitigation). You should also involve domain experts and social scientists to determine the potential sources of the lurking discriminatory proxies or correlations that may influence the outputs of the system. |
| | | UNSURE: Before taking any further steps in the proposed project, you should determine, through expert and stakeholder input where appropriate, whether your AI system will use unstructured data or a combination of structured and unstructured data and, if so, how to manage the risks that these data may generate inferences that are unreasonable, inequitable, or contain lurking discriminatory proxies or correlations. When this information is ascertained, you should return to your PCRA and revise it accordingly. |
| | | NOT APPLICABLE: No message |
| **Dataset linkage** | 32) Is there a possibility of deanonymizing or identifying rights-holders through data linkage with existing data, publicly available datasets, or data that could be easily obtained?<br><br>☐ YES<br><br>☐ NO<br><br>☐ UNSURE | YES:<br><br>**Major circumstantial risk factor**<br><br>• Where there is a possibility of deanonymizing or identifying rights-holders through data linkage with existing data, publicly available datasets, or data that could be easily obtained, this presents a ***major circumstantial risk factor*** for adverse impacts on the human rights and fundamental freedom of persons. |
| | ☐ NOT APPLICABLE<br><br>***Example*** | **Actions to take for your HUDERIA:**<br><br>No trigger message<br><br>**Goals, properties, and areas to focus on in your HUDERAC**:<br><br>• Data protection and privacy (data security). You should also involve experts to determine the potential sources of deanonymisation or identification through data linkage. |
| | | NO: No message |
| | | UNSURE: Before taking any further steps in the proposed project, you should determine, through expert and stakeholder input where appropriate, whether your AI system may deanonymize or identify rights-holders through data linkage with existing data, publicly available datasets, or data that could be easily obtained and, if so, how to manage the risks of this potential deanonymization or identification. When this information is ascertained, you should return to your PCRA and revise it accordingly. |
| | | NOT APPLICABLE: No message |
| **Data labelling and annotating practices** | 33) Will processes of labelling and annotating the data used to produce the AI system be | YES, WE HAVE PLANS IN PLACE TO DO THIS: No message |
| | transparently reported and made accessible for audit, oversight, and review by appropriate authorities and relevant stakeholders?<br><br>☐ YES, WE HAVE PLANS IN PLACE TO DO THIS<br><br>☐ WE HAD NOT CONSIDERED IT, BUT MAY DO THIS<br><br>☐ WE HAD NOT CONSIDERED IT, BUT ARE UNLIKELY TO DO THIS<br><br>☐ NO, WE ARE NOT PLANNING TO DO THIS<br><br>☐ UNSURE<br><br>☐ NOT APPLICABLE<br><br>***Example*** | WE HAD NOT CONSIDERED IT, BUT MAY DO THIS / WE HAD NOT CONSIDERED IT, BUT ARE UNLIKELY TO DO THIS / NO, WE ARE NOT PLANNING TO DO THIS:<br><br>**Moderate modifiable risk factor**<br><br>• Where processes of labelling and annotating the data used to produce the AI system are not transparently reported and made accessible for audit, oversight, and review by appropriate authorities and relevant stakeholders, this presents a ***moderate modifiable risk factor*** for adverse impacts on the human rights and fundamental freedom of persons.<br><br>**Actions to take for your HUDERIA:**<br><br>No trigger message<br><br>**Goals, properties, and areas to focus on in your HUDERAC**:<br><br>• Accountability and process transparency (traceability, auditability, accessibility, and responsible governance) |
| | | UNSURE: Before taking any further steps in the proposed project, you should determine, through expert and stakeholder input where appropriate, how to make the processes of labelling and annotating the data that will be used to produce your AI system transparent and accessible for audit, oversight, and review by appropriate |
| | | authorities and relevant stakeholders. When this information is ascertained, you should return to your PCRA and revise it accordingly. |
| | | NOT APPLICABLE: No message |
| | 34) Where human labellers and annotators are involved, will sufficient and transparently reported processes be put into place to mitigate potential labelling or annotation biases, especially in cases where these activities concern social and demographic categories that can import patterns of discrimination and proxies for protected characteristics?<br><br>☐ YES, WE HAVE PLANS IN PLACE TO DO THIS<br><br>☐ WE HAD NOT CONSIDERED IT, BUT MAY DO THIS<br><br>☐ WE HAD NOT CONSIDERED IT, BUT ARE UNLIKELY TO DO THIS<br><br>☐ NO, WE ARE NOT PLANNING TO DO THIS<br><br>☐ UNSURE<br><br>☐ NOT APPLICABLE | YES, WE HAVE PLANS IN PLACE TO DO THIS: No message |
| | | WE HAD NOT CONSIDERED IT, BUT MAY DO THIS / WE HAD NOT CONSIDERED IT, BUT ARE UNLIKELY TO DO THIS / NO, WE ARE NOT PLANNING TO DO THIS:<br><br>**Moderate modifiable risk factor**<br><br>• Where sufficient and transparently reported processes are not put into place to mitigate potential labelling or annotation biases where human labellers and annotators are involved, especially in cases where these activities concern social and demographic categories that can import patterns of discrimination and proxies for protected characteristics, this presents a ***moderate modifiable risk factor*** for adverse impacts on the human rights and fundamental freedom of persons.<br><br>**Actions to take for your HUDERIA:**<br><br>No trigger message |
| | ***Example*** | **Goals, properties, and areas to focus on in your HUDERAC**:<br><br>• Fairness (non-discrimination and bias mitigation), accountability and process transparency (accessibility, traceability, auditability, and responsible governance) |
| | | UNSURE: Before taking any further steps in the proposed project, you should determine, through expert and stakeholder input where appropriate, how to mitigate any potential labelling or annotation biases that may arise in the production of your AI system. When this information is ascertained, you should return to your PCRA and revise it accordingly. |
| | | NOT APPLICABLE: No message |
| | 35) If data labelling or annotation is partly or fully automated, will sufficient and transparently reported processes of human oversight be implemented to mitigate the negative impact of biases generated by automated labelling or annotation, especially in cases where the dataset includes social and demographic categories that can import patterns of discrimination and proxies for protected characteristics? | Yes: No message |
| | | WE HAD NOT CONSIDERED IT, BUT MAY DO THIS / WE HAD NOT CONSIDERED IT, BUT ARE UNLIKELY TO DO THIS / NO, WE ARE NOT PLANNING TO DO THIS:<br><br>**Moderate modifiable risk factor**<br><br>• Where sufficient and transparently reported processes of human oversight are not implemented to mitigate the negative impact of biases generated by automated labelling or annotation, |
| | ☐ YES, WE HAVE PLANS IN PLACE TO DO THIS<br><br>☐ WE HAD NOT CONSIDERED IT, BUT MAY DO THIS<br><br>☐ WE HAD NOT CONSIDERED IT, BUT ARE UNLIKELY TO DO THIS<br><br>☐ NO, WE ARE NOT PLANNING TO DO THIS<br><br>☐ UNSURE<br><br>☐ NOT APPLICABLE<br><br>***Example*** | especially in cases where the dataset includes social and demographic categories that can import patterns of discrimination and proxies for protected characteristics, this presents a ***moderate modifiable risk factor*** for adverse impacts on the human rights and fundamental freedom of persons.<br><br>**Actions to take for your HUDERIA:**<br><br>No trigger message<br><br>**Goals, properties, and areas to focus on in your HUDERAC**:<br><br>• Fairness (non-discrimination, bias mitigation), accountability and process transparency (traceability, auditability, accessibility, and responsible governance) |
| | | UNSURE: Before taking any further steps in the proposed project, you should determine, through expert and stakeholder input where appropriate, how to mitigate any potential labelling or annotation biases generated by automated data labelling or annotation. When this information is ascertained, you should return to your PCRA and revise it accordingly. |
| | | NOT APPLICABLE: No message |
| **Goal Setting and Problem Formulation Context** | | |
| **Decision to design** | 36) Will an evaluation be carried out as to whether building the AI system is the right approach given available resources and data, existing technologies and processes, the complexity of the use-contexts involved, and the nature of the policy or social problem that needs to be solved?<br><br>☐ YES, WE HAVE PLANS IN PLACE TO DO THIS<br><br>☐ WE HAD NOT CONSIDERED IT, BUT MAY DO THIS<br><br>☐ WE HAD NOT CONSIDERED IT, BUT ARE UNLIKELY TO DO THIS<br><br>☐ NO, WE ARE NOT PLANNING TO DO THIS<br><br>☐ UNSURE<br><br>***Example*** | YES, WE HAVE PLANS IN PLACE TO DO THIS: No message |
| | | WE HAD NOT CONSIDERED IT, BUT MAY DO THIS / WE HAD NOT CONSIDERED IT, BUT ARE UNLIKELY TO DO THIS / NO, WE ARE NOT PLANNING TO DO THIS:<br><br>**Moderate modifiable risk factor**<br><br>• Where an evaluation is not carried out as to whether building the AI system is the right approach given available resources and data, existing technologies and processes, the complexity of the use-contexts involved, and the nature of the policy or social problem that needs to be solved, this presents a ***moderate modifiable risk factor*** for adverse impacts on the human rights and fundamental freedom of persons.<br><br>**Actions to take for your HUDERIA:**<br><br>• You should take steps to ensure that an initial evaluation of this kind takes place and incorporate its results into your HUDERIA as part of your impact assessment process.<br><br>**Goals, properties, and areas to focus on in your HUDERAC**:<br><br>• Sustainability (reflection on context and impacts) |
| | | Unsure: Before taking any further steps in the proposed project, you should evaluate, through expert and stakeholder input where appropriate, whether building the AI system is the right approach given available resources and data, existing technologies and processes, the complexity of the use-contexts involved, and the nature of the policy or social problem that needs to be solved. When this information is ascertained, you should return to your PCRA and revise it accordingly. |
| **Definition of outcome** | 37) Will processes of formulating the problem to be solved by the AI system and of defining its target variable (or measurable proxy) be opened to input from stakeholder engagement and public scrutiny?<br><br>☐ YES, WE HAVE PLANS IN PLACE TO DO THIS<br><br>☐ WE HAD NOT CONSIDERED IT, BUT MAY DO THIS<br><br>☐ WE HAD NOT CONSIDERED IT, BUT ARE UNLIKELY TO DO THIS<br><br>☐ NO, WE ARE NOT PLANNING TO DO THIS<br><br>☐ UNSURE<br><br>***Example*** | YES, WE HAVE PLANS IN PLACE TO DO THIS: No message |
| | | WE HAD NOT CONSIDERED IT, BUT MAY DO THIS / WE HAD NOT CONSIDERED IT, BUT ARE UNLIKELY TO DO THIS / NO, WE ARE NOT PLANNING TO DO THIS:<br><br>**Moderate modifiable risk factor**<br><br>• Where processes of formulating the problem to be solved by the AI system and of defining its target variable (or measurable proxy) are not opened to input from stakeholder engagement and public scrutiny, this presents a ***moderate modifiable risk factor*** for adverse impacts on the human rights and fundamental freedom of persons.<br><br>**Actions to take for your HUDERIA:**<br><br>• You should determine, based on this preliminary risk analysis and your |
| | | stakeholder engagement process, the proportionate level of stakeholder involvement, and seek public input accordingly. This public input should include (1) determining the reasonableness, fairness, equity, and justifiability of the translation of the project's objective into the statistical and mathematical frame and (2) determining the alignment of that translation with the potential impacts of the system on human rights, fundamental freedoms, democracy, and the rule of law. The way you integrate this public input into the system's problem formulation and outcome definition should be incorporated into your HUDERIA as part of your impact assessment process.<br><br>**Goals, properties, and areas to focus on in your HUDERAC**:<br><br>• Sustainability (reflection on context and impacts and stakeholder engagement and involvement) |
| | | UNSURE: Before taking any further steps in the proposed project, you should evaluate the extent to which you need to incorporate public input into your process of ascertaining the AI system's problem formulation and outcome definition. To attain this objective, you should determine, based on this preliminary risk analysis and your stakeholder engagement process, the proportionate level of stakeholder involvement, and seek appropriate public input. When this |
| | | information is ascertained, you should return to your PCRA and revise it accordingly. |
| **Model Design & Development Context** | | |
| **AI model characteristics** | 38) If the algorithmic model(s) or technique(s) used by the AI system have a non-deterministic, probabilistic, evolving, or dynamic character that prevents or hinders the system's intended functionality from being formalized into specific and checkable design-time requirements (or that impairs commonly accepted methods of formal verification and validation), will the system interact with rights-holders in ways that could adversely impact their human rights and fundamental freedoms?<br><br>☐ YES<br>☐ NO<br>☐ UNSURE<br>☐ NOT APPLICABLE<br><br>***Example*** | YES:<br><br>**Major modifiable risk factor**<br><br>• Where the AI system directly interacts with rights-holders in ways that could adversely impact their human rights and fundamental freedoms where the algorithmic model(s) or technique(s) used by the system have a non-deterministic, probabilistic, evolving, or dynamic character that prevents or hinders the system's intended functionality from being formalized into specific and checkable design-time requirements (or that impairs commonly accepted methods of formal verification and validation), this presents a ***major modifiable risk factor*** for adverse impacts on the human rights and fundamental freedom of persons.<br><br>**Actions to take for your HUDERIA:**<br><br>No trigger message<br><br>**Goals, properties, and areas to focus on in your HUDERAC**:<br><br>• Safety (security, reliability, and robustness), data quality, and data |
| | | integrity as well as, where appropriate, fairness (non-discrimination and bias mitigation) |
| | | NO: No message |
| | | UNSURE: Before taking any further steps in the proposed project, you should determine, through expert and stakeholder input where appropriate, (1) whether the algorithmic model(s) or technique(s) used by the AI system will have a non-deterministic, probabilistic, evolving, or dynamic character that prevents or hinders the system's intended functionality from being formalized into specific and checkable design-time requirements (or that impairs commonly accepted methods of formal verification and validation), and (2) whether the system will interact with rights-holders in ways that could adversely impact their their human rights and fundamental freedoms. When this information is ascertained, you should return to your PCRA and revise it accordingly. |
| | | NOT APPLICABLE: No message |
| | 39) If the algorithmic model(s) or technique(s) used by the AI system have a complex, high-dimensional, or non-linear character that impairs or prevents the interpretability and explainability of the system, will the system interact with rights-holders in ways that could adversely impact their human rights and fundamental freedoms?<br><br>☐ YES<br><br>☐ NO<br><br>☐ UNSURE<br><br>☐ NOT APPLICABLE<br><br>***Example*** | YES:<br><br>**Major modifiable risk factor**<br><br>• Where the AI system directly interacts with rights-holders in ways that could adversely impact their human rights and fundamental freedoms where the algorithmic model(s) or technique(s) used by the AI system have a complex, high-dimensional, or non-linear character that impairs or prevents the interpretability and explainability of the system, this presents a ***major modifiable risk factor*** for adverse impacts on the human rights and fundamental freedom of persons.<br><br>**Actions to take for your HUDERIA:**<br><br>No trigger message<br><br>**Goals, properties, and areas to focus on in your HUDERAC**:<br><br>• Safety (security, reliability, and robustness) and explainability as well as, where appropriate, fairness (non-discrimination and bias mitigation) |
| | | NO: No message |
| | | UNSURE: Before taking any further steps in the proposed project, you should determine, through expert and stakeholder input where appropriate, (1) whether the algorithmic model(s) or technique(s) used by the AI system have a complex, high-dimensional, or non-linear character that impairs or prevents the interpretability and explainability of the system, and (2) whether the system will interact with rights-holders in ways that could adversely impact their physical, psychological, or moral integrity or their human rights and fundamental freedoms. When this information is ascertained, you should return to your PCRA and revise it accordingly. |
| | | NOT APPLICABLE: No message |
| **Pre-processing and feature engineering** | 40) Where feature engineering, whether automated or carried out by humans, involves the grouping, disaggregating, or excluding of input features related to protected or potentially sensitive characteristics (e.g. decisions about combining or separating categories of gender or ethnic groups) or proxies for these, will the production of the AI system incorporate processes to mitigate emergent forms of bias and to make the rationale behind these decisions transparent and accessible to impacted rights- | YES, WE HAVE PLANS IN PLACE TO DO THIS: No message |
| | | WE HAD NOT CONSIDERED IT, BUT MAY DO THIS / WE HAD NOT CONSIDERED IT, BUT ARE UNLIKELY TO DO THIS / NO, WE ARE NOT PLANNING TO DO THIS:<br><br>**Moderate modifiable risk factor**<br><br>• Where feature engineering, whether automated or carried out by humans, involves the grouping, disaggregating, or excluding of input features related to protected or potentially sensitive characteristics (e.g. decisions about combining or separating categories of |
| | holders and other relevant stakeholders?<br><br>☐ YES, WE HAVE PLANS IN PLACE TO DO THIS<br><br>☐ WE HAD NOT CONSIDERED IT, BUT MAY DO THIS<br><br>☐ WE HAD NOT CONSIDERED IT, BUT ARE UNLIKELY TO DO THIS<br><br>☐ NO, WE ARE NOT PLANNING TO DO THIS<br><br>☐ UNSURE<br><br>☐ NOT APPLICABLE<br><br>***Example*** | gender or ethnic groups) or proxies for these, the production of the AI system does not incorporate processes to mitigate any emergent forms of bias and to make the rationale behind these decisions transparent and accessible to impacted rights-holders and other relevant stakeholders<br><br>**Actions to take for your HUDERIA:**<br><br>No trigger message<br><br>**Goals, properties, and areas to focus on in your HUDERAC**:<br><br>Fairness (non-discrimination and bias mitigation) and accountability and process transparency (traceability, auditability, accessibility, and responsible governance). In particular, you should ensure that the feature engineering stage of the AI system's lifecycle incorporates processes to mitigate emergent forms of bias and to make the rationale behind these decisions transparent and accessible to impacted rights-holders |
| | | UNSURE: Before taking any further steps in the proposed project, you should determine, through expert and stakeholder input where appropriate, how to incorporate measures to mitigate any emergent forms of bias in the event that the feature engineering stage of an AI system's production, whether automated or carried out by |
| | | humans, involves the grouping, disaggregating, or excluding of input features related to protected or potentially sensitive characteristics or proxies for these. You should also determine how to incorporate and demonstrate such bias mitigation measures in your HUDERAC. When this information is ascertained, you should return to your PCRA and revise it accordingly. |
| | | NOT APPLICABLE: No message |
| **Model selection** | 41) Where complex or potentially opaque models are under consideration, will processes of model selection include appropriate and transparent considerations of the AI system's explainability by taking into account:<br><br>a. The normal expectations of intelligibility and accessibility that accompany the function the system will fulfil in the sector or domain in which it will operate<br>b. The availability of more interpretable algorithmic models or techniques in cases where the selection of an opaque model poses risks to the physical, psychological, or moral integrity of rights-holders or | YES, WE HAVE PLANS IN PLACE TO DO THIS: No message |
| | | WE HAD NOT CONSIDERED IT, BUT MAY DO THIS / WE HAD NOT CONSIDERED IT, BUT ARE UNLIKELY TO DO THIS / NO, WE ARE NOT PLANNING TO DO THIS:<br>**Moderate modifiable risk factor**<br>• Where complex or potentially opaque models are under consideration and processes of model selection do not include appropriate and transparent considerations of the AI system's explainability, this presents a ***moderate modifiable risk factor*** for adverse impacts on the human rights and fundamental freedom of persons.<br><br>**Actions to take for your HUDERIA:**<br><br>No trigger message |
| | to their human rights and fundamental freedoms<br>c. The availability of the resources and capacity that will be needed to responsibly provide supplementary methods of explanation (e.g. simpler surrogate models, sensitivity analysis, or relative feature important) in cases where an opaque model is deemed appropriate and selected?<br><br>☐ YES, WE HAVE PLANS IN PLACE TO DO THIS<br><br>☐ WE HAD NOT CONSIDERED IT, BUT MAY DO THIS<br><br>☐ WE HAD NOT CONSIDERED IT, BUT ARE UNLIKELY TO DO THIS<br><br>☐ NO, WE ARE NOT PLANNING TO DO THIS<br><br>☐ UNSURE<br><br>☐ NOT APPLICABLE<br><br>***Example 1***<br>***Example 2*** | **Goals, properties, and areas to focus on in your HUDERAC**:<br><br>• Explainability (responsible model selection). Model selection considerations should include (a) the normal expectations of intelligibility and accessibility that accompany the function the system will fulfil in the sector or domain in which it will operate; (b) The availability of more interpretable algorithmic models or techniques in cases where the selection of an opaque model poses risks to the physical, psychological, or moral integrity of rights-holders or to their human rights and fundamental freedoms; (c) The availability of the resources and capacity that will be needed to responsibly provide supplementary methods of explanation (e.g. simpler surrogate models, sensitivity analysis, or relative feature important) in cases where an opaque model is deemed appropriate and selected. |
| | | UNSURE: Before taking any further steps in the proposed project, you should determine, through expert and stakeholder input where appropriate, how to incorporate measures to include explainability considerations in your model selection process. When this information is ascertained, you should return to your PCRA and revise it accordingly. |
| | | NOT APPLICABLE: No message |
| | | |
| **Model Output & Implementation Context** | | |
| **Model inference** | 42) Will sufficient and transparently reported processes be implemented throughout the project's lifecycle to ensure that the inferences generated from the model's learning mechanisms are reasonable, fair, equitable, and do not contain discriminatory correlations or influences of lurking or hidden proxies for discriminatory features that may act as significant factors in the generation of its output?<br><br>☐ YES, WE HAVE PLANS IN PLACE TO DO THIS<br><br>☐ WE HAD NOT CONSIDERED IT, BUT MAY DO THIS<br><br>☐ WE HAD NOT CONSIDERED IT, BUT ARE UNLIKELY TO DO THIS<br><br>☐ NO, WE ARE NOT PLANNING TO DO THIS<br><br>☐ UNSURE | YES, WE HAVE PLANS IN PLACE TO DO THIS: No message |
| | | WE HAD NOT CONSIDERED IT, BUT MAY DO THIS / WE HAD NOT CONSIDERED IT, BUT ARE UNLIKELY TO DO THIS / NO, WE ARE NOT PLANNING TO DO THIS:<br><br>**Major modifiable risk factor**<br><br>• Sufficient and transparently reported processes are not implemented throughout the project's lifecycle to ensure that the inferences generated from the model's learning mechanisms are reasonable, fair, equitable, and do not contain discriminatory correlations or influences of lurking or hidden proxies for discriminatory features that may act as significant factors in the generation of its output<br><br>**Actions to take for your HUDERIA:**<br><br>No trigger message<br><br>**Goals, properties, and areas to focus on in your HUDERAC**: |
| | ☐ NOT APPLICABLE<br><br>***Example*** | • Fairness (non-discrimination and bias mitigation) and accountability and process transparency (traceability, auditability, accessibility, and responsible governance). You should involve domain experts and social scientists to determine the potential sources of the hidden discriminatory proxies or correlations that may influence the outputs of the system, and carry out comprehensive assessments of algorithmic bias and of the differential performance of the system as its outputs relate to vulnerable and marginalized groups. |
| | | UNSURE: Before taking any further steps in the proposed project, you should determine, through expert and stakeholder input where appropriate, how to ensure that the inferences generated from the model's learning mechanisms are reasonable, fair, equitable, and do not contain discriminatory correlations or influences of lurking or hidden proxies for discriminatory features that may act as significant factors in the generation of its output. You should also determine how to incorporate and demonstrate such measures in your HUDERAC. When this information is ascertained, you should return to your PCRA and revise it accordingly. |
| | | NOT APPLICABLE: No message |
| **Model verification and validation** | 43) Will sufficient and transparently reported processes of external peer review and evaluation by | YES, WE HAVE PLANS IN PLACE TO DO THIS: No message |
| | independent domain and technical experts be included in the evaluation, verification, and validation of the AI model?<br><br>☐ YES, WE HAVE PLANS IN PLACE TO DO THIS<br><br>☐ WE HAD NOT CONSIDERED IT, BUT MAY DO THIS<br><br>☐ WE HAD NOT CONSIDERED IT, BUT ARE UNLIKELY TO DO THIS<br><br>☐ NO, WE ARE NOT PLANNING TO DO THIS<br><br>☐ UNSURE<br><br>☐ NOT APPLICABLE<br><br>***Example*** | WE HAD NOT CONSIDERED IT, BUT MAY DO THIS / WE HAD NOT CONSIDERED IT, BUT ARE UNLIKELY TO DO THIS / NO, WE ARE NOT PLANNING TO DO THIS:<br><br>**Moderate modifiable risk factor**<br><br>• Sufficient and transparently reported processes of external peer review and evaluation by independent domain and technical experts are not included in the evaluation, verification, and validation of the AI model<br><br>**Actions to take for your HUDERIA:**<br><br>No trigger message<br><br>**Goals, properties, and areas to focus on in your HUDERAC**:<br><br>• Accountability and process transparency, traceability, auditability, accessibility, reproducibility, and responsible governance. You should make sure to incorporate peer review protocols from the earliest stage of the project lifecycle, so that your workflow documentation processes deliberately enable seamless external evaluation and reproducibility. |
| | | UNSURE: Before taking any further steps in the proposed project, you should determine, through expert and stakeholder input where appropriate, how to ensure that processes of external peer |
| | | review and evaluation by independent domain and technical experts be included in the evaluation, verification, and validation of the AI model. You should also determine how to incorporate and demonstrate such measures in your HUDERAC. When this information is ascertained, you should return to your PCRA and revise it accordingly. |
| | | NOT APPLICABLE: No message |
| | 44) Will processes for evaluating the trained AI model include transparently reported external validation?<br><br>☐ YES, WE HAVE PLANS IN PLACE TO DO THIS<br><br>☐ WE HAD NOT CONSIDERED IT, BUT MAY DO THIS<br><br>☐ WE HAD NOT CONSIDERED IT, BUT ARE UNLIKELY TO DO THIS<br><br>☐ NO, WE ARE NOT PLANNING TO DO THIS<br><br>☐ UNSURE<br><br>☐ NOT APPLICABLE | YES, WE HAVE PLANS IN PLACE TO DO THIS: No message |
| | | WE HAD NOT CONSIDERED IT, BUT MAY DO THIS / WE HAD NOT CONSIDERED IT, BUT ARE UNLIKELY TO DO THIS / NO, WE ARE NOT PLANNING TO DO THIS:<br><br>**Moderate modifiable risk factor**<br><br>• Where processes for evaluating the trained AI model do not include transparently reported external validation, this presents a ***moderate modifiable risk factor*** for adverse impacts on the human rights and fundamental freedom of persons.<br><br>**Actions to take for your HUDERIA:**<br><br>No trigger message |
| | ***Example*** | **Goals, properties, and areas to focus on in your HUDERAC**:<br><br>• Accountability and process transparency (traceability, auditability, reproducibility, accessibility, and responsible governance). You should make sure to consult domain experts (and, where appropriate, relevant users and practitioners) during the external validation process, so that you have a realistic understanding of your system's operating environment and of the real-world conditions that may hamper the performance of the trained model. |
| | | UNSURE: Before taking any further steps in the proposed project, you should determine, through expert and stakeholder input where appropriate, how to ensure that processes of external validation be included in the evaluation of the trained AI model. You should also determine how to incorporate and demonstrate such measures in your HUDERAC. When this information is ascertained, you should return to your PCRA and revise it accordingly. |
| | | NOT APPLICABLE: No message |
| | 45) Will processes of monitoring the AI system during its operation involve regular re-evaluations of | YES, WE HAVE PLANS IN PLACE TO DO THIS: No message |
| | performance that are sufficient to keep pace with real world changes that may cause concept drifts and shifts in underlying data distributions?<br><br>☐ YES, WE HAVE PLANS IN PLACE TO DO THIS<br><br>☐ WE HAD NOT CONSIDERED IT, BUT MAY DO THIS<br><br>☐ WE HAD NOT CONSIDERED IT, BUT ARE UNLIKELY TO DO THIS<br><br>☐ NO, WE ARE NOT PLANNING TO DO THIS<br><br>☐ UNSURE<br><br>☐ NOT APPLICABLE<br><br>***Example*** | WE HAD NOT CONSIDERED IT, BUT MAY DO THIS / WE HAD NOT CONSIDERED IT, BUT ARE UNLIKELY TO DO THIS / NO, WE ARE NOT PLANNING TO DO THIS:<br><br>**Moderate modifiable risk factor**<br><br>• Processes of monitoring the AI system during its operation do not involve regular re-evaluations of performance that are sufficient to keep pace with real world changes that may cause concept drifts and shifts in underlying data distributions<br><br>**Actions to take for your HUDERIA:**<br><br>• Make sure to incorporate into your impact assessment process—especially in its post-development stage iteration—an exploration and provisional determination of the timeframes that are appropriate for performance re-evaluation and impact re-assessment.<br><br>**Goals, properties, and areas to focus on in your HUDERAC**:<br><br>• Safety, accuracy and performance metrics, reliability, robustness, sustainability, change monitoring, and reflection on context and impacts. You should involve domain experts to determine the potential sources of distributional shifts or drifts and build transparently processes of dynamic |
| | | assessment, re-assessment, external validation, and monitoring into your project lifecycle. |
| | | UNSURE: Before taking any further steps in the proposed project, you should determine, through expert and stakeholder input where appropriate, how to ensure that processes of monitoring your AI system during its operation involve regular re-evaluations of performance that are sufficient to keep pace with real world changes that may cause concept drifts and shifts in underlying data distributions. You should also determine how to incorporate and demonstrate such measures in your HUDERAC. When this information is ascertained, you should return to your PCRA and revise it accordingly. |
| | | NOT APPLICABLE: No message |
| **Model accuracy and performance metrics** | 46) When performance metrics for the AI system are considered, determined, and reported, will the prioritization of error types (e.g. false positives/negatives) be<br><br>a. Informed by the specific context of the use case and by the potential effects of differential error rates on affected sub-populations (in particular, on vulnerable or protected groups) | YES, WE HAVE PLANS IN PLACE TO DO THIS: No message |
| | | WE HAD NOT CONSIDERED IT, BUT MAY DO THIS / WE HAD NOT CONSIDERED IT, BUT ARE UNLIKELY TO DO THIS / NO, WE ARE NOT PLANNING TO DO THIS:<br><br>**Moderate modifiable risk factor**<br><br>• When performance metrics for the AI system are considered, determined, and reported, the prioritization and reporting of error types (e.g. false positives/negatives) are (a) not informed |
| | b. Clearly and accessibly presented, so that the rationale behind the chosen metrics is made explicit and understandable in plain, non-technical language?<br><br>☐ YES, WE HAVE PLANS IN PLACE TO DO THIS<br><br>☐ WE HAD NOT CONSIDERED IT, BUT MAY DO THIS<br><br>☐ WE HAD NOT CONSIDERED IT, BUT ARE UNLIKELY TO DO THIS<br><br>☐ NO, WE ARE NOT PLANNING TO DO THIS<br><br>☐ UNSURE<br><br>☐ NOT APPLICABLE<br><br>***Example*** | by the specific context of the use case and the potential effects of differential error rates on affected sub-populations (in particular, on vulnerable or protected groups) and (b) not clearly and accessibly presented, so that the rationale behind the chosen metrics is made explicit and understandable in plain, non-technical language<br><br>**Actions to take for your HUDERIA:**<br><br>No trigger message<br><br>**Goals, properties, and areas to focus on in your HUDERAC**:<br><br>• Safety, accuracy and performance metrics, accessibility, explainability. You should ensure that the selection and reporting of error types are informed by the specific context of the use case and the potential effects of differential error rates on affected sub-populations. |
| | | UNSURE: Before taking any further steps in the proposed project, you should determine, through expert and stakeholder input where appropriate, how to ensure that the selection and reporting of error types are informed by the specific context of the use case and the potential effects of differential error rates on affected sub-populations. You should also determine how to incorporate and demonstrate such measures in your HUDERAC. When this information is |
| | | ascertained, you should return to your PCRA and revise it accordingly. |
| | | NOT APPLICABLE: No message |
| | 47) When performance metrics for the AI system are considered, determined, and presented, will the prioritization and reporting of metrics beyond accuracy (e.g. sensitivity, precision, specificity) be informed by the specific context of the use case and its performance needs (e.g. a system whose effective identification of rare events is more critical than its overall accuracy rate)?<br><br>☐ YES, WE HAVE PLANS IN PLACE TO DO THIS<br><br>☐ WE HAD NOT CONSIDERED IT, BUT MAY DO THIS<br><br>☐ WE HAD NOT CONSIDERED IT, BUT ARE UNLIKELY TO DO THIS<br><br>☐ NO, WE ARE NOT PLANNING TO DO THIS<br><br>☐ UNSURE<br><br>☐ NOT APPLICABLE | YES, WE HAVE PLANS IN PLACE TO DO THIS: No message |
| | | WE HAD NOT CONSIDERED IT, BUT MAY DO THIS / WE HAD NOT CONSIDERED IT, BUT ARE UNLIKELY TO DO THIS / NO, WE ARE NOT PLANNING TO DO THIS:<br><br>**Moderate modifiable risk factor**<br><br>• When performance metrics for the AI system are considered, determined, and reported, the prioritization and reporting of metrics beyond accuracy (e.g. sensitivity, precision, specificity) is not informed by the specific context of the use case and its performance needs (e.g. a system whose effective identification of rare events is more critical than its overall accuracy rate)<br><br>**Actions to take for your HUDERIA:**<br><br>No trigger message<br><br>**Goals, properties, and areas to focus on in your HUDERAC**: |
| | ***Example*** | • Safety, accuracy and performance metrics, accessibility, explainability. You should ensure that the selection and reporting of performance metrics for your AI system beyond accuracy are informed by its specific context and its performance needs. You should also determine how to incorporate and demonstrate this in your HUDERAC. |
| | | UNSURE: Before taking any further steps in the proposed project, you should determine, through expert and stakeholder input where appropriate, how to ensure that the selection and reporting of performance metrics for your AI system beyond accuracy (e.g. sensitivity, precision, specificity) are informed by the specific context of the use case and its performance needs. You should also determine how to incorporate and demonstrate such measures in your HUDERAC. When this information is ascertained, you should return to your PCRA and revise it accordingly. |
| | | NOT APPLICABLE: No message |
| **System-User Interface and Human Factors Context** | | |
| **Implementers or users of the system** | 48) Will the project lifecycle incorporate sufficient processes to ensure that the deployment of the system does not harm the physical, psychological, or moral | YES, WE HAVE PLANS IN PLACE TO DO THIS: No message |
| | | WE HAD NOT CONSIDERED IT, BUT MAY DO THIS / WE HAD NOT CONSIDERED IT, BUT ARE |
| | integrity of implementers or adversely impact their dignity, autonomy, and ability to make free, independent, and well-informed judgements?<br><br>☐ YES, WE HAVE PLANS IN PLACE TO DO THIS<br><br>☐ WE HAD NOT CONSIDERED IT, BUT MAY DO THIS<br><br>☐ WE HAD NOT CONSIDERED IT, BUT ARE UNLIKELY TO DO THIS<br><br>☐ NO, WE ARE NOT PLANNING TO DO THIS<br><br>☐ UNSURE<br><br>☐ NOT APPLICABLE<br><br>***Example*** | UNLIKELY TO DO THIS / NO, WE ARE NOT PLANNING TO DO THIS:<br><br>**Major modifiable risk factor**<br><br>• The project lifecycle does not incorporate sufficient processes to ensure that the deployment of the system does not harm the physical, psychological, or moral integrity of implementers or adversely impact their dignity, autonomy, and ability to make free, independent, and well-informed judgements<br><br>**Actions to take for your HUDERIA:**<br><br>• Make sure to incorporate into your impact assessment process thorough consideration of the possible effects of the AI system on the human rights and fundamental freedoms of its implementers. As affected rights-holders, implementers should be included as participants in the HUDERIA process. You should also integrate any mitigation measures for potential adverse impacts identified into your HUDERAC.<br><br>**Goals, properties, and areas to focus on in your HUDERAC**:<br><br>• Sustainability (reflection on context and impacts, and responsible implementation and user training) |
| | | UNSURE: Before taking any further steps in the proposed project, you should determine, through |
| | | expert and stakeholder input where appropriate, how to ensure that the deployment of the system does not harm the physical, psychological, or moral integrity of implementers or adversely impact their dignity, autonomy, and ability to make free, independent, and well-informed judgements. When this information is ascertained, you should return to your PCRA and revise it accordingly. |
| | | NOT APPLICABLE: No message |
| **Level of automation/level of human involvement and choice** | 49) Will implementers of the AI system be sufficiently trained so that they are able to fully understand<br>a. the strengths and limitation of the system and its outputs<br>b. the potential conditions of situational complexity, uncertainty, anomaly, or system failure that may dictate the need for the exercise of human judgment, common sense, and practical intervention?<br><br>☐ YES, WE HAVE PLANS IN PLACE TO DO THIS<br><br>☐ WE HAD NOT CONSIDERED IT, BUT MAY DO THIS | YES, WE HAVE PLANS IN PLACE TO DO THIS: No message |
| | | WE HAD NOT CONSIDERED IT, BUT MAY DO THIS / WE HAD NOT CONSIDERED IT, BUT ARE UNLIKELY TO DO THIS / NO, WE ARE NOT PLANNING TO DO THIS:<br><br>**Major modifiable risk factor**<br><br>• Implementers of the AI system are not sufficiently trained so that they are able to fully understand both the strengths and limitation of the system (and its outputs) and the potential conditions of situational complexity, uncertainty, anomaly, or system failure that may dictate the need for the exercise of human judgment, common sense, and practical intervention<br><br>**Actions to take for your HUDERIA:** |
| | ☐ WE HAD NOT CONSIDERED IT, BUT ARE UNLIKELY TO DO THIS<br><br>☐ NO, WE ARE NOT PLANNING TO DO THIS<br><br>☐ UNSURE<br><br>☐ NOT APPLICABLE<br><br>**Example** | No trigger message<br><br>**Goals, properties, and areas to focus on in your HUDERAC**:<br><br>• Responsible implementation and user training. |
| | | UNSURE: Before taking any further steps in the proposed project, you should determine, through expert and stakeholder input where appropriate, how to ensure that implementers of the AI system are sufficiently trained to fully understand both the strengths and limitation of the system (and its outputs) and the potential conditions of situational complexity, uncertainty, anomaly, or system failure that may dictate the need for the exercise of human judgment, common sense, and practical intervention. You should also determine how to incorporate and demonstrate such training measures in your HUDERAC. When this information is ascertained, you should return to your PCRA and revise it accordingly. |
| | | NOT APPLICABLE: No message |
| | 50) If the AI system has a high level of automation or operational 'autonomy' and interacts with rights-holders in ways that could adversely impact their physical, psychological, or moral integrity or | YES: No message |
| | | NO: STOP. |
| | harm their human rights and fundamental freedoms, will mechanisms of human control and intervention (e.g. human-in-the-loop or human-on-the-loop) be incorporated into the implementation of the system?<br><br>☐ YES<br><br>☐ NO<br><br>☐ UNSURE<br><br>☐ NOT APPLICABLE<br><br>***Example*** | **Prohibitive modifiable risk factor**<br><br>Where AI systems have high levels of automation or operational 'autonomy' and interact with rights-holders in ways that could adversely impact their physical, psychological, or moral integrity or harm their human rights and fundamental freedoms, implementation processes must incorporate appropriate mechanisms of human control and intervention. Systems that fail to do this pose too high a risk to impacted rights-holders and communities to be developed or implemented in that form. In revising your project, you should ensure that processes of human control and intervention are integrated into the system's deployment and determine how to demonstrate this in your HUDERAC. If you wish to continue with the PCRA anyway at this time, select yes. If not, please reconsider your project and return when you have ascertained that your AI system will have sufficient mechanisms that enable human control and intervention. |
| | | UNSURE: STOP. Where AI systems have high levels of automation or operational 'autonomy' and interact with rights-holders in ways that could adversely impact their physical, psychological, or moral integrity or harm their human rights and fundamental freedoms, implementation processes must incorporate appropriate mechanisms of human control and intervention. Systems that fail to do this pose too high a risk to impacted rights-holders and communities to be developed or implemented in that form. Before taking any further steps in the |
| | | proposed project, you should determine, through expert and stakeholder input where appropriate, how to ensure that, if your AI system has a high level of automation and operational 'autonomy' and interacts with rights-holders in ways that could adversely impact their physical, psychological, or moral integrity or harm their human rights and fundamental freedoms, processes of human control and intervention are appropriately integrated into the system's deployment. If you wish to continue with the PCRA anyway at this time, select yes. If not, please reconsider your project and return when you have ascertained that your AI system will have sufficient mechanisms that enable human control and intervention. |
| | | NOT APPLICABLE: No message |
| **Rights & Freedoms Context** | | |
| **Respect for and protection of human dignity** | 51) If the AI system interacts with rights-holders (users or decision subjects) in ways that could adversely impact their physical, psychological, or moral integrity or harm their dignity, will they be able to opt out of the interaction and revert to human intervention?<br><br>☐ YES<br><br>☐ NO | YES: No message |
| | | NO: STOP<br><br>**Prohibitive modifiable risk factor**<br><br>Where AI systems interact with rights-holders (users or decision subjects) in ways that could adversely impact their physical, psychological, or moral integrity or harm their dignity but they are not given the choice to opt out of the interaction |
| | ☐ UNSURE<br><br>☐ NOT APPLICABLE<br><br>***Example*** | and revert to human intervention, such systems are considered to pose too high a risk to impacted rights-holders and communities for you to proceed. In revising your project, you should pay close attention to ensuring the availability and accessibility of an opt out option for impacted rights-holders, and to demonstrating this in your HUDERAC. You should also pay close attention in your HUDERIA to assessing the impact of the system on human dignity. If you wish to continue with the PCRA anyway at this time, select yes. If not, please reconsider your project and return when you have ascertained that your AI system will have sufficient mechanisms that ensure the availability and accessibility of an opt out option for impacted rights-holders. |
| | | UNSURE: STOP Before taking any further steps in the proposed project, you should determine, through expert and stakeholder input where appropriate, whether, in cases where the AI system will interact with rights-holders (users or decision subjects), they will be given the choice to opt out of the interaction and revert to human intervention if they feel that their dignity risks being violated. If no option is given, the system is considered to pose too high a risk to impacted rights-holders and communities for you to proceed. In revising your project, you should pay close attention to ensuring the availability and accessibility of an opt out option for impacted rights-holders, and to demonstrating this in your HUDERAF. You should also pay close attention in your HUDERIA to assessing the impacts of the system on human dignity. If you wish to continue with the PCRA anyway at this time, select yes. If not, please reconsider your project and return |
| | | when you have ascertained that your AI system will have sufficient mechanisms that ensure the availability and accessibility of an opt out option for impacted rights-holders. |
| | | NOT APPLICABLE: No message |
| | 52) If the AI system interacts directly with rights-holders (users or decision subjects) in ways that could adversely impact their physical, psychological, or moral integrity or harm their dignity, will they have pre-knowledge of this interaction and be able to provide meaningful and informed consent to participate in the interaction?<br><br>☐ YES<br>☐ NO<br>☐ UNSURE<br>☐ NOT APPLICABLE<br><br>***Example*** | YES: No message |
| | | NO: STOP<br><br>**Prohibitive modifiable risk factor**<br><br>Where AI systems interact directly with rights-holders (users or decision subjects) in ways that could adversely impact their physical, psychological, or moral integrity or harm their dignity but they do not have pre-knowledge of this interaction and are not able to provide informed consent to participate in the interaction, such systems are considered to pose too high a risk to impacted rights-holders and communities for you to proceed. In revising your project, you should pay close attention to ensuring that impacted rights-holders are provided clear foreknowledge about interacting with the system and are able to provide meaningful and informed consent to participate in the interaction. This should be demonstrated in your HUDERAC. You should also pay close attention in your HUDERIA to assessing the impacts of the system on human dignity and human freedom and autonomy. If you wish to continue with the PCRA anyway at this |
| | | time, select yes. If not, please reconsider your project and return when you have ascertained that your AI system will have sufficient mechanisms to ensure that impacted rights-holders are provided clear foreknowledge about interacting with the system and are able to provide meaningful and informed consent to participate in the interaction. |
| | | UNSURE: STOP Before taking any further steps in the proposed project, you should determine, through expert and stakeholder input where appropriate, whether, in cases where the AI system interacts directly with rights-holders (users or decision subjects) in ways that could adversely impact their physical, psychological, or moral integrity or harm their dignity, they will have pre-knowledge of this interaction and be able to provide meaningful and informed consent to participate in the interaction. If not, the system is considered to pose too high a risk to impacted rights-holders and communities for you to proceed. In revising your project, you should pay close attention to ensuring that impacted rights-holders are provided clear foreknowledge about interacting with the system and are able to provide meaningful and informed consent to participate in the interaction. This should be demonstrated in your HUDERAC. You should also pay close attention in your HUDERIA to assessing the impacts of the system on human dignity, human freedom, and autonomy. If you wish to continue with the PCRA anyway at this time, select yes. If not, please reconsider your project and return when you have ascertained that your AI system will have sufficient mechanisms to ensure that impacted rights-holders are provided |
| | | clear foreknowledge about interacting with the system and are able to provide meaningful and informed consent to participate in the interaction. |
| | | NOT APPLICABLE: No message |
| **Protection of human freedom and autonomy** | 53) Will the project lifecycle incorporate sufficient and transparently reported processes to ensure the ability of rights-holders<br><br>a. to make free and well-informed judgements about the reasonableness and justifiability of the outputs of the AI system<br>b. to effectively contest and challenge decisions informed and/or made by that system<br>c. to demand that such decisions be reviewed by a person?<br><br>☐ YES, WE HAVE PLANS IN PLACE TO DO THIS<br><br>☐ WE HAD NOT CONSIDERED IT, BUT MAY DO THIS<br><br>☐ WE HAD NOT CONSIDERED IT, BUT ARE UNLIKELY TO DO THIS | YES, WE HAVE PLANS IN PLACE TO DO THIS: No message |
| | | WE HAD NOT CONSIDERED IT, BUT MAY DO THIS / WE HAD NOT CONSIDERED IT, BUT ARE UNLIKELY TO DO THIS / NO, WE ARE NOT PLANNING TO DO THIS:<br><br>**Major modifiable risk factor**<br><br>• Where project lifecycles do not incorporate sufficient processes to ensure the ability of rights-holders to make free, independent, and well-informed judgements about the reasonableness and justifiability of the outputs of the AI system, to effectively contest and challenge decisions informed and/or made by that system, and to demand that such decision be reviewed by a person, this presents a ***major modifiable risk factor*** for adverse impacts on the human rights and fundamental freedom of persons.<br><br>**Actions to take for your HUDERIA:** |
| | ☐ NO, WE ARE NOT PLANNING TO DO THIS<br><br>☐ UNSURE<br><br>☐ NOT APPLICABLE<br><br>***Example 1***<br>***Example 2*** | • Make sure to incorporate into your impact assessment process thorough consideration of the possible effects of the AI system on the freedom and autonomy of affected rights-holders and on their abilities to make free, independent, and well-informed decisions.<br><br>**Goals, properties, and areas to focus on in your HUDERAC**:<br><br>• Sustainability (reflection on context and impacts) and accountability and process transparency (traceability, auditability, accessibility, and responsible governance) |
| | | UNSURE: Before taking any further steps in the proposed project, you should determine, through expert and stakeholder input where appropriate, how you can incorporate sufficient processes in your project lifecycles to ensure the ability of rights-holders to make free, independent, and well-informed judgements about the reasonableness and justifiability of the outputs of the AI system, to effectively contest and challenge decisions informed and/or made by that system, and to demand that such decision be reviewed by a person. You should also determine how to incorporate and demonstrate such measures in your HUDERAC. When this information is ascertained, you should return to your PCRA and revise it accordingly. |
| | | NOT APPLICABLE: No message |
| **Non-discrimination, fairness, and equality** | 54) Will sufficient and transparently reported processes be implemented throughout the project's lifecycle to ensure that the AI system, in both its production and use, mitigates possible sources of bias and discriminatory patterns in each of the following:<br><br>a. The datasets used to train the system<br>b. The decisions made to build the system<br>c. The way the problem to which the system responds is understood, formulated, and framed<br>d. The way the target variable and its measurable proxy are defined<br>e. The way the system's algorithmic model(s) is selected, and its parameters tuned and adjusted<br>f. The way that model is trained, tested, and validated<br>g. The way the system is implemented, and the way users are trained to deploy it<br>h. The choices made about monitoring, updating, repurposing, or deprovisioning the system? | YES, WE HAVE PLANS IN PLACE TO DO THIS: No message |
| | | WE HAD NOT CONSIDERED IT, BUT MAY DO THIS / WE HAD NOT CONSIDERED IT, BUT ARE UNLIKELY TO DO THIS / NO, WE ARE NOT PLANNING TO DO THIS:<br>**Major modifiable risk factor**<br><br>• Where project lifecycles do not incorporate sufficient and transparently reported processes to ensure that AI systems, in both their production and use, mitigate possible sources of bias and discriminatory patterns, this presents a ***major modifiable risk factor*** for adverse impacts on the human rights and fundamental freedom of persons.<br><br>**Actions to take for your HUDERIA:**<br><br>• Make sure to incorporate into your impact assessment process thorough consideration of the principles and priorities of non-discrimination, fairness, and equality. You should also pay special attention in your HUDERIA to assessing the impacts of the system on historically marginalized and vulnerable groups and on groups with protected characteristics.<br><br>**Goals, properties, and areas to focus on in your HUDERAC**: |
| | ☐ YES, WE HAVE PLANS IN PLACE TO DO THIS<br><br>☐ WE HAD NOT CONSIDERED IT, BUT MAY DO THIS<br><br>☐ WE HAD NOT CONSIDERED IT, BUT ARE UNLIKELY TO DO THIS<br><br>☐ NO, WE ARE NOT PLANNING TO DO THIS<br><br>☐ UNSURE<br><br>☐ NOT APPLICABLE<br><br>***Example*** | • Fairness (non-discrimination and bias mitigation) and accountability and process transparency (traceability, auditability, accessibility, and responsible governance).<br><br>UNSURE: Before taking any further steps in the proposed project, you should determine, through expert and stakeholder input where appropriate, how to incorporate sufficient and transparently reported processes to ensure that your AI systems, in both its production and use, mitigates possible sources of bias and discriminatory patterns. You should also determine how to incorporate and demonstrate such measures in your HUDERAC. When this information is ascertained, you should return to your PCRA and revise it accordingly.<br><br>NOT APPLICABLE: No message |
| | 55) Will sufficient and transparently reported processes be implemented throughout the project's lifecycle to ensure that the system, in both its production and use, promotes diversity and inclusiveness<br><br>a. In the composition of innovation team building the system | YES, WE HAVE PLANS IN PLACE TO DO THIS: No message<br><br>WE HAD NOT CONSIDERED IT, BUT MAY DO THIS / WE HAD NOT CONSIDERED IT, BUT ARE UNLIKELY TO DO THIS / NO, WE ARE NOT PLANNING TO DO THIS:<br><br>**Major modifiable risk factor**<br><br>• Where project lifecycles do not incorporate sufficient and transparently reported processes to ensure that AI |
| | b. In the expertise, insight, and knowledge drawn upon to develop it<br>c. In the individuals and groups able to access its benefits?<br><br>☐ YES, WE HAVE PLANS IN PLACE TO DO THIS<br><br>☐ WE HAD NOT CONSIDERED IT, BUT MAY DO THIS<br><br>☐ WE HAD NOT CONSIDERED IT, BUT ARE UNLIKELY TO DO THIS<br><br>☐ NO, WE ARE NOT PLANNING TO DO THIS<br><br>☐ UNSURE<br><br>☐ NOT APPLICABLE<br><br>***Example*** | systems, in both their production and use, promote diversity and inclusiveness, this presents a ***major modifiable risk factor*** for adverse impacts on the human rights and fundamental freedom of persons.<br><br>**Actions to take for your HUDERIA:**<br><br>• Make sure to incorporate into your impact assessment process thorough consideration of the principles and priorities of diversity, inclusiveness, non-discrimination, fairness, and equality. You should also pay special attention in your Stakeholder Engagement Process to assessing team positionality and the need for diversification of team composition, expertise, insight, and knowledge.<br><br>**Goals, properties, and areas to focus on in your HUDERAC**:<br><br>• Fairness (non-discrimination, bias mitigation, diversity, and inclusiveness) and sustainability (reflection on context and impacts and stakeholder engagement and involvement) |
| | | UNSURE: Before taking any further steps in the proposed project, you should determine, through expert and stakeholder input where appropriate, how to ensure that your AI system, in both its production and use, promotes diversity and inclusiveness. You should also determine how to incorporate and demonstrate such measures in your HUDERAC. When this information is |
| | | ascertained, you should return to your PCRA and revise it accordingly. |
| | | NOT APPLICABLE: No message |
| **Data protection and privacy context** | 56) If the AI system will be processing personal data, will the project lifecycle incorporate transparently reported mechanisms that demonstrate its compliance with data protection and privacy law and the principles set out in the Council of Europe's Convention 108+?<br><br>☐ YES, WE HAVE PLANS IN PLACE TO DO THIS<br><br>☐ WE HAD NOT CONSIDERED IT, BUT MAY DO THIS<br><br>☐ WE HAD NOT CONSIDERED IT, BUT ARE UNLIKELY TO DO THIS<br><br>☐ NO, WE ARE NOT PLANNING TO DO THIS<br><br>☐ UNSURE<br><br>☐ NOT APPLICABLE | YES, WE HAVE PLANS IN PLACE TO DO THIS: No message |
| | | WE HAD NOT CONSIDERED IT, BUT MAY DO THIS / WE HAD NOT CONSIDERED IT, BUT ARE UNLIKELY TO DO THIS / NO, WE ARE NOT PLANNING TO DO THIS:<br><br>**Major modifiable risk factor**<br><br>• Where project lifecycles for AI systems that process personal data do not incorporate transparently reported mechanisms that demonstrate their compliance with data protection and privacy law and the principles set out in the Council of Europe's Convention 108+, this presents a ***major modifiable risk factor*** for adverse impacts on the human rights and fundamental freedom of persons.<br><br>**Actions to take for your HUDERIA:**<br><br>• Make sure to pay special attention in your HUDERIA to assessing the impacts of the system on data protection and the respect for private and family life, drawing on your |
| | ***Example*** | Data Protection Impact Assessment (DPIA), if applicable.<br><br>**Goals, properties, and areas to focus on in your HUDERAC**:<br><br>• Data protection and privacy (transparency, proportionality, consent or legitimate basis for processing, data quality, data security, purpose limitation, accountability, data minimisation, data protection and privacy by design, fairness, and lawfulness). You should consult data protection experts to examine and determine the measures that your project needs to take to process personal data lawfully and in accordance with the principles set out in Convention 108+, including the completion of a Data Protection Impact Assessment (DPIA)(undertaken either as part of transparent reporting or as a compliance mechanism or as both). You should include your DPIA in your HUDERAC, if applicable, to evidence your compliance and best practices. |
| | | UNSURE: Before taking any further steps in the proposed project, you should determine, through input from data protection experts, the measures that your project needs to take to process personal data lawfully and in accordance with the principles set out in Convention 108+, including the completion of a Data Protection Impact Assessment (DPIA)(undertaken either as part of transparent reporting or as a compliance mechanism or as both). When this information is |
| | | ascertained, you should return to your PCRA and revise it accordingly. |
| | | NOT APPLICABLE: No message |
| | 57) If the AI system will be processing personal data, will the project lifecycle incorporate transparently reported mechanisms that demonstrate its respect for the rights of data subjects and conformity to the additional obligations of data controllers and processors as set out in Articles 9 and 10 of Convention 108+?<br><br>☐ YES, WE HAVE PLANS IN PLACE TO DO THIS<br><br>☐ WE HAD NOT CONSIDERED IT, BUT MAY DO THIS<br><br>☐ WE HAD NOT CONSIDERED IT, BUT ARE UNLIKELY TO DO THIS<br><br>☐ NO, WE ARE NOT PLANNING TO DO THIS<br><br>☐ UNSURE<br><br>☐ NOT APPLICABLE | YES, WE HAVE PLANS IN PLACE TO DO THIS: No message |
| | | WE HAD NOT CONSIDERED IT, BUT MAY DO THIS / WE HAD NOT CONSIDERED IT, BUT ARE UNLIKELY TO DO THIS / NO, WE ARE NOT PLANNING TO DO THIS:<br>**Major modifiable risk factor**<br><br>• Where project lifecycles for AI systems that process personal data do not incorporate transparently reported mechanisms that demonstrate their respect for the rights of data subjects and conformity to the additional obligations of data controllers and processors as set out in Articles 9 and 10 of Convention 108+, this presents a ***major modifiable risk factor*** for adverse impacts on the human rights and fundamental freedom of persons.<br><br>**Actions to take for your HUDERIA:**<br><br>• Make sure to pay special attention in your HUDERIA to assessing the impacts of the system on data protection, the rights of the data subject, and the respect for |
| | ***Example*** | private and family life, drawing on your DPIA, if applicable.<br><br>**Goals, properties, and areas to focus on in your HUDERAC**:<br><br>• Data protection and privacy (respect for the rights of data subjects and conformity to additional obligations—Articles 9 and 10, Convention 108+) You should consult data protection experts to examine and determine the measures that your project needs to take to process personal data lawfully and in accordance with the principles set out in Convention 108+, including the completion of a Data Protection Impact Assessment (DPIA) (undertaken either as part of transparent reporting or as a compliance mechanism or as both). You should include your DPIA in your HUDERAC, if applicable, to evidence your compliance and best practices. |
| | | UNSURE: Before taking any further steps in the proposed project, you should determine, through input from data protection experts, the measures that your project needs to take to process personal data in accordance with obligations of data controllers and processors as set out in Articles 9 and 10 of Convention 108+. When this information is ascertained, you should return to your PCRA and revise it accordingly. |
| | | NOT APPLICABLE: No message |
| | 58) If the AI system will be processing sensitive data as defined in Convention 108+, will the project lifecycle incorporate transparently reported mechanisms that demonstrate its compliance with appropriate safeguards?<br><br>☐ YES, WE HAVE PLANS IN PLACE TO DO THIS<br><br>☐ WE HAD NOT CONSIDERED IT, BUT MAY DO THIS<br><br>☐ WE HAD NOT CONSIDERED IT, BUT ARE UNLIKELY TO DO THIS<br><br>☐ NO, WE ARE NOT PLANNING TO DO THIS<br><br>☐ UNSURE<br><br>☐ NOT APPLICABLE<br><br>***Example*** | YES, WE HAVE PLANS IN PLACE TO DO THIS: No message |
| | | WE HAD NOT CONSIDERED IT, BUT MAY DO THIS / WE HAD NOT CONSIDERED IT, BUT ARE UNLIKELY TO DO THIS / NO, WE ARE NOT PLANNING TO DO THIS:<br>**Major modifiable risk factor**<br><br>• Where project lifecycles for AI systems that process sensitive data as defined in Convention 108+ do not incorporate transparently reported mechanisms that demonstrate their compliance with appropriate safeguards, this presents a ***major modifiable risk factor*** for adverse impacts on the human rights and fundamental freedom of persons.<br><br>**Actions to take for your HUDERIA:**<br><br>• Make sure to pay special attention in your HUDERIA to assessing the impacts of the system on data protection, the rights of the data subject, and the respect for private and family life, drawing on your DPIA, if applicable.<br><br>**Goals, properties, and areas to focus on in your HUDERAC**:<br><br>• Data protection and privacy (responsible handling of sensitive data). You should consult data protection experts to |
| | | examine and determine the measures that your project needs to take to process personal data lawfully and in accordance with the principles set out in Convention 108+, including the completion of a Data Protection Impact Assessment (DPIA)(undertaken either as part of transparent reporting or as a compliance mechanism or as both). You should include your DPIA in your HUDERAC, if applicable, to evidence your compliance and best practices. |
| | | UNSURE: Before taking any further steps in the proposed project, you should determine, through input from data protection experts, the measures that your project needs to take to process sensitive data as defined in Convention 108+ in compliance with appropriate safeguards. When this information is ascertained, you should return to your PCRA and revise it accordingly. |
| | | NOT APPLICABLE: No message |
| | 59) If the AI system is designed with the purpose or function of individual-targeted curation, profiling, prediction, or behavioural steering, will affected rights-holders be able to obtain from the data controller sufficient information concerning: | YES: No message |
| | | NO: STOP. AI systems that are designed with the purpose or function of individual-targeted curation, profiling, prediction, or behavioural steering and that do not make available to affected rights-holders sufficient information concerning data use, output rationale, purpose, and conveyance of processing results are |
| | a. The use of their personal data and the categories used in the system's processing<br>b. An explanation of the rationale behind the output of the processing in plain, non-technical language<br>c. The purpose of the curation, profiling, prediction, classification, or behavioural steering<br>d. The categories of persons or bodies to whom personal data, the profile or the result of the processing may be communicated?<br><br>☐ YES<br>☐ NO<br>☐ UNSURE<br>☐ NOT APPLICABLE<br><br>***Example*** | considered to pose too high a risk to impacted rights-holders and communities for you to proceed. In revising your project, you should pay close attention to ensuring that impacted rights-holders are provided clear, accessible, and sufficient information concerning data use, output rationale, purpose, and conveyance of processing results You should consult data protection experts to examine and determine the measures that your project needs to take to accomplish this. If you wish to continue with the PCRA anyway at this time, select yes. If not, please reconsider your project and return when you have ascertained that your AI system will have sufficient mechanisms to ensure that impacted rights-holders are provided clear, accessible, and sufficient information concerning data use, output rationale, purpose, and conveyance of processing results. |
| | | UNSURE: STOP Before taking any further steps in the proposed project, you should determine, through input from data protection experts, whether, in the case that your AI system will be designed with the purpose or function of individual-targeted curation, profiling, prediction, or behavioural steering, you will make available to affected rights-holders sufficient information concerning data use, output rationale, purpose, and conveyance of processing results. If not, the system is considered to pose too high a risk to impacted rights-holders and communities for you to proceed. In revising your project, you should pay close attention to ensuring that impacted rights-holders are provided clear, accessible, and sufficient information concerning data use, output rationale, purpose, and conveyance of processing |
| | | results You should consult data protection experts to examine and determine the measures that your project needs to take to accomplish this. If you wish to continue with the PCRA anyway at this time, select yes. If not, please reconsider your project and return when you have ascertained that your AI system will have sufficient mechanisms to ensure that impacted rights-holders are provided clear, accessible, and sufficient information concerning data use, output rationale, purpose, and conveyance of processing results. |
| | | NOT APPLICABLE: No message |
| **Accountability and access to justice** | 60) Will sufficient and transparently reported processes be implemented throughout the project's lifecycle to ensure end-to-end accountability across the production and use of the AI system? Namely, will it ensure that the system<br><br>a. Is auditable by design, allowing for the end-to-end traceability and oversight of its processes of production and use<br>b. Establishes a continuous chain of human responsibility for all roles involved in the project lifecycle to allow for end-to- | YES: No message |
| | | NO:<br><br>**Major modifiable risk factor**<br><br>• Where project lifecycles do not incorporate sufficient and transparently reported processes to ensure end-to-end accountability across the production and use of AI systems, this presents a ***major modifiable risk factor*** for adverse impacts on the human rights and fundamental freedom of persons.<br><br>**Actions to take for your HUDERIA:**<br><br>No trigger message |
| | end answerability in the event that the human rights or fundamental freedoms of affected individuals have been negatively impacted<br>c. Enables designated public authorities and third parties, where appropriate, to assess its compliance with existing legislation, regulation, and standards instruments across the entire project lifecycle?<br><br>☐ YES<br>☐ NO<br>☐ UNSURE<br>☐ NOT APPLICABLE<br><br>***Example*** | **Goals, properties, and areas to focus on in your HUDERAC**:<br>• Accountability and process transparency (traceability, auditability, accessibility, and responsible governance).<br><br>UNSURE: Before taking any further steps in the proposed project, you should determine, through expert and stakeholder input where appropriate, how to incorporate transparently reported processes to ensure sufficient accountability across the production and use of the AI system. You should pay special attention to finding out how to assure, in your HUDERAC, the goals of accountability, transparency, and explainability and how to demonstrate this in a clear and accessible way. When this information is ascertained, you should return to your PCRA and revise it accordingly.<br><br>NOT APPLICABLE: No message |
| | 61) Will sufficient and transparently reported processes be implemented throughout the project's lifecycle to ensure that affected persons whose human rights or fundamental freedoms have been adversely impacted by | YES, WE HAVE PLANS IN PLACE TO DO THIS: No message<br><br>WE HAD NOT CONSIDERED IT, BUT MAY DO THIS / WE HAD NOT CONSIDERED IT, BUT ARE UNLIKELY TO DO THIS / NO, WE ARE NOT PLANNING TO DO THIS: |
| | the AI system have actionable redress and effective remedy by<br><br>a. Providing sufficient and meaningful information that indicates when the system is being used and how and where to complain in the event of an adverse impact on human rights and fundamental freedoms<br>b. Facilitating access of affected rights-holders to sufficient and meaningful information about the processes behind the design, development, and deployment of the system and about the rationale underlying the outcomes of its processing<br>c. Employing algorithmic models that are appropriately interpretable or explainable (especially regarding discriminatory proxies or inferences that may be embedded in trained machine learning systems) given the risks to human rights, fundamental freedoms, democracy, and the rule of law they may pose? | **Major modifiable risk factor**<br><br>• Where project lifecycles do not implement sufficient and transparently reported processes to ensure that affected individuals have actionable redress and effective remedy in cases where their human rights or fundamental freedoms have been adversely impacted, this presents a ***major modifiable risk factor*** for adverse impacts on the human rights and fundamental freedom of persons.<br><br>**Actions to take for your HUDERIA:**<br><br>• Make sure to pay close attention to assessing the impacts of the system on human freedom and autonomy and on the ability of rights-holders to make free, independent, and well-informed decisions.<br><br>**Goals, properties, and areas to focus on in your HUDERAC**:<br><br>• Accountability and process transparency (traceability, auditability, accessibility, and responsible governance), explainability (interpretability and accessible rationale explanation). This should include incorporating into your HUDERAC evidence that you will be able to (a) provide sufficient and meaningful information that indicates when the system is being used and how and where to complain in the event of an adverse |
| | ☐ YES, WE HAVE PLANS IN PLACE TO DO THIS<br><br>☐ WE HAD NOT CONSIDERED IT, BUT MAY DO THIS<br><br>☐ WE HAD NOT CONSIDERED IT, BUT ARE UNLIKELY TO DO THIS<br><br>☐ NO, WE ARE NOT PLANNING TO DO THIS<br><br>☐ UNSURE<br><br>☐ NOT APPLICABLE<br><br>***Example*** | impact on human rights and fundamental freedoms, (b) facilitate access of affected rights-holders to sufficient and meaningful information about the processes behind the design, development, and deployment of the system and about the rationale underlying the outcomes of its processing, and (c) employ algorithmic models that are appropriately interpretable or explainable (especially regarding discriminatory proxies or inferences that may be embedded in trained machine learning systems) given the risks to human rights, fundamental freedoms, democracy, and the rule of law they may pose. |
| | | UNSURE: Before taking any further steps in the proposed project, you should determine, through expert and stakeholder input where appropriate, how to incorporate transparently reported processes to ensure that affected individuals have actionable redress and effective remedy in cases where their human rights or fundamental freedoms have been adversely impacted. You should pay special attention to finding out how to assure, in your HUDERAC, the goals of accountability, transparency, and explainability and how to demonstrate this in a clear and accessible way. When this information is ascertained, you should return to your PCRA and revise it accordingly. |
| | | NOT APPLICABLE: No message |
| | 62) If the AI system is used in the field of justice and law enforcement, will meaningful information be provided to affected rights-holders about the existence and use of the system, its role within law enforcement and the judicial process, and the right to challenge the decisions informed or made thereby?<br><br>☐ YES, WE HAVE PLANS IN PLACE TO DO THIS<br><br>☐ WE HAD NOT CONSIDERED IT, BUT MAY DO THIS<br><br>☐ WE HAD NOT CONSIDERED IT, BUT ARE UNLIKELY TO DO THIS<br><br>☐ NO, WE ARE NOT PLANNING TO DO THIS<br><br>☐ UNSURE<br><br>☐ NOT APPLICABLE<br><br>***Example*** | YES, WE HAVE PLANS IN PLACE TO DO THIS: No message |
| | | WE HAD NOT CONSIDERED IT, BUT MAY DO THIS / WE HAD NOT CONSIDERED IT, BUT ARE UNLIKELY TO DO THIS / NO, WE ARE NOT PLANNING TO DO THIS:<br><br>**Major modifiable risk factor**<br><br>• Where AI system are being used in the field of justice and law enforcement and meaningful information is not provided to affected rights-holders about the existence and use of the system, its role within law enforcement and the judicial process, and the right to challenge the decisions informed or made thereby, this presents a ***major modifiable risk factor*** for adverse impacts on the human rights and fundamental freedom of persons.<br><br>**Actions to take for your HUDERIA:**<br><br>• Make sure to pay special attention in your HUDERIA to assessing the potential impacts of the system on individual dignity, freedom, and autonomy, and on how effects on legal processes and institutions may damage the rule of law.<br><br>**Goals, properties, and areas to focus on in your HUDERAC**: |
| | | • Accountability and process transparency (traceability, auditability, accessibility, and responsible governance) and explainability (interpretability and accessible rationale explanation) |
| | | UNSURE: Before taking any further steps in the proposed project, you should determine, through expert and stakeholder input where appropriate, how meaningful information can be provided to affected rights-holders about the existence and use of the system, its role within law enforcement and the judicial process, and the right to challenge the decisions informed or made thereby. You should pay special attention to finding out how to assure, in your HUDERAC, the goals of accountability, transparency, and explainability and how to demonstrate this in a clear and accessible way. When this information is ascertained, you should return to your PCRA and revise it accordingly. |
| | | NOT APPLICABLE: No message |
| | 63) If the AI system is used in the field of justice and law enforcement, will sufficiently and transparently reported processes be implemented throughout the project's lifecycle to ensure that its deployment is in line with the essential requirements of impacted individuals' right to a fair trial (equality of arms, right to a natural | YES, WE HAVE PLANS IN PLACE TO DO THIS: No message |
| | | WE HAD NOT CONSIDERED IT, BUT MAY DO THIS / WE HAD NOT CONSIDERED IT, BUT ARE UNLIKELY TO DO THIS / NO, WE ARE NOT PLANNING TO DO THIS:<br>**Major modifiable risk factor** |
| | judge established by law, the right to an independent and impartial tribunal, and respect for the adversarial process)?<br><br>☐ YES, WE HAVE PLANS IN PLACE TO DO THIS<br><br>☐ WE HAD NOT CONSIDERED IT, BUT MAY DO THIS<br><br>☐ WE HAD NOT CONSIDERED IT, BUT ARE UNLIKELY TO DO THIS<br><br>☐ NO, WE ARE NOT PLANNING TO DO THIS<br><br>☐ UNSURE<br><br>☐ NOT APPLICABLE<br><br>***Example*** | • Where AI system are being used in the field of justice and law enforcement and sufficiently and transparently reported processes are not implemented throughout the project's lifecycle to ensure that its deployment is in line with the essential requirements of impacted individuals' right to a fair trial (equality of arms, right to a natural judge established by law, the right to an independent and impartial tribunal, and respect for the adversarial process), this presents a ***major modifiable risk factor*** for adverse impacts on the human rights and fundamental freedom of persons.<br><br>**Actions to take for your HUDERIA:**<br><br>• Make sure to pay special attention in your HUDERIA to assessing the potential impacts of the system on individual dignity, freedom, and autonomy, and on how effects on legal processes and institutions may damage the rule of law.<br><br>**Goals, properties, and areas to focus on in your HUDERAC**:<br><br>• Accountability and process transparency (traceability, auditability, accessibility, and responsible governance) and explainability (interpretability and accessible rationale explanation) |
| | | UNSURE: Before taking any further steps in the proposed project, you should determine, through expert and stakeholder input where appropriate, how to incorporate transparently reported processes to ensure that its deployment is in line with the essential requirements of impacted individuals' right to a fair trial. You should pay special attention to finding out how to assure, in your HUDERAC, the goals of accountability, transparency, and explainability and how to demonstrate this in a clear and accessible way. When this information is ascertained, you should return to your PCRA and revise it accordingly. |
| | | NOT APPLICABLE: No message |
| **Social and economic rights** | 64) If the AI system is being used in the areas of employment decisions, recruitment, worker management, or the distribution of work, will the project lifecycle incorporate transparently reported processes to ensure that the system protects<br>a. The dignity of work and the right to just working conditions as set out in Article 2 of the European Social Charter<br>b. The right to safe and healthy working conditions as set out in Article 3 of the Europeans Social Charter | YES, WE HAVE PLANS IN PLACE TO DO THIS: No message |
| | | WE HAD NOT CONSIDERED IT, BUT MAY DO THIS / WE HAD NOT CONSIDERED IT, BUT ARE UNLIKELY TO DO THIS / NO, WE ARE NOT PLANNING TO DO THIS:<br>**Major modifiable risk factor**<br>• Where AI systems are being used in the areas of employment decisions, recruitment, worker management, or the distribution of work and project lifecycles do not incorporate transparently reported processes to ensure that the system protects the dignity of work, the right to just working conditions, the right to safe |
| | c. The right to organize as set out in Article 5 of the European Social Charter?<br><br>☐ YES, WE HAVE PLANS IN PLACE TO DO THIS<br><br>☐ WE HAD NOT CONSIDERED IT, BUT MAY DO THIS<br><br>☐ WE HAD NOT CONSIDERED IT, BUT ARE UNLIKELY TO DO THIS<br><br>☐ NO, WE ARE NOT PLANNING TO DO THIS<br><br>☐ UNSURE<br><br>☐ NOT APPLICABLE<br><br>***Example a***<br>***Example b***<br>***Example c*** | and healthy working conditions, and the right to organize as set out in the European Social Charter, this presents a ***major modifiable risk factor*** for adverse impacts on the human rights and fundamental freedom of persons.<br><br>**Actions to take for your HUDERIA:**<br><br>• Make sure to pay close attention, in your HUDERIA, to assessing the impacts of the system on social and economic rights, equality, dignity, and the right to physical, psychological, or moral integrity.<br><br>**Goals, properties, and areas to focus on in your HUDERAC**:<br><br>• Fairness (equality, non-discrimination, bias mitigation, diversity, and inclusiveness) and sustainability (reflection on context and impacts and stakeholder engagement and involvement) |
| | | UNSURE: Before taking any further steps in the proposed project, you should determine, through expert and stakeholder input where appropriate, how to incorporate transparently reported processes to ensure that the system protects the dignity of work, the right to just working conditions, the right to safe and healthy working conditions, and the right to organize as set out in the European Social Charter. When this information is ascertained, you should return to your PCRA and revise it accordingly. |

<table>
<tr><td rowspan="3"></td><td></td><td>NOT APPLICABLE: No message</td></tr>
<tr><td rowspan="2">65) If the AI system is being used in the context of social security decisions or social welfare administration, will sufficient and transparently reported processes be implemented throughout the project's lifecycle to ensure that the system protects the right to social security as set out in Article 12 of the European Social Charter?<br><br>☐ YES, WE HAVE PLANS IN PLACE TO DO THIS<br><br>☐ WE HAD NOT CONSIDERED IT, BUT MAY DO THIS<br><br>☐ WE HAD NOT CONSIDERED IT, BUT ARE UNLIKELY TO DO THIS<br><br>☐ NO, WE ARE NOT PLANNING TO DO THIS<br><br>☐ UNSURE<br><br>☐ NOT APPLICABLE<br><br>Example</td><td>YES, WE HAVE PLANS IN PLACE TO DO THIS: No message</td></tr>
<tr><td>WE HAD NOT CONSIDERED IT, BUT MAY DO THIS / WE HAD NOT CONSIDERED IT, BUT ARE UNLIKELY TO DO THIS / NO, WE ARE NOT PLANNING TO DO THIS:<br><br>Major modifiable risk factor<br><br>• Where AI systems are being used in the context of social security decisions or social welfare administration and project lifecycles do not incorporate transparently reported processes to ensure that the system protects the right to social security as set out in Article 12 of the European Social Charter, this presents a major modifiable risk factor for adverse impacts on the human rights and fundamental freedom of persons.<br><br>Actions to take for your HUDERIA:<br><br>• Make sure to pay close attention, in your HUDERIA, to assessing the impacts of the system on social and economic rights, equality, dignity, and the right to physical, psychological, or moral integrity.<br><br>Goals, properties, and areas to focus on in your HUDERAC:</td></tr>
</table>

| | | |
|---|---|---|
| | | • Fairness (equality, non-discrimination, bias mitigation, diversity, and inclusiveness) and sustainability (reflection on context and impacts and stakeholder engagement and involvement) |
| | | UNSURE: Before taking any further steps in the proposed project, you should determine, through expert and stakeholder input where appropriate, how to incorporate transparently reported processes to ensure that the system protects the right to social security as set out in Article 12 of the European Social Charter. When this information is ascertained, you should return to your PCRA and revise it accordingly. |
| | | NOT APPLICABLE: No message |
| | 66) If the AI system is being used to automate decisions regarding the provision of healthcare and medical assistance, will sufficient and transparently reported processes be implemented throughout the project's lifecycle to ensure that the system protects the rights to the protection of health and to social and medical assistance as set out in Articles 11 and 13 of the European Social Charter? | YES, WE HAVE PLANS IN PLACE TO DO THIS: No message |
| | | WE HAD NOT CONSIDERED IT, BUT MAY DO THIS / WE HAD NOT CONSIDERED IT, BUT ARE UNLIKELY TO DO THIS / NO, WE ARE NOT PLANNING TO DO THIS:<br><br>**Major modifiable risk factor**<br><br>• Where AI systems are being used to automate decisions regarding the provision of healthcare and medical assistance and project lifecycles do not incorporate transparently reported |
| | ☐ YES, WE HAVE PLANS IN PLACE TO DO THIS<br><br>☐ WE HAD NOT CONSIDERED IT, BUT MAY DO THIS<br><br>☐ WE HAD NOT CONSIDERED IT, BUT ARE UNLIKELY TO DO THIS<br><br>☐ NO, WE ARE NOT PLANNING TO DO THIS<br><br>☐ UNSURE<br><br>☐ NOT APPLICABLE<br><br>***Example*** | processes to ensure that the system protects the rights to the protection of health and to social and medical assistance as set out in Articles 11 and 13 of the European Social Charter, this presents a ***major modifiable risk factor*** for adverse impacts on the human rights and fundamental freedom of persons.<br><br>**Actions to take for your HUDERIA:**<br><br>• Make sure to pay close attention, in your HUDERIA, to assessing the impacts of the system on social and economic rights, equality, dignity, and the right to physical, psychological, or moral integrity.<br><br>**Goals, properties, and areas to focus on in your HUDERAC**:<br><br>• Fairness (equality, non-discrimination, bias mitigation, diversity, and inclusiveness) and sustainability (reflection on context and impacts and stakeholder engagement and involvement) |
| | | UNSURE: Before taking any further steps in the proposed project, you should determine, through expert and stakeholder input where appropriate, how to incorporate transparently reported processes to ensure that the system protects the right to the protection of health and to social and medical assistance as set out in Articles 11 and 13 of the European Social Charter. When this |
| | | information is ascertained, you should return to your PCRA and revise it accordingly. |
| | | NOT APPLICABLE: No message |

**Section 2: Risks of adverse impacts on the human rights and fundamental freedom of persons, democracy, and the rule of law**

In the following section, you will be asked to provide a preliminary estimation of the likelihood and gravity potential of the adverse impacts that your system could have on human rights, democracy, and the rule of law. Draw on the stakeholder analysis and the determination of salient rights you carried out as part of your initial PS reporting to think through the relevance, likelihood, and gravity potential of the harms indicated in the prompts. Each prompt will be formed as a statement with likelihood/probability options below.

| Principles/Priorities | Statement | Trigger messages for summary report |
|---|---|---|
| **Respect for and protection of human dignity** | 67) The AI system could interact with rights-holders in ways that create confusion or uncertainty about whether they are interacting with a computational technology.<br><br>☐ UNLIKELY<br><br>☐ POSSIBLE<br><br>☐ LIKELY | UNLIKELY, POSSIBLE, LIKELY, VERY LIKELY: (Probability answers to be tabulated in the risk index table) |
| | | UNSURE: Before taking any further steps in the proposed project, you should determine, through expert and stakeholder input where appropriate, whether the AI system you are planning to build will interact with rights-holders in ways that could create confusion or uncertainty about whether they are interacting with a computational technology. When this information is ascertained, |
| | ☐ VERY LIKELY<br>☐ UNSURE<br>☐ N/A<br><br>***Example*** | you should return to your PCRA and revise it accordingly. |
| | 68) If you responded with any of the options besides N/A or Unsure in the previous question, how would you characterize the gravity potential of the harm? (For details on gravity potential, please refer to the relevant section of the user's guide).<br>☐ CATASTROPHIC HARM<br>☐ CRITICAL HARM<br>☐ SERIOUS HARM<br>☐ MODERATE OR MINOR HARM | |
| | 69) The AI system could interact with rights-holders (users or decision subjects) in ways that expose them to humiliation (being put in a state of helplessness or insignificance; being dehumanized or losing a sense of individual identity).<br><br>☐ UNLIKELY<br>☐ POSSIBLE<br>☐ LIKELY<br>☐ VERY LIKELY<br>☐ UNSURE<br>☐ N/A<br><br>***Example*** | UNLIKELY, POSSIBLE, LIKELY, VERY LIKELY: (Probability answers to be tabulated in the risk index table)<br><br>UNSURE: Before taking any further steps in the proposed project, you should determine, through expert and stakeholder input where appropriate, whether the AI system you are planning to build could interact with rights-holders (users or decision subjects) in ways that expose them to humiliation (being put in a state of helplessness or insignificance; being dehumanized or losing a sense of individual identity). When this information is ascertained, you should return to your PCRA and revise it accordingly. |
| | 70) If you responded with any of the options besides N/A or Unsure in the previous question, how would you characterize the gravity potential of the harm? (For details on gravity potential, please refer to the relevant section of the user's guide).<br><br>☐ CATASTROPHIC HARM<br><br>☐ CRITICAL HARM<br><br>☐ SERIOUS HARM<br><br>☐ MODERATE OR MINOR HARM | |
| | 71) The AI system could interact with rights-holders (users or decision subjects) in ways that expose them to instrumentalization or objectification (treating an individual solely as exchangeable, as a statistical aggregate, as a means to an end, or as an object to be freely manipulated or steered).<br><br>☐ UNLIKELY<br><br>☐ POSSIBLE<br><br>☐ LIKELY | UNLIKELY, POSSIBLE, LIKELY, VERY LIKELY: (Probability answers to be tabulated in the risk index table)<br><br>UNSURE: Before taking any further steps in the proposed project, you should determine, through expert and stakeholder input where appropriate, whether the AI system you are planning to build could interact with rights-holders (users or decision subjects) in ways that expose them to instrumentalization or objectification (treating an individual solely as exchangeable, as a statistical aggregate, as a means to an end, or as an object to be freely manipulated or steered). When this information is ascertained, you should return to your PCRA and revise it accordingly. |
| | ☐ VERY LIKELY<br>☐ UNSURE<br>☐ N/A<br><br>***Example*** | |
| | 72) If you responded with any of the options besides N/A or Unsure in the previous question, how would you characterize the gravity potential of the harm? (For details on gravity potential, please refer to the relevant section of the user's guide).<br>☐ CATASTROPHIC HARM<br>☐ CRITICAL HARM<br>☐ SERIOUS HARM<br>☐ MODERATE OR MINOR HARM | |
| | 73) The AI system could interact with rights-holders (users or decision subjects) in ways that expose | UNLIKELY, POSSIBLE, LIKELY, VERY LIKELY: (Probability answers to be tabulated in the risk index table) |
| | them to displacement, redundancy, or a sense of worthlessness in regard to their participation in work life, creative life, or the life of the community.<br><br>☐ UNLIKELY<br>☐ POSSIBLE<br>☐ LIKELY<br>☐ VERY LIKELY<br>☐ UNSURE<br>☐ N/A<br><br>***Example*** | UNSURE: Before taking any further steps in the proposed project, you should determine, through expert and stakeholder input where appropriate, whether the AI system you are planning to build could interact with rights-holders (users or decision subjects) in ways that expose them to displacement, redundancy, or a sense of worthlessness in regard to their participation in work life, creative life, or the life of the community. When this information is ascertained, you should return to your PCRA and revise it accordingly. |
| | 74) If you responded with any of the options besides N/A or Unsure in the previous question, how would you characterize the gravity potential of the harm? (For details on gravity potential, please refer to the relevant section of the user's guide).<br><br>☐ CATASTROPHIC HARM<br>☐ CRITICAL HARM | |
| | ☐ SERIOUS HARM<br>☐ MODERATE OR MINOR HARM | |
| **Protection of human freedom and autonomy** | 75) The AI system could interact with rights-holders in ways that adversely affect or hinder their abilities to make free, independent, and well-informed decisions about their lives or about the system's outputs.<br>☐ UNLIKELY<br>☐ POSSIBLE<br>☐ LIKELY<br>☐ VERY LIKELY<br>☐ UNSURE<br>☐ N/A<br>***Example*** | UNLIKELY, POSSIBLE, LIKELY, VERY LIKELY: (Probability answers to be tabulated in the risk index table)<br><br>Unsure: Before taking any further steps in the proposed project, you should determine, through expert and stakeholder input where appropriate, whether the AI system you are planning to build will interact with rights-holders in ways that could adversely affect or hinder their abilities to make free, independent, and well-informed decisions about their lives or about the system's outputs. When this information is ascertained, you should return to your PCRA and revise it accordingly. |
| | 76) If you responded with any of the options besides N/A or Unsure in the previous question, how would you characterize the gravity potential of the harm? (For details on gravity potential, please refer to the relevant section of the user's guide).<br><br>☐ CATASTROPHIC HARM<br>☐ CRITICAL HARM<br>☐ SERIOUS HARM<br>☐ MODERATE OR MINOR HARM | |
| | 77) The AI system could interact with rights-holders in ways that adversely affect or hinder their capacities to flourish, to fully develop themselves, and to pursue their own freely determined life plans.<br><br>☐ UNLIKELY<br>☐ POSSIBLE<br>☐ LIKELY<br>☐ VERY LIKELY | UNLIKELY, POSSIBLE, LIKELY, VERY LIKELY: (Probability answers to be tabulated in the risk index table)<br><br>UNSURE: Before taking any further steps in the proposed project, you should determine, through expert and stakeholder input where appropriate, whether the AI system you are planning to build will interact with rights-holders in ways that could adversely affect or hinder their capacities to flourish, to fully develop themselves, and to pursue their own freely determined life plans. When this information is ascertained, you should return to your PCRA and revise it accordingly. |
| | ☐ UNSURE<br>☐ N/A<br><br>***Example*** | |
| | 78) If you responded with any of the options besides N/A or Unsure in the previous question, how would you characterize the gravity potential of the harm? (For details on gravity potential, please refer to the relevant section of the user's guide).<br><br>☐ CATASTROPHIC HARM<br>☐ CRITICAL HARM<br>☐ SERIOUS HARM<br>☐ MODERATE OR MINOR HARM | |
| | 79) The deployment of the system could result in the arbitrary deprivation of rights-holders' physical freedom or personal security, or the denial of their freedoms of expression, thought, conscience, or assembly. | UNLIKELY, POSSIBLE, LIKELY, VERY LIKELY: (Probability answers to be tabulated in the risk index table)<br><br>UNSURE: Before taking any further steps in the proposed project, you should determine, through expert and stakeholder input where appropriate, |
| | ☐ UNLIKELY<br>☐ POSSIBLE<br>☐ LIKELY<br>☐ VERY LIKELY<br>☐ UNSURE<br>☐ N/A<br><br>***Example*** | whether the AI system you are planning to build will interact with rights-holders in ways that could result in the arbitrary deprivation of rights-holders' physical freedom or personal security, or the denial of their freedoms of expression, thought, conscience, or assembly. When this information is ascertained, you should return to your PCRA and revise it accordingly. |
| | 80) If you responded with any of the options besides N/A or Unsure in the previous question, how would you characterize the gravity potential of the harm? (For details on gravity potential, please refer to the relevant section of the user's guide).<br><br>☐ CATASTROPHIC HARM<br>☐ CRITICAL HARM<br>☐ SERIOUS HARM<br>☐ MODERATE OR MINOR HARM | |
| **Prevention of harm and protection of the right to life and physical, psychological, and moral integrity** | 81) The AI system could interact with rights-holders in ways that deprive them of their right to life or their physical, psychological, or moral integrity.<br><br>☐ UNLIKELY<br>☐ POSSIBLE<br>☐ LIKELY<br>☐ VERY LIKELY<br>☐ UNSURE<br>☐ N/A<br><br>***Example*** | UNLIKELY, POSSIBLE, LIKELY, VERY LIKELY: (Probability answers to be tabulated in the risk index table)<br><br>UNSURE: Before taking any further steps in the proposed project, you should determine, through expert and stakeholder input where appropriate, whether the AI system you are planning to build will interact with rights-holders in ways that could deprive them of their right to life or their physical, psychological, or moral integrity. When this information is ascertained, you should return to your PCRA and revise it accordingly. |
| | 82) If you responded with any of the options besides N/A or Unsure in the previous question, how would you characterize the gravity potential of the harm? (For details on gravity potential, please refer to the relevant section of the user's guide).<br><br>☐ CATASTROPHIC HARM<br>☐ CRITICAL HARM | |
| | ☐ SERIOUS HARM<br>☐ MODERATE OR MINOR HARM | |
| | 83) The AI system could interact with the environment (either in the processes of its production or in its deployment) in ways that harm the biosphere or adversely impact the health of the planet.<br>☐ UNLIKELY<br>☐ POSSIBLE<br>☐ LIKELY<br>☐ VERY LIKELY<br>☐ UNSURE<br>☐ N/A<br>***Example*** | UNLIKELY, POSSIBLE, LIKELY, VERY LIKELY: (Probability answers to be tabulated in the risk index table)<br>UNSURE: Before taking any further steps in the proposed project, you should determine, through expert and stakeholder input where appropriate, whether the AI system you are planning to build will interact with the environment (either in the processes of its production or in its deployment) in ways that could harm the biosphere or adversely impact the health of the planet. When this information is ascertained, you should return to your PCRA and revise it accordingly. |
| | 84) If you responded with any of the options besides N/A or Unsure in the previous question, how would you characterize the gravity potential of the harm? (For details on gravity potential, please refer to the relevant section of the user's guide).<br><br>☐ CATASTROPHIC HARM<br>☐ CRITICAL HARM<br>☐ SERIOUS HARM<br>☐ MODERATE OR MINOR HARM | |
| **Non-discrimination, fairness, and equality** | 85) The AI system (either in the processes of its production or in its deployment) could result in discrimination, have discriminatory effects on impacted rights-holders, or perform differentially for different groups in discriminatory or harmful ways—including for intersectional groups where vulnerable or protected characteristics converge.<br><br>☐ UNLIKELY<br>☐ POSSIBLE<br>☐ LIKELY | UNLIKELY, POSSIBLE, LIKELY, VERY LIKELY: (Probability answers to be tabulated in the risk index table)<br><br>UNSURE: Before taking any further steps in the proposed project, you should determine, through expert and stakeholder input where appropriate, whether the AI (either in the processes of its production or in its deployment) could result in discrimination, have discriminatory effects on impacted rights-holders, or perform differentially for different groups in discriminatory or harmful ways—including for intersectional groups where vulnerable or protected characteristics converge. When this information is ascertained, you should return to your PCRA and revise it accordingly. |
| | ☐ VERY LIKELY<br>☐ UNSURE<br>☐ N/A<br><br>***Example*** | |
| | 86) If you responded with any of the options besides N/A or Unsure in the previous question, how would you characterize the gravity potential of the harm? (For details on gravity potential, please refer to the relevant section of the user's guide).<br><br>☐ CATASTROPHIC HARM<br>☐ CRITICAL HARM<br>☐ SERIOUS HARM<br>☐ MODERATE OR MINOR HARM | |
| | 87) The use of the AI system could expand existing inequalities in the communities it affects or augment historical patterns of inequity and | UNLIKELY, POSSIBLE, LIKELY, VERY LIKELY: (Probability answers to be tabulated in the risk index table) |
| | discrimination in these communities.<br><br>☐ UNLIKELY<br>☐ POSSIBLE<br>☐ LIKELY<br>☐ VERY LIKELY<br>☐ UNSURE<br>☐ N/A<br><br>***Example*** | UNSURE: Before taking any further steps in the proposed project, you should determine, through expert and stakeholder input where appropriate, whether the AI system you are planning to build could expand existing inequalities in the communities they affect or augment historical patterns of inequity and discrimination in these communities. When this information is ascertained, you should return to your PCRA and revise it accordingly. |
| | 88) If you responded with any of the options besides N/A or Unsure in the previous question, how would you characterize the gravity potential of the harm? (For details on gravity potential, please refer to the relevant section of the user's guide).<br><br>☐ CATASTROPHIC HARM<br>☐ CRITICAL HARM<br>☐ SERIOUS HARM | |
| | ☐ MODERATE OR MINOR HARM | |
| **Data protection and the right to respect for private and family life** | 89) The design, development, and deployment of the AI system could harm the rights to data protection enshrined in data protection and privacy law and in the Council of Europe's Convention 108+.<br><br>☐ UNLIKELY<br>☐ POSSIBLE<br>☐ LIKELY<br>☐ VERY LIKELY<br>☐ UNSURE<br>☐ N/A<br><br>***Example*** | UNLIKELY, POSSIBLE, LIKELY, VERY LIKELY: (Probability answers to be tabulated in the risk index table)<br><br>UNSURE: Before taking any further steps in the proposed project, you should determine, through expert and stakeholder input where appropriate, whether the design, development, and deployment of the AI system could harm the right to data protection enshrined in data protection and privacy law and in the Council of Europe's Convention 108+. When this information is ascertained, you should return to your PCRA and revise it accordingly. |
| | 90) If you responded with any of the options besides N/A or Unsure in the previous question, how would you characterize the gravity potential of the harm? (For details on gravity potential, please refer to the relevant section of the user's guide). | |
| | ☐ CATASTROPHIC HARM<br>☐ CRITICAL HARM<br>☐ SERIOUS HARM<br>☐ MODERATE OR MINOR HARM | |
| | 91) The AI system could intrude on or interfere with the private and family life of rights-holders in ways that prevent or impede them from maintaining a personal sphere that is independent from the transformative effects of AI technologies and in which they are at liberty to freely think, form opinions and beliefs, and develop their personal identities and intimate relationships without the influence of AI technologies.<br>☐ UNLIKELY<br>☐ POSSIBLE<br>☐ LIKELY<br>☐ VERY LIKELY<br>☐ UNSURE<br>☐ N/A | UNLIKELY, POSSIBLE, LIKELY, VERY LIKELY: (Probability answers to be tabulated in the risk index table)<br><br>UNSURE: Before taking any further steps in the proposed project, you should determine, through expert and stakeholder input where appropriate, whether the AI system you are planning to build could intrude on or interfere with the private and family life of rights-holders in ways that prevent or impede them from maintaining a personal sphere that is independent from the transformative effects of AI technologies and in which they are at liberty to freely think, form opinions and beliefs, and develop their personal identities and intimate relationships without the influence of AI technologies. When this information is ascertained, you should return to your PCRA and revise it accordingly. |

| | *Example* | |
|---|---|---|
| | 92) If you responded with any of the options besides N/A or Unsure in the previous question, how would you characterize the gravity potential of the harm? (For details on gravity potential, please refer to the relevant section of the user's guide).<br><br>☐ CATASTROPHIC HARM<br><br>☐ CRITICAL HARM<br><br>☐ SERIOUS HARM<br><br>☐ MODERATE OR MINOR HARM | |
| **Social and economic rights** | 93) The deployment of the AI system could harm the social and economic rights of affected persons, including the right to just working conditions, the right to safe and healthy working conditions, the right to organize, the right to social security, and the rights to the protection of health and to social and medical assistance as set out in the European Social Charter.<br><br>☐ UNLIKELY<br><br>☐ POSSIBLE | UNLIKELY, POSSIBLE, LIKELY, VERY LIKELY: (Probability answers to be tabulated in the risk index table)<br><br>UNSURE: Before taking any further steps in the proposed project, you should determine, through expert and stakeholder input where appropriate, whether the AI system you are planning to build could harm the social and economic rights of affected persons, including the right to just working conditions, the right to safe and healthy working conditions, the right to organize, the right to social security, and the rights to the protection of health and to social and medical assistance as set out in the European Social |
| | ☐ LIKELY<br>☐ VERY LIKELY<br>☐ UNSURE<br>☐ N/A<br>***Example*** | Charter. When this information is ascertained, you should return to your PCRA and revise it accordingly. |
| | 94) If you responded with any of the options besides N/A or Unsure in the previous question, how would you characterize the gravity potential of the harm? (For details on gravity potential, please refer to the relevant section of the user's guide).<br>☐ CATASTROPHIC HARM<br>☐ CRITICAL HARM<br>☐ SERIOUS HARM<br>☐ MODERATE OR MINOR HARM | |
| **Democracy** | 95) The use or misuse of the AI system could lead to interference with free and fair election processes or with the ability of impacted individuals to participate freely, fairly, and fully in the political life of the community through any of the following: | UNLIKELY, POSSIBLE, LIKELY, VERY LIKELY: (Probability answers to be tabulated in the risk index table)<br><br>UNSURE: Before taking any further steps in the proposed project, you should determine, through expert and stakeholder input where appropriate, whether the use or misuse of the AI system could lead to interference with free and fair election |
| | a. Mass deception, at local, national, or global levels caused by the deployment of the system<br>b. Mass manipulation, at local, national, or global levels enabled by the deployment of the system<br>c. Mass intimidation or behavioural control, at local, national, or global levels enabled by the deployment of the system<br>d. Mass personalized political targeting or profiling, at local, national, or global levels enabled by the deployment of the system<br><br>☐ UNLIKELY<br>☐ POSSIBLE<br>☐ LIKELY<br>☐ VERY LIKELY<br>☐ UNSURE<br>☐ N/A<br><br>***Example*** | processes or with the ability of impacted individuals to participate freely, fairly, and fully in the political life of the community (through mass deceptions, mass manipulation, mass intimidation or behavioural control, or mass personalized political targeting or profiling). When this information is ascertained, you should return to your PCRA and revise it accordingly. |
| | 96) If you responded with any of the options besides N/A or Unsure in the previous question, how would you characterize the gravity potential of the harm? (For details on gravity potential, please refer to the relevant section of the user's guide).<br><br>☐ CATASTROPHIC HARM<br>☐ CRITICAL HARM<br>☐ SERIOUS HARM<br>☐ MODERATE OR MINOR HARM | |
| | 97) The use or misuse of the AI system could lead to the mass dispersal, at local, national, or global levels, of misinformation or disinformation.<br><br>☐ UNLIKELY<br>☐ POSSIBLE<br>☐ LIKELY<br>☐ VERY LIKELY<br>☐ UNSURE<br>☐ N/A | UNLIKELY, POSSIBLE, LIKELY, VERY LIKELY: (Probability answers to be tabulated in the risk index table)<br><br>UNSURE: Before taking any further steps in the proposed project, you should determine, through expert and stakeholder input where appropriate, whether the use or misuse of the AI system could lead to the mass dispersal, at local, national, or global levels, of misinformation or disinformation. When this information is ascertained, you should return to your PCRA and revise it accordingly. |
| | ***Example*** | |
| | 98) If you responded with any of the options besides N/A or Unsure in the previous question, how would you characterize the gravity potential of the harm? (For details on gravity potential, please refer to the relevant section of the user's guide).<br><br>☐ CATASTROPHIC HARM<br>☐ CRITICAL HARM<br>☐ SERIOUS HARM<br>☐ MODERATE OR MINOR HARM | |
| | 99) The use or misuse of the AI system could lead to obstruction of informational plurality, at local, national, or global levels.<br><br>☐ UNLIKELY<br>☐ POSSIBLE<br>☐ LIKELY<br>☐ VERY LIKELY<br>☐ UNSURE | UNLIKELY, POSSIBLE, LIKELY, VERY LIKELY: (Probability answers to be tabulated in the risk index table)<br><br>UNSURE: Before taking any further steps in the proposed project, you should determine, through expert and stakeholder input where appropriate, whether the use or misuse of the AI system could lead to obstruction of informational plurality, at local, national, or global levels. If so, the system is considered to pose significant risks to the physical, psychological, or moral integrity or the human rights and fundamental freedoms of affected persons. When this information is |
| | ☐ N/A<br><br>***Example*** | ascertained, you should return to your PCRA and revise it accordingly. |
| | 100) If you responded with any of the options besides N/A or Unsure in the previous question, how would you characterize the gravity potential of the harm? (For details on gravity potential, please refer to the relevant section of the user's guide).<br><br>☐ CATASTROPHIC HARM<br>☐ CRITICAL HARM<br>☐ SERIOUS HARM<br>☐ MODERATE OR MINOR HARM | |
| | 101) The use or misuse of the AI system (in particular, information filtering models such as recommender systems, search engines, or news aggregators) could lead to an obstruction of the free and equitable flow of the legitimate and valid forms of information that are necessary for the meaningful democratic participation of impacted rights holders and for their ability to engage freely, fairly, and fully in collective problem-solving. | UNLIKELY, POSSIBLE, LIKELY, VERY LIKELY: (Probability answers to be tabulated in the risk index table)<br><br>UNSURE: Before taking any further steps in the proposed project, you should determine, through expert and stakeholder input where appropriate, whether the use or misuse of the AI system could lead to an obstruction of the free and equitable flow of the legitimate and valid forms of information that are necessary for the meaningful democratic participation of impacted rights holders and for their ability to engage freely, |
| | ☐ UNLIKELY<br>☐ POSSIBLE<br>☐ LIKELY<br>☐ VERY LIKELY<br>☐ UNSURE<br>☐ N/A<br><br>***Example*** | fairly, and fully in collective problem-solving. When this information is ascertained, you should return to your PCRA and revise it accordingly. |
| | 102) If you responded with any of the options besides N/A or Unsure in the previous question, how would you characterize the gravity potential of the harm? (For details on gravity potential, please refer to the relevant section of the user's guide).<br><br>☐ CATASTROPHIC HARM<br>☐ CRITICAL HARM<br>☐ SERIOUS HARM<br>☐ MODERATE OR MINOR HARM | |
| | 103) The use or misuse of the AI system (in particular, models that track, identify, de-anonymize or surveil rights-holders and social groups or enable the creation of social graphs) could lead to interference with or obstruction of impacted rights-holders' abilities to exercise their freedoms of expression, assembly, or association.<br><br>☐ UNLIKELY<br>☐ POSSIBLE<br>☐ LIKELY<br>☐ VERY LIKELY<br>☐ UNSURE<br>☐ N/A<br><br>***Example*** | UNLIKELY, POSSIBLE, LIKELY, VERY LIKELY: (Probability answers to be tabulated in the risk index table)<br><br>UNSURE: Before taking any further steps in the proposed project, you should determine, through expert and stakeholder input where appropriate, whether the use or misuse of the AI system (in particular, models that track, identify, de-anonymize or surveil rights-holders and social groups or enable the creation of social graphs) could lead to interference with or obstruction of impacted rights-holders' abilities to exercise their freedoms of expression, assembly, or association. When this information is ascertained, you should return to your PCRA and revise it accordingly. |
| | 104) If you responded with any of the options besides N/A or Unsure in the previous question, how would you characterize the gravity potential of the harm? (For details on gravity potential, please refer to the relevant section of the user’s guide). | |
| | ☐ CATASTROPHIC HARM<br>☐ CRITICAL HARM<br>☐ SERIOUS HARM<br>☐ MODERATE OR MINOR HARM | |
| **Rule of Law** | 105) The deployment of the AI system could harm impacted individuals' right to effective remedy or right to a fair trial (equality of arms, right to a natural judge established by law, the right to an independent and impartial tribunal, and respect for the adversarial process).<br><br>☐ UNLIKELY<br>☐ POSSIBLE<br>☐ LIKELY<br>☐ VERY LIKELY<br>☐ UNSURE<br>☐ N/A<br><br>***Example*** | UNLIKELY, POSSIBLE, LIKELY, VERY LIKELY: (Probability answers to be tabulated in the risk index table)<br><br>UNSURE: Before taking any further steps in the proposed project, you should determine, through expert and stakeholder input where appropriate, whether the deployment of the AI system could harm impacted individuals' right to a fair trial (equality of arms, right to a natural judge established by law, the right to an independent and impartial tribunal, and respect for the adversarial process). When this information is ascertained, you should return to your PCRA and revise it accordingly. |
| | 106) If you responded with any of the options besides N/A or Unsure in the previous question, how would you | |
| | characterize the gravity potential of the harm? (For details on gravity potential, please refer to the relevant section of the user's guide).<br><br>☐ CATASTROPHIC HARM<br>☐ CRITICAL HARM<br>☐ SERIOUS HARM<br>☐ MODERATE OR MINOR HARM | |

PCRA Output Example

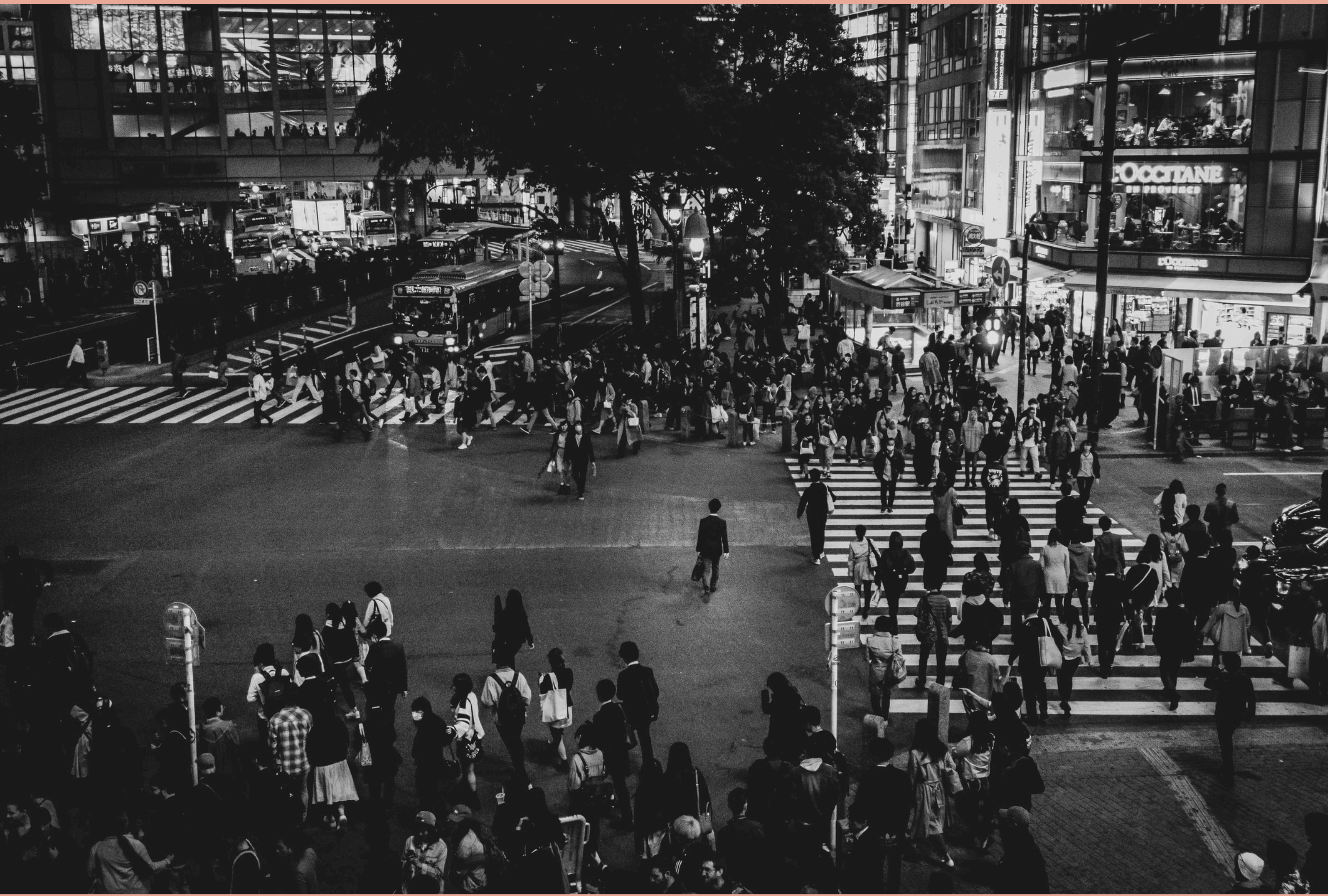

# Human Rights, Democracy, and the Rule of Law Preliminary Context-Based Risk Assessment Report

Thank you for completing the Human Rights, Democracy, and Rule of Law Preliminary Context-Based Risk Analysis (PCRA). This report has been generated based on your responses.

The report is divided into two sections corresponding to the two sections in the PCRA.

Responses from Section 1 are divided into three primary sections: Prohibitive Risks Factors, Major Risks Factors, and Moderate Risk Factors. Each of these sections detail best next steps and items to focus on in your impact assessment process (HUDERIA), as well as in your risk management and assurance practices (HUDERAC).

Section 2 provides a table that compiles the provisional risk index numbers (RINs) that have been calculated based upon your answers in the corresponding section of the PCRA. Each RIN falls with a numeric interval that establish a risk level for the potential adverse impact on the human rights and fundamental freedom of persons, democracy, and the rule of law. Under the table you will find recommendations for ***risk management actions*** for each potential harm identified as well as ***general recommendations for proportionate risk management practices and stakeholder engagement***.

The results of the PCRA are meant to be part of an iterative process which can be returned to and revisiting as you process with the other stages of the HUDERAF process.

# SECTION 1: IDENTIFYING RISK FACTORS

Section 1 produces a summary which compiles and organizes the answers that flagged up risk factors into the three risk factor categories: prohibitive, major, and moderate. Additionally, these three risk factor categories are further divided into modifiable or circumstantial. Those risk factors that emerge *externally* from the technical, sociotechnical historical, legal, economic, or political environments within which the design, development, and deployment of AI systems occur and that are thereby less controllable are called “circumstantial risk factors.” Those risk factors that emerge *internally* from the actual practices of producing and using AI technologies, and that are thus more controllable, are called “modifiable risk factors”.

The report provides prompts for each of these answers that direct you to specific actions to take in your impact assessment process and to specific goals, properties, and areas that you should focus on in your subsequent risk management and assurance processes to reduce and mitigate associated risks.

## Prohibitive Risk Factors:

**<u>Stop</u>** to evaluate the items in your assessment that have indicated the presence of prohibitive risk factors. These items must be prioritised and responded to before you proceed with any other project tasks or activities.

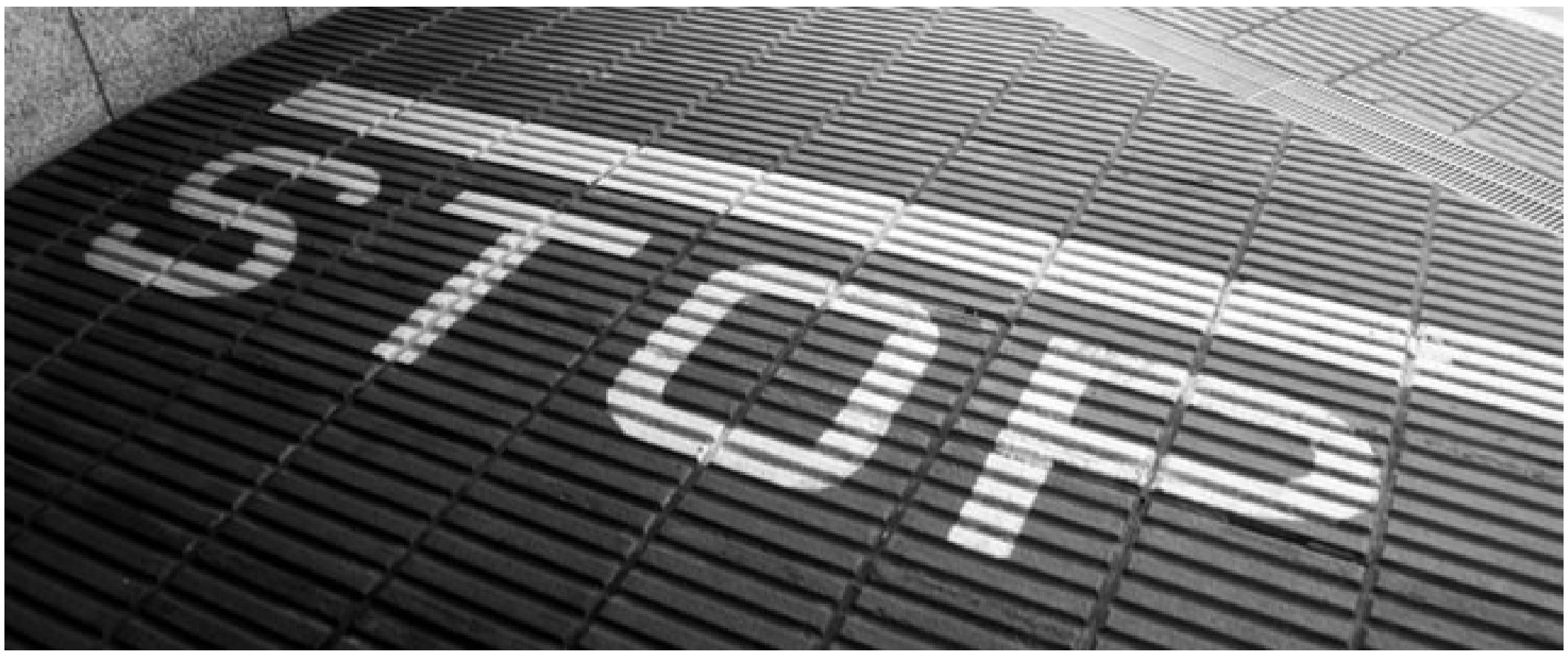

### *Because you answered No to Question 2:*

**STOP. Prohibitive circumstantial risk factor**

If you have not been able to establish a legal basis for the AI system, you should not proceed. Where appropriate, you should consult with experts to establish the lawfulness of your processing objective and proposed project before taking any next steps.

### *Because you answered Unsure to Question 6:*

STOP. Before taking any further steps in the proposed project, you should determine, through expert and stakeholder input where appropriate, whether the AI system you are planning to build is on the list of prohibited systems.

## <u>Major Risk Factors:</u>

### *Because you answered Yes to Question 1:*

**Major circumstantial risk factor:**
Where AI systems serve primary or critical functions in high impact or safety critical sectors, this presents a ***major circumstantial risk factor*** for adverse impacts on the human rights and fundamental freedom of persons.

- **Actions to take for your HUDERIA:** Make sure to focus upon considerations surrounding the prevention of harm and the respect of the right to life and to physical, psychological, and moral integrity
- **Goals, properties, and areas to focus on in your HUDERAC**: Safety (accuracy and system performance, security, reliability, and robustness), sustainability (reflection on context and impacts, change monitoring)

### ***Because you answered Yes to Question 5:***

**Major circumstantial risk factor**

Where AI systems perform a safety critical or high impact function independent of the sector in which they operate, this presents a ***major circumstantial risk factor*** for adverse impacts on the human rights and fundamental freedom of persons.

- **Actions to take for your HUDERIA:** Make sure to focus upon considerations surrounding the prevention of harm and the respect of the right to life and to physical, psychological, and moral integrity
- **Goals, properties, and areas to focus on in your HUDERAC**: Safety (accuracy and system performance, security, reliability, and robustness) and sustainability (reflect on contexts and impact, change monitoring)

### ***Because you answered Unsure to Question 7:***

Before taking any further steps in the proposed project, you should determine, through expert and stakeholder input where appropriate, whether the AI system could be used or repurposed in ways that fall under the list of prohibited systems which are considered to pose significant risks. When this information is ascertained, you should return to your PCRA and revise it accordingly.

## Moderate Risk Factors:

### ***Because you answered Yes to Question 3:***

**Moderate circumstantial risk factor**

Where the sector or domain in which the AI system will operate is historically highly regulated, this presents a ***moderate circumstantial risk factor*** for adverse impacts on the human rights and fundamental freedom of persons.

- **Actions to take for your HUDERIA:** Make sure to consider the regulatory context of your project and to reflect on the expectations of compliant and reasonable practices that arise in that context.
- **Goals, properties, and areas to focus on in your HUDERAC**: Accountability and process transparency, (traceability, accessibility, auditability, and responsible governance)

### ***Because you answered Unsure to Question 4:***

Before taking any further steps in the proposed project, you should determine, through expert and stakeholder input where appropriate, whether any other types of impact assessment for the specific use-case of the AI systems you are planning to develop are required by law or regulation. When this information is ascertained, you should return to your PCRA and revise it accordingly.

# SECTION 2: RISKS OF ADVERSE IMPACTS ON THE HUMAN RIGHTS AND FUNDAMENTAL FREEDOM OF PERSONS, DEMOCRACY, AND THE RULE OF LAW

Section 2 provides a table that compiles the provisional risk index numbers (RINs) that have been calculated based upon your answers in the corresponding section of the PCRA. As you were asked to provide a preliminary estimation of the likelihood of the potential adverse impacts that your system could have on human rights, democracy, and the rule of law, this estimation of likelihood is incorporated into the table and risk index calculation along with your responses to other questions such as the number of rights-holders affected.

The likelihood score that is indicated in the far-right column indicates a risk index number (RIN). An RIN has been calculated for each of the questions in Section 2 that were answered with likelihood estimates, but not for answers of "Unsure" or "Not Applicable".

RINs are calculated by adding the Severity Level (Gravity Potential + Number of Rights-Holders Affected) and Likelihood Level for each potential harm. Details about this calculation and its rationale can be found in the accompanying user's guide. Across all possible answers, RINs range from 2.5 to 10. Risk levels are established as intervals within this range:

| Low Risk | Moderate Risk | High Risk | Very High Risk |
|---|---|---|---|
| $RIN \leq 5$ | $5.5 \leq RIN \leq 6$ | $6.5 \leq RIN \leq 7.5$ | $RIN \geq 8$ |

These RINs correspond to preliminary recommendations for proportionate risk management and assurance practices and stakeholder engagement given the risk level of the prospective AI system. These recommendations can be found in the tables below the summary table generated from your responses to Section 2.

Risk management actions include recommendations on how to reduce the levels of risk present through a variety of methods such as expert and stakeholder consultations. These recommendations can be applied to your results by matching the RIN to the corresponding risk management recommendations.

Additional recommendations for establishing sufficient public transparency and accountability—as well as for ensuring adequate human rights diligence—are included for large-scale projects that could have macro-scale, long-term impacts on individuals and society. The second table includes recommendations for proportionate risk management practices and stakeholder engagement that can be applied to your results.

## Section 2 Output Table

| Adverse Impacts: | Gravity Potential | Rights-Holders Affected | Severity | Likelihood | Risk Index Number (RIN) | Risk Level |
|---|---|---|---|---|---|---|
| Anthropomorphic confusion over interaction with a computational system (dignity) | 3 | 1 | 4 | 2 | 6 | Moderate |
| Exposure to humiliation (dignity) | 3 | 1 | 4 | 1 | 5 | Low |
| Deprivation of rights-holders' abilities to make free, independent, and well-informed decisions about their lives or about the system's outputs, including the ability of rights-holders to effectively challenge decisions informed and/or made by that system and to demand that such decision be reviewed by a person (human freedom) | 3 | 1 | 4 | 1 | 5 | Low |
| Deprivation of the right to life or physical, psychological, or moral integrity (Prevention of harm) | 4 | 1 | 5 | 2 | 7 | High |
| Discrimination, discriminatory effects on impacted rights-holders, or differential performance (Non-discrimination) | 3 | 1 | 4 | 2 | 6 | Moderate |

**Distribution of RINs (risk index numbers)**

| Likelihood \ Severity | 1.5 | 2.5 | 3 | 3.5 | 4 | 4.5 | 5 | 5.5 | 6 |
|---|---|---|---|---|---|---|---|---|---|
| 1 | 2.5 | 3.5 | 4 | 4.5 | 5 | 5.5 | 6 | 6.5 | 7 |
| 2 | 3.5 | 4.5 | 5 | 5.5 | 6 | 6.5 | 7 | 7.5 | 8 |
| 3 | 4.5 | 5.5 | 6 | 6.5 | 7 | 7.5 | 8 | 8.5 | 9 |
| 4 | 5.5 | 6.5 | 7 | 7.5 | 8 | 8.5 | 9 | 9.5 | 10 |

See below a table of specific ***risk management actions*** for each of these potential impacts identified by your results:

| | |
|---|---|
| **For any $RIN \geq 8$ (very high)** | Further examination should be undertaken, through expert and stakeholder consultation, as to whether sufficient risk reduction is possible to make very high risks of harm tolerable or whether this is not feasible, and the risks are unacceptable. |
| **For any $6.5 \leq RIN \leq 7.5$ (high)** | Further examination should be undertaken, through expert and stakeholder consultation, as to whether the risks of harm indicated to be high are tolerable and can be appropriately reduced. Where likelihood = 1 (unlikely) and RPN indicates high risk, confirmation of low risk probability should also be undertaken through expert and stakeholder consultation. |
| **For any $5.5 \leq RIN \leq 6$ (moderate)** | Further examination should be undertaken, through expert and stakeholder consultation, as to whether the risks of harm indicated to be moderate are broadly accepted as such, are tolerable, and can be appropriately reduced. Where likelihood = 1 (unlikely) and RPN indicates moderate risk, confirmation of low risk probability should also be undertaken through expert and stakeholder consultation. |
| **For any $RIN \leq 5$ (low)** | Further examination should be undertaken, through expert and stakeholder consultation, as to whether the risks of harm indicated to be low are broadly accepted as such. |

See below for general recommendations for proportionate risk management practices and stakeholder engagement. These recommendations are based on a prioritization of a degree of risk management and human right diligence that is proportionate to the highest level of risk identified across an AI system's potential impacts.

| | |
|---|---|
| **Because the highest risk level in your results is $6.5 \leq RIN \leq 7.5$ (high)** | Full diligence in risk management and assurance practices is recommended to prioritize risk reduction and mitigation. This should be informed by the risk factors identified in section 1 of the PCRA and by the HUDERIA impact assessment process; Comprehensive stakeholder engagement across the project lifecycle (e.g., partnering with or empowering rights-holders as determined by the |
| | Stakeholder Engagement Process) is also recommended. |

# Stakeholder Engagement Process

**Purpose**

The purpose of the Stakeholder Engagement Process is to identify stakeholder salience and to facilitate proportionate rights-holder involvement and input throughout the project workflow. This process safeguards the equity and the contextual accuracy of the PCRA, HUDERIA and HUDERAC outcomes through stakeholder revisitation and evaluation.

Inputs

- PS Report
- PCRA Report
- Desk Research
- HUDERAC (after initial SEP iteration)

→

Steps

- 2.1 Stakeholder Analysis
- 2.2 Positionality Matrix
- 2.3 Engagement Objective
- 2.4 Engagement Method
- 2.5 Workflow Revisitation and Reporting

→

Outputs

- SEP Report informing subsequent engagement activities and HUDERAC process, providing validated list of salient stakeholder groups, validated engagement objectives and methods, and stakeholder insights.
- Updated PS Report

Stakeholder engagement is a core component of any human rights diligence process. It is not a one-off activity, but rather should be seen as an ongoing set of activities that occur throughout the project lifecycle.[44] Stakeholder engagement may take different forms at different points in a project. It is essential to ensure that rights-holders' views are incorporated at all stages and that any potential risks or adverse impacts are identified and mitigated.[45] In this sense, an iterative approach to stakeholder engagement is essential.[46] In particular, it is important to revisit and revise your stakeholder analysis to ensure that approaches taken continue to reflect the perspectives and interests of salient stakeholders, including vulnerable and underrepresented rights-holders.

A Stakeholder Engagement Process (SEP) involves 5 key activities:

**5 Key Activities within the Stakeholder Engagement Process:**

- Stakeholder Analysis
- Positionality Reflection
- Establishing Engagement Objectives
- Determining Engagement Method
- Workflow Revisitation and Reporting

44 See Esteves, A. M., Factor, G., Vanclay, F., Götzmann, N., & Moreira, S. (2017). Adapting social impact assessment to address a project's human rights impacts and risks. *Environmental Impact Assessment Review*, *67*, p. 77.
45 See Figure 1 in Götzmann, N., Bansal, T., Wrzoncki, E., Veiberg, C. B., Tedaldi, J., & Høvsgaard, R. (2020). Human rights impact assessment guidance and toolbox. The Danish Institute for Human Rights. p. 7.
46 See page 77 in (Esteves, et al 2017).

The initial four steps of the SEP take place internally (i.e. within the organisation or team undertaking the impact assessment) during the design phase of the project workflow. Throughout these steps, teams are to draw on the PCRA report, updated PS Report, and further desk research to answer series of questions and prompts. Step 5 of this process (Workflow Revisitation and Reporting) is undergone by conducting stakeholder engagement activities and providing project reports. This step facilitates stakeholder participation and input to revise approaches towards subsequent actions in the workflow. ***Note that, following its initial iteration as Step 5 of the SEP, the Workflow Revisiting and Reporting activity is repeated throughout the design, development, and deployment phases of your project to ensure ongoing stakeholder involvement across the workflow***.

While stakeholder engagement is vital throughout the process, this can take various forms (outlined below) and involve varying degrees of engagement that fit proportionality recommendations. In accordance with best practices in human rights due diligence, you should always aim for the greatest depth of engagement that is feasible within the scope and resource constraints of the project.

### Step 1: Carry Out Stakeholder Analysis

Stakeholder Salience Analysis is the process of evaluating the salience of identified stakeholders. It aims to help you understand the relevance of each identified stakeholder to your project and to its use contexts. It does this by providing a structured way for your team members to assess the relative interests, rights, vulnerabilities, and advantages of identified stakeholders as these interests, rights, vulnerabilities, and advantages may be impacted by, or may impact, the model or tool your team is planning to develop and deploy.

**Analysing Stakeholder Salience:**

Use the PCRA, the updated Project Summary Report, and further desk research (if necessary) to inform answers to the following questions, assessing relative stakeholder salience:

- *What stakeholder groups have existing interests in relation to the domain or sector in which the system/tool will be deployed?*
- *What stakeholder groups are most likely to be impacted by the deployment of the system or tool?*
- *What stakeholder groups have the greatest needs in relation to potential benefits/applications of the system/tool or the domain or sector in which it will be deployed?*
- *What stakeholder groups are most and least powerful?*
- *What stakeholder groups have existing influence within relevant communities, political processes, or in relation to the domain in which the system will be deployed?*
- *What stakeholder groups' influence is limited?*
-
- *What stakeholder groups' rights may be impacted in relation to the domain or sector in which the system/tool will be deployed?*

**Consider the following diagram**:

- What stakeholder groups possess both protected and contextually vulnerable characteristics?

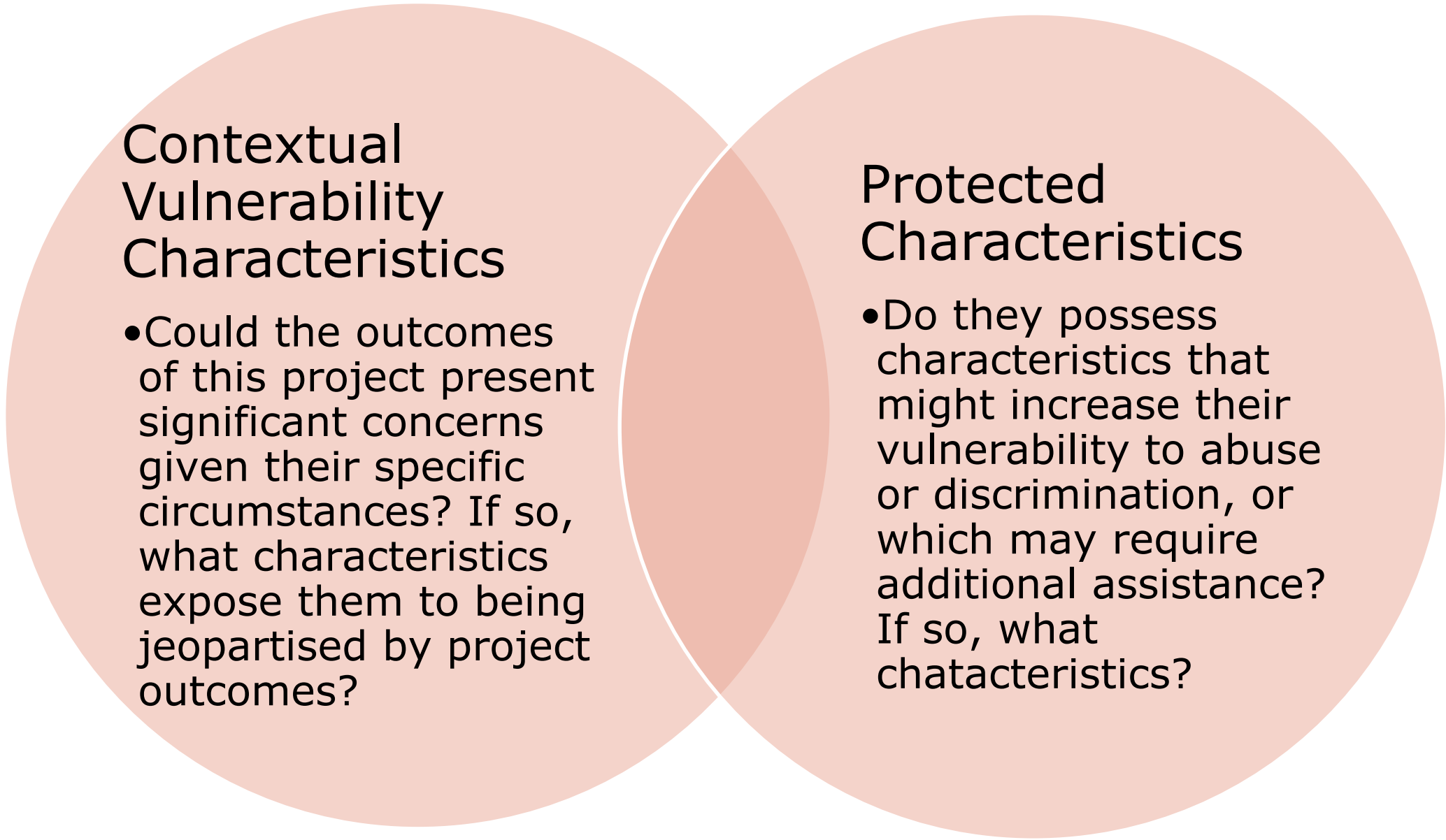

Analysis of stakeholder salience aims to identify which stakeholders are likely to be most impacted, vulnerable, and those that currently have least influence.[47] Special attention needs to be paid to engaging these stakeholders in subsequent stages.

47 See page 124 and 125 in (Götzmann et al 2020)

### Step 2: Engage in Positionality Reflection

When taking positionality into account, each of your team members should reflect on their own positionality matrix by answers the questions contained in the graphic below.

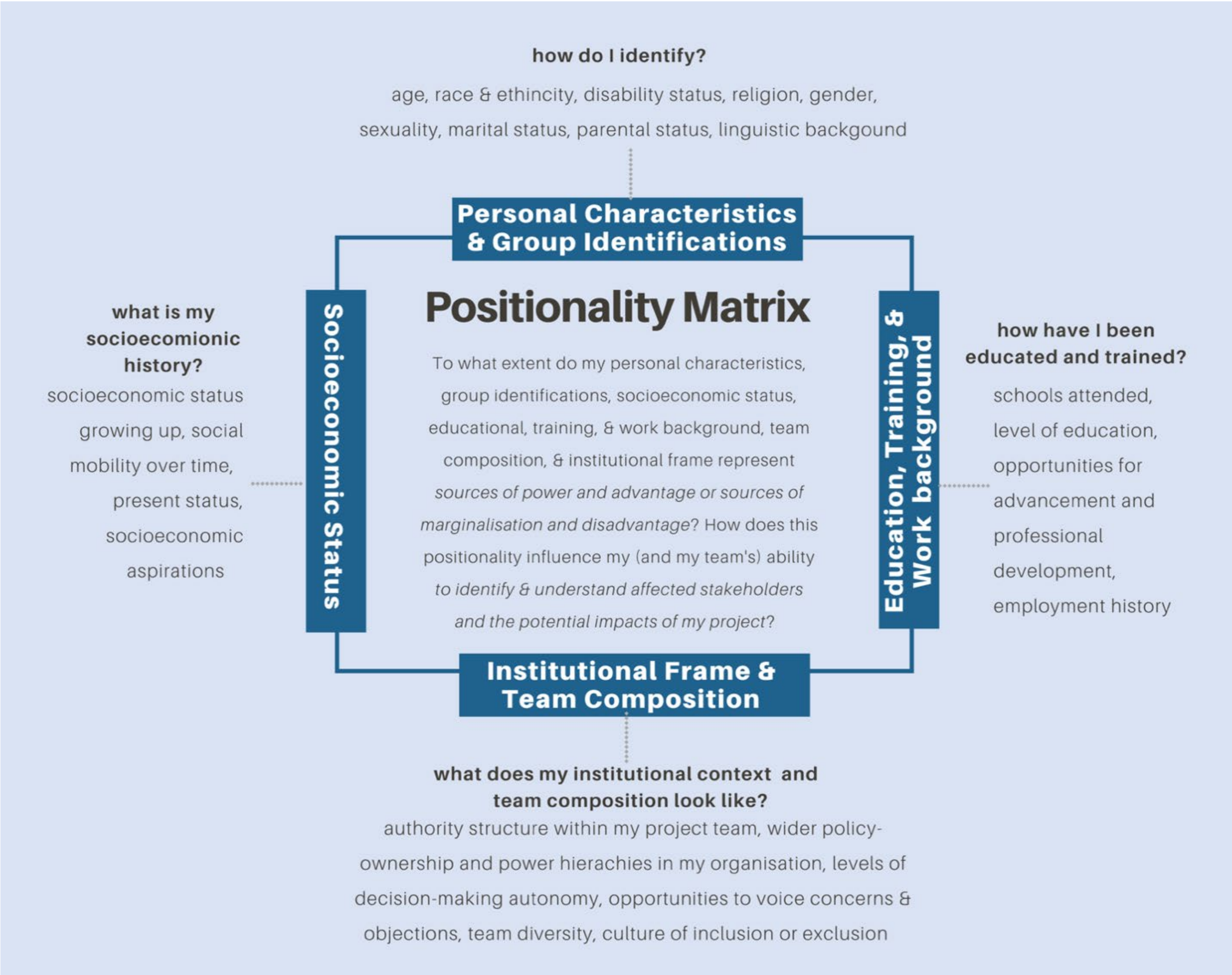

**Reflecting on positionality:**

After reflecting on their positionality matrices, team members should collaboratively answer the following questions:

a. *How does the positionality of team members relate to that of affected rights-holders and other stakeholders?*
b. *Are there ways that your position as a team could lead you to choose one option over another when assessing the risks that the prospective AI system poses to human rights, democracy and the rule of law?*
c. *What (if any) missing stakeholder viewpoints would strengthen your team's assessment of this system's potential impact on human rights and fundamental freedoms?*

### Step 3: Establish an Engagement Objective

Before selecting methods of engagement to use, you must first define objectives of your stakeholder engagement. Through this process you should articulate:

- Why you are engaging stakeholders
- The purpose and expected outcomes of engagement activities
- The level of influence stakeholders will have on engagement processes and outcomes

There are many ways a project team can engage with the stakeholders they have identified, and the objectives and methods of engagement that your team chooses will depend on several factors. These divide into three categories, which are presented here with accompanying descriptions:

| Factors determining the objectives of engagement | |
|---|---|
| **Team-based assessments of risks** | • Assessment of how stakeholder involvement should be made proportionate to the scope of a project's potential risks and hazards to human rights, democracy, and the rule of law |
| **Team-based assessments of positionality** | • Evaluation of team positionality—for instance, cases where the identity characteristics of team members do not sufficiently reflect or represent significantly impacted groups. How can the project team "fill the gaps" in knowledge, domain expertise, and lived experience through stakeholder participation? |
| **Establishment of stakeholder engagement goals** | • Determination of engagement objectives that enable the appropriate degree of stakeholder engagement and co-production in project evaluation and oversight processes<br>• Choosing participation goals from a spectrum engagement options (informing, partnering, consulting, empowering) that equip your project with a level of engagement which meets team-based assessments of risk and positionality. |

When weighing these three factors, your team should prioritise the establishment of a ***clear and explicit stakeholder participation goal*** and document this**.** This is crucial, because all stakeholder engagement processes can run the risk either of being cosmetic tools employed to legitimate projects without substantial and meaningful participation or of being insufficiently participative, i.e. of being one-way information flows or nudging exercises that serve as public relations instruments. The purpose of stakeholder involvement in human rights diligence is just the opposite: to amplify the participatory agency of affected individuals and organisations in impact assessment, risk management, and assurance processes.

To avoid such hazards of superficiality, your team should shore up its proportionate approach to stakeholder engagement with deliberate and precise goal-setting. There are a range of engagement options that can help your project obtain a level of citizen participation which meets team-based assessments of impact and positionality as well as practical considerations and stakeholder needs[48]:

## Levels of Stakeholder Engagement

| | DEGREE OF PARTICIPATION | MEANS OF PARTICIPATION | LEVEL OF AGENCY |
|---|---|---|---|
| EMPOWER | *Stakeholders are engaged with as decision-makers and are expected to gather pertinent information and be proactive in co-production.* | *Co-production exercises occur through citizens' juries, citizens' assemblies, and participatory co-design.Teams provide support for stakeholders' decision making.* | *Stakeholders exercise a high level of agency and control over agenda-setting and decision-making.* |
| PARTNER | *Stakeholders and teams share agency over the determination of areas of focus and decision making.* | *External input is sought out for collaboration and co-production. Stakeholders are collaborators in projects.They are engaged through focus groups.* | *Stakeholders exercise a moderate level of agency in helping to set agendas through collaborative decision-making* |
| CONSULT | *Stakeholders can voice their views on pre-determined areas of focus, which are considered in desicion-making.* | *Engagement occurs through online surveys or short phone interviews, door-to-door or in public spaces. Broader listening events can support consultations.* | *Stakeholder are included as sources of information input under narrow, highly controlled conditions of participation.* |
| INFORM | *Stakeholders are made aware of decisions and developments.* | *External input is not sought out. Information flows in one direction. This can be done through newsletters, the post, app notifications or community forums.* | *Stakeholders are treated as information subjects rather than active agents.* |

**Questions:**

- *Why are you engaging stakeholders?*
- *What are the expected outcomes of engagement activities?*
- *How will stakeholders be able to influence the engagement process and the outcomes?*

### Step 4: Determine Engagement Method

Once you have established your engagement objective, you are in a better position to assess which method or methods of stakeholder involvement are most

[48] Excellent resources on citizen and stakeholder engagement upon which these processes draw can be found here: https://www.thersa.org/globalassets/reports/2020/IIDP-citizens-assembly.pdf; https://www.local.gov.uk/sites/default/files/documents/New%20Conversations%20Guide%2012.pdf; https://datajusticelab.org/wp-content/uploads/2021/06/PublicSectorToolkit_english.pdf; https://www.communityplanningtoolkit.org/sites/default/files/Engagement.pdf.

appropriate to achieve your goals. Determining appropriate engagement methods necessitates that you (1) evaluate and accommodate of stakeholder needs, and (2) pay attention to practical considerations of resources, capacities, timeframes, and logistics that could enable or constrain the realisation of your objective:

| Factors determining engagement methods | |
|---|---|
| **Evaluation and accommodation of stakeholder needs** | • Identification of potential barriers to engagement such as constraints on the capacity of vulnerable stakeholder groups to participate, difficulties in reaching marginalised, isolated, or socially excluded groups, and challenges to participation that are presented by digital divides or information and communication gaps between public sector organisations and impacted communities<br>• Identification of strategies to accommodate stakeholder needs such as catering the location or media of engagement to difficult-to-reach groups, providing childcare, compensation, or transport to secure equitable participation, tailoring the provision of information and educational materials to the needs of participants |
| **Practical considerations of resources, capacities, timeframes, and logistics** | • the resources available for facilitating engagement activities<br>• the timeframes set for project completion<br>• the capacities of your organisation and team to properly facilitate public engagement<br>• the stages of project design, development, and implementation at which stakeholders will be engaged |

You and your team may face pitfalls when confronting any of these factors. For example, limits on available resources and tight timelines could be at cross-purposes with the degree of stakeholder involvement that is recommended by team-based assessments of potential hazards and positionality limitations. Likewise, the chosen degree of appropriate citizen participation may be unrealistic or out-of-reach given the engagement barriers that arise from high levels of stakeholder needs. In these instances, you should take a deliberate and reflective approach to deciding on how to balance engagement goals with practical considerations—always prioritizing the need for optimizing stakeholder input and rights-holder involvement in human rights diligence processes. And you should make explicit the rationale behind your choices and document this.

Reflecting the varied levels of engagement set out above, there are varied engagement methods which can be used. Stakeholder engagement is likely to require more than a single method and may combine approaches aimed at informing, consulting, partnering and empowering. However, stakeholder engagement must always contain at least one method pertaining to the stated engagement objective (i.e. a method pertaining to a lower level of engagement cannot account for the need for engagement aimed to be met through a higher engagement objective):

| Examples of Relevant Engagement Methods | | | |
|---|---|---|---|
| **Mode of Engagement** | **Degree of Engagement** | **Practical Strengths** | **Practical Weaknesses** |
| **Newsletters (email)**<br><br>Regular emails (e.g.: fortnightly or monthly) that contain updates, relevant news, and calls to action in an inviting format. | *Inform* | Can reach many people; can contain large amount of relevant information; can be made accessible and visually engaging. | Might not reach certain portions of the population; can be demanding to design and produce with some periodicity; easily forwarded to spam/junk folders without project team knowing (leading to overinflated readership stats). |
| **Letters (post)**<br><br>Regular letters (e.g. monthly) that contain the latest updates, relevant news and calls to action. | *Inform* | Can reach parts of the population with no internet or digital access; can contain large amount of relevant information; can be made accessible and visually engaging. | Might not engage certain portions of the population; Slow delivery and interaction times hamper the effective flow of information and the organisation of further engagement. |
| Posts in local community (e.g. local newspapers, community newsletters and noticeboards). | *Inform* | Can reach localised communities; can reach people who do not have digital access. | Limited to geographic region; requires physical access; less cost-effective than digital means. |
| **App notifications**<br><br>Projects can rely on the design of apps that are pitched to stakeholders who are notified on their phone with relevant updates. | *Inform* | Easy and cost-effective to distribute information to large numbers of people; rapid information flows bolster the provision of relevant and timely news and updates. | More significant initial investment in developing an app; will not be available to people without smartphones. |
| **Community fora**<br><br>Events in which panels of experts share their knowledge on issues and then stakeholders can ask questions. | *Inform* | Can inform people with more relevant information by providing them with the opportunity to ask questions; brings community together in a shared space of public communication. | More time-consuming and resource intensive to organise; might attract smaller numbers of people and self-selecting groups rather than representative subsets of the population; effectiveness is constrained by forum capacity. |
| **Online surveys**<br><br>Survey sent via email, embedded in a | *Consult* | Cost-effective; simple mass-distribution. | Risk of pre-emptive evaluative framework when designing questions; does not |
| website, shared via social media | | | reach those without internet connection or computer/smartphone access. |
| **Phone interviews**<br><br>Structured or semi-structured interviews held over the phone. | *Consult* | Opportunity for stakeholders to voice concerns more openly. | Risk of pre-emptive evaluative framework when designing questions; might exclude portions of the population without phone access or with habits of infrequent phone use. |
| **Door-to-door interviews**<br><br>Structured or semi-structured interviews held in-person at people's houses. | *Consult* | Opportunity for stakeholders to voice concerns more openly; can allow participants the opportunity to form connections through empathy and face-to-face communication. | Potential for limited interest to engage with interviewers; time-consuming; can be seen by interviewees as intrusive or burdensome. |
| **Feedback Forms**<br><br>Provided along with letters or posts or after events. | *Consult* | Provides mechanisms for stakeholders to respond to information provided; can be quick and easy for stakeholders to respond; can include fixed choice or open-ended responses. | Might limit responses to pre-defined questions; risk of pre-emptive evaluative framework when designing questions. |
| **In-person interviews**<br><br>Short interviews conducted in-person in public spaces. | Suggested for *Consult* | Can reach many people and a representative subset of the population if stakeholders are appropriately defined and sortition is used. | Less targeted; pertinent stakeholders must be identified by area; little time/interest to engage with interviewer; can be viewed by interviewees as time-consuming and burdensome. |
| **Focus group**<br><br>A group of stakeholders brought together and asked their opinions on a particular issue. Can be more or less formally structured. | *Consult* and *Partner* | Can gather in-depth information; can lead to new insights and directions that were not anticipated by the project team. | Subject to hazards of *group think* or peer pressure; complex to facilitate; can be steered by dynamics of differential power among participants. |
| **Citizen panel or assembly**<br><br>Large groups of people (dozens or even thousands) who are representative of a town/region. | *Inform, Partner* and *Empower* | Provides an opportunity for co-production of outputs; can produce insights and directions that were not anticipated by the project team; can provide an information base for conducting further outreach | Participant rolls must be continuously updated to ensure panels or assemblies remains representative of the population throughout their lifespan; resource-intensive for establishment and |
| | | (surveys, interviews, focus groups, etc.); can be broadly representative; can bolster a community's sense of democratic agency and solidarity. | maintenance; subject to hazards of *group think* or peer pressure; complex to facilitate; can be steered by dynamics of differential power among participants. |
| **Citizen jury**<br><br>A small group of people (between 12 and 24), representative of the demographics of a given area, come together to deliberate on an issue (generally one clearly framed set of questions), over the period of 2 to 7 days (involve.org.uk). | *Inform, Partner* and *Empower* | Can gather in-depth information; can produce insights and directions that were not anticipated by the project team; can bolster participants' sense of democratic agency and solidarity. Most effective when there is a clearly defined question which the jury is tasked with answering. | Subject to hazards of *group think*; complex to facilitate; Requires independent facilitation; Risk of pre-emptive evaluative framework; Small sample of citizens involved risks low representativeness of wider range of public opinions and beliefs. |

**Questions:**

Define an engagement method to be used for the Workflow Revisitation and Reporting. Consider:

- *What resources are available and what constraints will limit potential approaches?*
- *What accessibility requirements might stakeholders have?*
- *Will online or in-person methods (or a combination of both) be most appropriate to engage salient stakeholders?*
- *Which activities meet your team's engagement objective?*
- *How will your team make sure that the chosen method accommodates different types of stakeholders?*
- *How will each of the chosen activities feed useful information to consider in your stakeholder impact assessment?*

Outputs from the first four steps of the Stakeholder Engagement Process are used to create an initial Stakeholder Engagement Process Report, which is to be revised and updated at each subsequent iteration of Workflow Revisitation and Reporting.

### Step 5: Workflow Revisitation and Reporting

As indicated above, the Workflow Reporting and Revisitation step is conducted initially as a culmination of the Stakeholder Engagement Process, so that you can work directly with stakeholders to re-assess the first four steps of the SEP, and it provides an opportunity for you to revisit the PS Report and the project workflow with them. This initial stakeholder engagement will also flow into your impact assessment process; working through the PS Report and the project workflow with these stakeholders will help to prepare them for meaningful and well-informed involvement in your HUDERIA. The later iterations of Workflow Reporting and Revisitation (in the development and deployment phases of your project) will subsequently enable your project team to report to stakeholders the actions and evidence that are being compiled in the project assurance case (HUDERAC) in an accessible manner that facilitates feedback and input.

**Revisiting with Stakeholders as part of your SEP**

During Step 5 of the Stakeholder Engagement Process, it is essential to ensure that decisions and outcomes from steps 1 – 4 reflect the experiences and perspectives of salient stakeholders. Stakeholder engagement should involve rights-holders, duty bearers and other relevant parties. In instances where initial stakeholder mapping and risk analysis has identified no potential adverse impacts on rights-holders, a minimum level of stakeholder engagement should involve consultation with relevant organisations (e.g. civil society and human rights organisations) and rights-holders.

**Revisiting Steps 1 and 2 (Stakeholder Analysis and Positionality) with Stakeholders**

A useful strategy for revisiting Stakeholder Analysis and Positionality Reflection with rights-holders and other relevant parties is to merge these activities into engagement exercises that focus on identifying gaps and drawing on the lived experience of affected individuals and communities. This will allow you to:

- More fully involve initially-identified salient stakeholder groups
- Broaden identified stakeholders through a "snowball" approach
- Discuss team positionality with stakeholders (while being mindful of confidentiality needs and not disclosing sensitive or personal information about individuals).
- Map stakeholder salience through participatory processes which reflect stakeholder perspectives and interests.

Crucially, a comprehensive stakeholder analysis requires initial stakeholder engagement to identify potentially salient stakeholders who may not have been visible to the project team (particularly underrepresented rights-holders). Interviews or community engagement activities can be an effective mechanism to identify additional relevant stakeholders to engage. This "snowballing" technique is an effective mechanism to ensure that stakeholder analysis reflects varied perspectives and interests, recognising that there may not be consensus around which stakeholders are most salient.

**Revisiting Stakeholder Salience Questions, Engagement Objective and**

**Engagement Method with Stakeholders**

This step is undertaken through further discussion of stakeholder needs with stakeholders themselves. Particular attention should be paid here to vulnerable rights-holders. At this stage, the stakeholder engagement should focus on identifying existing mechanisms and constraints for engagement (i.e., networks, community groups, barriers to engagement, difficulties). This should include an assessment of any potential risks of participation (i.e. if taking part in an engagement activity could put vulnerable individuals at risk). Where there are potential risks associated with participation of vulnerable individuals, consideration must be given to alternative engagement methods or forms of representation (e.g. through civil society organisations).

Outputs from the above activities are used to update the SEP Report, informing approaches for the following engagement activity in the workflow (HUDERIA or Workflow Revisitation).

**Reporting to Stakeholders**

Communication and stakeholder evaluation of process outcomes at each phase is a vital component to build and sustain trust and to underpin ongoing stakeholder engagement. During the initial iteration of the stakeholder engagement process, you are to use the PS Report as a baseline to convey the project status in an accessible manner that facilitates input. After this initial iteration, you are to use the assurance case as a basis for updating the PS Report. The most up-to-date versions of the assurance case and Project Summary Report (which provides a synopsis of up-to-date understandings of the system, its domain and use contexts, identified stakeholders, the scope of impacts on human rights, democracy, and rule of law, and the project governance workflow), will enable you to share project reports that are relevant at each phase of the workflow.

Reporting to stakeholders must involve:

- Publication and dissemination of project outputs (taking account of accessibility requirements of diverse stakeholders);
- Facilitation of stakeholder feedback (e.g. through feedback forms, questionnaires or further engagement)' and incorporation of feedback documented through an update to the SEP report, in turn informing the subsequent HUDERIA and HUDERAC processes.

**Questions:**

- *What medium of reporting is likely to most effectively engage stakeholders?*
- *What are the accessibility requirements of stakeholders (e.g. taking account of physical and cognitive disabilities)?*
- *What mechanisms will be put in place to facilitate feedback from stakeholders, and how will feedback be addressed and/or incorporated? Consider how these mechanisms will yield feedback pertaining to*
- *stakeholders' validation of the system considering its domain and use contexts; the identified scope of impacts to human rights, democracy, and rule of law; and the project governance workflow.*

| **Stakeholder Engagement Process (SEP) Template** | |
|---|---|
| **Questions** | **Responses** |
| ***STEP 1: Stakeholder Analysis*** | |
| ANALYSYING STAKEHOLDER SALIENCE<br>Use the PCRA, the updated Project Summary Report, and further desk research (if necessary) to inform answers to the following questions, assessing relative stakeholder salience. (Refer to the Venn diagram in the related section as needed.): | |
| *What stakeholder groups have existing interests in relation to the domain or sector in which the system/tool will be deployed?* | |
| *What stakeholder groups are most likely to be impacted by the deployment of the system or tool?* | |
| *What stakeholder groups have the greatest needs in relation to potential benefits/applications of the system/tool or the domain or sector in which it will be deployed?* | |
| *What stakeholder groups are most and least powerful?* | |
| *What stakeholder groups have existing influence within relevant communities, political processes, or in relation to the domain in which the system will be deployed?* | |
| *What stakeholder groups' influence is limited?* | |
| *What stakeholder groups' rights may be impacted in relation to the domain or sector in which the system/tool will be deployed?* | |
| ***STEP 2: Positionality Reflection*** | |
| ENGAGE IN POSITIONALITY REFLECTION<br>When taking positionality into account, each of your team members should reflect on their own positionality by considering their individual responses to the questions posed in the Positionality Matrix provided. After reflecting on their positionality matrices, team members should collaboratively answer the following questions: | |
| *How does the positionality of team members relate to that of affected rights-holders and other stakeholders?* | |
| *Are there ways that your position as a team could lead you to choose one option over another when assessing the risks that the prospective AI system poses to human rights, democracy, and the rule of law?* | |
| *What (if any) missing stakeholder viewpoints would strengthen your team's assessment of this system's potential impact on human rights and fundamental freedoms?* | |
| ***STEP 3: Engagement Objective*** | |
| ESTABLISH AN ENGAGEMENT OBJECTIVE | |
| Refer to the Factors Determining the Objectives of Engagement and Levels of Stakeholder Engagement tables as needed. Be sure to outline a clear and explicit stakeholder participation goal. | |
| *Why are you engaging stakeholders?* | |
| *What is the purpose and what are the expected outcomes of engagement activities?* | |
| *How will stakeholders be able to influence the engagement process and the outcomes?* | |
| *What engagement objective do you believe would be appropriate for this project considering challenges or limitations to assessments related to positionality, and proportionality to the project's potential degree of impact?* | |
| ***STEP 4: Engagement Method*** | |
| DETERMINE AN ENGAGEMENT METHOD<br>Define an engagement method to be used for the Workflow Revisitation and Reporting. | |
| *What resources are available and what constraints will limit potential approaches?* | |
| *What accessibility requirements might stakeholders have?* | |
| *Will online or in-person methods (or a combination of both) be most appropriate to engage salient stakeholders?* | |
| *Which activities meet your team's engagement objective?* | |
| *How will your team make sure that the chosen method accommodates different types of stakeholders?* | |
| *How will each of the chosen activities feed useful information to consider in your stakeholder impact assessment?* | |
| ***STEP 5: Workflow Revisitation and Reporting*** | |
| REPORTING TO STAKEHOLDERS<br>Outline a stakeholder communication plan that includes receiving and incorporating feedback and reporting back to stakeholders throughout the Stakeholder Engagement Process. | |
| *What medium of reporting is likely to most effectively engage stakeholders?* | |
| *What are the accessibility requirements of stakeholders (e.g. taking account of physical and cognitive disabilities)?* | |
| *What mechanisms will be put in place to facilitate feedback from stakeholders, and how will feedback be addressed and/or incorporated? Consider how these mechanisms will yield feedback pertaining to stakeholders' validation of the system considering its domain and use contexts; the identified scope of impacts to human rights, democracy, and rule of law; and the project governance workflow.* | |

# Human Rights, Democracy, and the Rule of Law Impact Assessment

**Purpose**

The purpose of the Human Rights, Democracy and the Rule of Law Impact Assessment is to provide detailed evaluations of the potential and actual impacts that the design, development, and use of an AI system could have on human rights, fundamental freedoms and elements of democracy and the rule of law. This process contextualizes and corroborates potential harms identified through the PCRA, enables the identification and analysis of further harms through the dialogical integration of stakeholders, establishes an impact mitigation plan, and sets up access to remedy.

**Inputs**

- PS Report
- PCRA Report
- SEP Report

→

**Steps**

- 3.1 HUDERIA
- identification of potential impacts and assessment of the severity of potential impacts identified
- Co-design of Impact mitigation plan and set up of access to remedy
- Monitoring and impact re-assessment protocols established

→

**Outputs**

- HUDERIA Report informing subsequent HUDERAC process, providing a detailed assessment of harms posed at the point of assessment alongside stakeholder insights

Your Human Rights, Democracy, and the Rule of Law Impact Assessment (HUDERIA) provides you with the opportunity to draw on the learning and insights you have gained in your PCRA and SEP processes, and on the lived experience of engaged rights-holders, to delve more deeply into the potential impacts of your project on human rights, fundamental freedoms, and elements of democracy and the rule of law. Your HUDERIA should enable you

- To re-examine and re-evaluate the potential harms you have already identified in your PS Report and your PCRA
- To contextualize and corroborate these potential harms in dialogue with stakeholders
- To identify and analyse further potential harms by engaging in extended reflection and by giving stakeholders the chance to uncover new harms that have not yet been explored and to pinpoint gaps in the completeness and comprehensiveness of the previously enumerated harms
- To explore, with stakeholders, the severity (scope, scale, and remediability) of the potential adverse impacts, so that the risks of these can be better assessed, prioritized, managed, and mitigated
- To begin formulating impact mitigation measures (avoid, restore, remediate) based upon the severity of the identified harms
- To set up access to remedy for affected rights-holders and other relevant parties.

## Getting the HUDERIA Process Right

Before you launch into building the content of your HUDERIA, it will be important that you establish a responsible process for carrying out the impact assessment with the participation of affected rights-holders and other relevant stakeholders. To do this, you will need to ensure that your HUDERIA process aligns with five key criteria: ***participation, non-discrimination, empowerment, transparency, and accountability***.[49] Use the table below (adapted from the important work of the Danish Institute for Human Rights), to orient yourself to these criteria and to the guiding principles that will have to be operationalised to secure a responsible HUDERIA process:

| Key criteria and guiding principles for responsible HUDERIA processes | | |
|---|---|---|
| **Participation** | Meaningful participation of affected or potentially affected rights-holders is integrated during all stages of the impact assessment process, including planning and scoping; data collection and context analysis; impact analysis; impact prevention, mitigation and remediation; and reporting and evaluation. | • Engage a broad range of rights-holders including vulnerable and marginalised groups<br>• Consider the rights, freedoms, and involvement of rights-holders throughout the AI innovation ecosystem (designers, implementers, users, those potentially impacted by the product or service, those who could be displaced or deskilled by the product or service)<br>• Involve rights-holders or their proxies throughout the impact assessment process<br>• Involve rights-holders, duty-bearers and other relevant parties in designing measures to address impacts (e.g. through prevention, mitigation and remediation) and follow-up to evaluate the effectiveness of these measures<br>• Include rights-holder representatives or representative organisations, or rights-holder proxies in consultation and engagement, including consideration of the legitimacy of their claim to represent the relevant individuals and/or groups<br>• Ensure that engagement and participation in the impact assessment |

[49] These criteria and the table in which they are described are directly adapted from research by The Danish Institute for Human Rights (DIHR). In 2020, the DIHR produced their "Guidance on Human Rights Impact Assessment of Digital Activities: Introduction." In this paper, they present 10 key principles for a Human Rights Impact Assessment with five key criteria for the HRIA process. The criteria used here and the table in which they are elaborated have been drawn from their guidance. See Kernell, E. L., Veiberg, C. B., & Jacquot, C. (2020). "Guidance on Human Rights Impact Assessment of Digital Activities: Introduction." The Danish Institute for Human Rights.

| | | |
|---|---|---|
| | | is guided by the local context, including through using the impacted individuals' preferred mechanisms (e.g. modes of communication) where possible<br>• Make sure the assessment process is being undertaken at particular times to ensure equitable participation (e.g., when women are not at work and young people are not at school) |
| **Non-discrimination** | Engagement and consultation processes are inclusive, gender-sensitive, and account for the needs of individuals and groups at risk of vulnerability or marginalisation. | • Facilitate ongoing dialogue between rights-holders, duty-bearers and other relevant parties (e.g. through collaborative problem analysis and design of mitigation measures)<br>• To the extent digital and virtual means of engagement are utilised (e.g., online consultations and surveys), assess and address accessibility issues, particularly with regard to the most vulnerable rights-holders<br>• Involve women, men, and transgender people, including through gender-identity-aware engagement methods as necessary<br>• Take steps to ensure that the modes of engagement and participation address any barriers that may be faced by vulnerable and marginalised individuals (e.g., by offering transport or holding meetings in culturally appropriate locations, and considering 'technology barriers' for older persons, persons with disabilities, or persons who do not have access to the internet or to digital technologies) |
| **Empowerment** | Capacity building of individuals and groups at risk of vulnerability or marginalisation is undertaken to ensure their meaningful participation. | • Identify the needs of vulnerable and marginalised individuals and act to address these to ensure inclusive rights-holder involvement<br>• Include sufficient time in the assessment process for capacity building to allow individuals and groups to be meaningfully involved (e.g., to first present the digital products or services in a way that the audience understands, and to follow-up later with the same groups when they have had time to discuss and organise, in order to receive feedback and potential concerns)<br>• Make sure that rights-holders have access to independent and competent legal, technical, and other advice as |
| | | necessary, and, if not, include provisions for making such support available<br>• Provide for capacity building of rights-holders to know their rights (e.g., by thoroughly explaining the right to privacy before explaining how the AI product or service will be developed to ensure respect for the same right), as well as of duty-bearers to meet their human rights duties |
| **Transparency** | The impact assessment process is as transparent as possible to adequately engage affected or potentially affected rights-holders, without causing any risk to security and well-being of rights-holders or other participants (such as NGOs and human rights defenders). Impact assessment findings are appropriately publicly communicated. | • Provide for information sharing between stakeholders at relevant and regular intervals (such as through Workflow Revisitation and Reporting processes)<br>• Make available information about the AI project, product, or service to participating stakeholders that is adequate for giving a comprehensive understanding of potential implications and human rights impacts (such as through sharing the PS Report)<br>• Publicly communicate HUDERIA findings and impact management plans (action plans) to the greatest extent possible (e.g., published, with any reservations based on risk to rights-holders or other participants clearly justified)<br>• Secure a firm top-level management commitment with regard to transparency before the start of the HUDERIA process<br>• Communicate the phases of the impact assessment, including timeframes, to relevant stakeholders in a clear and timely manner<br>• Ensure that communication and reporting consider and respond to the local context (e.g., by making sure that information is made available in relevant languages and formats, in non-technical summaries and in physical and/or web-based formats that are accessible to stakeholders) |
| **Accountability** | The impact assessment team is supported by human rights expertise, and the roles and responsibilities for impact assessment, prevention, mitigation and management are | • Make sure that responsibility for the implementation, monitoring and follow-up of mitigation measures is assigned to particular individuals or functions within the organisation (e.g., data engineers are tasked with |
| | assigned and adequately resourced. The impact assessment identifies the entitlements of rights-holders and the duties and responsibilities of relevant duty-bearers (e.g., developers, companies buying digital products or services, those using or applying digital products and services, and government authorities). | changing the design to limit potential misuse)<br>• Make sure that sufficient resources are dedicated to undertaking the HUDERIA, as well as implementing the impact management plan (e.g., adequate time, as well as financial and human resources)<br>• Meaningfully and appropriately engage relevant duty-bearers in the impact assessment process, including in impact prevention, mitigation and remediation (e.g., data protection authorities are engaged since some systemic impacts can best be dealt with through data protection policies and regulation)<br>• Ensure that project team members who are undertaking the HUDERIA have the relevant interdisciplinary skills and expertise (including human rights, technical, legal, domain, language, and local knowledge) to undertake the impact assessment in the given context and with regard to the specific AI product or service (e.g., data engineers and software developers might need to be involved)<br>• Prioritize that the HUDERIA is carried out by a diverse and inclusive group of team members |

## Three Steps in the HUDERIA Process

Once you have set up your HUDERIA process in accordance with the criteria outlined above, your project team will be ready to begin working on the content of the impact assessment. Although the impact assessment process may take many different shapes—depending on the needs and resources of your organisation and the results of your PCRA and SEP—there are a few steps that should be common to any full HUDERIA process:

**Step 1. Project team prepares for and begins HUDERIA**

- Project team members prepare for the HUDERIA by revisiting Table 1 (Principles and Priorities, Corresponding Rights and Freedoms, and Elaborations), reviewing the outcome of Activity 3 of PS reporting, and revisiting PCRA and SEP results; Project team completes first part of the HUDERIA and the additional project team questions supplement.

**Step 2. Rights-holders prepared for and begin HUDERIA**

- Rights-holders are prepared to meaningfully engage in the HUDERIA through Workflow Revisiting and Reporting and by reviewing PS Report and content of Table 1 with facilitators; Rights-holders provide input on the HUDERIA questions from the first part through facilitators.

**Step 3. Integrate results of Steps 1 and 2 through rights-holder/duty-bearer dialogue**

- Results of Steps 1 and 2 are compared and integrated into rights-holder/ duty-bearer dialogue; Impact mitigation measures consonant with severity of potential impacts are explored; impact management plan is co-designed by project team and engaged rights-holders; access to remedy and operational grievance mechanisms are set up.

## Building the Content of your HUDERIA

As noted above, the content of the HUDERIA is assembled to allow your team (1) to re-examine and re-evaluate the potential harms you have already identified in your PS Report and your PCRA, (2) to contextualize and corroborate these potential harms in dialogue with stakeholders, (3) to identify and analyse further potential harms through extended project team reflection and by giving stakeholders the chance to uncover new harms and to pinpoint gaps in the completeness of the previously enumerated harms, (4) to explore, with stakeholders, the severity (scope, scale, and remediability) of the potential adverse impacts, so that the risks of these can be better assessed, prioritized, managed, and mitigated, and (5) to set up access to remedy for affected rights-holders and other relevant parties. In the interest of covering this spectrum of desiderata, the HUDERIA questions and prompts are organised around three categories:

**3 Categories of HUDERIA questions and prompts:**

1. **Identifying potential adverse impacts**
2. **Assessing the severity of potential impacts identified**
3. **Mitigating potential impacts and setting up access to remedy**

In the first part of the HUDERIA, the first two of these question types are applied to each of the 20 impact prompts from section 2 of the PCRA. Just as in the PCRA, these prompts are grouped under the appropriate principle/priority from Table 1, so that the potential harms to the rights, freedoms, or elements of democracy and the rule of law that are associated with each principle/priority can be considered together under the appropriate heading. This makes it easier for project team members and engaged rights-holders to identify potential gaps in the completeness of the previously enumerated adverse impacts (as these fall under each principle/priority) and to unearth corresponding harms that have not yet been recognized. It also allows considerations about the severity, mitigation, and remedy of contextually specific impacts to be based around the appropriate principle/priority and thus to include wider reflections on how each impact is situated within the more general area of concern. For instance, particular harms related to anthropomorphic deception, humiliation, and instrumentalization can be thought about as *adverse impacts on the dignity of affected rights-holders,* and associated considerations about the severity, mitigation, and remedy can thus occur, in part, through the wider-angled lens of concerns about protecting and respecting human dignity.

## *Identifying Potential Adverse Impacts*

For the HUDERIA questions that come under the category of ***Identifying potential adverse impacts***, each prompt from section 2 of the PCRA is now reformulated as an explanation-seeking "how, if at all," question that enables the project team members and stakeholders to interrogate *the ways in which* the right, freedom, or element of democracy and the rule of law associated with the principle/priority under consideration could be adversely impacted.[50] This enables more open-ended and flexible deliberation on the range of potential adverse impacts at the same time as it allows participants in the impact assessment process to zoom in on the specific contexts of the impacts to facilitate better, more granular understandings of their scope, scale, and remediability.

[50] N.B., for these questions, project team members should refer back to Activity 3, ***Scope Impacts to Human Rights, Democracy, and Rule of Law,*** in the PS Report and work from the previous answers given there.

The ***Identifying potential adverse impact*** questions also include a final gap-identifying "in what other ways, if at all" question that encourages project team members and engaged rights-holders to think beyond the pre-identified adverse impacts first introduced in the PCRA and to discern novel harms. Here is how this appears in the section on the **Protection of Human Freedom and Autonomy**:

| HUDERIA Questions | Responses |
|---|---|
| ***Protection of Human Freedom and Autonomy*** | |
| IDENTIFYING POTENTIAL ADVERSE IMPACTS | |
| *How, if at all, could the system adversely affect or hinder the abilities of rights-holders to make free, independent, and well-informed decisions about their lives or about the system's outputs?* | |
| *How, if at all, could the system adversely affect or hinder the capacities of rights-holders to flourish, to fully develop themselves, and to pursue their own freely determined life plans?* | |
| *How, if at all, could the deployment of the system result in the arbitrary deprivation of rights-holders' physical freedom or personal security, or the denial of their freedoms of expression, thought, conscience, or assembly* | |
| *In what other ways, if any, could the use of this system adversely impact the freedom and autonomy of affected rights-holders?* | |

### *Assessing the Severity of Potential Impacts*

As explained in the *United Nations Guiding Principles on Business and Human Rights* (UNGP), assessing the severity of potential human rights impacts involves consideration of their scale, scope, and remediability, where scale is defined as "the gravity or seriousness of the impact," scope as "how widespread the impact is, or the numbers of people impacted," and remediability, as the "ability to restore those affected to a situation at least the same as, or equivalent to, their situation before the impact".[51] The UNGP further notes: "Where it is necessary to prioritize

[51]. See Principle 14 of (UNGP, 2011), but also Council of Europe's "Recommendation CM/Rec(2020)1 of the Committee of Ministers to member States on the human rights

actions to address actual and potential adverse human rights impacts, business enterprises should first seek to prevent and mitigate those that are most severe or where delayed response would make them irremediable."[52]

The HUDERIA questions that appear in the ***Assessing the Severity of Potential Impacts*** section are intended to bring clarity to the prioritisation of impact mitigation and assurance actions by allowing the severity levels of an AI system's potential impacts to be differentiated, elucidated, and refined by relevant rights-holders or their proxies. The elements of ***differentiation*** and ***inclusive rights-holder participation*** are crucial here. Different human rights impacts may have varying degrees of relative severity and must therefore be carefully distinguished and understood both individually and in relation to each other. Likewise, understanding the scale, scope, and remediability of potential human rights impacts requires that the lived experience and contextual insights of a diverse and inclusive group of potentially affected rights-holders be included in the assessment process. This is because, as the United Nations Human Rights Office of the High Commissioner notes, "scale, scope and remediability may differ for different individuals or groups at heightened risks of becoming vulnerable or marginalized, and…there may be different risks faced by different groups, such as men and women."[53] Including a wide range of perspectives in the impact assessment process—especially from members of vulnerable and marginalized groups—allows for essential first-hand knowledge about how different groups may experience harms differently to be incorporated into assessments of potential severity.

### *Scale*

Assessing the scale of a potential adverse impact on human rights, fundamental freedoms and elements of democracy and the rule of law involves both consideration of the gravity or seriousness of the impact and a contextually sensitive understanding of the way that different groups (especially vulnerable, historically-discriminated-against, and marginalized groups) could suffer prejudice

---

impacts of algorithmic systems" and the UN Human Rights Office of the High Commissioner (UNHROHC) (2020) "Identifying and Assessing Human Rights Risks related to End-Use".

[52] See (UNGP 2011, p. 26). The commentary to this paragraph expands, "While business enterprises should address all their adverse human rights impacts, it may not always be possible to address them simultaneously. In the absence of specific legal guidance, if prioritization is necessary business enterprises should begin with those human rights impacts that would be most severe, recognizing that a delayed response may affect remediability. Severity is not an absolute concept in this context, but is relative to the other human rights impacts the business enterprise has identified." See also (UNHROHC 2020).

[53] (UNHROHC 2020, p. 7). The interpretive guidance for the UNGP underlines this point: "Depending on the operational context, the most severe human rights impact may be faced by persons belonging to groups that are at higher risk of vulnerability or marginalization, such as children, women, indigenous peoples, or people belonging to ethnic or other minorities. If the enterprise decides it needs to prioritize its responses to human rights impacts, it should take into account the vulnerability of such groups and the risk that a delayed response to certain impacts could affect them disproportionately." See: UNHROHC (2012) "The Corporate Responsibility to Protect Human Rights: An Interpretive Guide". p. 84.

in their exercise of the right or freedom under consideration in different ways and to different extents.

To orient your considerations of the gravity of potential adverse impacts, review the following Gravity Table[54]:

| **Gravity Level** | **Descriptions** |
|---|---|
| **Catastrophic Harm** | Catastrophic prejudices or impairments in the exercise of fundamental rights and freedoms that lead to the deprivation of the right to life; irreversible injury to physical, psychological, or moral integrity; deprivation of the welfare of entire groups or communities; catastrophic harm to democratic society, the rule of law, or to the preconditions of democratic ways of life and just legal order; deprivation of individual freedom and of the right to liberty and security; harm to the biosphere. |
| **Critical Harm** | Critical prejudices or impairments in the exercise of fundamental rights and freedoms that lead to the significant and enduring degradation of human dignity, autonomy, physical, psychological, or moral integrity, or the integrity of communal life, democratic society, or just legal order |
| **Serious Harm** | Serious prejudices or impairments in the exercise of fundamental rights and freedoms that lead to the temporary degradation of human dignity, autonomy, physical, psychological, or moral integrity, or the integrity of communal life, democratic society, or just legal order or that harm to the information and communication environment |
| **Moderate or Minor Harm** | Moderate or minor prejudices or impairments in the exercise of fundamental rights and freedoms that do not lead to any significant, enduring, or temporary degradation of human dignity, autonomy, physical, psychological, or moral integrity, or the integrity of communal life, democratic society, or just legal order |

In the HUDERIA questions contained in this subsection, you will be prompted to characterize the gravity of each potential adverse impact that has been identified.

[54] Note that the use of this table to help you think through the gravity level of a possible adverse human rights impact differs from the use of the Gravity Table in the PCRA, which asked that the potential gravity of a harm be estimated assuming its maximal impact. In the HUDERIA, both project team members and potentially impacted rights-holders should engage in contextually aware explorations of the actual gravity or seriousness of the potential impact.

You will also be asked to take into account the different ways and different extents to which individuals or groups who possess characteristics that could make them more vulnerable to the adverse impact could suffer the harm:

| ASSESSING THE SEVERITY OF POTENTIAL IMPACTS | |
|---|---|
| ***SCALE:*** *Before answering the questions below, review the descriptions of harm in Gravity Table contained in the accompanying user's guide.* | |
| *For each potential impact identified in the previous section, would you characterize it as a catastrophic, critical, serious, or minor harm?* | |
| *For each potential impact identified in the previous section, are there individuals or groups who possess characteristics that could make them more vulnerable to the impact? If so, what are these characteristics and could those who possess them suffer the harm more acutely or seriously than others?* | |
| *For each potential impact identified in the previous section, which individuals or groups could encounter the gravest impact on the right or freedom under consideration?* | |

Your responses to these questions (and the responses of potentially impacted rights-holders) will subsequently serve an important function during the impact mitigation planning stage when the redress and prioritization of harms is under consideration. At this point, however, your focus should be on (1) thinking through the varying contexts of the potential harms, (2) making sure that considerations about the differing vulnerabilities of rights-holders to suffering harm is integrated into the gravity assessment, and (3) ensuring that relevant rights-holders or their proxies are appropriately and sufficiently represented in this assessment process.

### *Scope*

In the Scope subsection, you will be asked to consider how extensive or widespread each of the potential adverse impacts identified could be in terms of the numbers of people potentially impacted, the timescale of the impacts, and the exposure of particular groups of affected rights-holders to harm:

| |
|---|
| ***SCOPE:*** *Before answering the questions below, review the answers to* Activity 1 (specifically in the Project section) of the PS Report and *questions 10 and 11 from the PCRA* (on numbers of rights-holders affected by the AI system and timescale of the system's potential impacts). |

| | |
|---|---|
| *For each potential impact identified in the previous section, does the answer given to question 10 of the PCRA seem to provide an accurate estimate of the scope of the impact? If not, how many rights-holders could encounter a deprivation of the right or freedom under consideration?* | |
| *For each potential impact identified in the previous section, are there groups who possess characteristics that could make them vulnerable to higher levels of exposure to the impact? If so, how much exposure could these groups face?* | |
| *For each potential impact identified in the previous section, consider the overall timescale of the AI system's impacts on the right or freedom under consideration. Are there cumulative or aggregate impacts of the system on rights-holders and their progeny that could expand the effects of the system beyond the scope of impact identified in the previous answers?* | |

The first question asks that you re-evaluate the answer given to question 10 of the PCRA (about the number of rights-holders potentially impacted by the system) in the context of the specific right or freedom under consideration. This is intended as a sense check. In some cases, the estimated total number of rights-holders potentially affected by a system's use may be greater than the number of those who could suffer from a particular harm, as, for instance, where a prejudice in the exercise of a right or freedom is suffered disproportionately by a specific group (e.g. a commercial insurance risk model used in healthcare that disproportionately mismeasures the level of chronic illness for members of a particular racial group because they are systematically underrepresented in the training dataset). In other instances, the estimated number of rights-holders potentially affected by a system's use may (if misidentified) be much less than the number of those who could suffer from a particular harm, as, for instance, where a system has knock-on effects over time that lead to wider legacy impacts (e.g. harms done to members of a particular racial group by a discriminatory predictive decision-support model used in medicine leads to the group's long-term reluctance and distrust in the medical system and much wider impacts on the welfare of its members).

The second question asks you to consider whether there are groups who possess characteristics that could make them vulnerable to higher levels of exposure to the impact under consideration. The term "level of exposure" here is understood as the proportion of a group that is adversely impacted by an AI system, where, in the case that a small fraction of the group is impacted, members have low levels of exposure and in the case that a very large fraction of the group is impacted, members have high levels of exposure. As an example, members of a group that is characterised by low socioeconomic status may have a high level of exposure to the potential adverse impacts of an AI model that is used to allocate benefits in a social-services setting, because a large portion of that groups needs to access the benefits of social services in order to survive.

The final question on scope asks you to consider the overall timescale of the AI system's impacts on the right or freedom under consideration, taking the answer given to question 11 of the PCRA as a starting point. In particular, it asks you to think through whether there may be "cumulative or aggregate impacts of the system on rights-holders and their progeny that could expand the effects of the system beyond the scope of impact identified in the previous answers". Identifying cumulative or aggregate impact can be much more difficult than identifying harms directly or proximately caused by the use of an AI system, and providing a full answer to this question, for the project team, may involve additional research and consultation with domain experts and other relevant stakeholders. This difficulty results from the fact that cumulative impacts are often incremental and more difficult to perceive, and they frequently involve complex contexts of multiple actors or projects operating in the same area or sector or affecting the same populations.[55] Some "big picture" questions to reflect on when assessing cumulative or aggregate impacts include:

- Could the production and use of the system contribute to wider adverse human rights impacts when its deployment is coordinated with (or occurs in tandem with) other systems that serve similar functions or purposes? For example, if the human rights impacts of a conversational AI system that serves a customer service function are considered in combination with the proliferation of many other similar systems in a given sector, concerns about wider cumulative impacts on labour displacement, the right to work, and other social and economic rights become relevant.

- Could the production and use of the system replicate, reinforce, or augment socio-historically entrenched legacy harms that create knock-on effects in impacted individuals and groups? For example, an AI system that uses facial analysis technology to match photographs to other pictures contained in databases scraped from social media websites could add to the legacy harms of companies that have used data recklessly and eroded public trust regarding the respect of privacy and data protection rights in the digital sphere, thereby creating wider chilling effects on elements of open communication, information sharing, and interpersonal connection that are essential components for the sustainability of democratic forms of life.

- Could the production and use of the system be understood to contribute to wider aggregate adverse impacts on the biosphere and on planetary health when its deployment is considered in combination with other systems that may have similar environmental impacts? For example, a complex AI system that involves moderate levels of energy consumption in training, operating, or data storage may be seen to contribute to significant environmental impact when considered alongside the energy consumption of similar systems across AI innovation ecosystems.

[55] This explanation of cumulative impacts draws heavily on Götzmann, N., Bansal, T., Wrzoncki, E., Veiberg, C. B., Tedaldi, J., & Høvsgaard, R. (2020). Human rights impact assessment guidance and toolbox. The Danish Institute for Human Rights. pp. 86-88.

*Remediability*

Questions in the Remediability subsection seek information about the degree of reparability or restoration that is possible for affected rights-holders as the result of efforts to overcome the adverse impact under consideration. As noted above, remediability involves the "ability to restore those affected to a situation at least the same as, or equivalent to, their situation before the impact." Much as with considerations surrounding the gravity of a potential impact, gaining an understanding of how remediable or reversable a harm is will depend on knowledge both about the specific context of the harm and about the affected rights-holders who are subjected to it. Members of different groups may have different levels of resilience, depending on their positions in society and the circumstances of the harm (with vulnerable and marginalised groups often possessing less resilience than other dominant, privileged, or majority groups). Those seeking to understand the degree of remediability for any specific adverse impact on fundamental rights and freedoms must therefore pay close attention to who could be harmed and in what contexts.

Establishing the degree of remediability for a potential adverse impact involves considerations about the effort needed to overcome and (potentially) reverse the harm[56]:

| Degree of Remediability | Effort |
|---|---|
| **Very Low** | Suffered harm may be irreversible and may not be overcome (e.g., long-term psychological or physical ailments, death, etc.) |
| **Low** | Suffered harm can be overcome albeit with serious difficulties and enduring effects (e.g., economic loss, property damage, worsening health, loss of social trust, deterioration of confidence in the legal order, etc.) |
| **Medium** | Suffered harm can be overcome despite some difficulties (e.g., extra costs, fear, lack of understanding, stress, minor physical ailments, etc.) |
| **High** | Suffered harm can be overcome without any problem (e.g., time spent amending information, annoyances, irritations, etc.) |

You will be prompted to use this Table as a starting point, to answer these questions:

[56] This Table is adapted from Table 6 in Mantelero & Esposito *(2021)* – An evidence-based methodology for human rights impact assessment (HRIA) in the development of AI data-intensive systems.

| **REMEDIABILITY:** *Before answering the questions below, review the descriptions of harm in the Remediability Table contained in the accompanying user's guide.* | |
|---|---|
| *For each potential impact identified in the previous section, how remediable is the adverse effect on impacted right-holders? (Refer to the Remediability Table for categories of remediability to consider)* | |
| *For each potential impact identified in the previous section, are there groups who possess characteristics that could make their members vulnerable to lower levels of remediability for the impact (i.e., more subject to difficult-to-overcome or irreversible harm)? If so, to what degree is the impact on members of each group irreversible or irremediable?* | |

### *Additional Questions for the Project Team*

At the end of Part I of the HUDERIA, you'll find a supplemental section which contains additional questions about human rights diligence to be answered by the project team. The purpose of these questions is to establish that all human rights impacts that the project is linked to across its value chain have been appropriately considered and assessed[57]:

| **ADDITIONAL QUESTIONS FOR THE PROJECT TEAM** | |
|---|---|
| *These questions are meant to serve as a supplement to Part I of the HUDERIA. They are intended to be answered by the project team and then to be revisited during the dialogue between duty-bearers and rights-holders in Step 3 of the HUDERIA* | |
| *Has a thorough assessment of the human rights compliant business practices of all businesses, parties and entities involved in the value chain of the AI product or service been undertaken? This would include all businesses, parties, and entities directly linked to your business lifecycle through supply chains, operations,* | |

[57] For further guidance on due diligence processes for responsible organisational conduct as it relates to the value chain see OECD (2018), *Due Diligence Guidance for Responsible Business Conduct*.

| | |
|---|---|
| *contracting, sales, consulting, and partnering. (If not, do you have plans to do this?)* | |
| *Do you have processes in place and appropriate resources dedicated to maintaining human rights diligence with regard to all businesses, parties, and entities directly linked to your business lifecycle?* | |
| *Do you require your vendors and suppliers to demonstrate compliance with legal protections in place to protect rights-holders from modern slavery, human trafficking, and labour exploitation?* | |

## Mitigating Potential Impacts and Setting Up Access to Remedy

Once sections on ***Identifying Potential Adverse Impacts***, ***Assessing the Severity of Potential Impacts Identified,*** and ***Additional Questions for the Project Team*** have been completed, the project team and engaged rights-holders should be brought together to discuss and compare results and to begin impact prevention and mitigation prioritization and planning.[58]

Diligent impact prevention and mitigation planning begins with a scoping and prioritization stage. Project team members and engaged rights-holders should go through all of the identified potential adverse impacts and map out the interrelations and interdependencies between them as well as surrounding social factors (such as contextually specific rights-holder vulnerabilities and precariousness) that could make impact mitigation more challenging. Where prioritization of prevention and mitigation actions is necessary (for instance, where delays in addressing a potential harm could reduce its remediability), decision-making should be steered by the relative severity of the impacts under consideration. As a general rule, while impact prevention and mitigation planning may involve prioritization of actions, ***all potential adverse impacts on human rights, fundamental freedoms, and elements of democracy and the rule of law must be addressed***.

### *The mitigation hierarchy*

When deciding upon the range of available actions that can be taken to prevent or mitigate potential adverse impacts, project team members and engaged rights should refer to what many different types of impact assessment processes call the "mitigation hierarchy" (avoid, reduce, restore, compensate). During the initial iteration of the HUDERIA—i.e at the early, design stage of the project lifecycle—the impacts under consideration will be *potential impacts*, namely, impacts that have not yet happened, so mitigation options of "avoid" and "reduce" will be more

[58] The following material in this section draws heavily on (Götzmann et al 2020).

relevant, whereas in later iterations of HUDERIA monitoring, revisitation and re-evaluation—i.e. during the deployment stage—actual adverse impacts may arise, making mitigation options of "restore" and "compensate" relevant alongside "avoid" and "reduce". Descriptions of the elements of the mitigation hierarchy are as follows:

| Mitigation Hierarchy | Description |
|---|---|
| **Avoid** | Making changes to the design, development, or deployment processes behind the production and use of the AI system (or to the AI system itself), at the outset, to avoid the potential adverse impact. |
| **Reduce** | Implementing actions in the design, development, or deployment processes behind the production and use of the AI system (or making changes to the AI system itself) to minimise potential or actual impacts. |
| **Restore** | Taking actions to restore or rehabilitate affected rights-holders to a situation at least the same as, or equivalent to, their situation before the impact |
| **Compensate** | Compensating in kind or by other means, where other mitigation approaches are neither possible nor effective. |

The use of the term "mitigation ***hierarchy***" is prescriptive and meaningful here in the sense that it is directing actions to give precedence to avoiding potential adverse impacts altogether, in the first instance, and then to reducing and remediating them. It is also notable that, at later stages of the project lifecycle, where options of restoration and compensation become more relevant, more than one of these mitigation options may be at play contemporaneously (as, for instance, where an affected rights-holder needs to be restored simultaneously as immediate actions to minimise further harms must be taken). At all events, decisions about which mitigation action(s) to take should be guided, first and foremost, by human rights considerations, and choices made to avoid and reduce adverse impacts should take first position and be kept distinct from choices to compensate or remunerate impacted rights-holders for suffered harms.

### *Co-designing and implementing an impact management plan*

Once potential adverse impacts have been mapped out and organised, and mitigation actions have been considered, project team members and engaged rights-holders should begin co-designing an impact mitigation plan (IMP). The IMP will become the part of your HUDERIA that specifies the actions and processes needed to address the adverse impacts which have been identified and that assigns responsibility for the completions of these tasks and processes.[59] As such, the IMP will serve a crucial documenting function, which, as part of the HUDERIA, will then become an evidentiary element of the argument-based assurance practices carried out in your HUDERAC. The IMP should include:

- A summary of combined impact findings from the project team and engaged rights-holders with a detailed mapping of potential adverse impacts and any significant interrelationships between them

- A clear presentation of the measures and actions that will be taken and, where needed, a presentation of the prioritization of actions accompanied by an explanation of prioritization rationale (The project team and engaged rights-holders should also draw on recommendations generated in the results of Section 1 of the PCRA when setting out needed measures and actions. This will feed directly into the goal definition and property determination aspects of your HUDERAC.)

- For any potential adverse impacts identified in direct links of the business or organisation to businesses, parties and entities involved in the value chain of the AI product or service, a presentation of plans to exercise leverage or alter relationships to mitigate impacts

- A clarification of the roles and responsibilities of the various actors involved in impact mitigation, management, and monitoring

- (From the project team directly): a statement that indicates the allotted commitment of time and resources that will be made available to carry out the IMP

- A plan for monitoring impact mitigation efforts and for re-assessing and re-evaluating the HUDERIA during subsequent development and deployment phases of the project lifecycle

- An accessible presentation of access to remedy and operational-level grievance mechanisms that will be available to impacted rights-holders if they suffer any harms identified in the HUDERIA or any other adverse impacts that were not detected or recognized in the impact assessment process

[59] Daniel M. Franks (2011), 'Management of the social impacts of mining', in P. Darling (Ed), *SME Mining Engineering Handbook* (3rd edn), pp.1817-1825; Daniel M. Franks and Frank Vanclay (2013), 'Social impact management plans: Innovation in corporate and public policy', *Environmental Impact Assessment Review*, 43, p.57. Cited in (Götzmann et al 2020, p. 95).

## Revisiting the HUDERIA across the Project Lifecyle

Carrying out a HUDERIA at the beginning of an AI innovation project is only a first, albeit critical, step in a much longer, end-to-end process of responsive evaluation and re-assessment. In your impact assessment process, you must pay continuous attention both to the dynamic and changing character of the AI production and implementation lifecycle and to the shifting conditions of the real-world environments in which your AI system will be embedded.

There are two factors that necessitate this demand for responsiveness in sustainable AI innovation:

1. **Production and implementation factors**: Choices made at any point along the design, development, and deployment workflow may impact prior decisions and assessments—leading to a need for re-assessment, reconsideration, and amendment. For instance, design and development choices could be made that were not anticipated in the initial impact assessment (such choices might include adjusting the target variable, choosing a more complex algorithm, or grouping variables in ways that may impact specific groups). These changes may influence how an AI system performs or how it impacts affected individuals and groups. Processes of AI model design, development, and deployment are also iterative and frequently bi-directional, and this often results in the need for revision and update. For these reasons, responsible and sustainable AI design, development, and use must remain agile, attentive to change, and at-the-ready to move back and forth across the decision-making pipeline as downstream actions affect upstream choices and evaluations.
2. **Environmental factors:** Changes in project-relevant social, regulatory, policy or legal environments (occurring during the time in which the system is in production or use) may have a bearing on how well the model works and on how the deployment of the system impacts the rights and freedoms of affected individuals and groups. Likewise, domain-level reforms, policy changes, or changes in data recording methods may take place in the population of concern in ways that affect whether the data used to train the model accurately portrays phenomena, populations, or related factors in an accurate manner. In the same vein, cultural or behavioral shifts may occur within affected populations that alter the underlying data distribution and hamper the performance of a model, which has been trained on data collected prior to such shifts. All of these alterations of environmental conditions can have a significant effect on how an AI system performs and on the way it impacts affected rights-holders and communities.

Bearing these factors in mind, it becomes clear that revisitation of your HUDERIA at relevant points across your project lifecycle will play a pivotal role in its continued efficacy and reliability. As part of the IMP component of your initial HUDERIA activities, a plan is established for monitoring impact mitigation efforts and for re-assessing and re-evaluating your HUDERIA during subsequent development and deployment phases of the project lifecycle. Such follow-up processes of re-assessment and re-evaluation should include an appropriate stakeholder engagement aspect, which is proportionate to the risks posed by the system. Such processes should also remain as responsive as possible to the way that system is interacting with its operating environments and with impacted rights-holders. In rapidly evolving or changing contexts, there may arise a need for more frequent re-assessment and re-evaluation interventions, and your project team must remain flexible and ready to amend previous iterations of the HUDERIA and IMP.

Additionally, in the monitoring phase—during system deployment—considerations of actual adverse impacts are added to considerations of potential impacts. This means that options for mitigating actual impacts (i.e. restoration and compensation) will now be included as options in the mitigation hierarchy. Keep in mind, however, that the avoidance, reduction, and remediation of any adverse human rights impacts should be given primacy in mitigation efforts and that compensation cannot replace these mitigation actions.

| Human Rights, Democracy, and the Rule of Law Impact Assessment (HUDERIA) Template | |
|---|---|
| **Questions** | **Responses** |
| ***Respect for and protection of human dignity*** | |
| IDENTIFYING POTENTIAL ADVERSE IMPACTS | |
| *How, if at all, could this system prompt confusion or uncertainty in rights-holders about whether they are interacting with an AI system rather than a human being?* | |
| *How, if at all, could this system expose rights-holders to humiliation (i.e put them in a state of helplessness or insignificance; dehumanize them or deprive them of a sense of individual identity)?* | |
| *How, if at all, could the system expose rights-holders to instrumentalization or objectification (treating them solely as exchangeable, as statistical aggregates, as means to ends, or as objects to be freely manipulated or steered)?* | |
| *How, if at all, could this system expose rights-holders to displacement, redundancy, or a sense of* | |
| *worthlessness in regard to their participation in work life, creative life, or the life of the community?* | |
| *In what other ways, if any, could the use of this system adversely impact the dignity of affected rights-holders?* | |
| ASSESSING THE SEVERITY OF POTENTIAL IMPACTS | |
| ***SCALE:*** *Before answering the questions below, review the descriptions of harm in Impact Gravity Table contained in the accompanying user's guide.* | |
| *For each potential impact identified in the previous section, would you characterize it as a catastrophic, critical, serious, or moderate/minor harm?* | |
| *For each potential impact identified in the previous section, are there individuals or groups who possess characteristics that could make them more vulnerable to the impact? If so, could these individuals or groups experience the harm more intensely or seriously than others?* | |
| *For each potential impact identified in the previous section, which individuals or groups could encounter the gravest prejudice in their exercise of the right or freedom under consideration?* | |
| ***SCOPE:*** *Before answering the questions below, review the answers to* Activity 1 (specifically in the Project section) of the PS Report and *questions 10 and 11 from the PCRA* (on numbers of rights-holders affected by the AI system and timescale of the system's potential impacts). | |
| *For each potential impact identified in the previous section, does the answer given to question 10 of* | |
| *the PCRA seem to provide an accurate estimate of the scope of the impact? If not, how many rights-holders could encounter a deprivation of the right or freedom under consideration?* | |
| *For each potential impact identified in the previous section, are there groups who possess characteristics that could make them vulnerable to higher levels of exposure to the impact? If so, how much exposure could these groups face?* | |
| *For each potential impact identified in the previous section, consider the overall timescale of the AI system's impacts on the right or freedom under consideration. Are there cumulative or aggregate impacts of the system on rights-holders and their progeny that could expand the effects of the system beyond the scope of impact identified in the previous answers?* | |
| ***REMEDIABILITY:*** *Before answering the questions below, review the descriptions of harm in Remediability Table contained in the accompanying user's guide.* | |
| *For each potential impact identified in the previous section, how remediable is the potential adverse effect on impacted right-holders? (Refer to the Remediability Table for categories of remediability to consider)* | |
| *For each potential impact identified in the previous section, are there groups who possess characteristics that could make them vulnerable to lower levels of remediability for the impact (i.e. more subject to difficult-to-overcome or* | |
| *irreversible harm)? If so, to what degree is the impact irreversible or irremediable?* | |
| ***Protection of Human Freedom and Autonomy*** | |
| IDENTIFYING POTENTIAL ADVERSE IMPACTS | |
| *How, if at all, could the system adversely affect or hinder the abilities of rights-holders to make free, independent, and well-informed decisions about their lives or about the system's outputs?* | |
| *How, if at all, could the system adversely affect or hinder the capacities of rights-holders to flourish, to fully develop themselves, and to pursue their own freely determined life plans?* | |
| *How, if at all, could the deployment of the system result in the arbitrary deprivation of rights-holders' physical freedom or personal security, or the denial of their freedoms of expression, thought, conscience, or assembly?* | |
| *In what other ways, if any, could the use of this system adversely impact the freedom or autonomy of affected rights-holders?* | |
| ASSESSING THE SEVERITY OF POTENTIAL IMPACTS | |
| ***SCALE:*** *Before answering the questions below, review the descriptions of harm in Impact Gravity Table contained in the accompanying user's guide.* | |
| *For each potential impact identified in the previous section, would you characterize it as a catastrophic, critical, serious, or moderate/minor harm?* | |
| *For each potential impact identified in the previous section, are there individuals or groups who possess characteristics that could make them more vulnerable to the impact? If so, what are these characteristics and could those who possess them suffer the harm more acutely or seriously than others?* | |
| *For each potential impact identified in the previous section, which individuals or groups could encounter the gravest impact on the right or freedom under consideration?* | |
| ***SCOPE:*** *Before answering the questions below, review the answers to* Activity 1 (Project Section) of the PS Report and answers to *questions 10 and 11 from the PCRA* (on numbers of rights-holders affected by the AI system and timescale of the system's potential impacts). | |
| *For each potential impact identified in the previous section, does the answer given to question 10 of the PCRA seem like accurate and realistic representation of the potential scope of the impact? If not, how many rights-holders could encounter a deprivation of the right or freedom under consideration?* | |
| *For each potential impact identified in the previous section, are there groups who possess characteristics that could make them vulnerable to* | |
| *higher levels of exposure to the impact? If so, how much exposure could these groups face?* | |
| *For each potential impact identified in the previous section, consider the overall timescale of the AI system's impacts on the right or freedom under consideration (answer to question 11 of the PCRA). Are there cumulative or aggregate impacts of the system on rights-holders or their progeny that could extend the effects of the system beyond the scope of impact identified in the previous answers?* | |
| ***REMEDIABILITY:*** *Before answering the questions below, review the Remediability Table contained in the accompanying user's guide.* | |
| *For each potential impact identified in the previous section, how remediable is the potential adverse effect on affected right-holders? (Refer to the Remediability Table for categories of remediability to consider)* | |
| *For each potential impact identified in the previous section, are there groups who possess characteristics that could make them vulnerable to lower levels of remediability for the impact (i.e. more subject to difficult-to-overcome or irreversible harm)? If so, what degree of irreversible harm do these groups face?* | |
| ***Prevention of harm and protection of the right to life and physical, psychological, and moral integrity*** | |
| IDENTIFYING POTENTIAL ADVERSE IMPACTS | |
| *How, if at all, could interaction with the AI system deprive rights-holders of their right to life or their physical, psychological, or moral integrity?*<br><br>*How, if at all, could the AI system's interactions with the environment (either in the processes of its production or in its deployment) harm the biosphere or adversely impact the health of the planet?* | |
| *In what other ways, if any, could the use of this system adversely impact the right to life or physical, psychological, and moral integrity of affected rights-holders?* | |
| ASSESSING THE SEVERITY OF POTENTIAL IMPACTS | |
| ***SCALE:*** *Before answering the questions below, review the descriptions of harm in Impact Gravity Table contained in the accompanying user's guide.* | |
| *For each potential impact identified in the previous section, would you characterize it as a catastrophic, critical, serious, or moderate/minor harm?* | |
| *For each potential impact identified in the previous section, are there individuals or groups who possess characteristics that could make them more vulnerable to the impact? If so, what are these characteristics and could those who possess them suffer the harm more acutely or seriously than others?* | |
| *For each potential impact identified in the previous section, which individuals or groups could* | |
| *encounter the gravest impact on the right or freedom under consideration?* | |
| ***SCOPE:*** *Before answering the questions below, review the answers to* Activity 1 (Project Section) of the PS Report and answers to *questions 10 and 11 from the PCRA* (on numbers of rights-holders affected by the AI system and timescale of the system's potential impacts). | |
| *For each potential impact identified in the previous section, does the answer given to question 10 of the PCRA seem like accurate and realistic representation of the potential scope of the impact? If not, how many rights-holders could encounter a deprivation of the right or freedom under consideration?* | |
| *For each potential impact identified in the previous section, are there groups who possess characteristics that could make them vulnerable to higher levels of exposure to the impact? If so, how much exposure could these groups face?* | |
| *For each potential impact identified in the previous section, consider the overall timescale of the AI system's impacts on the right or freedom under consideration (answer to question 11 of the PCRA). Are there cumulative or aggregate impacts of the system on rights-holders or their progeny that could extend the effects of the system beyond the scope of impact identified in the previous answers?* | |
| ***REMEDIABILITY:*** *Before answering the questions below, review the Remediability Table contained in the accompanying user's guide.* | |
| *For each potential impact identified in the previous section, how remediable is the potential adverse effect on affected right-holders? (Refer to the Remediability Table for categories of remediability to consider)* | |
| *For each potential impact identified in the previous section, are there groups who possess characteristics that could make them vulnerable to lower levels of remediability for the impact (i.e. more subject to difficult-to-overcome or irreversible harm)? If so, what degree of irreversible harm do these groups face?* | |
| ***Non-discrimination, fairness, and equality*** | |
| IDENTIFYING POTENTIAL ADVERSE IMPACTS | |
| *How, if at all, could the AI system (either in the processes of its production or in its deployment) result in discrimination, have discriminatory effects on impacted rights-holders, or perform differentially for different groups in discriminatory or harmful ways—including for intersectional groups where vulnerable or protected characteristics converge?* | |
| *How, if at all, could use of the AI system expand existing inequalities in the communities it affects or augment historical patterns of inequity and discrimination in these communities?* | |
| *In what other ways, if any, could the use of this system cause or contribute to inequity or inequality?* | |
| ASSESSING THE SEVERITY OF POTENTIAL IMPACTS | |
| ***SCALE:*** *Before answering the questions below, review the descriptions of harm in Impact Gravity Table contained in the accompanying user's guide.* | |
| *For each potential impact identified in the previous section, would you characterize it as a catastrophic, critical, serious, or moderate/minor harm?* | |
| *For each potential impact identified in the previous section, are there individuals or groups who possess characteristics that could make them more vulnerable to the impact? If so, what are these characteristics and could those who possess them suffer the harm more acutely or seriously than others?* | |
| *For each potential impact identified in the previous section, which individuals or groups could encounter the gravest impact on the right or freedom under consideration?* | |
| ***SCOPE:*** *Before answering the questions below, review the answers to* Activity 1 (Project Section) of the PS Report and answers to *questions 10 and 11 from the PCRA* (on numbers of rights-holders affected by the AI system and timescale of the system's potential impacts). | |
| *For each potential impact identified in the previous section, does the answer given to question 10 of the PCRA seem like accurate and realistic* | |
| *representation of the potential scope of the impact? If not, how many rights-holders could encounter a deprivation of the right or freedom under consideration?* | |
| *For each potential impact identified in the previous section, are there groups who possess characteristics that could make them vulnerable to higher levels of exposure to the impact? If so, how much exposure could these groups face?* | |
| *For each potential impact identified in the previous section, consider the overall timescale of the AI system's impacts on the right or freedom under consideration (answer to question 11 of the PCRA). Are there cumulative or aggregate impacts of the system on rights-holders or their progeny that could extend the effects of the system beyond the scope of impact identified in the previous answers?* | |
| ***REMEDIABILITY:*** *Before answering the questions below, review the Remediability Table contained in the accompanying user's guide.* | |
| *For each potential impact identified in the previous section, how remediable is the potential adverse effect on affected right-holders? (Refer to the Remediability Table for categories of remediability to consider)* | |
| *For each potential impact identified in the previous section, are there groups who possess characteristics that could make them vulnerable to lower levels of remediability for the impact (i.e. more subject to difficult-to-overcome or* | |
| *irreversible harm)? If so, what degree of irreversible harm do these groups face?* | |
| ***Data protection and the right to respect of private and family life*** | |
| IDENTIFYING POTENTIAL ADVERSE IMPACTS | |
| *How, if at all, could the design, development, and deployment of the AI system harm the rights to data protection enshrined in data protection and privacy law and in the Council of Europe's Convention 108+?* | |
| *How, if at all, could the AI system intrude on or interfere with the private and family life of rights-holders in ways that prevent or impede them from maintaining a personal sphere that is independent from the transformative effects of AI technologies and in which they are at liberty to freely think, form opinions and beliefs, and develop their personal identities and intimate relationships without the influence of AI technologies?* | |
| ASSESSING THE SEVERITY OF POTENTIAL IMPACTS | |
| ***SCALE:*** *Before answering the questions below, review the descriptions of harm in Impact Gravity Table contained in the accompanying user's guide.* | |
| *For each potential impact identified in the previous section, would you characterize it as a catastrophic, critical, serious, or moderate/minor harm?* | |
| *For each potential impact identified in the previous section, are there individuals or groups who* | |
| *possess characteristics that could make them more vulnerable to the impact? If so, what are these characteristics and could those who possess them suffer the harm more acutely or seriously than others?* | |
| *For each potential impact identified in the previous section, which individuals or groups could encounter the gravest impact on the right or freedom under consideration?* | |
| ***SCOPE:*** *Before answering the questions below, review the answers to* Activity 1 (Project Section) of the PS Report and answers to *questions 10 and 11 from the PCRA* (on numbers of rights-holders affected by the AI system and timescale of the system's potential impacts). | |
| *For each potential impact identified in the previous section, does the answer given to question 10 of the PCRA seem like accurate and realistic representation of the potential scope of the impact? If not, how many rights-holders could encounter a deprivation of the right or freedom under consideration?* | |
| *For each potential impact identified in the previous section, are there groups who possess characteristics that could make them vulnerable to higher levels of exposure to the impact? If so, how much exposure could these groups face?* | |
| *For each potential impact identified in the previous section, consider the overall timescale of the AI system's impacts on the right or freedom under consideration (answer to question 11 of the PCRA). Are there cumulative or aggregate impacts of the* | |
| *system on rights-holders or their progeny that could extend the effects of the system beyond the scope of impact identified in the previous answers?* | |
| ***REMEDIABILITY:*** *Before answering the questions below, review the Remediability Table contained in the accompanying user's guide.* | |
| *For each potential impact identified in the previous section, how remediable is the potential adverse effect on affected right-holders? (Refer to the Remediability Table for categories of remediability to consider)* | |
| *For each potential impact identified in the previous section, are there groups who possess characteristics that could make them vulnerable to lower levels of remediability for the impact (i.e. more subject to difficult-to-overcome or irreversible harm)? If so, what degree of irreversible harm do these groups face?* | |
| ***Social and economic rights*** | |
| IDENTIFYING POTENTIAL ADVERSE IMPACTS | |
| *How, if at all, could the deployment of the AI system harm the social and economic rights of affected persons, including the right to just working conditions, the right to safe and healthy working conditions, the right to organize, the right to social security, and the rights to the protection of health and to social and medical assistance as set out in the European Social Charter?* | |
| *In what other ways, if any, could the use of this system adversely impact other relevant social and economic rights?* | |
| ASSESSING THE SEVERITY OF POTENTIAL IMPACTS | |
| ***SCALE:*** *Before answering the questions below, review the descriptions of harm in Impact Gravity Table contained in the accompanying user's guide.* | |
| *For each potential impact identified in the previous section, would you characterize it as a catastrophic, critical, serious, or moderate/minor harm?* | |
| *For each potential impact identified in the previous section, are there individuals or groups who possess characteristics that could make them more vulnerable to the impact? If so, what are these characteristics and could those who possess them suffer the harm more acutely or seriously than others?* | |
| *For each potential impact identified in the previous section, which individuals or groups could encounter the gravest impact on the right or freedom under consideration?* | |
| ***SCOPE:*** *Before answering the questions below, review the answers to* Activity 1 (Project Section) of the PS Report and answers to *questions 10 and 11 from the PCRA* (on numbers of rights-holders affected by the AI system and timescale of the system's potential impacts). | |
| *For each potential impact identified in the previous section, does the answer given to question 10 of the PCRA seem like accurate and realistic* | |
| *representation of the potential scope of the impact? If not, how many rights-holders could encounter a deprivation of the right or freedom under consideration?* | |
| *For each potential impact identified in the previous section, are there groups who possess characteristics that could make them vulnerable to higher levels of exposure to the impact? If so, how much exposure could these groups face?* | |
| *For each potential impact identified in the previous section, consider the overall timescale of the AI system's impacts on the right or freedom under consideration (answer to question 11 of the PCRA). Are there cumulative or aggregate impacts of the system on rights-holders or their progeny that could extend the effects of the system beyond the scope of impact identified in the previous answers?* | |
| ***REMEDIABILITY:*** *Before answering the questions below, review the Remediability Table contained in the accompanying user's guide.* | |
| *For each potential impact identified in the previous section, how remediable is the potential adverse effect on affected right-holders? (Refer to the Remediability Table for categories of remediability to consider)* | |
| *For each potential impact identified in the previous section, are there groups who possess characteristics that could make them vulnerable to lower levels of remediability for the impact (i.e. more subject to difficult-to-overcome or* | |
| *irreversible harm)? If so, what degree of irreversible harm do these groups face?* | |
| ***Democracy*** | |
| IDENTIFYING POTENTIAL ADVERSE IMPACTS | |
| *How, if at all, could the use or misuse of the AI system lead to interference with free and fair election processes or with the ability of impacted individuals to participate freely, fairly, and fully in the political life of the community through any of the following: mass deception, mass manipulation, mass intimidation or behavioural control, or mass personalized political targeting or profiling, at the local, national, or global levels?* | |
| *How, if at all, could the use or misuse of the AI system lead to the mass dispersal, at local, national, or global levels, of misinformation or disinformation?* | |
| *How, if at all, could the use or misuse of the AI system lead to obstruction of informational plurality, at local, national, or global levels?* | |
| *How, if at all, could the use or misuse of the AI system (in particular, information filtering models such as recommender systems, search engines, or news aggregators) lead to an obstruction of the free and equitable flow of the legitimate and valid forms of information that are necessary for the meaningful democratic participation of impacted* | |
| *rights holders and for their ability to engage freely, fairly, and fully in collective problem-solving?* | |
| *How, if at all, could the use or misuse of the AI system (in particular, models that track, identify, de-anonymize or surveil rights-holders and social groups or enable the creation of social graphs) lead to interference with or obstruction of impacted rights-holders' abilities to exercise their freedoms of expression, assembly, or association?* | |
| *In what other ways, if any, could the use of this system adversely impact democratic principles or the preconditions of democratic ways of life?* | |
| ASSESSING THE SEVERITY OF POTENTIAL IMPACTS | |
| ***SCALE:*** *Before answering the questions below, review the descriptions of harm in Impact Gravity Table contained in the accompanying user's guide.* | |
| *For each potential impact identified in the previous section, would you characterize it as a catastrophic, critical, serious, or moderate/minor harm?* | |
| *For each potential impact identified in the previous section, are there individuals or groups who possess characteristics that could make them more vulnerable to the impact? If so, what are these characteristics and could those who possess them suffer the harm more acutely or seriously than others?* | |
| *For each potential impact identified in the previous section, which individuals or groups could* | |
| *encounter the gravest impact on the right or freedom under consideration?* | |
| ***SCOPE:*** *Before answering the questions below, review the answers to* Activity 1 (Project Section) of the PS Report and answers to *questions 10 and 11 from the PCRA* (on numbers of rights-holders affected by the AI system and timescale of the system's potential impacts). | |
| *For each potential impact identified in the previous section, does the answer given to question 10 of the PCRA seem like accurate and realistic representation of the potential scope of the impact? If not, how many rights-holders could encounter a deprivation of the right or freedom under consideration?* | |
| *For each potential impact identified in the previous section, are there groups who possess characteristics that could make them vulnerable to higher levels of exposure to the impact? If so, how much exposure could these groups face?* | |
| *For each potential impact identified in the previous section, consider the overall timescale of the AI system's impacts on the right or freedom under consideration (answer to question 11 of the PCRA). Are there cumulative or aggregate impacts of the system on rights-holders or their progeny that could extend the effects of the system beyond the scope of impact identified in the previous answers?* | |
| ***REMEDIABILITY:*** *Before answering the questions below, review the Remediability Table contained in the accompanying user's guide.* | |
| *For each potential impact identified in the previous section, how remediable is the potential adverse effect on affected right-holders? (Refer to the Remediability Table for categories of remediability to consider)* | |
| *For each potential impact identified in the previous section, are there groups who possess characteristics that could make them vulnerable to lower levels of remediability for the impact (i.e. more subject to difficult-to-overcome or irreversible harm)? If so, what degree of irreversible harm do these groups face?* | |
| ***Rule of Law*** | |
| IDENTIFYING POTENTIAL ADVERSE IMPACTS | |
| *How, if at all, could the deployment of the AI system harm impacted individuals' rights to effective remedy and to a fair trial (equality of arms, right to a natural judge established by law, the right to an independent and impartial tribunal, and respect for the adversarial process)?* | |
| *In what other ways, if any, could the use of this system adversely impact the rule of law or the precondition of a just legal order?* | |
| ASSESSING THE SEVERITY OF POTENTIAL IMPACTS | |
| ***SCALE:*** *Before answering the questions below, review the descriptions of harm in Impact Gravity Table contained in the accompanying user's guide.* | |
| *For each potential impact identified in the previous section, would you characterize it as a catastrophic, critical, serious, or moderate/minor harm?* | |
| *For each potential impact identified in the previous section, are there individuals or groups who possess characteristics that could make them more vulnerable to the impact? If so, what are these characteristics and could those who possess them suffer the harm more acutely or seriously than others?* | |
| *For each potential impact identified in the previous section, which individuals or groups could encounter the gravest impact on the right or freedom under consideration?* | |
| ***SCOPE:*** *Before answering the questions below, review the answers to* Activity 1 (Project Section) of the PS Report and answers to *questions 10 and 11 from the PCRA* (on numbers of rights-holders affected by the AI system and timescale of the system's potential impacts). | |
| *For each potential impact identified in the previous section, does the answer given to question 10 of the PCRA seem like accurate and realistic representation of the potential scope of the impact? If not, how many rights-holders could encounter a deprivation of the right or freedom under consideration?* | |
| *For each potential impact identified in the previous section, are there groups who possess characteristics that could make them vulnerable to* | |
| *higher levels of exposure to the impact? If so, how much exposure could these groups face?* | |
| *For each potential impact identified in the previous section, consider the overall timescale of the AI system's impacts on the right or freedom under consideration (answer to question 11 of the PCRA). Are there cumulative or aggregate impacts of the system on rights-holders or their progeny that could extend the effects of the system beyond the scope of impact identified in the previous answers?* | |
| ***REMEDIABILITY:*** *Before answering the questions below, review the Remediability Table contained in the accompanying user's guide.* | |
| *For each potential impact identified in the previous section, how remediable is the potential adverse effect on affected right-holders? (Refer to the Remediability Table for categories of remediability to consider)* | |
| *For each potential impact identified in the previous section, are there groups who possess characteristics that could make them vulnerable to lower levels of remediability for the impact (i.e. more subject to difficult-to-overcome or irreversible harm)? If so, what degree of irreversible harm do these groups face?* | |
| **ADDITIONAL QUESTION FOR THE PROJECT TEAM** | |
| *These questions are meant as a supplement to Part I of the HUDERIA. They are intended to be answered by the project team and then to be revisited during the dialogue between duty-bearers and rights-holders in Step 3 of the HUDERIA* | |
| *Has a thorough assessment of the human rights compliant business practices of all businesses, parties and entities involved in the value chain of the AI product or service been undertaken? This would include all businesses, parties, and entities directly linked to your project lifecycle through supply chains, operations, contracting, sales, consulting, and partnering. (If not, do you have plans to do this?)* | |
| *Do you have processes in place and appropriate resources dedicated to maintaining human rights diligence with regard to all businesses, parties, and entities directly linked to your project lifecycle?* | |
| *Do you require your vendors and suppliers to demonstrate compliance with legal protections in place to protect rights-holders from modern slavery, human trafficking, and labour exploitation?* | |

| **MITIGATING POTENTIAL IMPACTS AND SETTING UP ACCESS TO REMEDY** |
|---|
| A SUMMARY OF COMBINED IMPACT FINDINGS |
| Provide a summary of combined impact findings from the project team and engaged rights-holders with a detailed mapping of potential adverse impacts and any significant interrelationships between them. |
| |
| PRESENTATION OF THE MEASURES AND ACTIONS TO BE TAKEN |
| Present the measures and actions that will be taken and, where needed, a presentation of the prioritization of actions accompanied by an explanation of prioritization rationale (The project team and engaged rights-holders should also draw on recommendations generated in the results of Section 1 of the PCRA when setting out needed measures and actions). |
| |
| PRESENTATION OF PLANS TO EXERCISE LEVERAGE OR ALTER RELATIONSHIPS |
| For any potential adverse impacts identified in direct links of the business or organisation to businesses, parties and entities involved in the value chain of the AI product or service, present plans to exercise leverage or alter relationships to mitigate impacts. |
| |
| CLARIFICATION OF ROLES AND RESPONSIBILITES |
| Clarify the roles and responsibilities of the various actors involved in impact mitigation, management, and monitoring. |
| |
| |
| STATEMENT OF COMMITMENT OF TIME AND RESOURCES FOR IMP |
| (From the project team directly): Provide a statement that indicates the allotted commitment of time and resources that will be made available to carry out the IMP. |
| |
| PLAN FOR MONITORING IMPACT MITIGATION EFFORTS |
| Formulate a plan for monitoring impact mitigation efforts and for re-assessing and re-evaluating the HUDERIA during subsequent development and deployment phases of the project lifecycle. |
| |
| ACCESSIBLE PRESENTATION OF ACCESS TO REMEDY AND OPERATIONAL-LEVEL GRIEVANCE MECHANISMS |
| Provide an accessible presentation of the access to remedy and operational-level grievance mechanisms that will be available to impacted rights-holders if they suffer any harms identified in the HUDERIA or any other adverse impacts that were not detected or recognized in the impact assessment process. |
| |

# Human Rights, Democracy, and Rule of Law Assurance Case (HUDERAC)

**Purpose**

The Purpose of the Human Rights, Democracy, and Rule of Law Assurance Case is to help project teams build a structured argument that provides assurance to stakeholders that claims about an AI system are warranted given available evidence. The process of developing an assurance case assists internal reflection and deliberation, promoting the adoption of best practices and conveying the integration of these into design, development, and deployment lifecycles to impacted stakeholders.

Inputs

- PS Report
- PCRA Report
- SEP Report
- HUDERIA Report
- Additional Evidence*

* There is a wide variety of additional evidential artefacts that may feed into an assurance case (see chapter).

→

Steps

- 4.1 Establishing goals requisite for risk management and impact mitigation and determining the properties needed to assure these goals
- 4.2 Taking actions to operationalize these properties
- 4.3 Compiling evidence of these actions

→

Outputs

- HUDERAC
  - Document demonstrating how system goals have been operationalized by reference to relevant (evidence-backed) properties of the system or project lifecycle.

## Section Outline

1. Introduction
2. What is an Assurance Case?
3. Structure of a HUDERAC
4. Components of a HUDERAC
   a. Goal(s)
   b. Properties and (Project or System Property) Claims
   c. Evidence
5. Building a HUDERAC
   a. HUDERAC Template
   b. Goals, Actions, and Claims

## Introduction

Assurance is a process of establishing trust.

For a stakeholder to trust that an AI system respects human rights and fundamental freedoms, democratic principles, and the rule of law—especially in novel, high risk, or safety critical domains—they will typically require assurance that their trust is placed judiciously.[60]

Providing assurance can take the form of compiling clear and accessible evidence that a risk has been sufficiently mitigated[61], or that mechanisms have been put in place to prevent negative impacts from occurring.[62]

In the context of the HUDERAF, trust and assurance are built on and emerge from the foundations of the preliminary context-based risk analysis (PCRA), the stakeholder engagement process (SEP) and the human rights, democracy, and rule of law impact assessment (HUDERIA). These three stages, therefore, are part of the human rights, democracy, and rule of law assurance framework (HUDERAF) that is detailed in this section.

In addition to outlining this framework and showing how it relates to the three other stages, this section also presents the final practical mechanism of this process: a human rights, democracy, and rule of law assurance case (HUDERAC).

### What is an Assurance Case?

In its simplest form, an assurance case is a form of documentation that demonstrates how a goal (or set of goals) has been established by reference to relevant (evidence-backed) properties of the system or project lifecycle.[63] However, the process of developing an assurance case can,

- assist internal reflection and deliberation by providing a systematic and structured means for evaluating how the development of AI systems or products impact human rights, democracy, and the rule of law;
- provide a deliberate means for the anticipation and pre-emption of potential risks and adverse impacts through mechanisms of end-to-end assessment and redress;
- facilitate transparent communication between developers and affected stakeholders;

---

60 High-Level Expert Group on AI. (2019). 'Ethics Guidelines for Trustworthy AI'. European Commission. https://ec.europa.eu/digital-single-market/en/news/ethics-guidelines-trustworthy-ai
61 Cleland, G. M., Habli, I., Medhurst, J., & Health Foundation. (2012). Evidence: Using safety cases in industry and healthcare.
62 Sujan, M., Furniss, D., Grundy, K., Grundy, H., Nelson, D., Elliott, M., White, S., Habli, I., & Reynolds, N. (2019). Human factors challenges for the safe use of artificial intelligence in patient care. BMJ Health & Care Informatics, 26(1), e100081. https://doi.org/10.1136/bmjhci-2019-100081
63 GSN Community. (2018). GSN Community Standard (Version 2). The Assurance Case Working Group. https://scsc.uk/r141B:1?t=1

- support mechanisms and processes of documentation (or, reporting) to ensure accountability (e.g. audits, compliance);
- and build trust and confidence by promoting the adoption of best practices (e.g. standards for warranted evidence) and by conveying the integration of these into design, development, and deployment lifecycles to impacted stakeholders.

In the specific context of the human rights, democracy, and rule of law assurance framework, we can define an assurance case as:

> A structured argument that provides assurance to another party (or parties) that a particular claim (or set of related claims) about a property of an AI system is warranted given the available evidence.

There are three *primary* components that are required to build an assurance case: (a) a concise statement about the top-level normative goal (or goals) of the project, (b) a set of claims about the project or system properties required to operationalise the goal, and (c) a body of supporting evidence that demonstrates how the properties and goals have been realised.

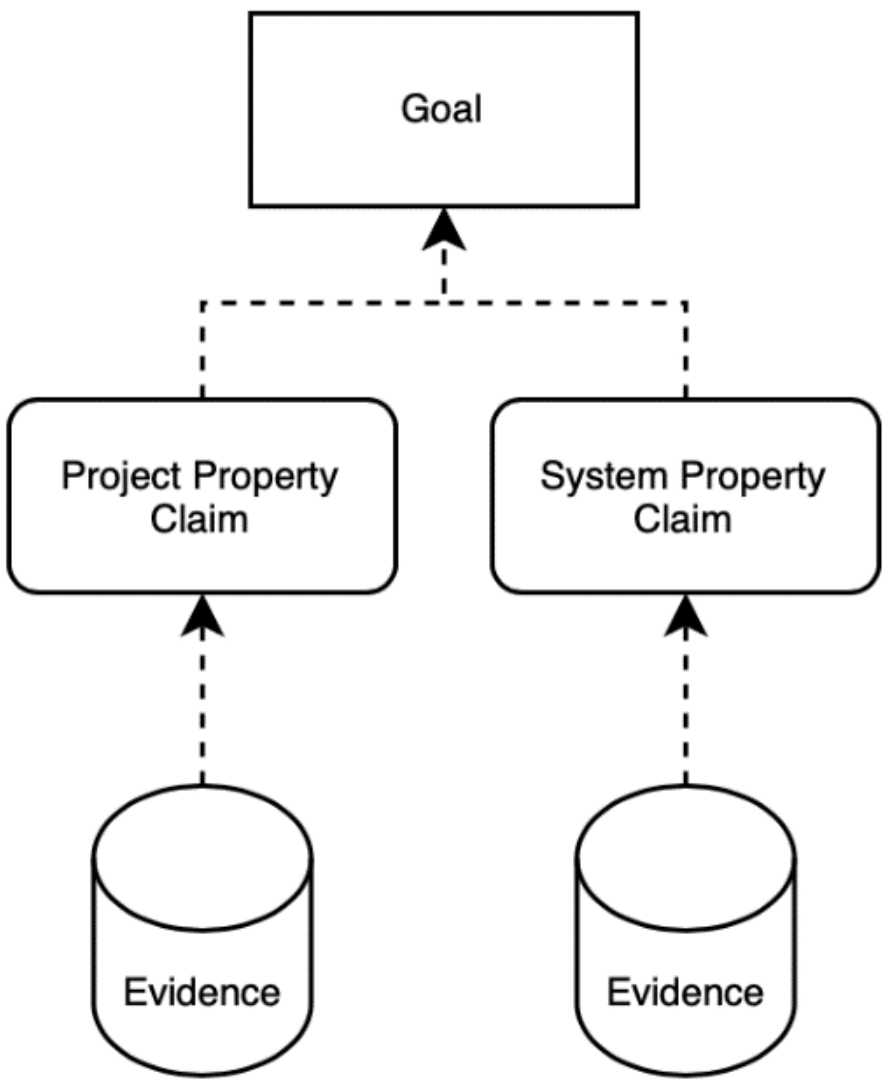

*Table 1 – the main elements of an assurance case*

Additional components can be added to provide additional context or information, but we will just focus on these three for present purposes. **Error! Reference source not found.** shows a simplified overview of how these primary components fit together.

Assurance cases are widely used in safety-critical domains, such as energy or aviation, where systems and manufacturing processes are closely inspected and assessed to ensure they are safe to operate and use in their intended context.[64] Different approaches to the development of an assurance case will incorporate additional elements beyond those presented in **Error! Reference source not found.**. However, each of the different approaches will typically have a *goal-structured* focus, and include property claims about the system or project, which are backed up by evidence.[65]

Collectively, these elements form a *structured argument*, which is designed to provide assurance that a goal (or set of goals) has been sufficiently established (e.g., safety, explainability, data integrity).[66] We can, therefore, refer to the development of this structured argument as an *argument-based* form of assurance.

## Structure of a HUDERAC

Whereas the PCRA, SEP, and HUDERIA serve as mechanisms for *identifying* and *evaluating* possible harms that could arise from the design, development, and deployment of an AI system, an assurance case serves to *demonstrate* what actions have been taken to prevent such harms from occurring. An assurance case, therefore, is the primary means by which the basis for trust in the system is *communicated* to and *scrutinised* by diverse stakeholders, including rights holders, procurers, regulators, and policymakers.

Consider the following example. A developer has completed the PCRA and worked with stakeholders to further analyse and evaluate possible risks, as outlined in the SEP. They have also carried out the HUDERIA and have identified a set of actions that they need to perform when designing, developing, and deploying their AI system. One of the actions the developer has taken is to engage independent domain experts to help evaluate their training data and ensure that it has the following properties: accurate, representative, and up-to-date.

Presuming the domain experts are in agreement and believe the training data possess such properties, would this justify the developer making the following claim?

> The training data were accurate, representative, and up-to-date.

The developer may be confident that these properties of their training data have been established, but what matters here is that the developer can *justify* this confidence to other stakeholders. This is what it means to provide assurance.

---

64 Hawkins, R. et al., (2021) 'Guidance on the Assurance of Machine Learning in Autonomous Systems'. University of York: Assuring Autonomy International Programme (AAIP) https://www.york.ac.uk/media/assuring-autonomy/documents/AMLASv1.1.pdf
65 Bloomfield, R., & Bishop, P. (2010). Safety and Assurance Cases: Past, Present and Possible Future – an Adelard Perspective. In C. Dale & T. Anderson (Eds.), Making Systems Safer (pp. 51–67). Springer London. https://doi.org/10.1007/978-1-84996-086-1_4
66 Walton, D. (2009). Argumentation Theory: A Very Short Introduction. In G. Simari & I. Rahwan (Eds.), Argumentation in Artificial Intelligence (pp. 1–22). Springer US. https://doi.org/10.1007/978-0-387-98197-0_1

So, how does one provide assurance that is both warranted and justified?

The *structure* of an assurance case helps establish a justified (or warranted) basis for believing and claiming that a goal has been met. The directionality of **Error! Reference source not found.**, for instance, depicts a process of inferential support in which the evidence supports a claim about some system or project property, and the set of property claims subsequently help realise the goal. **Error! Reference source not found.** adapts the original figure to present a partial example oriented towards a goal of safety.

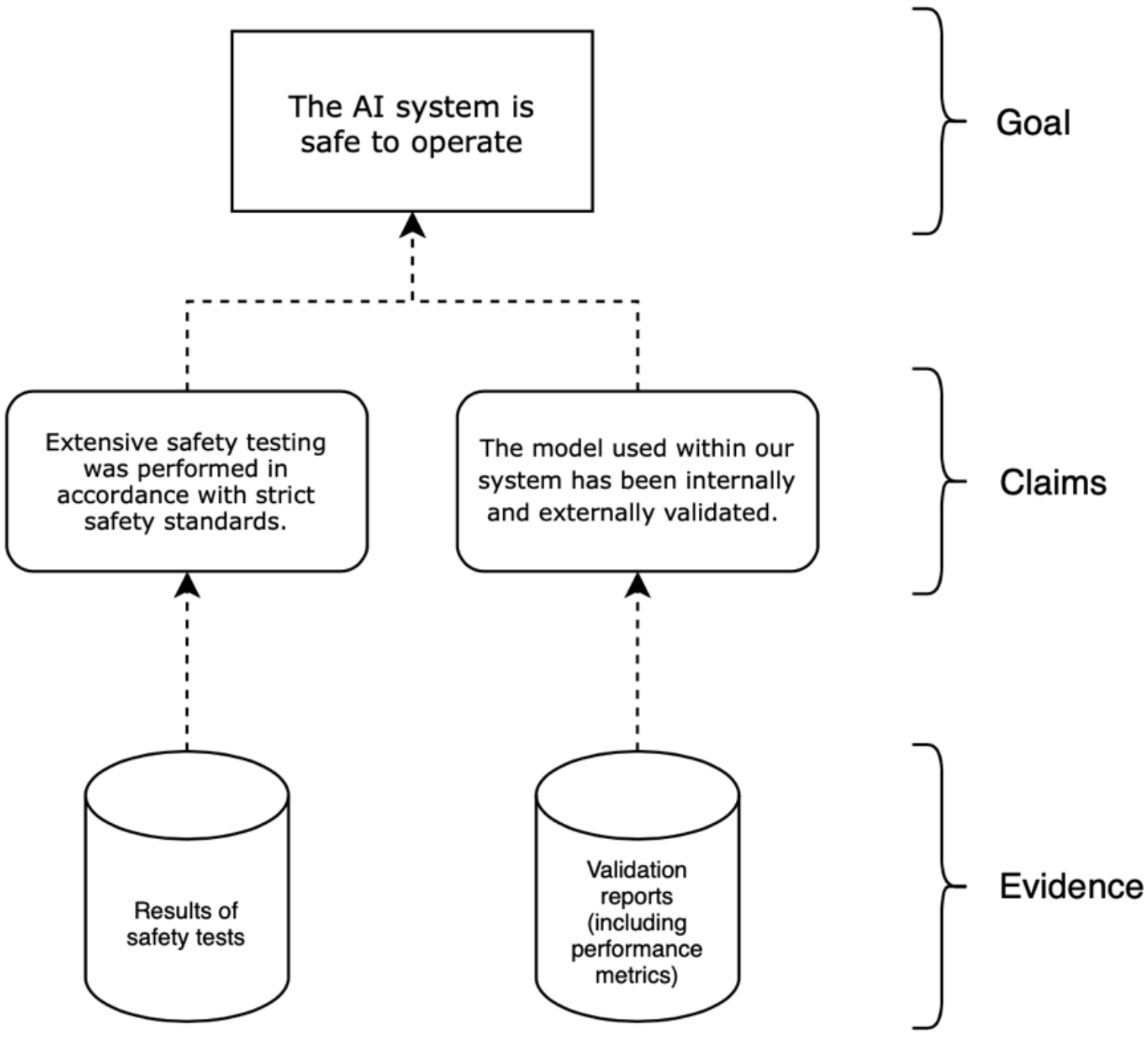

*Figure 1 - An example of a partial assurance case*

The inferential structure of an assurance case helps establish trust that some goal has been obtained. However, the structure is not sufficient on its own. In addition to the structural aspects of an assurance case, we must also consider the *substantive* aspects that help establish warrant.

## Components of a HUDERAC

Let's look at the specific components of a human rights, democracy, and rule of law assurance case in more detail. Specifically, we will discuss the goal(s), properties, and evidence.

### *Goal(s)*

The HUDERAF sets important constraints on the types of goals that may be selected. For instance, the PCRA, SEP, or HUDERIA all help to identify how specific fundamental rights and freedoms may be violated, how an AI system is likely to harm vulnerable users, or how certain democratic principles are at risk in the deployment of a particular AI system. Each of these prior stages are orientated towards some goal or set of goals, even if the goal is not made explicit. One function of an assurance case, therefore, is to make these goals explicit and relate them to key human rights and freedoms, or principles of democratic governance and the rule of law.

To aid this process, we can start from the following list of goals that are referred to as the SAFE-D goals (see Box 1). We have provided a general description of the meaning for each of these goals, including reference to ancillary and neighbouring concepts.

SAFETY is of paramount importance for ensuring the sustainable development, deployment, and use of an AI system. From a technical perspective, this requires the system to be secure, robust, and reliable. And from a social sustainability perspective, this requires the practices behind the system's production and use to be informed by ongoing consideration of the risk of exposing affected rights-holders to harms, continuous reflection on project context and impacts, ongoing stakeholder engagement and involvement, and change monitoring of the system from its deployment through to its retirement or deprovisioning.

ACCOUNTABILITY can include specific forms of process transparency (e.g., as enacted through process logs or external auditing) that may be necessary for mechanisms of redress, or broader processes of responsible governance that seek to establish clear roles of responsibility where transparency may be inappropriate (e.g., confidential projects).

FAIRNESS is inseparably connected with sociolegal conceptions of equity and justice, which may emphasize a variety of features such as non-discrimination, equitable outcomes, or procedural fairness through bias mitigation, but also social and economic equality, diversity, and inclusiveness.

EXPLAINABILITY is a key condition for autonomous and informed decision-making in situations where AI systems interact with or influence human judgement and decision-making. Explainability goes beyond the ability to merely interpret the outcomes of an AI system; it also depends on the ability to provide an accessible and relevant information base about the processes behind the outcome.

DATA QUALITY, INTEGRITY, PROTECTION AND PRIVACY must all be established to be confident that the (data-driven) AI system has been developed on secure grounds.

- 'DATA QUALITY' captures the *static* properties of data, such as whether they are (a) *relevant* to and *representative* of the domain and use context, (b) *balanced* and *complete* in terms of how well the dataset represents the underlying data generating process, and (c) *up-to-date* and *accurate* as required by the project.
- 'DATA INTEGRITY' refers to more *dynamic* properties of data stewardship, such as how a dataset evolves over the course of a project lifecycle. In this manner, data integrity requires (a) *contemporaneous* and *attributable* records from the start of a project (e.g., process logs; research statements), (b) ensuring *consistent* and *verifiable* means of data analysis or processing during development, and (c) taking steps to establish *findable*, *accessible*, *interoperable*, and *reusable* records towards the end of a project's lifecycle.

- 'DATA PROTECTION AND PRIVACY' reflect ongoing developments and priorities as set out in relevant legislation and regulation of data practices as they pertain to fundamental rights and freedoms, democracy, and the rule of law. For example, the right for data subjects to have inaccurate personal data rectified or erased.

*Box 1 – SAFE-D Goals*

Each of these goals plays an important role in upholding human rights, democracy, and the rule of law. For example, 'DATA PROTECTION'[67] and 'EXPLAINABILITY' are both required to ensure that an individual's autonomy is protected, whereas 'FAIRNESS' and 'DATA INTEGRITY' are both key components of social equality.

The specific meaning of the goals, however, will be delineated in different ways depending on the use context or domain of the AI system, and the processes and results of the SEP. For example, depending on the AI system, SAFETY could relate to the *physical* safety of patients in the context of an AI system used in healthcare, or *environmental* safety for an AI system used in agriculture or forestry. Moreover, a group of stakeholders may wish to prioritise responsible governance over transparency, in relation to ACCOUNTABILITY where a project requires greater (and justifiable) levels of secrecy.

This means that the top-level goals will need to be further specified, both by detailing the intended use context for the AI system[68] and by making specific claims about relevant properties of the AI system or project.

## Properties and (Project or System Property) Claims

Each of the SAFE-D goals has a variety of lower-level properties associated with them. These properties ought to be established in either the project or the system if the goal is said to have been obtained. In a HUDERAC, a corresponding claim about the (project or system) property is, therefore, required. The following properties can be identified for each of the SAFE-D goals.

### *Safety*

- **Sustainability:** the goal of safety must be achieved with an eye towards the sustainability of a safe system. This goes beyond environmental sustainability (e.g., the ecological footprint of the project and system). It

67 To indicate that a concept is being employed as a goal we will use small caps formatting (e.g., 'fairness'.)
68 The 'project summary' output from the PCRA is likely to help here, as it will typically contain a lot of this information.

also includes an understanding of the long-term use context and impact of the system, and the resources needed to ensure the system continues to operate safely over time. For instance, sustainability may depend upon sufficient *change monitoring* processes that establish whether there has been a substantive change in the underlying data distributions or social operating environment. Sustainability also involves engaging and involving rights-holders in the design and assessment of AI systems that could impact their human rights and fundamental freedoms.

- **Security:** Security encompasses the protection of several operational dimensions of an AI system when confronted with possible adversarial attack. A secure system is capable of maintaining the integrity of the information that constitutes it. This includes protecting its architecture from the unauthorised modification or damage of any of its component parts. A secure system also remains continuously functional and accessible to its authorised users and keeps confidential and private information secure even under hostile or adversarial conditions.
- **Robustness:** The objective of robustness can be thought of as the goal that an AI system functions reliably and accurately under harsh conditions. These conditions may include adversarial intervention, implementer error, or skewed goal-execution by an automated learner (in reinforcement learning applications). The measure of robustness is therefore the strength of a system's integrity the soundness of its operation in response to difficult conditions, adversarial attacks, perturbations, data poisoning, and undesirable reinforcement learning behaviour.
- **Reliability:** The objective of reliability is that an AI system behaves exactly as its designers intended and anticipated. A reliable system adheres to the specifications it was programmed to carry out. Reliability is therefore a measure of consistency and can establish confidence in the safety of a system based upon the dependability with which it operationally conforms to its intended functionality.
- **Accuracy and Performance Metrics:** In machine learning, the accuracy of a model is the proportion of examples for which it generates a correct output. This performance measure is also sometimes characterised conversely as an error rate or the fraction of cases for which the model produces an incorrect output. As a performance metric, accuracy should be a central component to establishing and nuancing the approach to safe AI. Specifying a reasonable performance level for the system may also often require refining or exchanging of the measure of accuracy. For instance, if certain errors are more significant or costly than others, a metric for total cost can be integrated into the model so that the cost of one class of errors can be weighed against that of another.

### *Accountability*

- **Traceability:** Traceability refers to the process by which all stages of the data lifecycle from collection to deployment to system updating or deprovisioning are documented in a way that is accessible and easily understood. This may include not only the parties within the organisation involved but also the actions taken at each stage that may impact the individuals who use the system.

- **Answerability:** Answerability depends upon a human chain of responsibility. Answerability responds to the question of *who is accountable* for an automation supported outcome.
- **Auditability:** Whereas the property of answerability responds to the question of *who is accountable* for an automation supported outcome, the notion of auditability answers the question of *how the designers and implementers of AI systems are to be held accountable*. This aspect of accountability has to do with *demonstrating and evidencing* both the responsibility of design and use practices and the justifiability of outcomes.
- **Clear Data Provenance and Data Lineage:** Clear provenance and data lineage consists of records that are accessible and simultaneously detail how data was collected and how it has been used and altered throughout the stages of pre-processing, modelling, training, testing, and deploying.
- **Accessibility:** Accessibility involves ensuring that information about the processes that took place to design, develop, and deploy an AI system are easily accessible by individuals. This not only refers to suitable means of explanation (clear, understandable, and accessible language) but also the mediums for delivery.
- **Reproducibility:** Related to and dependant on the above four properties, reproducibility refers to the ability for others to reproduce the steps you have taken throughout your project to achieve the desired outcomes and where necessary to replicate the same outcomes by following the same procedure.
- **Responsible Governance:** Responsible governance ensures accountability and responsibility for the processes that occur throughout the data lifecycle. This includes the identification and assignment of a data protection officer, as well as clearly identifying data controllers and processors. This may also include the creation of an independent oversight board to ensure these individuals are held accountable and the processes are well-documented.

### *Fairness*

- **Bias Mitigation:** It is not possible to eliminate bias entirely. However, effective bias mitigation processes can minimise the unwanted and undesirable impact of systematic deviations, distortions, or disparate outcomes that arise to a project governance problem, interfering factor, or from insufficient reflection on historical social or structural discrimination.
- **Diversity and Inclusiveness:** A significant component of fairness aware design is ensuring the inclusion of diverse voices and opinions in the design and development process through the participation of a more representative range of stakeholders. This includes considering whether values of civic participation, inclusion, and diversity been adequately considered in articulating the purpose and setting the goals of the project.

Consulting with internal organisational stakeholders is also necessary to strengthen the openness, inclusiveness, and diversity of the project.

- **Non-Discrimination:** your system should not create or contribute to circumstances whereby members of protected groups are treated differently or less favourably than other groups because of their respective protected characteristic.
- **Equality:** the outcome or impact of a system should either maintain or promote a state of affairs in which every individual has equal rights and liberties, and equal access or opportunities to whatever good or service the AI system brings about.

### *Explainability*

- **Interpretability:** Interpretability consists of the ability to know how and why a model performed the way it did in a specific context and therefore to understand the rationale behind its decision or behaviour.
- **Responsible Model Selection:** The normal expectations of intelligibility and accessibility that accompany the function the system will fulfil in the sector or domain in which it will operate. The availability of more interpretable algorithmic models or techniques in cases where the selection of an opaque model poses risks to the physical, psychological, or moral integrity of rights-holders or to their human rights and fundamental freedoms. The availability of the resources and capacity that will be needed to responsibly provide supplementary methods of explanation (e.g. simpler surrogate models, sensitivity analysis, or relative feature important) in cases where an opaque model is deemed appropriate and selected.
- **Accessible Rationale Explanation:** the reasons that led to a decision—especially one that is automated—delivered in an accessible and non-technical way.
- **Responsible Implementation and User Training:** training users to operate the AI system may include
    - conveying basic knowledge about the nature of machine learning,
    - explaining the limitations of the system,
    - educating users about the risks of AI-related biases, such as decision-automation bias or automation-distrust bias[69], and
    - encouraging users to view the benefits and risks of deploying these systems in terms of their role in helping humans to come to judgements, rather than replacing that judgement.

69 Decision-automation bias occurs when users of AI decision-support systems may become hampered in their critical judgment and situational awareness as a result of an overconfidence in the objectivity, or certainty of the AI system. At the other extreme, automation-distrust bias occurs when users of an automated decision-support system tend to disregard its contributions to evidence-based reasoning as a result of their distrust or scepticism about AI technologies in general (see ICO & ATI (2020) for further information).

### *Data Quality*

- **Source Integrity and Measurement Accuracy:** effective bias mitigation begins at the very commencement of data extraction and collection processes. Both the sources and instruments of measurement may introduce discriminatory factors into a dataset. When incorporated as inputs in the training data, biased prior human decisions and judgments—such as prejudiced scoring, ranking, interview-data or evaluation—will become the 'ground truth' of the model and replicate the bias in the outputs of the system in order to secure discriminatory non-harm, as well as ensuring that the data sample has optimal source integrity. This involves securing or confirming that the data gathering processes involved suitable, reliable, and impartial sources of measurement and sound methods of collection.
- **Timeliness and Recency:** if datasets include outdated data then changes in the underlying data distribution may adversely affect the generalisability of the trained model. Provided these distributional drifts reflect changing social relationship or group dynamics, this loss of accuracy with regard to the actual characteristics of the underlying population may introduce bias into an AI system. In preventing discriminatory outcomes, timeliness and recency of all elements of the data that constitute the datasets must be scrutinised.
- **Relevance, Appropriateness, and Domain Knowledge:** The understanding and utilisation of the most appropriate sources and types of data are crucial for building a robust and unbiased AI system. Solid domain knowledge of the underlying population distribution and of the predictive or classificatory goal of the project is instrumental for choosing optimally relevant measurement inputs that contribute to the reasonable determination of the defined solution. Domain experts should collaborate closely with the technical team to assist in the determination of the optimally appropriate categories and sources of measurement.
- **Adequacy of Quantity and Quality:** this property involves assessing whether the data available is comprehensive enough to address the problem set at hand, as determined by the use case, domain, function, and purpose of the system. Adequate quantity and quality should address sample size, representativeness, and availability of features relevant to problem.
- **Balance and Representativeness:** a balanced and representative dataset is one in which the distribution of features that are included, and the number of samples within each class is similar to the underlying distribution that exists in the overall population.

### *Data Integrity*[70]

- **Attributable:** Data should clearly demonstrate who observed and recorded it, when it was observed and recorded, and who it is about.

70 Adapted from SL Controls. (n.d.)

- **Consistent, Legible and Accurate:** Data should be easy to understand, recorded permanently and original entities should be preserved. Data should be free from errors and conform with the protocol. Consistency includes ensuring data is chronological (e.g., has a date and time stamp that is in the expected sequence).
- **Complete:** All recorded data requires an audit trail to show nothing has been deleted or lost.
- **Contemporaneous:** Data should be recorded as it was observed, and at the time it was executed.
- **Responsible Data Management:** Responsible data management ensures that the team has been trained on how to manage data responsibly and securely, identifying possible risks and threats to the system and assigning roles and responsibilities for how to deal with these risks if they were to occur. Policies on data storage and public dissemination of results should be discussed within the team and with stakeholders, as well as being clearly documented.
- **Data Traceability and Auditability:** Any changes or revisions to the dataset (e.g., additions, augmentations, normalisation) that occur after the original collection should be clearly traceable and well-documented to support any auditing.

*Data Protection and Privacy*

- **Consent (or legitimate basis) for processing:** Each Party shall provide that data processing can be carried out on the basis of the free, specific, informed and unambiguous consent of the data subject or of some other legitimate basis laid down by law. The data subject must be informed of risks that could arise in the absence of appropriate safeguards. Such consent must represent the free expression of an intentional choice, given either by a statement (which can be written, including by electronic means, or oral) or by a clear affirmative action and which clearly indicates in this specific context the acceptance of the proposed processing of personal data. Mere silence, inactivity or pre-validated forms or boxes should not, therefore, constitute consent. No undue influence or pressure (which can be of an economic or other nature) whether direct or indirect, may be exercised on the data subject and consent should not be regarded as freely given where the data subject has no genuine or free choice or is unable to refuse or withdraw consent without prejudice. The data subject has the right to withdraw the consent he or she gave at any time (which is to be distinguished from the separate right to object to pro- cessing).
- **Data Security:** Each Party shall provide that the controller, and, where applicable the processor, takes appropriate security measures against risks such as accidental or unauthorised access to, destruction, loss, use, modification or disclosure of personal data. Each Party shall provide that the controller notifies, without delay, at least the competent supervisory authority within the meaning of Article 15 of this Convention, of those data breaches which may seriously interfere with the rights and fundamental freedoms of data subjects.

- **Data Minimisation:** Personal data being processed is adequate (sufficient to properly fulfil the stated purpose), relevant (has a rational link to that purpose), and limited to what is necessary  do not hold more data than needed for that purpose).
- **Transparency:** The transparency of AI systems can refer to several features, both of their inner workings and behaviours, as well as the systems and processes that support them. An AI system is transparent when it is possible to determine how it was designed, developed, and deployed. This can include, among other things, a record of the data that were used to train the system, or the parameters of the model that transforms the input (e.g., an image) into an output (e.g, a description of the objects in the image). However, it can also refer to wider processes, such as whether there are legal barriers that prevent individuals from accessing information that may be necessary to understand fully how the system functions (e.g., intellectual property restrictions).
- **Proportionality:** delivering a just outcome in ways that are proportionate to the cost, complexity, and resources available. In a similar vein, the term 'proportionality' can also be used as an evaluative notion, such as in the case of a data protection principle that states only personal data that are necessary and adequate for the purposes of the task are collected.
- **Purpose Limitation:** The purposes for data processing must be outlined and documented from the beginning and made available to all individuals through privacy information. Personal data must adhere to the original purpose unless it is compatible with the original purpose, additional consent is received, or there is an obligation or function set out in law.
- **Accountability:** Appropriate measures and records must be in place to demonstrate compliance and responsibility for how data has been processed in alignment with the other principles.
- **Lawfulness, fairness, and transparency:** These three principles necessitate 'lawful basis' for the collection and use of personal data. Personal data must be used in a fair manner that is not unduly detrimental, unexpected, or misleading. Any processes in which data is used should not be in breach of any other laws, and teams must be clear, open, and honest with individuals about how their personal data is being used.
- **Respect for the rights of data subjects:** respect for the rights of data subjects requires putting in place adequate mechanisms or undertaking necessary actions so as to ensure that the rights of data subjects as defined under Convention 108+ and GDPR are upheld. Where necessary, this includes the responsible handling of sensitive data.

Providing assurance for one or more of the above properties requires the demonstration and documentation of relevant actions that have been taken at different stages of a project's lifecycle. These actions are captured through *claims* about the *properties*, which may apply to either the *project* or *system*. For example, the action of security testing a system may give rise to a property that is salient for either the *reliability* or *robustness* of the system. Devising a claim about the property of the system that has resulted from such an action may be a pre-requisite for arguing that a system is SAFE TO OPERATE.

The set of *system or project property claims* will collectively help specify how the goal is to be understood in the context of the project lifecycle. Figure 2 provides an example of this using the goal of explainability.

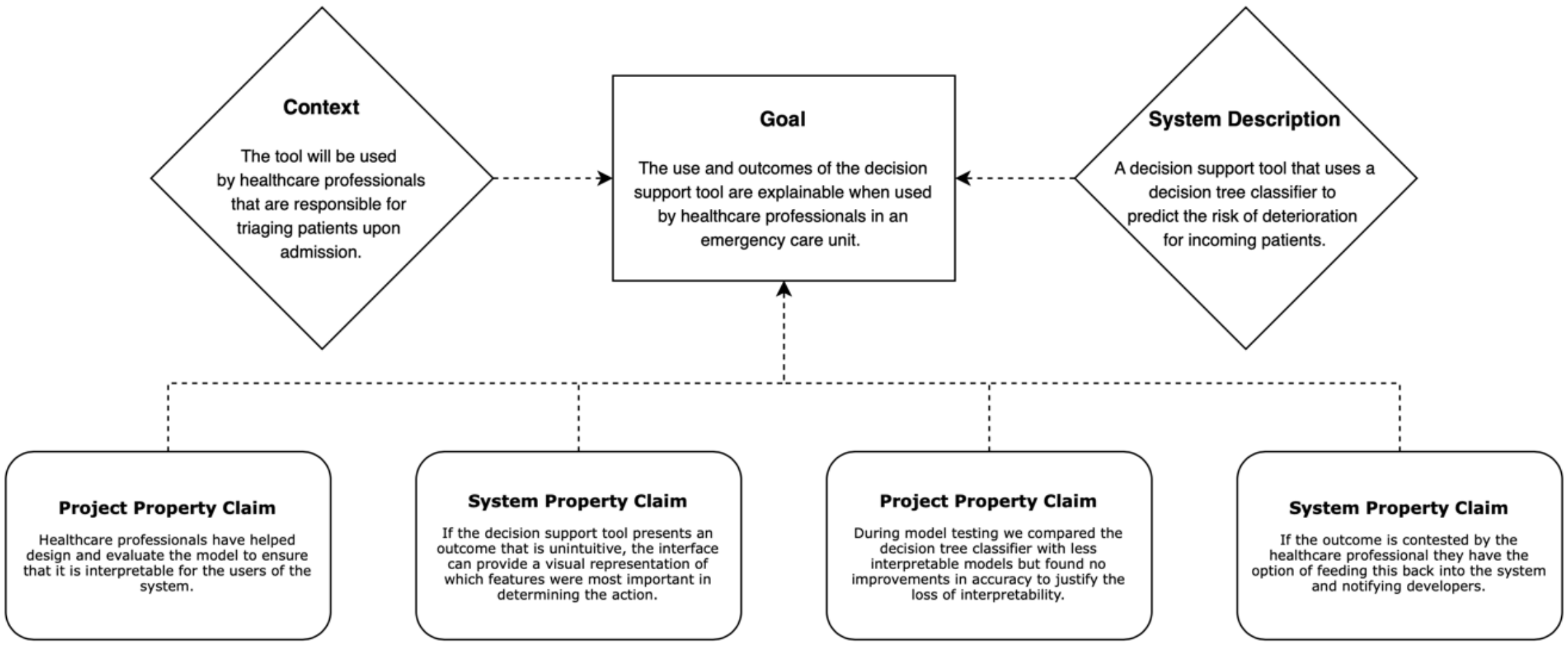

*Figure 2 – an example of a partial assurance case with an 'explainability' goal and supporting claims*

The PCRA, SEP, and HUDERIA all play an important role in helping to identify necessary and (jointly) sufficient claims. For instance, the SEP is designed to be maximally inclusive of differing perspectives, as it is through this open dialogue and engagement that developers will be likely to identify additional properties that need to be established in their project or system.

## The section below titled 'A HUDERAC Template

The process of building a HUDERAC is a three-stage process:

> Reflect ⇒ Act ⇒ Justify

Each stage can be broken into a series of steps, which help build a HUDERAC. The following sections outline these steps and provide a list of goals, actions, and claims that provide illustrative examples.

### *Reflect*

**STEPS:** Reflection is an anticipatory process, so the steps taken in this stage can be thought of as deliberative prompts:

1. What are the goals of your system?
2. How are these goals defined?
3. Which stakeholders have participated in the identification and defining of these goals?

4. What properties need to be implemented in the project or system to ensure that these goals are achieved?
5. Which actions ought to be taken to establish these properties within the project or system?

**OUTPUTS:** The HUDERAC outputs of this section will be a preliminary set of goals and claim.

### *Act*

**STEPS:** Salient actions are carried out at all stages of the HUDERAF, as detailed in the process map at the start of this document. A detailed process log and secure repository of documentation should be maintained to keep track and record these actions, including any associated artefacts (e.g., data protection impact assessment, report of model performance evaluation).

As the HUDERAF is split into three stages—design, development, and deployment—there is a high-level structure that can help with the organisation and documentation of relevant actions:

1. What actions have been undertaken during (project) design that have generated salient evidence for your goals and claims?
2. What actions have been undertaken during (model) development that have generated salient evidence for your goals and claims?
3. What actions have been undertaken during (system) deployment that have generated salient evidence for your goals and claims?

**OUPUTS:** A well maintained repository (or process log) of documentation and evidential artefacts, which demonstrate that specific actions were undertaken at specific times throughout the project lifecycle, by specific team members.

### *Justify*

**STEPS:** Once the goals, claims, and evidence have been established and documented, the final step is to justify that your evidence base is sufficient to warrant the claims that are being made about the properties of your project or system. This connection is a vital step and can expose weaknesses in the overall case being developed.

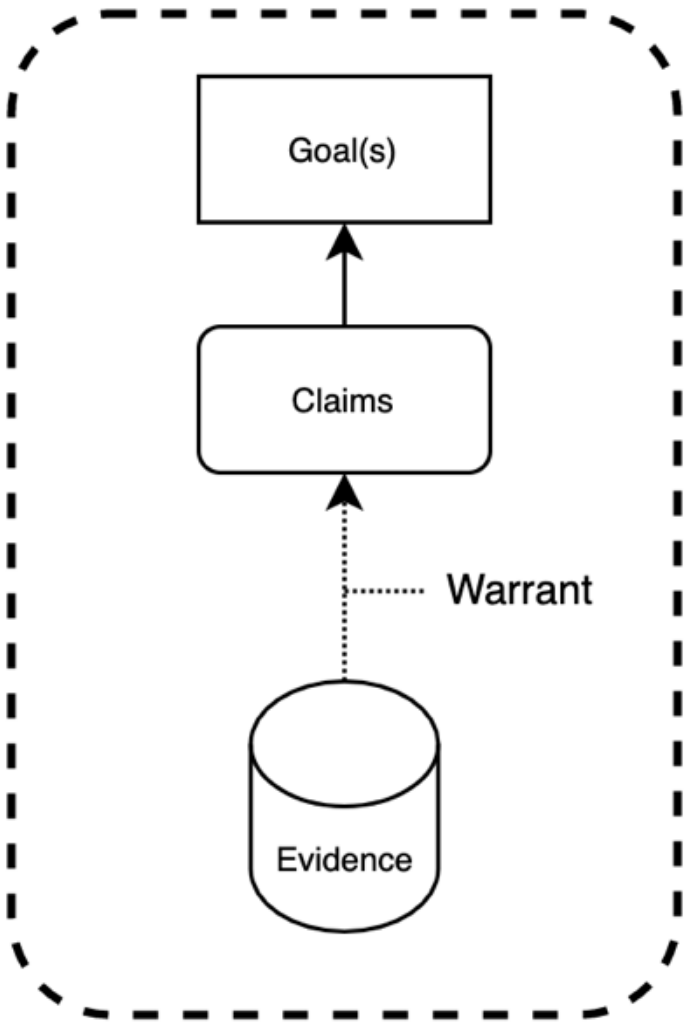

To help evaluate the evidence, the following questions can be instructive:

1. Which stakeholders, identified in your stakeholder engagement plan, can support the evaluation of your evidence and overall case?
2. Is any evidence missing from your case?
3. Are the collection of property claims jointly sufficient to support your top-level goal?

**OUPUTS:** A HUDERAC that can be presented and communicated to relevant stakeholder groups for assessment, evaluation, or auditing.

Goals, Actions and Claims' provides a set of example claims that are grouped according to the relevant SAFE-D goal and with reference to specific actions and respective properties.

### *Evidence*

A property claim that is made about the system or project will typically require supporting evidence that the relevant action was performed. For instance, consider the following claim,

> During model testing we compared the decision tree classifier with less interpretable models but found no improvements in accuracy to justify the loss of interpretability.

This claim helps justify why a specific model was selected. However, an auditor may want to see the results of the model comparison process with specific reference to the performance metrics that were used. Providing this *evidence* is an important justificatory step that helps ground the assurance case.

Generating this evidence will occur naturally over the course of a project's lifecycle, and the specific steps set out in the workflow summary have also been carefully designed to give rise to relevant evidential artefacts. Each decision or action taken, whether as part of the SEP or during a specific development stage (e.g., exploratory data analysis) is an action that is likely to or could create documentation that can serve as evidence for a property claim. It is not, therefore, necessary to compile all the evidence for an assurance case as an additional step at the end of a project's lifecycle. Instead, the HUDERAF has been designed to naturally accommodate the iterative process of building an assurance case as a project evolves. In the same way that an iterative approach to stakeholder engagement is recommended, therefore, the process of building a HUDERAC is also best approached as a continuous and iterative process.

## Building a HUDERAC

Building a HUDERAC is about connecting the evidence gathered during project design, model development, and system deployment to specific property claims, in order to form a structured argument that some goal (or set of goals) has been met. To simplify this process, we can think of the HUDERAF as an end-to-end process of reflection, action, and justification that gives rise to a HUDERAC.

Early-stage project design activities, such as the PCRA or SEP, are *reflective* and *anticipatory*—identifying risks and possible impacts early on before they lead to harm. They also help map out a project governance process that can identify which *actions* need to be taken to assure properties that are relevant to the goals of the system, and which may have been co-defined with stakeholders. The *justificatory* step of compiling evidence, therefore, is linked to the final production of a HUDERAC.

Figure 3 provides a high-level schematic of this process of reflection, action, and justification, showing how specific elements of an assurance case emerge from each of the stages.

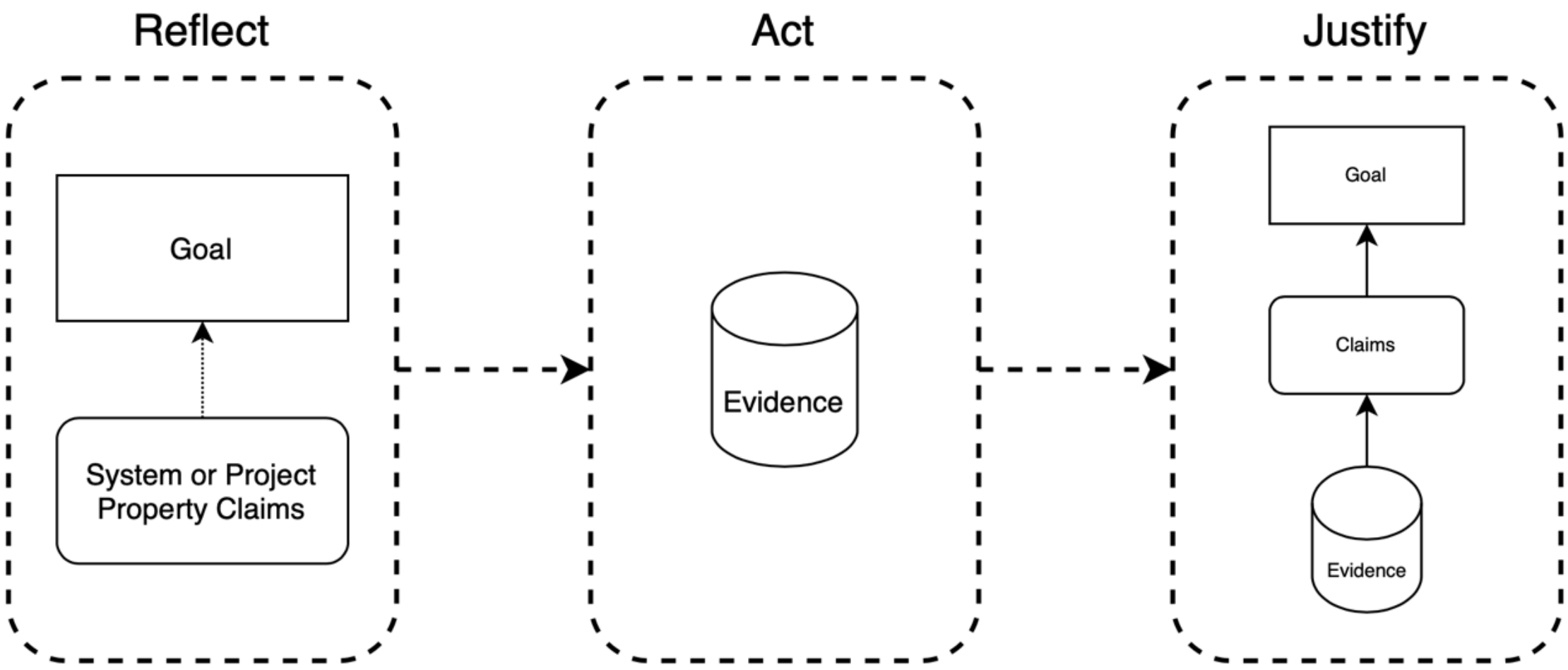

*Figure 3 – a high-level schematic of the reflect, act, and justify (HUDERAF) process*

The process of justification depends on compiling the relevant evidence and linking it to the specific claims. However, whether an assurance case is judged to be justified is a *relational process*. Although a developer can "make their case" for why a specific goal has been achieved, it must also be presented to relevant stakeholders (e.g., regulators, procurers) to *evaluate its justifiability*. In principle, this is an ongoing process as different groups of stakeholders may wish to contest or examine different property claims or question specific forms of evidence.

For example, let's pretend that a SEP report is used as evidence for a claim that a specific group of stakeholders were engaged and approved of a particular design choice for the respective AI system. This could seem, *prima facie*, like a reasonable evidential basis. However, let's also pretend that the stakeholders were only sent letters with the proposed design plans for the AI system and given a week to respond with objections, and that any non-responses would be treated as indicating approval. As we discussed earlier, this approach to stakeholder engagement is known as "informing", and is the weakest method of engagement. Whether it is accepted as reliable evidence is, therefore, likely to depend on wider, contextual factors.

The process of justifying a HUDERAC should, therefore, be approached as a process of building and maintaining a living document, and as part of the overall dialogical and reflective process that defines the HUDERAF.

In the next sections we provide a template for the development of a HUDERAC. This template serves a procedural function insofar as it offers a practical mechanism for project teams. It does not provide guidance on the substantive content that ought to be included within a particular HUDERAC, as this can only be determined by carrying out the actual stages of the HUDERAF (i.e., PCRA, SEP and HUDERIA).

## A HUDERAC Template

The process of building a HUDERAC is a three-stage process:

> Reflect ⇒ Act ⇒ Justify

Each stage can be broken into a series of steps, which help build a HUDERAC. The following sections outline these steps and provide a list of goals, actions, and claims that provide illustrative examples.

### *Reflect*

**STEPS:** Reflection is an anticipatory process, so the steps taken in this stage can be thought of as deliberative prompts:

6. What are the goals of your system?
7. How are these goals defined?
8. Which stakeholders have participated in the identification and defining of these goals?
9. What properties need to be implemented in the project or system to ensure that these goals are achieved?
10. Which actions ought to be taken to establish these properties within the project or system?

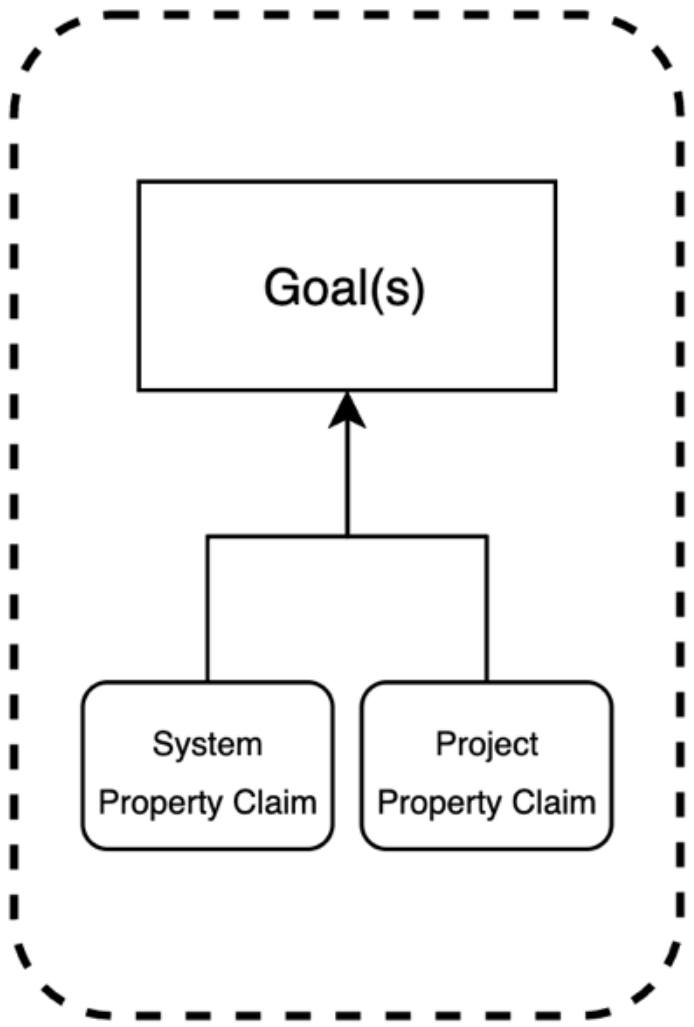

**OUTPUTS:** The HUDERAC outputs of this section will be a preliminary set of goals and claim.

### *Act*

**STEPS:** Salient actions are carried out at all stages of the HUDERAF, as detailed in the process map at the start of this document. A detailed process log and secure repository of documentation should be maintained to keep track and record these actions, including any associated artefacts (e.g., data protection impact assessment, report of model performance evaluation).

As the HUDERAF is split into three stages—design, development, and deployment—there is a high-level structure that can help with the organisation and documentation of relevant actions:

4. What actions have been undertaken during (project) design that have generated salient evidence for your goals and claims?
5. What actions have been undertaken during (model) development that have generated salient evidence for your goals and claims?
6. What actions have been undertaken during (system) deployment that have generated salient evidence for your goals and claims?

**OUPUTS:** A well maintained repository (or process log) of documentation and evidential artefacts, which demonstrate that specific actions were undertaken at specific times throughout the project lifecycle, by specific team members.

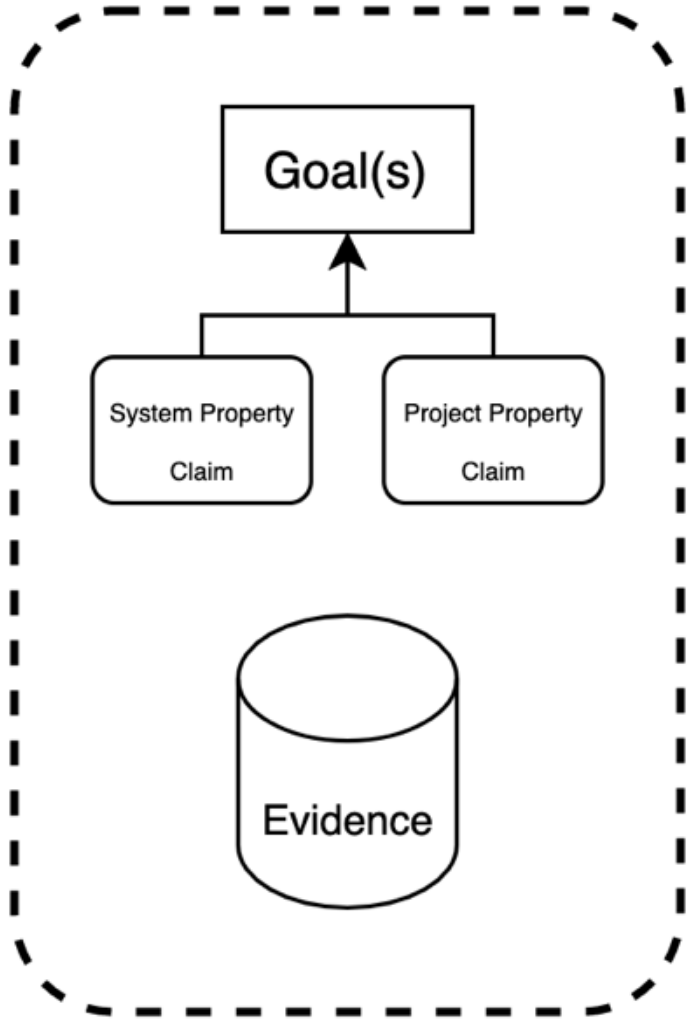

### *Justify*

**STEPS:** Once the goals, claims, and evidence have been established and documented, the final step is to justify that your evidence base is sufficient to warrant the claims that are being made about the properties of your project or system. This connection is a vital step and can expose weaknesses in the overall case being developed.

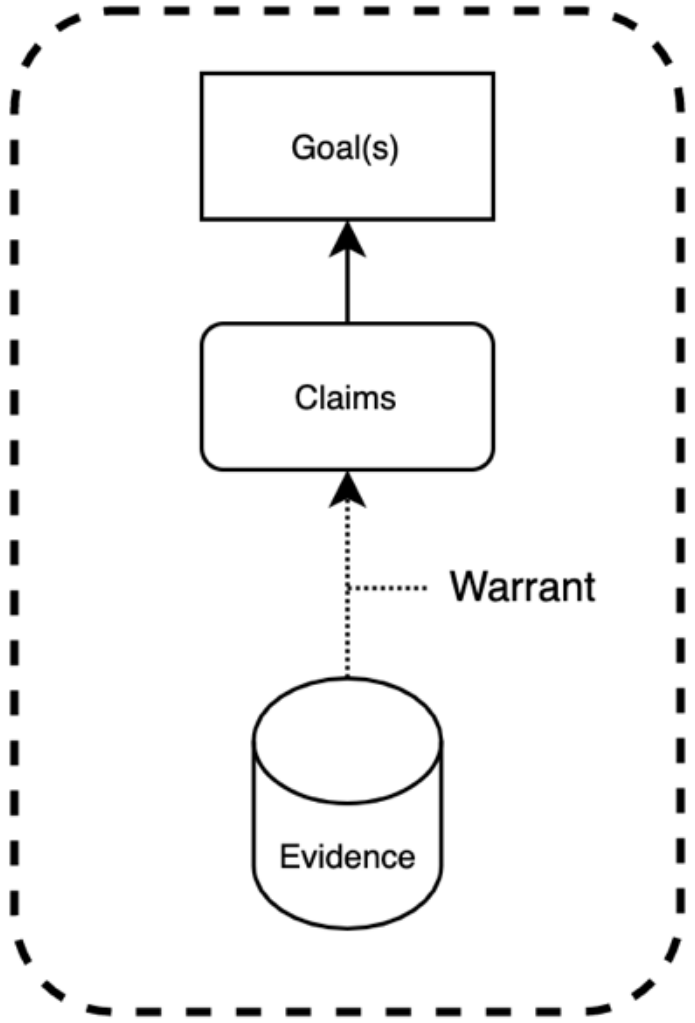

To help evaluate the evidence, the following questions can be instructive:

4. Which stakeholders, identified in your stakeholder engagement plan, can support the evaluation of your evidence and overall case?
5. Is any evidence missing from your case?
6. Are the collection of property claims jointly sufficient to support your top-level goal?

**OUPUTS:** A HUDERAC that can be presented and communicated to relevant stakeholder groups for assessment, evaluation, or auditing.

### *Goals, Actions and Claims*

It is not possible to offer a comprehensive set of claims that need to be established for each of the SAFE-D goals and properties, as the task of specifying the goals and operationalising the relevant properties is highly contextual.

However, Table 2 provides a set of examples to show how this may occur in different contexts. The claims are organised according to where in the design, development, and deployment lifecycle they occur. We do not give a claim for every stage, but do provide at least one example for every property, and also offer further references in footnotes to wider literature for each goal.

Finally, it is important to note that some of the properties may relate to and serve multiple goals, but are presented in connection with one in this table.

| **Goals** | | **Claims and Actions** | | |
|---|---|---|---|---|
| | | Design | Development | Deployment |
| **SAFETY**[71] | Sustainability | Previous safety cases for similar technologies were consulted during the planning of our system to anticipate and identify possible risks. | Key decision-makers collaborated with system developers to set reasonable expectations about what is achievable. | The system requires ongoing monitoring from a human-on-the-loop to minimise the risk of physical harm that could arise from its operation. |
| | Robustness | | The model used within our system has been internally and externally validated. External validation has been carried out across a wide range of environments to ensure the system is robust. | |

[71] CDT, n.d.; Leslie, 2019; Diakopoulos, et al., 2017; Amodei, et al., 2016; Auernhammer et al., 2019; Demšar & Bosnić, 2018; Google, 2019; Göpfert et al., 2018; Irving & Askell, 2019; Kohli, et al.., 2019; Kolter & Madry, n.d.; Marcus, 2018; Muñoz-González, et al., 2017; Nicolae, et al., 2018; Ortega & Maini, 2018; Ranjan, et al., 2017; Ratasich, et al., 2019; Salay & Czarnecki, 2018; Shi, et al., 2018; Song, et al., 2018; Warde-Farley & Goodfellow, 2016; Webb, et al., 2018; Zantedeschi, et al., 2017; Zhao, et al., et al., 2018; Zhang, et al., 2019; Varshney & Alemzadeh, 2017

| | | | | |
|---|---|---|---|---|
| | Security | | Our model was evaluated and developed to minimise its vulnerability to inversion attacks (NB: model inversion attacks attempt to reconstruct training data from model parameters). | An independent security analyst has carried out extensive penetration testing on our systems to ensure that sensitive data will not be revealed to non-trusted parties. |
| | Accuracy and Performance Metrics | The dataset was augmented using synthetic data, in collaboration with a domain and subject matter expert, to address known gaps and ensure the training data were representative and complete. | Diverse stakeholder groups were involved with the model's development to ensure it performed reliably within and between each group. | |
| | Reliability | | The runtime environment of our model is well controlled and understood. Our model training was oriented towards performance in this environment. | Model reliability has been evaluated and optimised using sensitivity analyses and perturbations to training data to minimise the risk of encountering novel data in the runtime environment. |
| **ACCOUNTABILITY**[72] | Traceability | The origin of our data sources and collection processes have been sufficiently documented to support reproducibility. | | A complete description of all data collected by our system is available through our privacy policy, and the data types are linked to their specific uses. |

[72] AI Now Institute, 2018; Leslie, 2019; Binns, 2018; Cavoukian, et al., 2010; CDT (n.d.); Diakopoulos, et al., 2015; Diakopoulos, et al., 2017; Kroll et al., 2016; Malgieri & Comandé, 2017; O'Sullivan et al., 2019; Reed, 2018; Stahp & Wright, 2018; Wachter, et al, (2017a); Zook, et al., 2017; ACM US Public Policy Council, 2017; Ananny & Crawford, 2018; Antunes, et al., 2018; Burrell, 2016; Citron, 2008; Citron & Pasquale, 2014; Crawford & Schultz, 2014; Edwards & Veale, 2017; Kemper & Kolkman, 2019; Turilli & Floridi, 2009; Weller, 2017; Lepri, et al., 2018; Mittelstadt, et al., 2016; Selbst, et al., 2019; Suresh & Guttag, 2019; Veale, et al., 2018; Davidson & Freire, 2008; Buhmann, et al., 2020; Cech, 2020; Cobbe, et al., 2021; Diakopoulos, 2014; Diakopoulos, 2015; Diakopoulos, 2016; FAT/ML, 2016; Fink, 2018; European Parliamentary Research Service, 2019; Hamilton, et al., 2014; Hutchinson, et al., 2021; Kacianka & Pretschner, 2021; Kaminski, 2018; Wieringa, 2020; Young, et al., 2019; Zicari, et al., 2021; Tagiou, et al., 2019; Shah, 2018; ICO, 2020; Rosenblat, et al., 2014; Rosenbaum & Fichman, 2019; Reddy, et al., 2019

| | | | | |
|---|---|---|---|---|
| | Auditability | | All relevant details of our model training, testing, and validation have been recorded on an accessible team repository (e.g., predictor selection processes, baselines for model comparison, performance metrics). | |
| | Clear Data Provenance and Data Lineage | | System logs are stored securely and kept for a period of up to 1 year to support any internal or external audit or review. | |
| | Accessibility | All identified stakeholders were consulted prior to the development of our system to help critically evaluate our project plans and ensure they were intelligible. | | Where our system is deployed in public spaces, clear messaging is established to ensure all individuals are aware that the system is in operation. |
| | Reproducibility | | | An extensive set of documentation has been made available to any individual who wishes to deploy our model in their own system, including suggestions about appropriate/inappropriate use cases. |
| | Responsible Governance | A data protection officer has identified and confirmed the appropriate controllers and processers of the data. | | A member of the team is tasked with monitoring use of the system for compliance with relevant standards and is authorised to revoke use if found to be misused and/or harmful. |
| | Answerability | | Members of the development team are free and empowered | |
| | | | to object to potentially harmful design decisions without fear of reprisal. | |
| **FAIRNESS**[73] | Non-Discrimination | Members of identity and demographic groups that are most at risk of harm by AI systems are consulted about design intentions to help identify and remove discriminatory effects of the AI system. | During pre-processing, steps were taken to balance the distribution of the data sets' classes.<br><br>Post-processing techniques were used to minimise our classifier's correlation with protected attributes. | Users are constrained by design or policy from using the system to profile or target persons based on protected characteristics.<br><br>Persons affected by use of the system have avenues of recourse, ability to contest system outputs, and demand human intervention. |
| | Equality | An equality impact assessment was performed to identify and mitigate potential discriminatory harms that could arise through the deployment of this system. | The system has been thoroughly tested to ensure its benefits and harms are distributed across identities and groups in an equitable manner. | The system has been evaluated by independent domain experts to ensure that it does not impinge on expressive rights.<br><br>The system will not make use of personal data or reveal personal data to others without the unambiguous consent of the subject. |
| | Diversity and Inclusiveness | The problem being addressed by the system was formulated in a multistakeholder process to ensure the inclusion of a broad range of perspectives and potential concerns. | | The system has been tested by independent domain experts to ensure that it will function similarly for differently situated persons. |

[73] Diakopoulos, et al., 2017; Leslie, 2019; Donovan, et al., 2018; Binns, 2017; Binns, et al., 2018; Holstein, et al., 2018; Lepri, et al., 2018; Custers, 2013; Custers & Schermer, 2014; European Commission Expert Group on FAIR Data, 2018; L'heureux, et al., 2017; Ruggieri, et al., 2010; Hajian, et al., 2016; Kamiran & Calders, 2012; Lehr & Ohm, 2017; Passi & Barocas, 2019; Singhal & Jena, 2013; van der Aalst, et al., 2017; Corbett-Davies, et al., 2017; Dwork, et al., 2012; Grgić-Hlača, et al., 2017; Kusner, et al., 2017; Verma & Rubin, 2018; Zafar, et al., 2015; Žliobaitė , 2017

| | | | | |
|---|---|---|---|---|
| | | | | |
| | Bias Mitigation | Data were collected, extracted, or acquired with the meaningful consent of the data subjects.<br><br>Independent domain experts analysed the data to determine acceptable levels of bias. | | Users have been fully trained to identify and mitigate "human factors" concerns associated with system use. |
| **EXPLAINABILITY**[74] | Accessible Rationale Explanation | Data sets are open and accessible for analysis by independent auditors. | Our system was tested with a diverse set of end users to ensure that its outputs were accessible at the time they were needed and delivered relevant and interpretable information. | Previously identified stakeholders were consulted again towards the end of the project lifecycle to evaluate whether they were happy with the accessibility and comprehensiveness of the explanations being offered by the system. |
| | Responsible Implementation and User Training | | User testing was conducted prior to full deployment of the system. This involved a representative sample of users who were asked to ensure they could satisfactorily interpret the outputs of the system and provide suitable explanations | The interface through which users will interact with our system, has been developed to meet universal design standards and promote accessibility for all users, including those with visual or cognitive impairments. |

[74] CDT, n.d.; Diakopoulos, et al., 2017; Leslie, 2019; Janssen & Kuk, 2016; Wachter, et al, (2017a); Wachter et al., 2017b; Demšar & Bosnić, 2018; Edwards & Veale, 2017; Adadi & Berrada, 2018; Bathaee, 2018; Bibal & Frénay, 2016; Bracamonte, 2019; Burrell, 2016; Card, 2017; Caruana, et al., 1999; Chen, et al., 2018; Doshi-Velez & Kim, 2017; Doshi-Velez, et al., 2017; Dosilovic, et al., 2018; Eisenstadt & Althoff, 2018; Feldmann, 2018; Gilpin, et al., 2018; Guidotti, et al., 2018; Kleinberg, et al., 2017; Kroll, 2018; Lakkaraju, et al., 2016; Lepri, et al., 2018; Lipton, 2016; Lipton & Steinhardt, 2018; Lundberg & Lee, 2017, Mittelstadt, et al., 2019; Moinar, 2018; Murdoch, et al., 2019; Olhede & Wolfe, 2018; Park, et al., 2016; Pedreschi, et al., 2018; Pedreschi, et al., 2019; Poursabzi-Sangdeh, et al., 2018; Ribeiro, et al., 2016b; Rudin, 2018; Rudin & Ustun, 2018; Shmueli, 2010; Shaywitz, 2018; Simonite, 2017; Sokol & Flach, 2018; ICO & ATI, 2020

| | | | | |
|---|---|---|---|---|
| | | | about its functions to affected persons. | |
| | Responsible Model Selection | | Features were selected to optimise for both interpretability and predictive power. | |
| | Interpretability | A range of stakeholder engagement session were held to facilitate accessible participation and identify specific needs and challenges of the respective groups. | | |
| **DATA QUALITY**[75] | Source Integrity and Measurement Accuracy | An automated script has been established that automatically flags for human checking any unexpected deviations, missing data, or unexpected formatting. | | Mechanisms have been established that monitor continuous training data, and test/verify their continued accuracy. |
| | Timeliness and Recency | Only data that were collected within the previous 3 months were used to ensure the training data were up-to-date. | | New data are collected and used to retrain/revalidate the model every month. |
| | Relevance, Appropriateness, and Domain Knowledge | Domain experts were consulted about the findings from our exploratory data analysis, and verified the relevance of the input variables. | | |
| | Adequacy of Quantity and Quality | | External validation of the model was carried out prior to full deployment to verify whether the training data were adequate to | |

[75] Abiteboul, et al., 2015; Dai, et al., 2018; Leslie, 2019

| | | | | |
|---|---|---|---|---|
| | | | stand-in for data encountered in novel settings. | |
| | Balance and Representativeness | | Cross-validation was carried out to determine the generalisability of training data samples and models. | Automated triggers have been setup to frequently check whether the model is still representative of the original data generation process. |
| **Data Integrity**[76] | Attributable | Metadata is stored alongside the datasets to help aid and improve data analysis. | | |
| | Consistent, Legible and Accurate | Multiple copies of the dataset were stored securely and used to verify consistency throughout the project. | | |
| | Contemporaneous | Data collection was carried out using a bespoke application to avoid timely transcription and ensure the data were captured in a relevant and labelled format. | | New data that is collected in the runtime environment of the system is gathered in a structured format and with accompanying metadata. |
| | Responsible Data Management | Our project team completed tailored training on data governance prior to the start of the project. | | Data is stored in an interoperable and reusable format to promote replicability and transparency. |
| | Data Traceability and Auditability | | Training and testing data splits have been fully documented. | |

[76] Ambacher, et al., 2007; Faudeen, 2017; Stoyanovich, et al., 2017; SL Controls (n.d.)

| | | | | |
|---|---|---|---|---|
| **Data Protection and Privacy**[77] | Consent (or legitimate basis) for processing | | | |
| | Data Security | Before data is copied or transferred outside of our secure development environment, a multi-party approval system is triggered to minimise the risk of security leaks. | | |
| | Data Minimisation | Domain experts and stakeholders were consulted to identify the minimum level of data that needed to be collected to ensure satisfactory performance of the system. | | |
| | Transparency | | | A document is available on our website that explains how our model works and also provides details of the relevant input features and outputs. |
| | Proportionality | | Model comparison was undertaken to determine that all personal data that are collected were necessary for the purposes of the task by iteratively removing different types and recording the decrease in model performance. | |
| | Purpose Limitation | | | Before using the system a user is provided with a privacy policy outlining the purposes for all data |

[77] ICO, 2017; Council of Europe, 2018; Cavoukian, et al., 2010; ICO, 2021; Malgieri & Comandé, 2017; Veale, et al., 2018; Wachter et al., 2017b; Alper, et al., 2018; Antignac, et al., 2016; Kaminski, 2018

| | | | | |
|---|---|---|---|---|
| | | | | that is collected and required to acknowledge informed consent to proceed. |
| | Accountability | | | Mechanisms have been established that allow users to request full data erasure if they no longer want to use the system. |
| | Lawfulness, Fairness, and Transparency | The system is designed to function without the need to collect any personal information. | Personal data for each of the records were collected to verify that none of the extracted features contained hidden proxies. Once this assessment had been carried out, the personal data were destroyed. | |
| | Respect for the rights of data subjects | | The model selection process was constrained by the requirement that any model must be interpretable to ensure their right to be informed is respected. | |

*Table 2 – List of goals and example claims and actions that could be used in a HUDERAC*

# Appendix 1: Summary of Processes, Steps, and User Activities

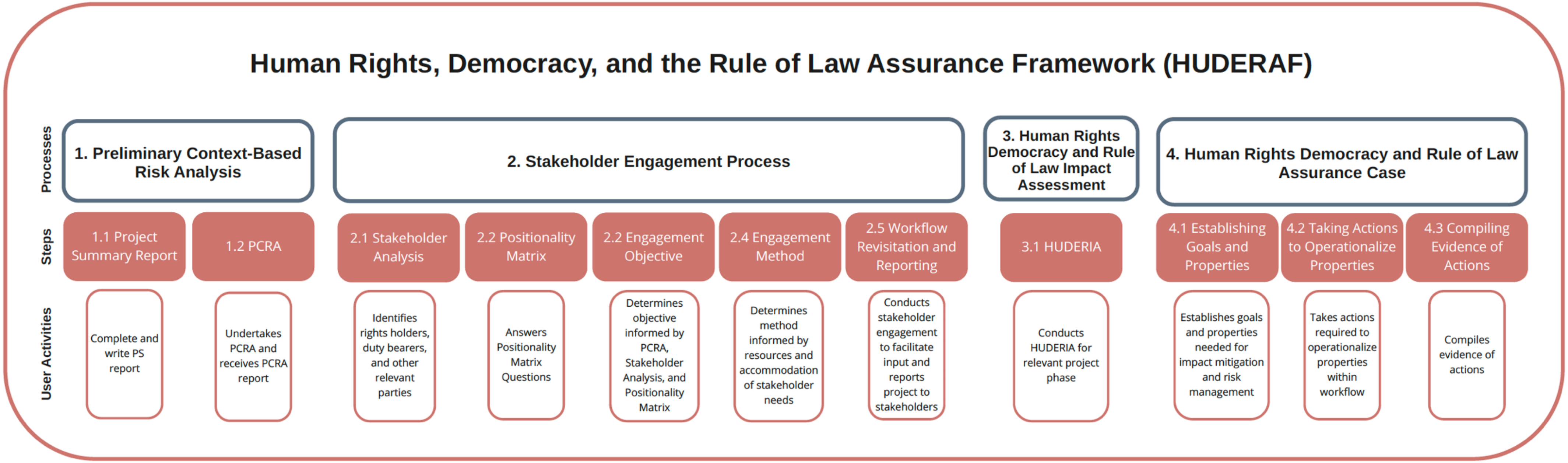

# Appendix 2: HUDERAF Project Lifecycle Process Map

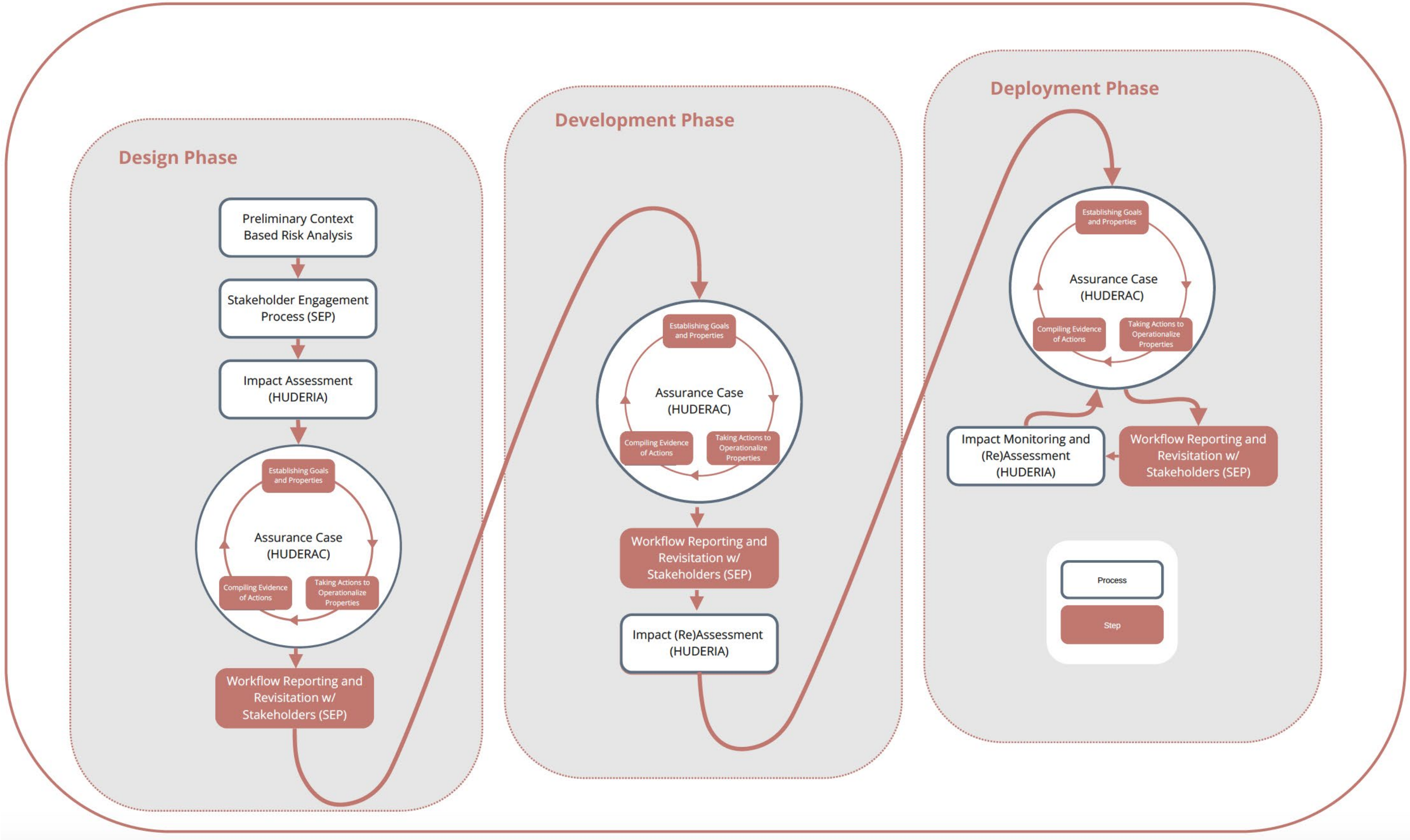

# Appendix 3: A Sociotechnical Approach to the ML/AI Project Lifecycle[78]

There are many ways of carving up a project lifecycle for a ML/AI system or other data-driven technology (hereafter shortened to just 'project lifecycle'). For instance, Sweenor et al. (2020) break it into four stages: Build, Manage, Deploy & Integrate, Monitor.[79] Similarly, (Ashmore, Calinescu, and Paterson 2019) identify four stages, which have a more specific focus on data science: data management, model learning, model verification, and model deployment. Furthermore, there are also well-established methods that seek to govern common tasks within a project lifecyle, such as data mining (e.g. CRISP-DM or SEMMA).

The multiplicity of approaches is likely a product of the evolution of diverse methods in data mining/analytics, the significant impact of ML on research and innovation, and the specific practices and considerations inherent to each of the various domains where ML techniques are applied. While there are many benefits of existing frameworks (e.g. carving up a complex process into smaller components that can be managed by a network of teams or organisations), they do not tend to focus on the wider social or ethical aspects that interweave throughout the various stages of a ML lifecycle. Figure 2, therefore, presents a model of the ML lifecycle, which we have designed to support the process of building an ethical assurance case, while remaining faithful to the importance of technical requirements and challenges and also supporting more open, reflective, and participatory forms of deliberation.

[78] This Appendix reproduces a section of Christopher Burr and David Leslie (2021), "Ethical Assurance: A practical approach to the responsible design, development, and deployment of data-driven technologies" Forthcoming, *The Alan Turing Institute*.

[79] These four stages are influenced by an 'ML OPs' perspective (Sweenor et al. 2020). The term 'MLOps' refers to the application of DevOps practices to ML pipelines. The term is often used in an inclusive manner to incorporate traditional statistical or data science practices that support the ML lifecycle, but are not themselves constitutive of machine learning (e.g. exploratory data analysis), as well as deployment practices that are important within business and operational contexts (e.g. monitoring KPIs).

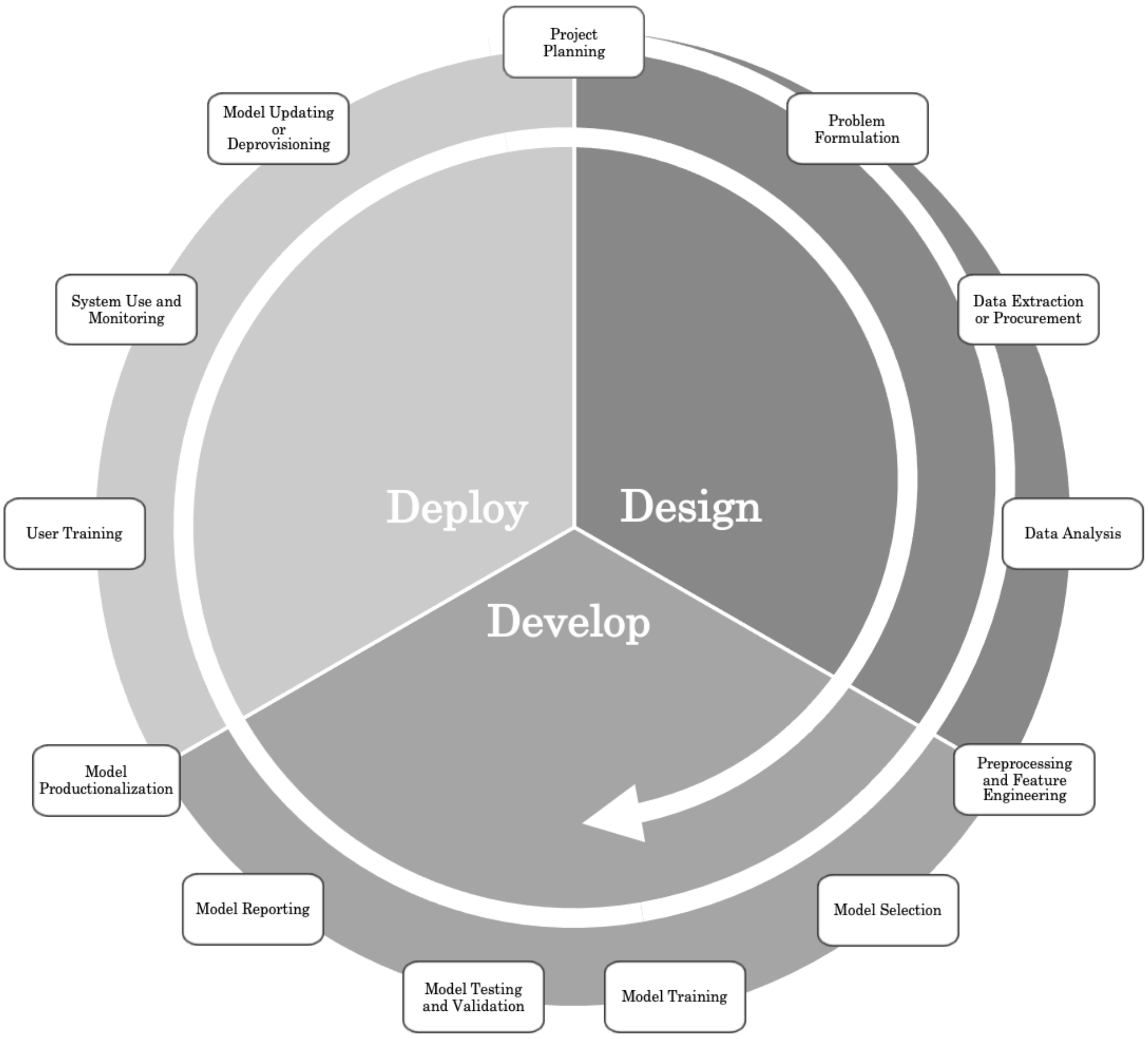

*The Project Lifecycle—the overarching stages of design, development, and deployment (for a typical data-driven project) can be split into indicative tasks and activities. In practice, both the stages and the tasks will overlap with their neighbours, and may be revisited where a particular task requires an iterative approach. The spiral indicates that this is a diachronic, macroscopic process that evolves and develops over time, and as the deployment stage finishes, a new iteration is likely to begin.*

To begin, the inner circle breaks the project lifecycle into a process of (project) design, (model) development, and (system) deployment. These terms are intended to be maximally inclusive. For example, the *design stage* encompasses any project task or decision-making process that scaffolds or sets constraints on later project stages (i.e. design system constraints). Importantly, this includes ethical, social, and legal constraints, which we will discuss later.

Each of the stages shades into its neighbours, as there is no clearly delineated boundary that differentiates certain *project* design activities (e.g. data extraction and exploratory analysis) from *model* design activities (e.g. preprocessing and feature engineering, model selection). As such, the design stage overlaps with the *development stage*, but the latter extends to include the actual process of training, testing, and validating a ML model. Similarly, the process of productionizing a model within its runtime environment can be thought of as both a development and deployment activity. And, so, the *deployment stage* overlaps with the 'development' stage, and also overlaps with the 'design' stage, as the deployment of a system should be thought of as an ongoing process (e.g. where new data are used to continuously train the ML model, or, the decision to de-provision a model may require the planning and design of a new model, if the older (legacy) system becomes outdated). For these reasons, the project lifecycle is depicted as a spiral. However, despite the unidirectional nature of the arrows, we also acknowledge that ML/AI research and innovation is frequently an iterative process. Therefore, the singular direction is only present at a macroscopic level of abstraction (i.e., the *overall* direction of progress for a project), and allows for some inevitable back and forth between the stages at the microscopic level.

The three higher-level stages can be thought of as a useful heuristic for approaching the project lifecycle. However, each higher-level stage subsumes a wide variety of tasks and activities that are likely to be carried out by different individuals, teams, and organisations, depending on their specific roles and responsibilities (e.g. procurement of data). Therefore, it is important to break each of the three higher-level stages into their (typical) constituent parts, which are likely to vary to some extent between specific projects or within particular organisations. In doing so, we expose a wide range of diverse tasks, each of which give rise to a variety of ethical, social, and legal challenges. The following sections provides an illustrative overview of these stages and tasks, as well as a *non-exhaustive* sample of the associated challenges.

## (Project) Design Tasks and Processes

### *Project Planning*

Rather than using ML/AI as a "hammer" to go looking for nails, it is best to have a clear idea in mind of what the project's goals are at the outset. This can help to avoid a myopic focus on a narrow class of ML/AI-based "solutions," and also helps create space for a diversity of approaches—some of which may not require ML/AI at all. Project planning, therefore, can comprise a wide variety of tasks, including, but not limited to:

- an assessment of whether building an AI model is the right approach given available resources and data, existing technologies and processes already in place, the complexity of the use-contexts involved, and the nature of the policy or social problem that needs to be solved (Leslie et al 2021a);
- an analysis of user needs in relation to the prospective AI model and whether a solution involving the latter provides appropriate affordances in keeping with user needs and related functional desiderata;
- mapping of key stages in the project to support governance and business tasks (e.g. scenario planning);
- an assessment of resources and capabilities within a team, which is necessary for identifying any skills gaps,
- a contextual assessment of the target domain and of the expectations, norms, and requirements that derive therefrom;

- stakeholder analysis and team positionality reflection to determine the appropriate level and scope of community engagement activities (Leslie et al 2021b);
- stakeholder impact assessment, supported by affected people and communities, to identify and evaluate possible harms and benefits associated with the project (e.g. socioeconomic inequalities that may be exacerbated as a result of carrying out the project), to gain social license and public trust, and also feed into the process of problem formulation in the next stage;
- wider impact assessments—both where required by statute and done voluntarily for transparency and best practice (e.g. equality impact assessments, data protection impact assessments, human rights impact assessment, bias assessment)

### *Problem Formulation*

Here, 'problem' refers both to a well-defined computational process (or a higher-level abstraction of the process) that is carried out by the algorithm to map inputs to outputs and to the wider practical, social, or policy issue that will be addressed through the translation of that issue into the statistical or mathematical frame. For instance, on the computational side, a convolutional neural network carries out a series of successive transformations by taking (as input) an image, encoded as an array, in order to produce (as output) a decision about whether some object is present in the image. On the practical, social, and policy side, there will be a need to define the computational "problem" being solved in terms of the algorithmic system's embeddedness in the social environment and to explain how it contributes to (or affects) the wider sociotechnical issue being considered. In the convolutional neural network example, the system being produced may be a facial recognition technology that responds to a perceived need for the biometric identification of criminal suspects by matching face images in a police database. The social issue of wanting to identify suspects is, in this case, translated into the computational mechanism of the computer vision system. But, beyond this, diligent consideration of the practical, social, or policy issue being addressed by the system will also trigger, *inter alia*, reflection on the complex intersection of potential algorithmic bias, the cascading effects of sociohistorical patterns of racism and discrimination, wider societal and community impacts, and the potential effects of the use of the model on the actors in the criminal justice systems who will become implementers and subjects of the technology.

Sociotechnical considerations are also important for determining and evaluating the choice of target variables used by the algorithm, which may ultimately be implemented within a larger automated decision-making system (e.g. in a verification system). The task of formulating the problem allows the project team to get clear on what input data will be needed, for what purpose, and whether there exists any representational issues in, for example, how the target variables are defined. It also allows for a project team (and impacted stakeholders) to reflect on the reasonableness of the measurable proxy that is used as a mathematical expression of the target variable, for instance, whether being taken into care within six months of a visit from child protective services is a reasonable proxy for a child's being "at risk" in a predictive risk model for children's social care. The semantic openness and contestability of formulating problems and defining target variables in ML/AI innovation lifecycles is why stakeholder engagement, which helps bring a diversity of perspectives to project design, is so vital, and why this stage is so closely connected with the interpretive burdens of the project planning stage (e.g. discussion about legal and ethical concerns regarding permissible uses of personal or sensitive information).

### *Data Extraction or Procurement*

Ideally, the project team should have a clear idea in mind (from the planning and problem formulation stages) of what data are needed prior to extracting or procuring them. This can help mitigate risks associated with over-collection of data (e.g. increased privacy or security concerns) and help align the project with values such as *data minimisation* (ICO and Institute 2020). Of course, this stage may need to be revisited after carrying out subsequent tasks (e.g. preprocessing, model testing) if it is clear that insufficient or imbalanced data were collected to achieve the project's goals. Where data is procured, questions about provenance arise (e.g. legal issues, concerns about informed consent of human data subjects). Generally, responsible data extraction and procurement require the incorporation of domain expertise into decision-making so that desiderata of data minimisation as well as of securing relevant and sufficient data can be integrated into design choices.

### *Data Analysis*

Exploratory data analysis is an important stage for hypothesis generation or uncovering possible limitations of the dataset that can arise from missing data, in turn identifying the need for any subsequent augmentation of the dataset to deal with possible class imbalances. However, there are also risks that stem from cognitive biases (e.g. confirmation bias) that can create cascading effects that effect downstream tasks (e.g. model reporting).

## (Model) Development Tasks and Processes

### *Preprocessing and Feature Engineering*

Pre-processing and feature engineering is a vital but often lengthy process, which overlaps with the design tasks in the previous section and shares with them the potential for human choices to introduce biases and discriminatory patterns into the ML/AI workflow. Tasks at this stage include data cleaning, data wrangling or normalisation, and data reduction or augmentation. It is well understood that the methods employed for each of these tasks can have a significant impact on the model's performance (e.g. deletion of rows versus imputation methods for handling missing data). As Ashmore et al. (2019) note, there are also various desiderata that motivate the tasks, such as the need to ensure the dataset that will feed into the subsequent stages is relevant, complete, balanced, and accurate. At this stage, human decisions about how to group or disaggregate input features (e.g. how to carve up categories of gender or ethnic groups) or about which input features to exclude altogether (e.g. leaving out deprivation indicators in a predictive model for clinical diagnostics) can have significant downstream influences on the fairness and equity of an ML/AI system.

### *Model Selection*

This stage determines the model type and structure that will be produced in the next stages. In some projects, model selection will result in multiple models for the purpose of comparison based on some performance metric (e.g. accuracy). In other projects, there may be a need to first of all implement a pre-existing set of formal models into code. The class of relevant models is likely to have been highly constrained by many of the previous stages (e.g. available resources and skills, problem formulation), for instance, where the problem demands a supervised learning algorithm instead of an unsupervised learning algorithm; or where explainability considerations require a more interpretable model (e.g. a decision tree).

### *Model Training*

Prior to training the model, the dataset will need to be split into training and testing sets to avoid model overfitting. The *training set* is used to fit the ML model, whereas the *testing set* is a hold-out sample that is used to evaluate the fit of the ML model to the underlying data distribution. There are various methods for splitting a dataset into these components, which are widely available in popular package libraries (e.g. the scikit-learn library for the Python programming language). Again, human decision-making at this stage about the training-testing split and about how this shapes desiderata for external validation—a subsequent process where the model is validated in wholly new environments—can be very consequential for the trustworthiness and reasonableness of the development phase of an ML/AI system.

### *Model Validation and Testing*

The testing set is typically kept separate from the training set, in order to provide an unbiased evaluation of the final model fit on the training dataset. However, the training set can be further split to create a validation set, which can then be used to evaluate the model while also *tuning model hyperparameters*. This process can be performed repeatedly, in a technique known as (k-fold) cross-validation, where the training data are resampled (*k*-times) to compare models and estimate their performance in general when used to make predictions on unseen data. This type of validation is also known as 'internal validation,' to distinguish it from external validation, and, in a similar way to choices made about the training-testing split, the manner in which it is approached can have critical consequences for how the performance of a system is measured against the real-world conditions that it will face when operating "in the wild."

### *Model Reporting*

Although the previous stages are likely to create a series of artefacts while undertaking the tasks themselves, model reporting should also be handled as a separate stage to ensure that the project team reflect on the future needs of various stakeholders and end users. While this stage is likely to include information about the performance measures used for evaluating the model (e.g. decision thresholds for classifiers, accuracy metrics), it can (and should) include wider considerations, such as intended use of the model, details of the features used, training-testing distributions, and any ethical considerations that arise from these decisions (e.g. fairness constraints, use of politically sensitive demographic features).[80]

## (System) Deployment Tasks and Processes

### *Model Productionalization*

Unless the end result of the project is the model itself, which is perhaps more common in scientific research, it is likely that the model will need to be implemented within a larger system. This process, sometimes known as 'model operationalisation,' requires understanding (a) how the model is intended to function in the proximate system (e.g. within an agricultural decision support system used to predict crop yield and quality) and (b) how the model will impact—and be impacted by—the functioning of the wider sociotechnical environment that the tool is embedded within (e.g. a decision support tool used in healthcare for patient triaging that may exacerbate existing health inequalities within the wider community). Ensuring the model works within the proximate system can be a complex programming and software engineering task, especially if it is expected that the model will be updated continuously in its runtime environment. But, more importantly, understanding how to ensure the model's sustainability given its embeddedness in complex and changing sociotechnical environments requires active and contextually-informed monitoring, situational awareness, and vigilant responsiveness.

### *User Training*

Although the performance of the model is evaluated in earlier stages, the model's impact cannot be entirely evaluated without consideration of the human factors that affect its performance in real-world settings. The impact of human cognitive biases, such as algorithmic aversion[81] must also be considered, as such biases can lead to over- and under-reliance on the model (or system), in turn negating any potential benefits that may arise from its use. Understanding the social and environmental context is also vital, as sociocultural norms may contribute to how training is received, and how the system itself is evaluated (see Burton et al. 2020).

### *System Use and Monitoring*

Depending on the context of deployment, it is likely that the performance of the model could degrade over time. This process of *model drift* is typically caused by increasing variation between how representative the training dataset was at the time of development and how representative it is at later stages, perhaps due to changing social norms (e.g. changing patterns of consumer spending, evolving linguistic norms that affect word embeddings). As such, mechanisms for monitoring the model's performance should be instantiated within the system's runtime protocols to track model drift, and key thresholds should be determined at early stages of a project (e.g. during project planning or in initial impact assessment) and revised as necessary based on monitoring of the system's use.

[80] There is some notable overlap between this stage of the project lifecyle and the ethical assurance methodology, as some approaches to model reporting often contain similar information that is used in building an ethical assurance case (Mitchell et al. 2019; Ashmore, Calinescu, and Paterson 2019), specifically in the process of establishing evidential claims and warrant (see §§4.3.3-4.3.4).

[81] Algorithmic aversion refers to the reluctance of human agents to incorporate algorithmic tools as part of their decision-making processes due to misaligned expectations of the algorithm's performance (see Burton et al. 2020).

### *Model Updating or De-provisioning*

As noted previously, model updating can occur continuously if the architecture of the system and context of its use allows for it. Otherwise, updating the model may require either revisiting previous stages to make planned adjustments (e.g. model selection and training), or if more significant alterations are required the extant model may need to be entirely de-provisioned, necessitating a return to a new round of project planning.

# Glossary

**Access to effective remedy:** A core component of human rights impact assessments. This requires organisations to ensure appropriate steps are in place to prevent, investigate, punish, and redress negative impacts. Stakeholder engagement is required to ensure that remedies reflect the needs and interests of impacted stakeholders.

**Accessibility:** Ensuring that information about the processes that took place to design, develop, and deploy an AI system are easily accessible by individuals. This not only refers to suitable means of explanation (clear, understandable, and accessible language) but also the mediums for delivery.

**Accuracy and Performance Metrics:** In machine learning, the accuracy of a model is the proportion of examples for which it generates a correct output. This performance measure is also sometimes characterised conversely as an error rate or the fraction of cases for which the model produces an incorrect output. As a performance metric, accuracy should be a central component to establishing and nuancing the approach to safe AI. Specifying a reasonable performance level for the system may also often require refining or exchanging of the measure of accuracy. For instance, if certain errors are more significant or costly than others, a metric for total cost can be integrated into the model so that the cost of one class of errors can be weighed against that of another

**Adequate quantity and quality:** This property involves assessing whether the data available is comprehensive enough to address the problem set at hand. Adequate quantity and quality should address sample size, representativeness, and availability of features relevant to problem.

**Arbitrary deprivation:** Where a physical or expressive freedom, or the personal security of a rightsholder is removed or withheld, such removal or withholding is *arbitrary* if it is not justified by a publicly legitimatised legal framework or social practice that reflects the principles and intent of human rights doctrine. Where such deprivation occurs through legislation or other policy instrument, hallmarks of legitimate deprivation include legal due process and the availability of an effective remedy for affected rights-holders to contest the lawfulness of the deprived freedom or security.

**Attributable:** Data should clearly demonstrate who observed and recorded it, when it was observed and recorded, and who it is about.

**Auditability:** Whereas the notion of answerability responds to the question of who is accountable for an automation supported outcome, the notion of auditability answers the question of how the designers and implementers of AI systems are to be held accountable. This aspect of accountability has to do with demonstrating both the responsibility of design and use practices and the justifiability of outcomes.

**Automated Decision:** The selection of an action or a recommendation made using computational processes. Automated decisions describe those that either augment or replace decisional work typically performed by humans alone. Most commonly, automated decisions are predictions about persons or conditions in the world derived from machine learning analysis of data about past events and its similarity to a given set of conditions.

**Automated Decision System:** "An automated decision system (ADS) augments or replaces human decision-making by using computational processes to produce answers to questions either as discrete classifications (e.g., yes, no; male, female, non-binary; malignant, benign) or continuous scores (e.g., degree of creditworthiness, risk of crime occurrence, projected tumour growth). Most ADS produce predictions about persons or conditions using machine learning and other computational logic by calculating the probability that a given condition is met. Typically, an automated decision system is ""trained"" on historical data looking for patterns of relationships between data points (e.g., the relationship between barometer readings, ambient temperature, and snowfall). An automated decision is made by comparing known patterns with existing inputs to estimate how closely they match (e.g., weather prediction based on the similarity between today's climate readings and those from the past). Examples of ADS include algorithms that calculate credit scores and biometric recognition systems that attempt to identify individual people based on physical traits, such as facial features.

**Clear Provenance and data lineage:** Clear provenance and data lineage consists of records that are accessible and simultaneously detail how data was collected and how it has been used and altered throughout the stages of pre-processing, modelling, training, testing, and deploying.

**Complete:** All recorded data requires an audit trail to show nothing has been deleted or lost.

**Concept drifts and shifts:** Once trained, most machine learning systems operate on static models of the world that have been built from historical data which have become fixed in the systems’ parameters. This freezing of the model before it is released ‘into the wild’ makes its accuracy and reliability especially vulnerable to changes in the underlying distribution of data. When the historical data that have crystallised into the trained model’s architecture cease to reflect the population concerned, the model’s mapping function will no longer be able to accurately and reliably transform its inputs into its target output values. These systems can quickly become prone to error in unexpected and harmful ways. There has been much valuable research done on methods of detecting and mitigating concept and distribution drift, and you should consult with your technical team to ensure that its members have familiarised themselves with this research and have sufficient knowledge of the available ways to confront the issue. In all cases, you should remain vigilant to the potentially rapid concept drifts that may occur in the complex, dynamic, and evolving environments in which your AI project will intervene. Remaining aware of these transformations in the data is crucial for safe AI, and your team should actively formulate an action plan to anticipate and to mitigate their impacts on the performance of your system.

**Consistent:** Consistency includes ensuring data is chronological, i.e., has a date and time stamp that is in the expected sequence.

**Contemporaneous:** Data should be recorded as it was observed, and at the time it was executed.

**Continuous learning:** A type of machine learning where the underlying algorithm or model are adjusted continuously while deployed.

**Critical Functions:** "Activities that are considered “critical” are economic and social activities of which “the interruption or disruption would have serious consequences on: 1) the health, safety, and security of citizens; 2) the effective functioning of services essential to the economy and society, and of the government; or 3) economic and social prosperity more broadly” (OECD, 2019[2]); (OECD, 2019[4]). It is important to note that not all systems in a critical sector are critical. For example, the administrative time tracking systems of a hospital or a bank are not considered to be critical systems.

**Data controller:** The individual or organisation which, alone or jointly with others, determines how and why personal data will be processed.

**Data Protection Impact Assessment:** "Where a type of processing in particular using new technologies, and taking into account the nature, scope, context and purposes of the processing, is likely to result in a high risk to the rights and freedoms of natural persons, the controller shall, prior to the processing, carry out an assessment of the impact of the envisaged processing operations on the protection of personal data. Data Protection Impact Assessments should contain at least: a systematic description of the envisaged processing operations and the purposes of the processing, including, where applicable, the legitimate interest pursued by the controller; an assessment of the necessity and proportionality of the processing operations in relation to the purposes; an assessment of the risks to the rights and freedoms of data subjects referred to in paragraph 1; and the measures envisaged to address the risks, including safeguards, security measures and mechanisms to ensure the protection of personal data and to demonstrate compliance with this GDPR regulation taking into account the rights and legitimate interests of data subjects and other persons concerned."

**Data subject:** An identified or identifiable natural person, who can be identified, directly or indirectly, by information such as a name or identity number, or by a combination of characteristics specific to that individual.

**Discrimination:** According to the UN Human Rights Committee overseeing the International Covenant on Civil and Political Rights, discrimination “should be understood to imply any distinction, exclusion, restriction or preference which is based on any ground such as race, colour, sex, language, religion, political or other opinion, national or social origin, property, birth or other status, and which has the purpose or effect of nullifying or impairing the recognition, enjoyment or exercise by all persons, on an equal footing, of all rights and freedoms.” Other definitions highlight the effects of systemic rather than explicit discrimination in which social and legal structures and/or historical conditions produce discriminatory effects even where a specific intention to discriminate is not present.

**Disproportionate Impact:** Where a system produces effects distributed across groups characterised by identity or demographic categories and such distributions are experienced in greater or lesser degree by members of one or more group. Disproportionality is most often considered when effects are especially harmful to the interests or life chances of marginalised groups.

**Duty-Bearers:** Actors who have human rights duties or responsibilities towards rights-holders. These include the company operating a project or conducting its activities, business suppliers and contractors, joint-venture and other business partners, state actors such as local government authorities and regional and national government departments and agencies.

**Equality Impact Assessment:** A formal evaluative process of identifying rightsholders whose right to equal status in society, access to opportunities, or treatment may be adversely affected by a technical system, policy, or practice in relation to other rightsholders, the extent to which they are affected, and the availability of redress or remediation.

**European Social Charter:** A Council of Europe treaty that guarantees fundamental social and economic rights as a counterpart to the European Convention on Human Rights, which refers to civil and political rights. It guarantees a broad range of everyday human rights related to employment, housing, health, education, social protection and welfare.

**Feature engineering:** Features are the input variables that are used during model training and are transformed by an algorithm. For example, a house price estimation tool used by mortgage valuators may use the following features, 'number of rooms', 'location', 'age of building', 'parking space' and 'size (sq. ft)' to predict the market value of the property. Features are often selected on the basis of prior domain knowledge but can also be engineered on the basis of their predictive or explanatory value. For instance, a machine learning algorithm could be fed raw data and trained to extract the most predictive features. In these cases, the features may not be easily recognisable using traditional concepts or labels.

**Flourish:** Human flourishing is a term used to refer to self-actualization or an individual's achievement of a desirable life goal and is most typically associated with well-being. For some theories of well-being, flourishing is a goal-oriented process that is concerned with an individual's psychology or emotional state. For other theories, the term refers to a broader social good that depends on the cultivation of social relationships.

**Fundamental Rights and Freedoms:** The first international standard for fundamental rights and freedoms was established by the United Nation's adoption of the Universal Declaration of Human Rights (UDHR), a document establishing basic rights aimed to safeguard the inviolable dignity of every person, which would become the basis for the many treaties, conventions, and charters on human rights that have been adopted worldwide up to the present. These treaties, charters, and conventions include: The European Convention on Human Rights (ECHR), which placed obligations on governments to protect ordinary people against human rights violations. The European Social Charter (ESC), which extends basic rights to include social and economic rights covering health, working conditions, housing, migrant labour, gender equality, and social security, The International Covenant on Economic, Social and Cultural Rights (ICESCR) (taking together the International Covenant on Civil and Political Rights and the International Covenant on Economic, Social and Cultural Rights)included freedom from torture, right to a fair trial, non-discrimination, and privacy rights, and extending basic rights to include rights to just working conditions, health, living standards, education, and social security. The most recent document establishing fundamental rights and freedoms is the Charter of Fundamental Rights of the European Union (CFR), which codified a basic set of civil, political, social, economic, and cultural rights for citizens of the European Union in EU law, including rights pertaining to human dignity, fundamental freedoms, equality, solidarity, and economic rights, and rights to participation in the life of the community.

**Harm to dignity**: An innate worth or status that is not earned and cannot be forfeited. The dignity of rightsholders is harmed where, as subjects of action or inaction, they experience the involuntary diminishment of their fundamental worth, self-respect, or personal autonomy, or else are treated as objects, humiliated, or otherwise degraded.

**Hidden proxy:** A variable which represents some property or feature of an object or person that is closely correlated with another (hidden) property or feature. For instance, the variable 'profession' is likely to have a high degree of mutual information with the variable 'salary'. In this example, the 'salary' variable is the hidden proxy if it is not directly recorded, but the 'profession' variable is present in the dataset. Hidden proxies can be problematic when they represent protected characteristics that were not intended to be in the dataset.

**High-dimensional:** In the context of data, dimensionality refers to the number of features (or variables) within a dataset. High-dimensional, therefore, is used to indicate a large number of features (e.g., genetics datasets). This is not to be confused with the size of the data as determined by the number of records (e.g., number of people).

**Highly Regulated:** A domain of practice can be considered highly regulated where multiple laws and policies are targeted specifically to a significant number of potential activities and relations of the domain. By example, most countries have highly regulated financial sectors with government agencies or departments dedicated to the sector, strict guidelines for the conduct of business, a range of consumer protections, and substantial penalties for rule violation.

**Human Rights Impact Assessment:** A formal evaluative process of identifying rightsholders whose human rights or fundamental freedoms may be adversely affected by a technical system, policy, or practice, the extent to which they are affected, and the availability of redress or remediation.

**Human-in-the-loop design:** The outputs of an AI model can be used as part of a wider process in which a human considers the output, as well as other information available to them, and then acts (makes a decision) based on this.

**Informational plurality:** A condition or environment where information (broadly defined) is available from multiple voices and perspectives and/or where there are multiple, differentiated sources of information.

**Intelligibility/Understandability:** Capable of being understood or comprehended, not requiring specific technical expertise.

**Interpretability:** Consists of the ability to know how and why a model performed the way it did in a specific context and therefore to understand the rationale behind its decision or behaviour.

**Labelling bias:** Labelling (or annotation) bias is variation in the meaning of a label used to represent a feature in training data. This commonly occurs when a specific label does not mean the same thing for all data subjects or rights holders (e.g., an emotion label that is applied to images of faces). Variation in a label's meaning is problematic when the specific label is strongly predictive of an outcome, as this can create situations where certain groups are miscategorised or receive inaccurate outcomes.

**Legal basis:** Derived from identifiable law or doctrine that can assert the lawfulness of an action or set of actions.

**Major risk:** A degree of risk in which the effects of a project or condition could be devastating but may be manageable if enacted with extreme care and close supervision.

**Marginalisation:** To treat a person, group, or concept as insignificant or peripheral. In race and social justice discourse, members of society are marginalised by holding an identity or being placed in a demographic category that is not attached to the dominant side of the prevailing structure of power and by which they experience oppression.

**Mass deception:** To deceive someone is to present (false) information as if it were true and to elicit some related behaviour on the basis of this information.

**Mass manipulation:** To manipulate someone is to get them to act in a particular way, perhaps by withholding salient information or influencing how they evaluate some state of the world.

**Measurement errors:** Any difference in the recorded value for some variable and its real-world value (e.g., measurement of the variable height). Measurement error can occur as a result of poor data collection processes, badly calibrated measuring devices, or from a range of biases. In the case of measurement bias, error can arise when the measurement scale does not capture data pertaining to the subjects in an equal manner (e.g., a self-reported pain scale that obscures underlying differences between two sub-groups).

**Moderate risk:** A degree of risk in which potential harms are directly or indirectly associated with risks of adverse impacts on the human rights and fundamental freedoms of affected persons, democracy, and the rule of law but that provide opportunities for risk reduction and mitigation that make the risks posed broadly acceptable.

**Non-deterministic**: A process is non-deterministic if its outcome cannot be predicted (or determined), due to a lack of knowledge about how its causal structure transforms the initial conditions. A non-deterministic algorithm is one where an input variable can take on the same value but provide different results.

**Non-Linear:** A nonlinear system is one in which the change in the output is not directly proportional to the change of the input. As a result, the dynamics of the system cannot be represented graphically using a straight line.

**Other Relevant Parties:** Stakeholders other than rights-holders and duty-bearers, which may include individuals or organisations representing the interests of rights holders and official representations at international, national and local levels (e.g., the UN, national human rights institutions, NGOs or civil society organisations).

**Performance Metrics:** Metrics used to define how good the algorithm is at reaching the correct decision that range beyond accuracy measures.

**Physical or Mental Integrity:** The core attributes of an individual's overall bodily integrity. Bodily integrity is an inviolable right that an individual has regarding the self-determination of their own body and mind.

**Precision:** The number of correctly identified positive predictions divided by the number of all incorrect and correct positive predictions. Precision helps us to understand how accurate the model is – out of those predicted positive, how many of them are actually positive? This is an especially useful metric when the importance of false positives is high.

**Pre-designated High-Risk or Safety Critical Sector:** Annex III of the European Union Proposed Rules on Artificial Intelligence indicates 'high-risk' sectors as those concerned with education, emergency services, employment, financial services, public benefits, law enforcement, immigration, border control, and the administration of justice and democratic processes.

**Probabilistic:** An event is probabilistic if the extent to which it occurs is considered to be subject to chance variation. In many cases, this is a reflection of our own uncertainty about the event's occurrence, rather than assuming the existence of a non-deterministic process.

**Prohibited AI system:** As outlined in the European Union Proposed Rules on Artificial Intelligence (Art. 5), prohibited AI systems are those that include manipulation, exploitation, social scoring, or real-time remote biometric ID.

**Prohibitive Risk:** A degree of risk significant enough to trigger the precautionary principle and precipitate pre-emptive measures to prevent adverse impacts on the human rights and fundamental freedoms of affected persons, democracy, and the rule of law. Pre-emptive measures are appropriate where the severity, scale, and irremediableness of the potential harm outweigh levels of risk reduction and mitigation. May also refer to a degree of risk in which the effects of a project or condition would be so devastating upon their occurrence that it would be irrational, irresponsible, or immoral to proceed.

**Proportionality:** A legal principle that refers to the idea of delivering a just outcome in ways that are proportionate to the cost, complexity, and resources available. In a similar vein, it can also be used as an evaluative notion, such as in the case of a data protection principle that states only personal data that are necessary and adequate for the purposes of the task are collected.

**Protected Characteristics:** Identities, categories, or other descriptive features of natural persons defined by human rights, civil rights, or data protection law, as well as other doctrines concerned with legal rights and equal protection. Such characteristics are 'protected' from arbitrary or capricious use in the allocation or withholding of rights and social goods. Examples in European law and regulation include racial or ethnic origin, political opinion, religion or beliefs, trade union membership, genetic or health status, and sexual orientation. Other commonly protected characteristics include skin tone, gender-identity, and cognitive or physical ability.

**Relevance, Appropriateness, and Domain Knowledge:** The understanding and utilisation of the most appropriate sources and types of data are crucial for building a robust and unbiased AI system. Solid domain knowledge of the underlying population distribution and of the predictive or classificatory goal of the project is instrumental for choosing optimally relevant measurement inputs that contribute to the reasonable determination of the defined solution. Domain experts should collaborate closely with the technical team to assist in the determination of the optimally appropriate categories and sources of measurement.

**Reliability:** The objective of reliability is that an AI system behaves exactly as its designers intended and anticipated. A reliable system adheres to the specifications it was programmed to carry out. Reliability is therefore a measure of consistency and can establish confidence in the safety of a system based upon the dependability with which it operationally conforms to its intended functionality.

**Rights-Holders:** All individuals are human rights-holders. These include workers and families, contractor (goods and services providers) employees and families, impacted community members, including individuals of all genders, children, indigenous peoples, migrant workers, ethnic minorities and so forth (both within the geographic vicinity of operations but also impacted downstream, trans-boundary or neighbouring communities), human rights defenders, and consumers. Within the HUDERIA, the primary focus is on rights-holders who are, or may be, adversely affected by a project.

**Robustness:** The objective of robustness can be thought of as the goal that an AI system functions reliably and accurately under harsh conditions. These conditions may include adversarial intervention, implementer error, or skewed goal-execution by an automated learner (in reinforcement learning applications). The measure of robustness is therefore the strength of a system's integrity the soundness of its operation in response to difficult conditions, adversarial attacks, perturbations, data poisoning, and undesirable reinforcement learning behaviour.

**Security:** The goal of security encompasses the protection of several operational dimensions of an AI system when confronted with possible adversarial attack. A secure system is capable of maintaining the integrity of the information that constitutes it. This includes protecting its architecture from the unauthorised modification or damage of any of its component parts. A secure system also remains continuously functional and accessible to its authorised users and keeps confidential and private information secure even under hostile or adversarial conditions.

**Sensitivity/Recall:** The number of correct predictions divided by the number of all samples that should have been identified as positive. Recall helps us to understand how many actual positives are captured through the model labelling it positive. This is an especially useful metric when the importance of false negatives is high.

**Snowball Approach:** An approach to sampling where a researcher or organisation begins with a small number of contacts or stakeholders who have been identified as relevant. During initial engagement with these stakeholders' recommendations are sought for other potentially relevant stakeholders to engage. As this process continues the sample of stakeholders increases and diversifies.

**Socio-technical System:** A system that couples human (or social) behaviour to the functioning of a technical system, and in doing so gives rise to novel (and emergent) functions that are not reducible to either the human or technical elements. By intervening in human behaviours, attitudes, or their relations to the world, the technical system restructures human behaviour. The socio-technical perspective is one that considers the human desires or goals a technology is meant to, or does, achieve. We can also describe as socio-technical those systems whose very existence, implementation, or effects implicate human political, economic, or social relations. For example, surveillance systems adopted by law enforcement agencies are socio-technical because their adoption and use have political dimensions; the selected targets of police surveillance are affected more acutely than others by the use of surveillance technologies based on the historical choices made by government and law enforcement officials. From this socio-technical perspective, surveillance technologies participate in relations between people and the centres of power in society.

**Stakeholder:** Individuals or groups that (1) have interests or rights that may be affected by the past, present, and future decisions and activities of an organisations; (2) may have the power or authority to influence the outcome of such decisions and activities; (3) possess relevant characteristics that put them in positions of advantage or vulnerability with regard to those decisions and activities.

**Target Variable:** A feature within data defined as the output that a machine learning model tries to predict. Each of the other features considered by the model are used to predict the target variable. For example, for an ML model built to predict what tomorrow's temperature may be, the target variable would be tomorrow's temperature, features in input data such as precipitation levels, humidity, amongst others would be used by the model to predict this target variable.

**Timeliness and Recency (up-to-date):** If datasets include outdated data, then changes in the underlying data distribution may adversely affect the generalisability of the trained model. Provided these distributional drifts reflect changing social relationship or group dynamics, this loss of accuracy with regard to the actual characteristics of the underlying population may introduce bias into an AI system. In preventing discriminatory outcomes, timeliness and recency of all elements of the data that constitute the datasets must be scrutinised.

**Traceability:** Refers to the process by which all stages of the data lifecycle from collection to deployment to system updating or deprovisioning are documented in a way that is accessible and easily understood. This may include not only the parties within the organisation involved but also the actions taken at each stage that may impact the individuals who use the system.

**Vulnerability:** Refers to being "at a higher risk of being unable to anticipate, cope with, resist, and recover from project-related risks and/or adverse impacts." Such risk may be corporeal, as in being subject to heightened risk of physical or psychological harm, or non-corporeal, as in being at heightened risk of diminished economic or social status. Vulnerable individuals or groups may include women, children, the elderly, the poor, ethnic, religious, cultural or linguistic minorities, or indigenous groups.

**Vulnerable Group:** Refers a degree of risk from harm associated membership in a particular group or category. For example, people living in poverty are members of a vulnerable group (i.e., "the poor") because they share characteristics that put them at greater risk of particular forms of harm than persons belonging to more affluent groups.